\ifdefined\pdfminorversion
\fi
\PassOptionsToPackage{expansion=false}{microtype}
\documentclass[11pt,letterpaper,logo]{mystyle}
\usepackage[numbers,sort&compress]{natbib}
\usepackage{mathtools}
\usepackage{amsthm}
\usepackage{amssymb}
\usepackage{graphicx}
\usepackage{booktabs}
\usepackage{array}
\usepackage{tabularx}
\usepackage{wrapfig}
\newcommand{\wraplift}{\vspace*{-14pt}}
\IfFileExists{longtable.sty}{\usepackage{longtable}}{}
\usepackage{multirow}
\IfFileExists{listings.sty}{\usepackage{listings}}{
  \usepackage{verbatim}
  \newcommand{\lstdefinestyle}[2]{}
  \newcommand{\lstset}[1]{}
  
}
\usepackage{algorithm}
\usepackage{algpseudocode}
\usepackage{enumitem}
\usepackage{microtype}
\usepackage{url}
\usepackage{hyperref}
\IfFileExists{fontawesome5.sty}{\usepackage{fontawesome5}}{}

\providecommand{\coloremojicode}[1]{%
  \raisebox{-0.15em}{\includegraphics[height=1.25em]{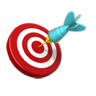}}}
\definecolor{PrimitiveBlue}{HTML}{263E62}
\definecolor{PrimitiveTeal}{HTML}{147D85}
\definecolor{PrimitiveGray}{HTML}{F3F5F7}
\definecolor{PrimitiveInk}{HTML}{1D2836}
\newcommand{\ours}{HEELO}
\newcommand{\papertitle}{The Last Mile to the Real World: Agent-Native Primitives as Reusable Building Blocks for Solving Real-World Tasks}
\newcommand{\paperrunningtitle}{The Last Mile to the Real World}
\newcommand{\sid}[1]{\textsc{#1}}

\newcommand{\Lib}{\mathcal{L}}
\newcommand{\Pre}{\operatorname{Pre}}
\newcommand{\Obl}{\operatorname{Obl}}
\newcommand{\pass}{\textup{\textsc{pass}}}
\newcommand{\fail}{\textup{\textsc{fail}}}
\newcolumntype{Y}{>{\raggedright\arraybackslash}X}
\newcolumntype{P}[1]{>{\raggedright\arraybackslash}p{#1}}
\theoremstyle{definition}
\newtheorem{definition}{Definition}[section]
\theoremstyle{plain}
\newtheorem{proposition}{Proposition}[section]
\setlist[itemize]{leftmargin=*,itemsep=2pt,topsep=3pt}
\setlist[enumerate]{leftmargin=*,itemsep=2pt,topsep=3pt}
\lstdefinestyle{manifest}{
  basicstyle=\ttfamily\footnotesize,
  breaklines=true,
  columns=fullflexible,
  keepspaces=true,
  showstringspaces=false,
  frame=single,
  rulecolor=\color{PrimitiveBlue!40},
  backgroundcolor=\color{PrimitiveGray},
  xleftmargin=4pt,
  xrightmargin=4pt,
  aboveskip=7pt,
  belowskip=7pt
}
\DeclareMathSizes{10.95}{10.95}{7}{5}
\hypersetup{colorlinks=true,linkcolor=PrimitiveBlue,citecolor=PrimitiveTeal,urlcolor=PrimitiveBlue}
\makeatletter
\IfFileExists{placeins.sty}{\usepackage{placeins}}{
  \newcommand{\FloatBarrier}{%
    \par
    \begingroup
      \let\@elt\relax
      \edef\primitive@pendingfloats{\@botlist\@deferlist\@dbldeferlist}%
      \ifx\primitive@pendingfloats\@empty
      \else
        \newpage
        \edef\primitive@pendingfloats{\@botlist\@deferlist\@dbldeferlist}%
        \ifx\primitive@pendingfloats\@empty\else\clearpage\fi
      \fi
    \endgroup}}
\makeatother
\IfFileExists{makecell.sty}{\usepackage{makecell}}{
  \newcommand{\makecell}[2][c]{\begin{tabular}[#1]{@{}c@{}}#2\end{tabular}}}
\definecolor{TabRule}{HTML}{22406A}
\definecolor{TabHead}{HTML}{D6E2F0}
\definecolor{TabHeadText}{HTML}{1B3358}
\definecolor{TabGroup}{HTML}{EDF0F4}
\definecolor{TabGroupText}{HTML}{44505E}
\definecolor{TabOurs}{HTML}{F1F5FA}
\definecolor{TabBest}{HTML}{D2EBDB}
\definecolor{TabSecond}{HTML}{FBE6C8}
\definecolor{TabUp}{HTML}{2E7D4F}
\definecolor{TabDown}{HTML}{B5403A}
\definecolor{TabDelta}{HTML}{5B6876}
\newcommand{\hdr}[1]{{\color{TabHeadText}\textbf{\boldmath #1}}}
\newcommand{\best}[1]{\cellcolor{TabBest}\textbf{#1}}
\newcommand{\second}[1]{\cellcolor{TabSecond}\underline{#1}}
\newcommand{\tabgroup}[2]{\rowcolor{TabGroup}\multicolumn{#1}{l}{\color{TabGroupText}\textit{#2}}}
\newcommand{\tabours}{\rowcolor{TabOurs}}
\newcommand{\hdrgapfill}{\noalign{\begingroup\color{TabHead}%
  \hrule height\dimexpr\aboverulesep+\cmidrulewidth+\belowrulesep\relax\endgroup
  \vskip-\dimexpr\aboverulesep+\cmidrulewidth+\belowrulesep\relax}}
\newcommand{\stg}[1]{\texttt{\mdseries #1}}

\makeatletter
\newcommand{\sn}[1]{\research@sn#1\@nil}
\def\research@sn#1#2\@nil{\ifx-#1\textminus#2\else#1#2\fi}
\makeatother
\newcommand{\textexp}[1]{\textsuperscript{\sn{#1}}}
\makeatletter
\newcommand{\dpp}[1]{\research@dpp#1\@nil}
\def\research@dpp#1#2\@nil{{\scriptsize\,\ifx-#1\textcolor{TabDown}{(\textminus#2)}\else\ifx+#1\textcolor{TabUp}{(+#2)}\else\textcolor{TabDelta}{(#1#2)}\fi\fi}}
\makeatother
\newsavebox{\researchtablebox}
\makeatletter
\def\research@fsize{9}% font size of the current table, in pt
\newcommand{\fitresearchtable}[1]{%
  \sbox{\researchtablebox}{#1}%
  \ifdim\wd\researchtablebox>\linewidth
    \ifdim 8\wd\researchtablebox>\research@fsize\linewidth
      \PackageWarning{research-style}{A table is shrunk below 8pt}%
    \fi
    \resizebox{\linewidth}{!}{\usebox{\researchtablebox}}%
  \else
    \usebox{\researchtablebox}%
  \fi}
\newif\ifresearchpanel
\def\research@panelsize{8.5}
\newcommand{\researchpanelsize}[1]{\def\research@panelsize{#1}}
\newcommand{\researchtablesetup}{%
  \arrayrulecolor{TabRule}%
  \setlength{\heavyrulewidth}{0.9pt}%
  \setlength{\lightrulewidth}{0.45pt}%
  \setlength{\cmidrulewidth}{0.35pt}%
  \ifresearchpanel\ifdim\research@panelsize pt<9pt
    \fontsize{8.5}{10}\selectfont
    \setlength{\tabcolsep}{3pt}%
    \def\research@fsize{8.5}%
  \else
    \fontsize{9}{10.5}\selectfont
    \setlength{\tabcolsep}{3.4pt}%
    \def\research@fsize{9}%
  \fi\else
    \fontsize{9}{10.5}\selectfont
    \setlength{\tabcolsep}{3.4pt}%
    \def\research@fsize{9}%
  \fi
  \setlength{\extrarowheight}{0.2pt}%
  \setlength{\aboverulesep}{0.4ex}%
  \setlength{\belowrulesep}{0.4ex}%
  \setlength{\cmidrulekern}{0.25em}%
}
\newenvironment{researchtable}
  {\ifresearchpanel\let\research@next\relax
   \else\def\research@next{\table[!htbp]}\fi
   \research@next}
  {\ifresearchpanel\par\else\endtable\fi}
\makeatother

\DeclareCaptionFont{research}{\fontsize{9}{11.4}\selectfont}
\DeclareCaptionFont{tabnavy}{\color{TabRule}}
\DeclareMathSizes{9}{9}{7}{5}
\DeclareMathSizes{8.5}{8}{6}{5}
\DeclareMathSizes{8}{7}{5}{5}
\DeclareMathSizes{10}{10}{7}{5}
\DeclareMathSizes{11.5}{11.5}{7}{5}

\NewDocumentCommand{\fillpar}{s O{.6} O{0}}{{\IfBooleanT{#1}{\tolerance=1000 \emergencystretch=2em }%
  \looseness=#3\relax\parfillskip=0pt plus #2\linewidth\relax\par}}
\NewDocumentCommand{\fillcap}{O{.6} O{0}}{{\looseness=#2\relax\parfillskip=0pt plus #1\linewidth\relax
  \ifhmode\unskip\nobreak\vrule width0pt height0pt depth\dp\strutbox\fi\par}}
\NewDocumentEnvironment{filldisp}{O{.6} O{0}}{\looseness=#2\relax\parfillskip=0pt plus #1\linewidth\relax}{\ignorespacesafterend}

\titleformat{\section}
  {\normalfont\fontsize{13}{16}\selectfont\bfseries\color{black}}
  {\thesection.}{0.55em}{#1}
\titleformat{name=\section,numberless}
  {\normalfont\fontsize{13}{16}\selectfont\bfseries\color{black}}
  {}{0pt}{#1}
\titleformat{\subsection}
  {\normalfont\fontsize{11.5}{14}\selectfont\bfseries\color{black}}
  {\thesubsection.}{0.55em}{#1}
\titleformat{\subsubsection}
  {\normalfont\fontsize{11}{13}\selectfont\bfseries\color{black}}
  {\thesubsubsection.}{0.55em}{#1}
\titleformat{\paragraph}[runin]
  {\normalfont\bfseries}{\theparagraph}{0pt}{#1}
\titlespacing*{\section}{0pt}{14pt plus 3pt minus 2pt}{6pt}
\titlespacing*{\subsection}{0pt}{11pt plus 2pt minus 2pt}{5pt}
\titlespacing*{\subsubsection}{0pt}{10pt plus 2pt minus 2pt}{4pt}
\titlespacing*{\paragraph}{0pt}{6pt plus 1pt minus 1pt}{0.6em}

\makeatletter
\ifdefined\pdfsavepos
  \newcommand{\runningsection}[1]{%
    \pdfsavepos
    \protected@write\@auxout{}{\string\rs@at{\thepage}{\noexpand\the\pdflastypos}{#1}}%
    \markright{#1}}
\else
  \newcommand{\runningsection}[1]{\markright{#1}}
\fi
\def\rs@at#1#2#3{%
  \@ifundefined{rs@first@#1}{\global\@namedef{rs@first@#1}{#3}%
    \global\@namedef{rs@y@#1}{#2}}{}%
  \global\@namedef{rs@last@#1}{#3}}
\newcount\rs@k
\def\rs@search{%
  \ifnum\rs@k>\@ne \global\advance\rs@k\m@ne
    \expandafter\@firstofone\else\expandafter\@gobble\fi
  {\@ifundefined{rs@last@\the\rs@k}{\rs@search}{\@nameuse{rs@last@\the\rs@k}}}}
\newcommand{\primitiverunninghead}{%
  \ifdefined\pdfsavepos
    \global\rs@k=\c@page
    \@ifundefined{rs@first@\the\rs@k}{\rs@search}{%
      \ifdim\@nameuse{rs@y@\the\rs@k}sp>\dimexpr\paperheight-1in-\voffset
          -\topmargin-\headheight-\headsep-0.5\textheight\relax
        \@nameuse{rs@first@\the\rs@k}%
      \else\rs@search\fi}%
  \else\rightmark\fi}
\makeatother
\renewcommand{\sectionmark}[1]{}
\renewcommand{\subsectionmark}[1]{}
\renewcommand{\headrule}{\color{black!35}\DefaultHeadRule}

\fancypagestyle{firststyle}{
  \fancyhf{}
  \fancyhead[L]{\ifthenelse{\boolean{logo}}
    {\logoimg[12pt]{1 1 1 1}{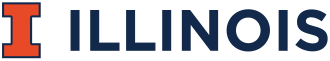}\hspace{12pt}\raisebox{-1.5pt}{\includegraphics[height=15pt]{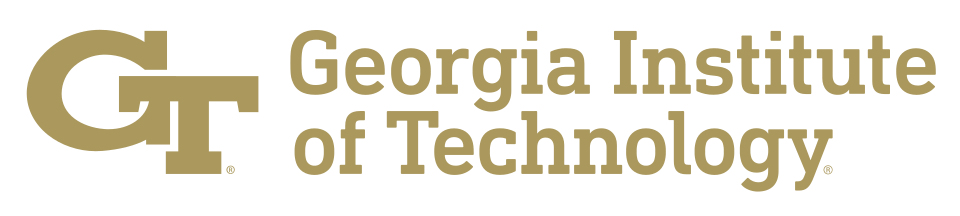}}}{}}
  \fancyfoot[R]{\footerfont\thepage}
  
  \renewcommand{\headrule}{\color{black}\DefaultHeadRule}
  
}
\makeatletter
\newcommand{\wrapguard}{\par
  \ifnum\c@WF@wrappedlines>\@ne
    \penalty\interlinepenalty
    \dimen@\numexpr\c@WF@wrappedlines-\@ne\relax\baselineskip
    \@tempdimb\pagegoal\advance\@tempdimb-\pagetotal
    \ifdim\pagetotal>\z@\ifdim\dimen@<\@tempdimb\vspace*{\dimen@}\fi\fi
  \fi
  \ifvoid\WF@box\WFclear\else
    \global\setbox\WF@box\hbox to\linewidth{\hss\box\WF@box\hss}%
    \WFclear\vskip\intextsep
  \fi}
\makeatother
\renewcommand{\titlefont}{\centering\color{black}\normalfont\fontsize{17}{19}\selectfont}

\renewcommand{\maketitle}{%
  \vspace*{5.5mm}%
  \primitiveTemplateMakeTitle
}

\ifdefined\pdfminorversion\else
  \AtBeginDvi{}
\fi

\newtcolorbox{abstractbox}{colback=white,colframe=LARGBlue!55,boxrule=0.8pt,arc=3mm,fontupper=\small,
  left=5mm,right=5mm,top=1.5mm,bottom=3mm,title={\large Abstract},fonttitle=\bfseries,
  coltitle=black,colbacktitle=white,halign title=center,titlerule=0pt,toptitle=1.5mm}
\tcbset{titlebox/.style={colback=white,colframe=white,boxrule=0pt,arc=0mm,left=5mm,right=5mm,top=1mm,bottom=1mm}}

\runningtitle{The Last Mile to the Real World}
\title{%
\textbf{The Last Mile to the Real World: Agent-Native Primitives as Reusable Building Blocks for Solving Real-World Tasks}\\
\vspace{0.3em}
{\fontsize{11.5}{13.5}\selectfont\scshape\color{black}
  $\lozenge$~Foundation~$\cdot$~Adaptation~$\cdot$~Composition~$\cdot$~Evolution~$\lozenge$}
\vspace{-0.8em}
}
\date{\vspace{-3ex}}
\hypersetup{
  pdftitle={\papertitle},
  pdfauthor={Haibo Jin, Peng Kuang, Xiaopeng Yuan, Lucheng Fu, Jerry Wang, Xucheng Yu, Yaoning Yu, Dehao Wu, Runguo Li, Colin Su, Jingxuan Sun, Xuyang Liu, Haojing Luo, and Haohan Wang},
  pdfsubject={Agent-native primitives and harness engineering for real-world tasks: LLM tool use, repository construction, formal theorem proving, knowledge-intensive tasks, multi-agent systems, software engineering, large information spaces, and agent harnesses}
}
\begin{document}
% Fancy author block of the user-supplied template (materials/Primitive_as_Service/authors.tex).
% The class wraps \author to skip an empty \author{}; restore the authblk command for this block.
\makeatletter
\ifdefined\primitiveRawAuthor\let\author\primitiveRawAuthor\fi
\makeatother
\providecolor{Maroon}{cmyk}{0,0.87,0.68,0.32}
\providecommand{\authorletter}{\ifdefined\faEnvelope\faEnvelope[regular]\else$\ast$\fi}
\author{
   \normalfont
   Haibo~Jin$^{\textcolor{Maroon}{1}}$\hspace{0.85em}
   Peng~Kuang$^{\textcolor{Maroon}{1}}$\hspace{0.85em}
   Xiaopeng~Yuan$^{\textcolor{Maroon}{1}}$\hspace{0.85em}
   Lucheng~Fu$^{\textcolor{Maroon}{2\dagger}}$\hspace{0.85em}
   Jerry~Wang$^{\textcolor{Maroon}{1}}$\hspace{0.85em}
   Xucheng~Yu$^{\textcolor{Maroon}{1}}$\hspace{0.85em}
   Yaoning~Yu$^{\textcolor{Maroon}{1}}$\\
   Dehao~Wu$^{\textcolor{Maroon}{1}}$\hspace{0.85em}
   Runguo~Li$^{\textcolor{Maroon}{1}}$\hspace{0.85em}
   Colin~Su$^{\textcolor{Maroon}{1}}$\hspace{0.85em}
   Jingxuan~Sun$^{\textcolor{Maroon}{1}}$\hspace{0.85em}
   Xuyang~Liu$^{\textcolor{Maroon}{1}}$\hspace{0.85em}
   Haojing~Luo$^{\textcolor{Maroon}{3}}$\hspace{0.85em}
   Haohan~Wang$^{\textcolor{Maroon}{1}}$\textsuperscript{\authorletter}
   \\
   \vspace{5pt}
   \small

   \raisebox{1.0ex}{\includegraphics[height=1.1ex]{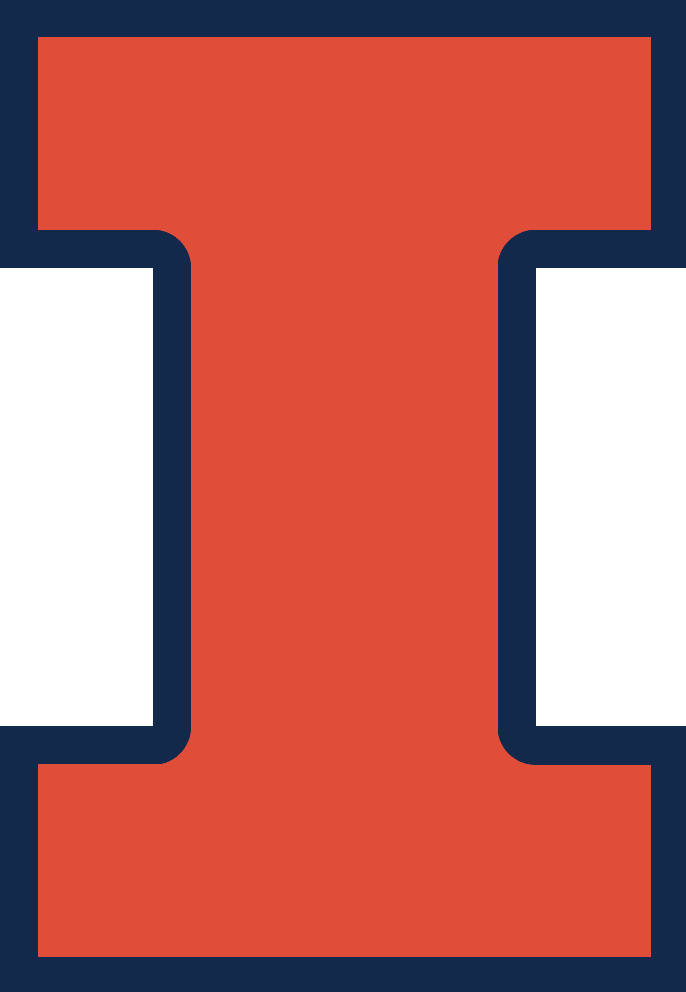}}$^{\textcolor{Maroon}{1}}$University of Illinois Urbana-Champaign \quad \raisebox{0.9ex}{\includegraphics[height=1.25ex]{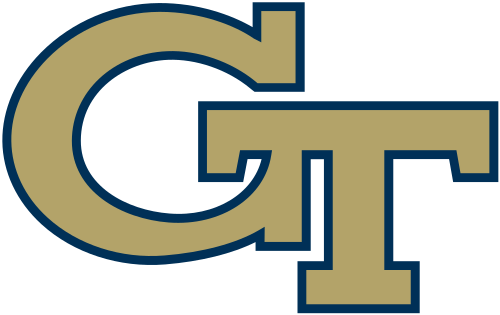}}$^{\textcolor{Maroon}{2}}$Georgia Institute of Technology \quad \raisebox{0.9ex}{\includegraphics[height=1.25ex]{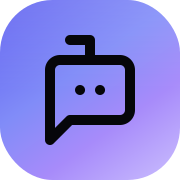}}$^{\textcolor{Maroon}{3}}$heelo.dev\\
   $^{\textcolor{Maroon}{\dagger}}$Work done at UIUC \quad \textsuperscript{\authorletter}\,\textit{Corresponding Author}
}

\begin{abstract}
\centering\raisebox{-0.2ex}{\includegraphics[height=1.75ex]{figs/heelo_icon.png}}~\textbf{Website}: \href{https://heelo.dev}{https://heelo.dev}
\end{abstract}
\maketitle
\begin{abstractbox}
How can an agent handle a real-world task end to end? Despite recent progress in LLM agents, frontier models paired with the harnesses built for them still fall short on public benchmarks of real-world tasks: on Terminal-Bench~4.0 at maximum reasoning effort, GPT-5.6 Sol with Codex resolves 37.3\% of the tasks and Claude Opus 5 with Claude Code resolves 51.8\%. These tasks fail at the step from reaching a resource to applying it, which we call the \textbf{last mile to the real world}: a tool call must ground its arguments, often in an earlier call's output, each result must reach the steps that consume it, and a failure must be traced back to the work it implicates. Existing approaches put the resource within reach but leave the competence to apply it with the caller, so each harness rebuilds that competence. To address this gap, we introduce \textbf{agent-native primitives}, reusable units that equip a resource with a resident LLM, which applies it in context and communicates with other primitives, and \textbf{\ours{}} (\textbf{H}arness \textbf{E}n\-gi\-neer\-ing around \textbf{E}volv\-ing Agent-Native Prim\-i\-tives for the \textbf{L}ast Mile to Real-W\textbf{O}rld Tasks), a harness engineering framework that builds each task's harness from a library of such primitives. \ours{} composes the primitives a task needs into a task-specific graph, verifies the result against evidence from the task's environment and revises the primitives that a failure implicates, and admits new or improved primitives into the library after validation. Results proved in our companion studies bound activation by the task's demand, however large the library, and identify when per-task composition beats the best fixed harness. Each task domain, however, places its own demand on the last mile, so we instantiate \ours{} with eight primitive families, each built as the solution to one kind of real-world task, and evaluate them on more than twenty benchmarks. On each comparator's backbone, Dev-Primitives raise Terminal-Bench~4.0 resolution from 37.3\% under Codex to 55.5\%. Over the mini-SWE-agent seed harness, Harness Primitives composed per task raise Pass@1 from 73.0\% to 80.5\% on 100 SWE-bench Verified tasks and from 60.0\% to 72.2\% on 45 Terminal-Bench~2.0 tasks, above the 79.0\% and 56.7\% of Codex. Tool Primitives raise ToolBench pass rate from 55.8\% to 75.1\%, Code Primitives raise repository delivery from 0.3180 to 0.5134, Agent Primitives raise mean accuracy over eight math, code, and QA benchmarks from 58.8\% for a single agent to 75.3\%, and Navigation Primitives raise GAIA-Text-103 accuracy from 45.6\% to 58.3\%. In formal theorem proving, Math Primitives place an annotated reference method in the top five for 95.0\% of held-out Putnam problems, against 84.2\% for the strongest reranker. Further analyses show that the benefit comes from applying each resource inside its primitive and that the task decides which primitives run. Beyond these benchmarks, we built heelo.dev, a platform of agent-native primitives for real-world actions. On its 46 real-world tasks, Claude Opus 5.5 with Claude Code and GPT-6.1 Sol with Codex complete 9 each on their own and 39 and 35 when connected to \ours{} through our MCP server, with $1.9\times$ and $2.3\times$ less time and $1.55\times$ lower cost per completed task. \ours{} provides a unified framework for building agents that handle real-world tasks end to end.

\par\smallskip
\coloremojicode{1F3AF}~\textbf{Keywords}: LLM Agents, Agent-Native Primitives, Harness Engineering, Real-World Tasks, LLM Tool Use
\end{abstractbox}
\clearpage
\begin{figure}[p]
\centering
\includegraphics[width=\linewidth]{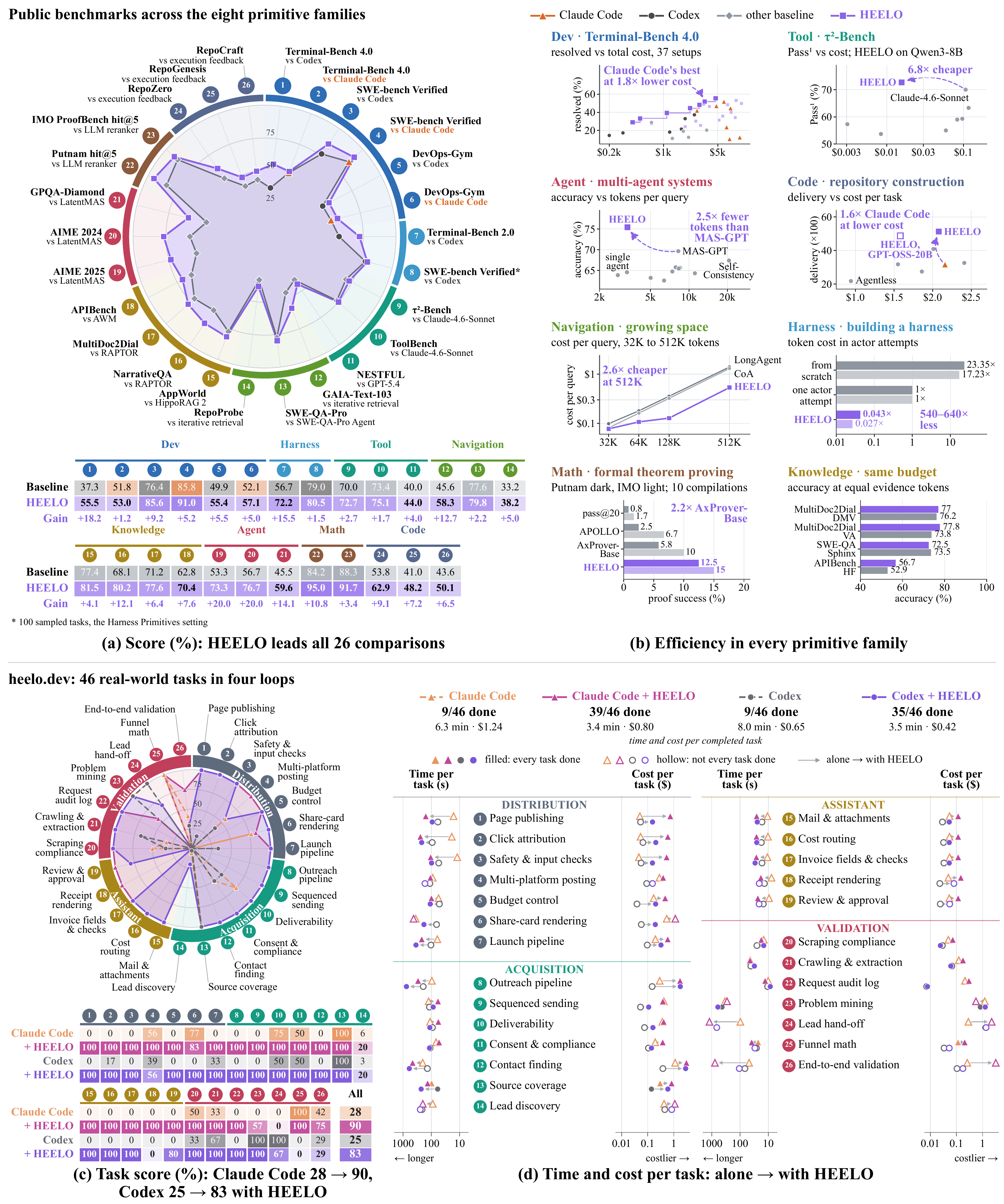}
\caption{\textbf{\ours{} on public benchmarks and on the real-world tasks of heelo.dev.} (a)~Each spoke compares \ours{} with one baseline: Codex or Claude Code on the same backbone for Dev and Harness Primitives, and otherwise the strongest baseline of the setting. (b)~Efficiency per family. (c,~d)~Claude Opus 5.5 with Claude Code and GPT-6.1 Sol with Codex run the 46 tasks on their own and connected to \ours{} through MCP; in (d), time and cost average a capability's tasks, and a marker is filled when every task of its capability is done.\fillcap}
\label{fig:heelo-results}
\end{figure}
\clearpage

\section{Introduction}
\label{sec:introduction}
\runningsection{Introduction}

A central objective of LLM agents is to complete real-world tasks end to end: to act on the world through tools, code, proofs, documents, and other agents until the task's own acceptance criteria are met~\citep{yao2023react,schick2023toolformer,wang2024executable}. Recent advances in language models and in the harnesses built around them have improved this ability, as GPT acts through Codex~\citep{openai2025codex} and Claude through Claude Code~\citep{anthropic2025claudecode}. Through these harnesses an agent reaches tools and APIs~\citep{qin2023toolllm,patil2023gorilla}, code in large repositories~\citep{yang2024swe,wang2025openhands}, formal proof libraries~\citep{demoura2021lean,mathlib2020}, documents and memories~\citep{lewis2020rag,packer2023memgpt}, and other agents~\citep{wu2023autogen,hong2023metagpt}. Meanwhile, benchmarks increasingly judge agents by whether the real task succeeds: the tool calls fulfil the request, the repository's tests pass, the answer is correct~\citep{qin2023toolllm,jimenez2023swebench,merrill2026terminal,ICLR2024_25ae35b5}. Existing studies have made resources easier to reach and to reuse through tool learning~\citep{schick2023toolformer,qin2023toolllm,liu2024toolace,patil2025bfcl}, retrieval and memory~\citep{lewis2020rag,sarthi2024raptor,shinn2023reflexion,zhao2024expel}, skill and code libraries~\citep{wang2023voyager,wang2024awm,wang2024trove}, and multi-agent frameworks, workflow search, and harness generators~\citep{hong2023metagpt,zhuge2024gptswarm,hu2024adas,zhang2024aflow,zhang2026jitagentscalingharnessintelligence}, providing an important foundation for agents that act in practical environments.

Despite this progress, frontier models in the harnesses built for them still fall short on public benchmarks of real-world tasks. On Terminal-Bench~4.0 at maximum reasoning effort, GPT-5.6 Sol with Codex resolves 37.3\% of the tasks and Claude Opus 5 with Claude Code resolves 51.8\%, and on the four stages of DevOps-Gym the same two pairs average 49.9\% and 52.1\% success. In the airline domain of $\tau^2$-Bench~\citep{barres2025tau}, Claude-4.6-Sonnet succeeds in all four trials on half of the tasks. Tasks as ordinary as booking a flight under an airline's policy or filling in a form whose later fields depend on earlier answers fail at the step from reaching a resource to applying it, which we call the \textbf{last mile to the real world}. Its difficulty shows in three aspects. \textbf{Application in context}: a reached resource must be applied to the task's own objects, as a tool call must ground every argument, often in an earlier call's output, and respect the rules of its environment~\citep{basu2025nestful,lu2025toolsandbox,yao2024tau}, retrieved code must be adapted to its target and to the components that depend on it~\citep{zhang2023repocoder,ding2023crosscodeeval}, a proof method applies only where its conditions hold~\citep{bundy1988use,melis1999knowledge}, and an answer drawn from documents must stay faithful to its source~\citep{liu-etal-2024-lost,asai2024self}. \textbf{Composition}: what one step returns or requires must reach the steps that consume it, in a form they can use. \textbf{Evidence-driven revision}: a failure observed in the task's environment must be traced back to the work it implicates, so that only this work is redone. Together, these aspects extend the last mile from a single call to the whole harness. However, existing approaches keep the resource but leave the competence to apply it with the caller, or fold it into one monolithic agent, so the harness rebuilds this competence for every task: what the caller must read grows with the library, what it learns ends with the task, and its results are hard to check. This raises a natural question: \textbf{how can an agent handle a real-world task end to end?}\fillpar[.6][1]

To address this gap, we argue that \textbf{the competence to apply a resource should travel with the resource itself}, and we introduce \textbf{agent-native primitives}, each a reusable resource equipped with a resident LLM. The primitive extends reusable software components, temporally extended options, and design by contract~\citep{szyperski2002component,sutton1999options,meyer1992contract} to LLM-based units. The resident LLM is what makes the primitive agent-native: it gives the resource a natural-language interface, applies it to each request in context, and communicates requirements to other primitives. Passive artifacts thus become active participants that talk to one another, and the primitive's interface lets a harness find and invoke it without reading the resource.

Primitives alone do not determine which of them a task needs, how their results reach one another, which to revise when the task fails, or which new primitives to admit. We therefore introduce \textbf{\ours{}} (\textbf{H}arness \textbf{E}n\-gi\-neer\-ing around \textbf{E}volv\-ing Agent-Native Prim\-i\-tives for the \textbf{L}ast Mile to Real-W\textbf{O}rld Tasks), a harness engineering framework that builds each task's harness, everything the task needs around the model, from an evolving library of primitives through three mechanisms. First, library selection and composition finds candidates through their interfaces or their own relevance assessments and composes the few the task needs into a task-specific graph that carries each result and message to its consumer. Second, evidence-based execution and improvement runs the graph in rounds, verifies the result against evidence from the task's environment, and traces each failure to the primitives it implicates, so that revision redoes their work and keeps the rest. Third, validation and publishing improves the library between tasks, admitting a new or improved primitive once evidence such as tests, a compiler, or development runs confirms it. Results proved in our companion studies show that need-driven activation stays bounded by the task's demand, however large the library grows, and identify when per-task composition beats the best fixed harness.

However, each task domain places its own demand on the last mile: cooperating agents must combine their work, a repository change must reach every component it affects, a search must follow what remains unresolved, and a harness must fit its task. We therefore instantiate \ours{} with eight primitive families, each built as the solution to one kind of real-world task: Tool Primitives for LLM tool use~\citep{jin2026toolprimitives}, Code Primitives for repository construction~\citep{jin2026codeprimitives}, Math Primitives for formal theorem proving~\citep{anonymous2026library}, Knowledge Primitives for knowledge-intensive tasks~\citep{jin2026sourcelearn}, Agent Primitives for multi-agent systems~\citep{jin2026agentprimitives}, Dev-Primitives for software engineering~\citep{jin2026devprimitives}, Navigation Primitives for large information spaces~\citep{jin2026antman}, and Harness Primitives for agent harnesses~\citep{jin2026stitch}. Each family realizes the definition above and the mechanisms its task needs, so the eight companion studies become instances of one framework. For these families, we build 25,519 Tool Primitives, 16,464 of them wrapping live APIs; 1,424 Code Primitives, each validated against the tests it carries; and 82 Math Primitives whose Lean artifacts compile. We also introduce a benchmark of 522 repository-reconstruction tasks drawn from real Python projects in seven domains.

Extensive experiments on more than twenty benchmarks demonstrate that, on the same backbones, Dev-Primitives outperform Codex by 18.2 points on Terminal-Bench~4.0, and Harness Primitives composed per task exceed the mini-SWE-agent seed harness by 7.5 and 12.2 points and Codex by 1.5 and 15.5 points on 100 SWE-bench Verified and 45 Terminal-Bench~2.0 tasks, for a composition cost of 2.7\% to 4.3\% of one actor attempt. Code Primitives surpass Claude Code on its backbone by 56.2\% in repository delivery and keep 95.1\% of their delivery at 24.0\% lower cost with GPT-OSS-20B adaptation and diagnosis. Tool Primitives on Qwen3-8B lead every reference model on $\tau^2$-Bench at up to $7.7\times$ lower cost than GPT-5.4 and complete $2.6$--$3.8\times$ as many of 50 curated real-world tasks as GPT-5.4, Claude-4.6-Sonnet, and Gemini-3.1-Pro calling the same tools. Agent Primitives on Qwen3-8B exceed the strongest prior multi-agent system by 7.5 points with ${\sim}3$--$4\times$ fewer tokens than a text-based one, Navigation Primitives lead the strongest GAIA-Text-103 baseline by 12.7 points, and with GPT-5.6 Luna, Knowledge Primitives gain 14.3 points over Hybrid RAG on average and Math Primitives lead the strongest reranker by 10.8 points in Putnam hit@5. Further analysis across families reveals that the gain lies inside the primitive and that the task decides which primitives run: Tool Primitives more than double the ToolBench pass rate of budget-matched self-refinement over raw schemas, and active Navigation Primitives grow $1.23\times$ while distractors grow the information space $16\times$ around the same evidence.

Beyond these research benchmarks, we build heelo.dev, a platform through which agents such as Claude Code, Codex, or Cursor invoke real-world actions as agent-native primitives, from publishing pages and sending outreach to reading invoices and scraping sources. Its 46 real-world tasks cover the 26 capabilities of four end-to-end loops, with the pass criteria of each task fixed before its first run. On their own, Claude Opus 5.5 with Claude Code and GPT-6.1 Sol with Codex, both at medium reasoning effort, complete 9 of the tasks each, with mean task scores of 28\% and 25\%. Connected to \ours{} through our MCP server, Claude Code completes 39 and Codex 35, with scores of 90\% and 83\%. Per completed task, Claude Code with \ours{} needs 3.4 minutes and \$0.80 instead of 6.3 minutes and \$1.24 on its own, and Codex with \ours{} 3.5 minutes and \$0.42 instead of 8.0 minutes and \$0.65. \ours{} thus provides a unified framework for handling real-world tasks end to end. Our main contributions are summarized as follows:
\begin{itemize}
\item We introduce \textbf{agent-native primitives} under one definition that pairs a reusable resource, from tools, code components, and proof methods to source models, multi-agent patterns, and repository files, with a resident LLM that applies it in context and messages other primitives. We further build libraries of 25,519 Tool Primitives, 1,424 validated Code Primitives, and 82 Math Primitives with Lean-checked scaffolds.
\item We propose \textbf{\ours{}}, one harness engineering framework that unifies the eight families: each task's harness is a graph composed from an evolving library and revised where the task's evidence implicates it, and the library grows between tasks with validated primitives, so that a new kind of task can join as a new primitive family.
\item Experiments on more than twenty benchmarks across eight kinds of real-world tasks show that every family improves on its reference baseline: primitives lead Codex by 18.2 points on Terminal-Bench~4.0 and Claude Code by 56.2\% in repository delivery, and cost up to $7.7\times$ less than GPT-5.4 on $\tau^2$-Bench. Cross-family analyses locate the gain inside the primitive and show that the task decides which primitives run.
\item We build \textbf{heelo.dev}, a platform that exposes real-world actions, from flights and calendars to email and browser automation, as agent-native primitives to agents such as Claude Code, Codex, or Cursor. On its 46 real-world tasks in four end-to-end loops, Claude Code and Codex complete 9 tasks each on their own and 39 and 35 when connected to \ours{} through MCP, with $1.9\times$ and $2.3\times$ less time and $1.55\times$ lower cost per completed task.\fillpar
\end{itemize}

\section{Related Work}
\label{sec:related}
\runningsection{Related work}

\textbf{Tool Learning and Interfaces.}
LLM agents learn to call tools from self-annotated data~\citep{schick2023toolformer}, large collections of real-world APIs~\citep{patil2023gorilla,qin2023toolllm}, and synthesized single- and multi-turn function-calling data~\citep{liu2024apigen,liu2024toolace,yin2025magnetmultiturntoolusedata}, while compact models make function calling robust on device~\citep{lin2024hammer}. Benchmarks span structured calls and tool-augmented dialogue~\citep{patil2025bfcl,li2023api}, large-scale, compositional, and nested call sequences~\citep{guo2024stabletoolbench,basu2024api,basu2025nestful}, and stateful, conversational environments~\citep{yao2024tau,barres2025tau,lu2025toolsandbox,chen2025acebench}, while AgentDojo evaluates tool-using agents under adversarial prompt injection~\citep{debenedetti2024agentsdojo}. OpenAPI, the Model Context Protocol, Agent2Agent, and Agent Skills make services, tools, agents, and packaged instructions discoverable~\citep{openapi2024spec,mcp2025spec,a2a2026spec,agentskills2025spec}. Across these settings, the caller binds each call itself: it writes schema-compliant arguments and grounds each value in the intended object, often an earlier call's output. Tool Primitives move schema binding into the tool, so that the caller states its need in natural language and a resident LLM conditioned on the tool's schema produces the typed call.

\textbf{Code Reuse and Repositories.}
Software reuse made interfaces, dependencies, and adaptation cost central to building systems from components~\citep{krueger1992software,szyperski2002component}. Learned systems discover libraries and abstractions~\citep{ellis2021dreamcoder,grand2024lilo,stengel2024regal}, make tools for later tasks~\citep{wang2024trove,cai2024large,yuan2024craft}, retrieve code and documentation during generation~\citep{lu2022reacc,zhou2022docprompting,zhang2023repocoder}, and distill reusable procedures from agent experience~\citep{qiu2025agentdistill,li2025codeskill}. Repository-level agents inspect, edit, and execute code~\citep{yang2024swe,wang2025openhands,wang2024executable}, follow explicit localization and repair stages~\citep{xia2024agentless,zhang2024autocoderover}, or divide work among task graphs, roles, and coordinated developers~\citep{chen2024coder,tao2024magis,wang2026codeteam}. Evaluation ranges from cross-file and repository-level generation~\citep{ding2023crosscodeeval,li2024deveval} and issue repair~\citep{jimenez2023swebench} to long-horizon evolution~\citep{le2025swe}, repository generation and whole-repository migration~\citep{ding2025nl2repo,hong2026swe}, and terminal and DevOps workflows~\citep{merrill2026terminal,tang2026devops}. In these systems, the reused artifact stays fixed while a central agent decides how to use it, and the units of reasoning and their connections are designed in advance or kept inside one trajectory~\citep{liu2026context,wang2026improving}. Code Primitives make adaptation an operation of the reused component, and Dev-Primitives route requirements along the repository's own dependencies.

\textbf{Retrieval, Memory, and Knowledge.}
Retrieval-augmented generation brings external content into model computation~\citep{guu2020retrieval,karpukhin-etal-2020-dense,lewis2020rag}, adaptive methods decide when and what to retrieve~\citep{trivedi2023interleaving,jiang2023active,asai2024self,jeong2024adaptive,yan2024corrective}, and long-context models use evidence unevenly across positions~\citep{liu-etal-2024-lost}. Reading-comprehension benchmarks ask a model to combine what it reads, through discrete reasoning over paragraphs in DROP~\citep{dua2019drop} and multi-hop reasoning in HotpotQA~\citep{yang-etal-2018-hotpotqa}. Beyond flat passages, structured representations organize a source into summary trees~\citep{sarthi2024raptor}, graphs~\citep{edge2024local,guo2025lightrag,li2024graphreader}, associative or global memories~\citep{gutierrez2024hipporag,gutierrez2025rag,qian2025memorag}, and task-conditioned structures~\citep{li2025structrag}. Agent memories keep interactions~\citep{park2023generative,zhong2024memorybank,packer2023memgpt,anokhin2024arigraph,xu2026mem}, feedback~\citep{shinn2023reflexion}, insights~\citep{zhao2024expel}, skills~\citep{wang2023voyager}, and workflows or procedures~\citep{wang2024awm,cao2026remember} across tasks. In large information spaces, retrieval chooses its next query from the remaining gaps~\citep{zhu-etal-2025-mitigating,su-etal-2024-dragin,li-etal-2026-s2g}, and multi-agent readers divide long inputs~\citep{zhao-etal-2024-longagent,zhang2024chainagentslargelanguage} or retrieval, filtering, and planning~\citep{pmlr-v267-chen25an,chang-etal-2025-main,hu2025owl} among workers. What persists in these methods is the corpus, a structure built from it, or experience from interaction, and when readers divide long inputs among workers, coordination grows with how the space is partitioned, regardless of what the query requires. By contrast, Knowledge Primitives keep a revisable model of one authoritative source and commit only the revisions that the source supports, and Navigation Primitives let unresolved information needs decide which primitives run.

\textbf{Formal Theorem Proving.}
Lean and Mathlib give provers a checked proof language and a large reusable library~\citep{demoura2021lean,mathlib2020}. LLM provers generate whole proofs or repair them from verifier feedback~\citep{xin2024deepseek,first2023baldur,liu2025proofaug}, and agentic loops alternate generation with Lean-guided repair~\citep{ospanov2026apollo,requena2026minimalagentautomatedtheorem}. Retrieval-augmented proving, premise selection, and declaration search supply formal facts~\citep{yang2023leandojo,zhu2026premise,gao2026leansearch}, and lemma libraries grow across problems~\citep{wang2024lego}. Such retrieval ranks a fact, a lemma, or a prior proof by relevance, and LLM rerankers that reason about fit report a score or an ordering~\citep{sun2023chatgpt,qin2024large,weller2025rank1,zhuang2026rank}, leaving the prover to decide how to use a candidate. Proof planning describes methods by their conditions and effects~\citep{bundy1988use,melis1999knowledge}, and adaptation-guided retrieval prefers cases that adapt to a new problem~\citep{leake1997case,smyth1998adaptation}. Math Primitives bring this view to LLM provers: a Method Contract carries a method's required conditions, intended action, and residual obligations, and each contract proposes its use on the goal before ranking.

\textbf{Multi-Agent Systems and Communication.}
Multi-agent frameworks organize LLMs by assigning predefined roles~\citep{hong2023metagpt,qian2024chatdev}, programming conversations among customizable agents~\citep{wu2023autogen}, recruiting collaborators iteratively~\citep{chen2023agentverse}, or letting agents debate~\citep{du2023improving,liang2024encouraging}. Other methods design the team automatically: DyLAN selects agents by estimated value~\citep{liu2024a}, GPTSwarm optimizes collaboration graphs~\citep{zhuge2024gptswarm}, and MAS-GPT generates a configuration from the query~\citep{ye2025mas}. Recent methods communicate below text by sharing a learned latent space~\citep{zheng2025thought}, projecting or propagating key--value caches~\citep{fu2025cache,zou2025latent}, or aggregating experts through cross-attention~\citep{feinashley2025mixturethoughtslearningaggregate}. For cache messages, rotary embeddings make a key's position part of its meaning~\citep{su2024roformer}. These systems take the agent or the team as the unit of design, and the latent methods mainly improve how information passes between agents. Agent Primitives make a recurring multi-agent computation pattern the reusable unit and implement its computation structure through latent KV cache communication, with all agents on one LLM so that their inputs and outputs stay aligned.\fillpar[0.6][1]

\textbf{Agent Harnesses and Workflows.}
The harness around a model decides how observations, memory, retries, and checks shape its actions: ReAct interleaves reasoning with actions~\citep{yao2023react}, Reflexion and Self-Refine feed verbal feedback into later attempts~\citep{shinn2023reflexion,madaan2023self}, and DSPy compiles declarative pipelines of model calls~\citep{khattab2024dspy}. Codex and Claude Code, the platform harnesses built for GPT and Claude, supply the tools and execution loop around these frontier models~\citep{openai2025codex,anthropic2025claudecode}. Design methods search over agent programs and workflows~\citep{hu2024adas,zhang2024aflow} or edit and evaluate candidate harnesses to select one for all tasks~\citep{zhang2026selfharnessharnessesimprove,lee2026metaharnessendtoendoptimizationmodel}. Others adapt the harness at test time, specializing it for each case from retrieved execution experience~\citep{huang2026memoharnessagentharnesseslearn} or rewriting harness code and generating task-specific modules~\citep{nie2026tthetesttimeharnessevolution,zhang2026jitagentscalingharnessintelligence}. These methods fix one harness across tasks, specialize it through instructions that the actor must follow, or write new control code at test time, which must then be implemented, debugged, and validated. Harness Primitives validate each control operation when it enters the library, so that a task's harness is selected and compiled from implemented parts before the actor runs.

\textbf{Key Differences.}
Across these lines of work, the units (a function, a code component, a lemma, a passage or memory, a role or team, a harness) keep a resource but leave the competence to apply it with the caller or fix it in an agent design. An agent-native primitive keeps that competence with the resource by equipping it with a resident LLM: the primitive owns the resource, exposes an interface through which a harness finds and invokes it, applies the resource to a natural-language request in context, and communicates with other primitives; where a domain offers an exact procedure, a program replaces the resident LLM. \ours{} selects and composes primitives per task, revises only what the evidence of a failure implicates, and publishes improved primitives through a validation path separate from task execution. Eight primitive families instantiate this design on eight kinds of real-world tasks. Overall, primitives let the harness of a real-world task be engineered around reusable, validated units.

\section{Methodology}
\label{sec:method}\label{sec:model}\label{sec:architecture}
\runningsection{Methodology}

\begin{figure}[!t]
\centering
\includegraphics[width=\linewidth]{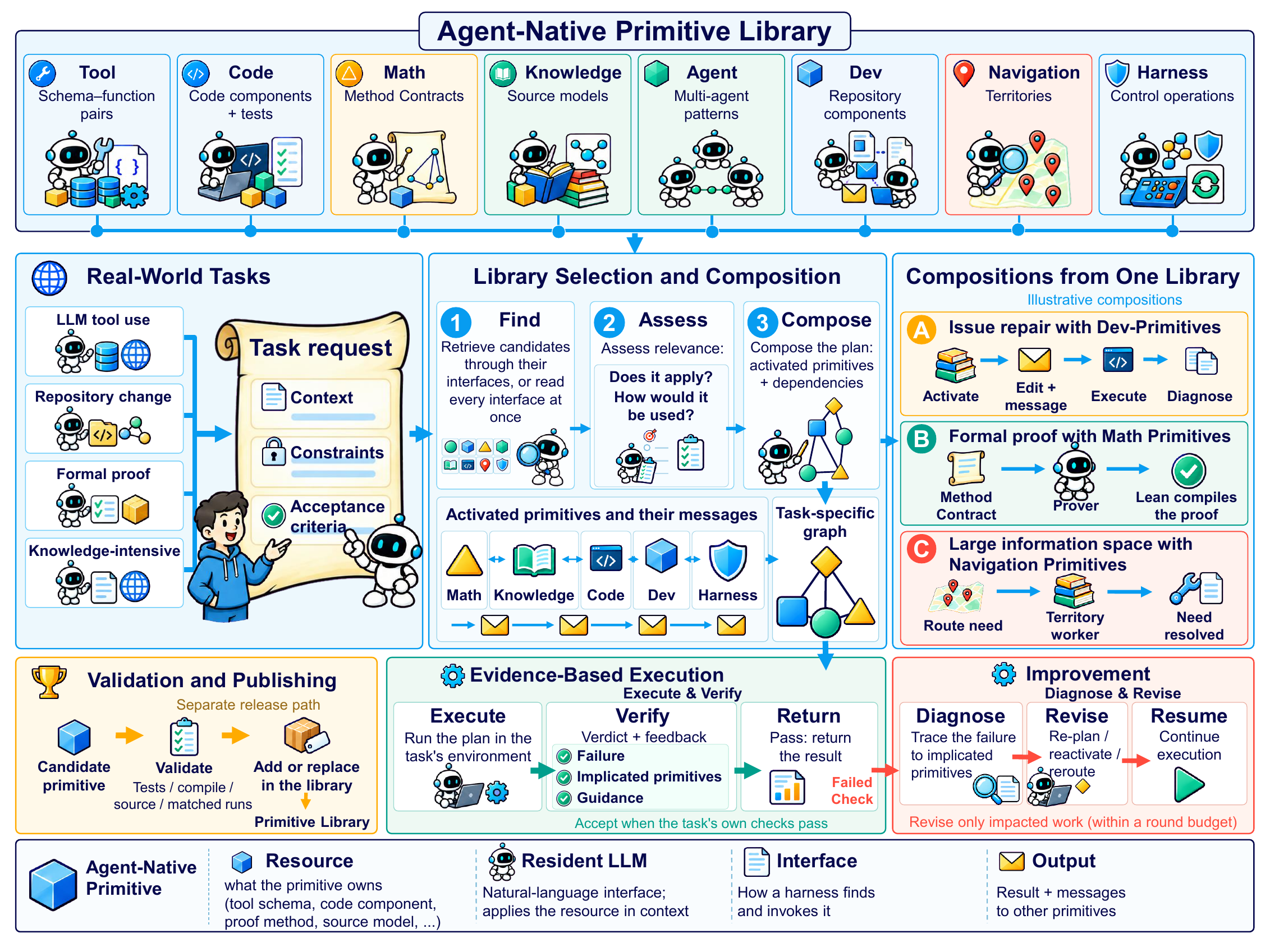}
\caption{\textbf{Overview of \ours{}.} \emph{Top:} the agent-native primitive library, one box per family with the resource its primitives own. \emph{Middle:} a real-world task enters as a task request with its context, constraints, and acceptance criteria; library selection and composition finds candidates through their interfaces, lets each assess its own relevance, and composes a task-specific graph whose activated primitives exchange messages, and three illustrative compositions are drawn on the right. \emph{Bottom:} evidence-based execution runs and verifies the plan in the task's environment, improvement traces a failure to the implicated primitives and revises only that work within a round budget, and validation and publishing adds a validated primitive to the library on a separate path. The bottom strip shows what an agent-native primitive holds: a resource, a resident LLM that serves as its natural-language interface, an interface through which a harness finds and invokes it, and an output of results and messages to other primitives.}
\label{fig:overview}
\end{figure}

Figure~\ref{fig:overview} gives an overview of \ours{}. We first state the problem of building a harness for a real-world task and define the agent-native primitive from which the harness is built. We then describe the three mechanisms that build and run each task's harness and improve the library between tasks, the eight primitive families that instantiate the method on eight kinds of real-world tasks, and three properties of the selection that decides which primitives run, each proved in a companion study.

\subsection{Problem Formulation}
\label{sec:formulation}

A real-world task $q$ arrives with a context $c_q$: its inputs, its environment, its constraints, and the acceptance criteria by which it is judged (the task request of Figure~\ref{fig:overview}). An agent acts on the task through a \emph{harness} $H$, the control program between the model and the environment. The harness decides which operations run, with which inputs, in which order, and when the run stops. Running $H$ on $q$ produces a trajectory $\tau$. The task's own acceptance procedure, such as its tests, its proof checker, or its answer key, scores the trajectory as $R_q(\tau)\in[0,1]$, and we write $R(q,H)=\mathbb{E}_{\tau}[R_q(\tau)\mid H]$.

A fixed harness, such as one general-purpose coding agent, uses the same $H$ for every task. Following our companion study on Harness Primitives~\citep{jin2026stitch}, we build the harness of each task from a library $\Lib$ of reusable units:
\begin{filldisp}\begin{equation}
H_q=F(q,\,c_q;\,\Lib)\in\mathcal{H}(c_q),\qquad
H_q^{*}\in\arg\max_{H\in\mathcal{H}(c_q)}\ R(q,H),
\label{eq:objective}
\end{equation}\end{filldisp}
where $\mathcal{H}(c_q)$ is the set of harnesses compatible with $c_q$. The construction procedure $F$ approximates $H_q^{*}$ from the units of $\Lib$, within a budget on model calls or revision rounds. While $F$ builds and runs $H_q$, it keeps a context $\mathcal{C}_t$ for round $t$, which starts as $\mathcal{C}_0=c_q$ and accumulates results and feedback.

The last mile lies inside Eq.~\eqref{eq:objective} at two levels. Each unit of $\Lib$ must be applied to the task's own objects, and $F$ must decide which units the task needs and how their results combine. \ours{} answers the first level with agent-native primitives and the second with the mechanisms that realize $F$.

\subsection{Agent-Native Primitives}
\label{sec:primitive-definition}\label{sec:agent-native}

A unit of the library helps a task only once it is applied in context. A tool call must ground its arguments, a retrieved component must be adapted to its target, a proof method must be instantiated on the goal, and an answer must stay faithful to its source. A caller that applies every unit itself rebuilds this competence for each task and must hold the details of every unit it uses in one context~\citep{jin2026toolprimitives,jin2026devprimitives}. An agent-native primitive keeps this competence with the unit by equipping the unit with a resident LLM that applies it.

\begin{definition}[Agent-native primitive]
\label{def:primitive}\label{def:agent-native}
An agent-native primitive $P_i$ equips a reusable resource $a_i$ with a resident LLM $\mathcal{M}$ and exposes an interface $d_i$ through which a harness finds and invokes it. Invoked with a natural-language request $x_i$ in a context $c_i$, the resident LLM, conditioned on the resource, applies $a_i$ and returns a result $r_i$ together with the messages $m_i=\{m_{i\to j}\}_j$ that the primitive addresses to other primitives $P_j$: the requirements that $r_i$ places on them, or task-relevant information they need, with $m_i=\varnothing$ when none is needed:
\begin{equation}
(r_i,\,m_i)=P_i(x_i;\,c_i)=\mathcal{M}\bigl([\,a_i;\,c_i;\,x_i\,]\bigr),
\label{eq:primitive}\csname ltx@label\endcsname{eq:agent-native}
\end{equation}
where $[\cdot]$ denotes the input context of the resident LLM.
\end{definition}

The resident LLM is what makes the primitive agent-native. It reads the resource and serves as its natural-language interface, sparing the caller from holding the resource's details in its own context. The primitive's messages carry requirements and task-relevant information to other primitives. Resources that were passive artifacts thus become active participants that reason locally and communicate with one another. The resource $a_i$ can be anything the task relies on, from a tool's schema and function to a repository component, a proof method, or a source model, and the interface $d_i$ is what a harness reads in its place, such as a tool's schema descriptor or the selection side of a Method Contract. One base model can serve as the resident LLM of every primitive of a library, conditioned on a different resource each time, so a primitive is defined by the resource it owns. Where a domain offers an exact procedure, as for the control operations of Harness Primitives, implemented code applies the resource and receives its request as parameters. A resource that a downstream prover or solver consumes, such as the execution side of a Method Contract or a source model, is applied by that model.

\textbf{Relevance Assessment.}
Similarity between a request and an interface does not show that a primitive applies. A primitive can therefore assess its own relevance before it is selected~\citep{jin2026codeprimitives,anonymous2026library}:
\begin{equation}
(s_i,\,u_i)=\mathcal{M}^{\mathrm{assess}}(P_i,\,x,\,\mathcal{C}_t),
\label{eq:assess}
\end{equation}
where $s_i$ judges whether $P_i$ applies to a request $x$ derived from the task, and $u_i$ proposes how $P_i$ would be used. Run over the whole library at once, this assessment removes the dependence on a pool that similarity retrieval chooses first (Proposition~\ref{prop:coverage}).

\textbf{Carried Validation.}
A resource may carry its own means of validation, such as tests or a schema. The primitive applies them before it returns and reports a structured failure when they fail. Carried validation shows that the primitive's own step succeeded. Whether the result serves the task is for the harness to decide.

\textbf{Library.}
Primitives form a library $\Lib=\{P_i\}_{i=1}^{N}$. A library is either built ahead of the tasks that draw on it and reused across them, or induced from a task's own data, such as a repository or an information space~\citep{jin2026devprimitives,jin2026antman}. A library built ahead of its tasks does not change while a task runs. An induced library follows the task's data and can change as the task edits that data.

\subsection{Harness Engineering around Agent-Native Primitives}
\label{sec:harness-engineering}

Primitives determine how each resource is applied. They do not determine which primitives a task needs, how their results reach one another, or what to change when the task still fails. Two mechanisms realize the construction procedure $F$ of Eq.~\eqref{eq:objective}: library selection and composition builds the harness, and evidence-based execution and improvement runs and revises it. A third mechanism, validation and publishing, improves the library between tasks (Figure~\ref{fig:overview}).\fillpar[.5]

\textbf{Library Selection and Composition.}\phantomsection\label{sec:composition-state}
A primitive sees only the request it is handed, so it cannot decide alone whether the task needs it. A library also usually holds more primitives than one task uses. COMPOSE therefore finds candidates through their interfaces, either by retrieval or by reading every interface at once. It reads their relevance assessments where primitives provide them and builds a plan
\begin{equation}
\Pi_t=\operatorname{COMPOSE}(q,\,\mathcal{C}_t;\,\Lib)=\bigl(\{(i,\,x_i)\}_{i\in\mathcal{A}_t},\ \mathcal{E}_t\bigr),
\label{eq:compose}
\end{equation}
where $\mathcal{A}_t\subseteq\{1,\dots,N\}$ is the activation set, $x_i$ is the local request of $P_i$, and $\mathcal{E}_t\subseteq\mathcal{A}_t\times\mathcal{A}_t$ holds the dependencies along which results and messages flow. The plan is the task-specific graph of Figure~\ref{fig:overview}. Only the activated primitives run, so the cost of applying primitives follows $|\mathcal{A}_t|$, however large $N$ is. A dependency carries an earlier result to the request that consumes it, a message to the primitive that must act on it, or internal state between models that share parameters. A plan may also hand a selected resource to the prover or solver that consumes it. A need that no primitive covers is left to the harness itself.

\textbf{Evidence-Based Execution and Improvement.}\phantomsection\label{sec:loop}
A primitive can succeed at its own step while the task still fails, and the task's environment reports such a failure without naming the primitive that should change. \ours{} therefore runs the plan in rounds that end in evidence and revises what the evidence implicates. In round $t$, the activated primitives act, and verification judges the integrated result:
\begin{equation}
\begin{aligned}
(r_i,\,m_i)&=P_i\bigl(x_i;\,c_i^{t}\bigr),\qquad i\in\mathcal{A}_t,\\
o_t&=\operatorname{EXECUTE}\bigl(S_t\bigr),\qquad
(v_t,\,\phi_t)=\operatorname{VERIFY}(q,\,\Pi_t,\,S_t,\,o_t),\qquad
\phi_t=(e_t,\,\mathcal{I}_t,\,u_t).
\end{aligned}
\label{eq:act}\csname ltx@label\endcsname{eq:verify}
\end{equation}
Here $c_i^{t}$ holds what the plan routes to $P_i$: the results of its predecessors in $\mathcal{E}_t$, the messages $m_{\to i}=\{m_{j\to i}\}_{j\in\mathcal{A}_t}$ addressed to it, and the part of the task state it needs. Each primitive changes only what it owns. Messages reach their recipients before the round ends, so a requirement found while one primitive works can still be met in the same round by the primitives that depend on it. The task state $S_t$ integrates the round's results into the task's artifacts, such as a repository, a proof, or an answer. EXECUTE runs what the task can run and records evidence $o_t$, such as build status, test outcomes, compiler messages, and traces. Where nothing runs, the results themselves are the evidence.\fillpar[0.6][1]

VERIFY turns the integrated state and its evidence into a verdict $v_t\in\{\pass,\fail\}$ and structured feedback $\phi_t$. The feedback names the observed failures $e_t$, the primitives $\mathcal{I}_t\subseteq\{1,\dots,N\}$ that they implicate, and revision guidance $u_t$. A failed build or test sets $v_t=\fail$, and no model judgment overrides it. Verification can also judge each step as soon as it completes, or run inside the plan as a primitive of its own. On a failure, the feedback enriches the context, and COMPOSE revises the plan:
\begin{equation}
\mathcal{C}_{t+1}=\mathcal{C}_t\cup\{\phi_t\},\qquad
\Pi_{t+1}=\operatorname{COMPOSE}\bigl(q,\,\mathcal{C}_{t+1};\,\Lib,\,\Pi_t,\,\phi_t\bigr).
\label{eq:revise}
\end{equation}
Revision changes only what $\phi_t$ implicates. It re-plans the affected steps, reactivates the implicated primitives or activates those the evidence newly implicates, or redirects an unresolved need. It keeps the work that the new evidence does not contradict. Revision can also precede execution: a plan that can be checked before it runs is corrected from its own diagnostics within a bounded number of attempts, and a fallback plan is kept if correction fails. The loop ends when $v_t=\pass$ or after $B$ revision rounds and returns the current result.

\textbf{Validation and Publishing.}\phantomsection\label{sec:building}
A library built ahead of its tasks is reused across them, so what enters it must be validated first, independently of the tasks it will serve. New and improved primitives may be mined from existing artifacts, harvested from runs whose results were verified in context, or proposed from failed trajectories. They enter the library only on the separate path of Figure~\ref{fig:overview}:
\begin{equation}
\Lib_{k+1}=\Lib_k\oplus\bigl\{\tilde P\in\tilde{\mathcal{P}}_k:\ \operatorname{VALIDATE}(\tilde P)=\pass\bigr\},
\label{eq:publish}
\end{equation}
where $\tilde{\mathcal{P}}_k$ holds the candidates of stage $k$, and $\oplus$ adds a new primitive or replaces the one that $\tilde P$ revises. VALIDATE checks each candidate with the evidence that its domain affords, such as its tests, a compiler, its source, or matched runs on development tasks. An output produced inside a task, such as an adapted component, enters the library only through this path. Together, Eqs.~\eqref{eq:compose}--\eqref{eq:publish} define the per-task harness and the evolving library that make up \ours{}.

\subsection{Eight Primitive Families}
\label{sec:variants}

The method above is shared across tasks, but each task domain places its own demand on the last mile: cooperating agents must exchange their intermediate state, a repository change must reach the components that depend on it, a search must follow the needs that remain unresolved, and a harness must fit its task. We therefore instantiate \ours{} with eight primitive families, each built as the solution to one kind of real-world task, and the primitives of a family own the resource that its task applies, from the tools of LLM tool use and the components of repository construction to the proof methods of formal theorem proving and the control operations of agent harnesses. The families differ in how they specialize the method, as summarized below in each family's own terms and notation, so that a symbol such as $F$, $M$, or $R$ keeps the meaning of its own study within its paragraph. We provide the full framework of each family, with its overview figure, in Appendix~\ref{app:families}.\fillpar

\textbf{Tool Primitives.}\phantomsection\label{sec:var-tool}
A Tool Primitive~\citep{jin2026toolprimitives} wraps one ToolFace tool: its resource $a_i$ is the schema--function pair $(s_i,f_i)$, and its interface $d_i$ is the schema descriptor. The shared model conditioned on the schema, $\mathcal{P}_i(x;c)=\mathcal{M}([s_i;c;x])$, maps the request $x$ to the arguments of $s_i$ and invokes $f_i$, returning the result with a summary or a structured failure. HEART is the harness engineering framework around Tool Primitives, and its COMPOSE has three parts: request decomposition asks the user for missing details and writes an invocation plan, semantic search over schema descriptors retrieves candidate tools for each step, and value resolution grounds each step's values in the query and earlier results; each step's result is the context $c$ of the next. VERIFY is per-step result validation on four criteria, and targeted revision re-plans from its feedback within $B=3$ rounds. ToolFace holds 25,519 tools with manually authored schemas, each inducing a Tool Primitive.

\textbf{Code Primitives.}\phantomsection\label{sec:var-code}
A Code Primitive~\citep{jin2026codeprimitives} owns an executable component $a_i=(C_i,I_i,D_i,V_i,X_i)$ of implementation, contract, dependency closure, carried tests, and provenance; the contract $I_i$ is its interface $d_i$, and a resident LLM $M_i$ assesses and adapts the component. LEGO constructs repositories from these primitives and task-specific code, with Retrieval and Activation over the CodeFace library as its COMPOSE: DECOMPOSE writes capability requirements, the top two candidates for each requirement assess their relevance (Eq.~\eqref{eq:assess}), and MATCH activates the smallest mutually compatible set. In Primitive Adaptation and Collaboration, each activated primitive adapts its own component into a target-specific variant $\widetilde{P}_i$, reruns $V_i$, and sends its consumers the requirements $\mu_{i\to j}$ the rewrite imposes. Execution Feedback extends the target's tests with synthesized ones, DIAGNOSE realizes VERIFY, and revision reactivates the implicated primitives within five rounds per module. CodeFace, mined from repositories and validated in isolation, grows with implementations harvested from verified construction runs.

\textbf{Math Primitives.}\phantomsection\label{sec:var-math}
A Math Primitive~\citep{anonymous2026library} packages a reusable proof method as a Method Contract $a_i$ with parameters $x_c$, required conditions $\Pre_c$, a target $G_c$ with its action, a Lean scaffold $S_c$, and residual obligations $\Obl_c$. The contract's selection side is the interface $d_i$, and its execution side goes to the prover. COMPOSE is self-advertised method selection: in one batched call, each contract in the library proposes its use on the goal as a target, an action, and required facts (Eq.~\eqref{eq:assess}). A commitment gate demotes a contract that claims the main role with a vague or unsupported proposal to a supporting role, a tiered ranking then orders the contracts, and an optional runoff re-ranks the top ten. The prover applies the top-ranked contract; Lean compiles each attempt, and its errors drive whole-proof repair within ten compilations. The 82 contracts are built once from Putnam 2000--2014 solutions and admitted after their Lean artifacts compile (Eq.~\eqref{eq:publish}).

\textbf{Knowledge Primitives.}\phantomsection\label{sec:var-knowledge}
Each authoritative source $D$ has one Knowledge Primitive~\citep{jin2026sourcelearn}, whose resource $a_i$ is the \emph{source model} $M$ of $D$, organized around its entities with units citing $D$, and a task grounded in $D$ invokes it. SourceLearn constructs and progressively refines these source models, and at task time it realizes COMPOSE as \emph{source-model activation} $A_B(M,q)$, the whole model if it fits $B=24$k tokens and its task-relevant regions otherwise; the solver $F$ applies the activated model beside evidence retrieved from $D$, $\hat{y}=F\bigl(q,R(D,q),A_B(M,q)\bigr)$. The source model stays fixed during testing and improves on the publishing path of Eq.~\eqref{eq:publish}: Self-Directed Source Learning deepens unresolved aspects of the initial model and connects knowledge fragmented across entities, and Task-Guided Source Learning revises the regions implicated by incorrectly answered guidance tasks and recalibrates $M$ under a learned representation policy. Grounded source reconstruction realizes VALIDATE, rejecting a rewrite unless the source supports it and previously supported meaning is preserved.

\textbf{Agent Primitives.}\phantomsection\label{sec:var-agent}
An Agent Primitive owns a recurring multi-agent computation pattern as its resource $a_i$ and encapsulates it as a reusable latent operator behind the interface of a standard LLM agent~\citep{jin2026agentprimitives}; the three primitives are Review, Voting and Selection, and Planning and Execution. An LLM Organizer performs COMPOSE: guided by a Knowledge Pool of 45 query--MAS-structure pairs, it maps each query to a primitive composition plan in code, and executing the plan forms a primitive-based MAS. Internal agents follow task-agnostic role prompts on one shared LLM and exchange KV caches within and across primitives, re-indexing the sender's keys for RoPE. Verification and revision happen inside the primitives, where a Critic's feedback drives the Solver's refinement in Review and a Selector judges candidates in Voting and Selection. The primitives come from decomposing existing MAS designs into recurring computation patterns, and the library stays fixed.

\textbf{Dev-Primitives.}\phantomsection\label{sec:var-dev}
A Dev-Primitive~\citep{jin2026devprimitives} wraps a repository component $a_i$, by default a file, with a resident LLM, and activation reads the repository in place of an interface $d_i$. The primitive edits its own artifact where needed and sends directed messages to the components its reasoning implicates, $(a_i',m_i)=\mathcal{M}([a_i;x_i;\mathcal{C}_i])$, where $\mathcal{C}_i$ holds the messages it receives. HERMES, the harness engineering framework for software engineering via Dev-Primitives, instantiates them on demand: its COMPOSE, Dynamic Primitive Activation, localizes the relevant components, expands along their dependencies, and assigns each a local objective $x_i$. The activated primitives collaborate in one round, and task-visible checks run on the modified repository. VERIFY is Bug Diagnosis, which traces a failure to the components it implicates, and activation re-plans from this feedback within $B=3$ revision rounds. The repository induces the library, and a persistent file created during the task becomes a new Dev-Primitive.

\textbf{Navigation Primitives.}\phantomsection\label{sec:var-navigation}
A Navigation Primitive~\citep{jin2026antman} is a scope-specialized worker whose resource $a_i$ is one territory of the information space and whose interface $d_i$ is the territory's WorkerCard. The library is induced per query: a deterministic partition splits the space into territories, bounded searchable regions fixed during execution. ANTMAN, an adaptive coordination framework for these workers, keeps the context $\mathcal{C}_t$ as a revisable Need Graph $G_t$ holding each need's status, evidence, attempts, and progress, and composes one need at a time: WorkerCard retrieval and an orchestrator decision over $G_t$ route an unresolved need to at most one worker, which searches its territory and reports evidence, progress, and uncertainty. Need-resolution assessment in the update $G_{t+1}=U(G_t,r_t)$ realizes VERIFY, judging the need resolved, partial, or unresolved, and task-local recovery reframes a stalled need, reroutes it, or sends it to a fallback path, within $\rho$ routings per need.

\textbf{Harness Primitives.}\phantomsection\label{sec:var-harness}
In a Harness Primitive~\citep{jin2026stitch}, the resource $a_i$ is an implemented control operation, fixed at test time, and the interface $d_i$ combines an application scope, stating when it helps, with a composition contract of inputs, outputs, setup, and dependencies. The framework that composes these primitives is STITCH, and its COMPOSE acts before the harness sends its first request to the actor, the task-solving model: the harness composer $\pi$ reads the whole catalog and either keeps the seed graph or returns a composition intent, whose primitives the deterministic compiler $K$ wires into $G_{\mathrm{seed}}$ with their contract dependencies, $\Pi_0=K\bigl(G_{\mathrm{seed}},\pi(q,c_q;G_{\mathrm{seed}},\Lib);\Lib,c_q\bigr)$. An invalid candidate returns its diagnostics to the composer for bounded correction, and the seed harness is kept if correction fails. During the run, verification and repair primitives return failed checks to the actor. A development agent mines the library from failed trajectories, accepting candidates that pass mechanism tests and improve matched runs (Eq.~\eqref{eq:publish}).\fillpar

\subsection{Theoretical Analysis}
\label{sec:analysis}

COMPOSE decides which primitives run, so it governs both what a harness costs and what it gains. The companion studies prove three properties of this choice, which we restate in the notation of this section: how many primitives a task activates, what choosing them per task gains over a fixed harness, and what limits a shortlist. We give the proofs, with the correspondence to each study's notation, in Appendix~\ref{app:proofs}.

\textbf{Demand-Bounded Activation.}\phantomsection\label{sec:analysis-activation}
Suppose that COMPOSE activates a primitive only by routing one unresolved need of the task to it, and routes each need at most $\rho$ times, reroutes included. Let $D(q)$ count the needs realized while $q$ runs, with a reframed or newly introduced need counted as new. Let $K(q)$ count the routing events, $\mathcal{A}_q=\bigcup_t\mathcal{A}_t$ be the set of primitives activated during the run, and $\Lib_q$ be the library available to $q$.

\begin{proposition}[{Demand-bounded activation~\citep{jin2026antman}}]
\label{prop:activation}
Under these conditions,
\begin{equation}
|\mathcal{A}_q|\le K(q)\le\rho\,D(q),\qquad\text{and consequently}\qquad|\mathcal{A}_q|\le\min\bigl\{|\Lib_q|,\ \rho\,D(q)\bigr\}.
\label{eq:activation}
\end{equation}
\end{proposition}

The companion study on Navigation Primitives proves the bound for workers that each search one territory of an information space. If realized demand stays bounded, $D(q)\le\bar D$, and $\rho$ does not depend on the library, then $|\mathcal{A}_q|\le\rho\bar D$: growth of the library alone does not force the active part of such a harness to grow.

\textbf{Task-Conditioned Composition.}\phantomsection\label{sec:analysis-adaptive}
The companion study on Harness Primitives asks, under a simplified additive model, when choosing primitives per task beats one fixed harness. Tasks $q$ follow a distribution $\mathcal{Q}$, and each primitive $P_i$ helps on a scope $\Omega_i$ of tasks with prevalence $\mu_i=\Pr_{q\sim\mathcal{Q}}(q\in\Omega_i)$. Activating $P_i$ yields a gain $g_i(q)>0$ on $\Omega_i$ and a penalty $\kappa_i(q)>0$ off it, with conditional means $\bar g_i$ and $\bar\kappa_i$. Under valid execution, the harness with activation vector $\mathbf{z}\in\{0,1\}^N$ attains $R(q,\mathbf{z})=R_0(q)+\sum_{i=1}^{N}z_i\Delta_i(q)$, where $R_0(q)$ is the performance without primitives and $\Delta_i(q)$ is $g_i(q)$ on $\Omega_i$ and $-\kappa_i(q)$ otherwise. All combinations are admissible, and $R$ lies in $[0,1]$. A harness $H$ earns $J(q,H)=\mathbf{1}[H\text{ executes validly}]\cdot R(q,H)$, with $R(q,H)$ as in Section~\ref{sec:formulation}. For a composer whose harnesses $H_{\mathrm{comp}}$ execute validly, with final activations $z_i$, the utility-weighted false-positive and false-negative rates are $\alpha_i=\mathbb{E}[z_i\kappa_i(q)\mid q\notin\Omega_i]/\bar\kappa_i$ and $\beta_i=\mathbb{E}[(1-z_i)g_i(q)\mid q\in\Omega_i]/\bar g_i$.

\begin{proposition}[{Fixed versus task-conditioned composition~\citep{jin2026stitch}}]
\label{prop:fixed-vs-adaptive}
Assume $\mu_i\in(0,1)$, $\bar g_i>0$, and $\bar\kappa_i>0$ for all $i$, and let $z_i^*(q)=\mathbf{1}[q\in\Omega_i]$ be the oracle task-conditioned activation. (i)~The best task-independent activation $\mathbf{z}^*_{\mathrm{fixed}}\in\arg\max_{\mathbf{z}}\mathbb{E}_{q}[R(q,\mathbf{z})]$ falls short of $\mathbf{z}^*(q)$ by the positive gap $\Gamma_{\mathrm{fixed}}$ below. (ii)~Task-conditioned composition outperforms the best fixed harness $H^*_{\mathrm{fixed}}$, which activates $\mathbf{z}^*_{\mathrm{fixed}}$, if and only if its utility-weighted selection loss is smaller than this gap:
\begin{equation}
\begin{gathered}
\Gamma_{\mathrm{fixed}}=\mathbb{E}_q\bigl[R(q,\mathbf{z}^*(q))\bigr]-\mathbb{E}_q\bigl[R(q,\mathbf{z}^*_{\mathrm{fixed}})\bigr]=\sum_{i=1}^{N}\min\bigl(\mu_i\bar g_i,\ (1-\mu_i)\bar\kappa_i\bigr)>0,\\[2pt]
\mathbb{E}\bigl[J(q,H_{\mathrm{comp}})\bigr]>\mathbb{E}\bigl[J(q,H^*_{\mathrm{fixed}})\bigr]\iff\sum_{i=1}^{N}\bigl[\mu_i\beta_i\bar g_i+(1-\mu_i)\alpha_i\bar\kappa_i\bigr]<\Gamma_{\mathrm{fixed}}.
\end{gathered}
\label{eq:fixed-gap}\csname ltx@label\endcsname{eq:selection-loss}
\end{equation}
\end{proposition}

A fixed harness that includes $P_i$ pays its expected out-of-scope penalty $(1-\mu_i)\bar\kappa_i$, and one that excludes it forgoes its expected in-scope gain $\mu_i\bar g_i$; $\Gamma_{\mathrm{fixed}}$ sums the smaller of the two. The composer need not be perfect: under the additive model and valid execution, building each task's harness from the library beats the best fixed harness built from the same primitives if and only if its selection loss stays below this gap.

\textbf{Coverage and Retrieval Ceiling.}\phantomsection\label{sec:analysis-coverage}
Let $\mathcal{U}_q\subseteq\Lib$ be the nonempty set of primitives that serve task $q$, which the study of self-advertised method selection takes to be the annotated reference methods of a goal. The set is random given the information $X$ that a selector reads: the task with its context and the interfaces of the library. With information $Z$, the best achievable expected hit@$k$ of a size-$k$ shortlist $S\subseteq\Lib$ is $V_k(Z)=\mathbb{E}\bigl[\max_{|S|=k}\Pr(S\cap\mathcal{U}_q\neq\varnothing\mid Z)\bigr]$.

\begin{proposition}[{Coverage and retrieval ceiling~\citep{anonymous2026library}}]
\label{prop:coverage}
(i)~If $T$ holds the features that a similarity-only selector computes from $X$, then $V_k(X)\ge V_k(T)$; the inequality can be strict when distinct tasks share the same similarity representation but need different shortlists, and the full interfaces distinguish them. (ii)~Consider a two-stage selector that takes the top $m$ primitives $\mathcal{S}_m(q)$ under a similarity-only selector and then orders them by any procedure, placing the remaining primitives after them. For every $k\le m$ and every task $q$,
\begin{equation}
\text{hit@}k\le\mathbf{1}\bigl[\mathcal{S}_m(q)\cap\mathcal{U}_q\neq\varnothing\bigr],\qquad
\text{recall@}k\le\frac{|\mathcal{S}_m(q)\cap\mathcal{U}_q|}{|\mathcal{U}_q|},
\label{eq:coverage-ceiling}
\end{equation}
for any model in the second stage. (iii)~The expected hit@$k$ objective $S\mapsto\Pr(S\cap\mathcal{U}_q\neq\varnothing\mid X)$ is monotone and submodular, and greedy selection with the true marginal gains attains at least $1-(1-1/k)^k\ge 1-1/e$ of the best size-$k$ coverage.\fillpar
\end{proposition}

The study of self-advertised method selection proves these results for selecting proof methods. An assessment run over the whole library, as in Eq.~\eqref{eq:assess}, escapes the ceiling of Eq.~\eqref{eq:coverage-ceiling}; its coverage is then limited by the information in $X$ and by how accurately the model judges applicability. Expected hit@$k$ is submodular: once a shortlist holds a primitive that is likely to serve the task, a near-duplicate adds little, so the best size-$k$ shortlist need not consist of the $k$ individually most likely primitives.

\section{Experiments}
\label{sec:instantiations}
\label{sec:evidence}
\runningsection{Experiments}

We evaluate \ours{} through its eight primitive families, each on the benchmarks of its own task and against the baselines of its companion study. Three families use the terms of their own studies: the resource of a Math Primitive is a Method Contract, that of a Knowledge Primitive is a source model, and a Navigation Primitive is a territory-backed worker. We order the families by the demand that their tasks place on the last mile, from calling a tool that needs no change to fitting the harness itself to the task. Each family keeps the notation of its paper; the budget $B$, for instance, counts revision rounds in HEART and HERMES, Lean compilations for Math Primitives, and tokens in SourceLearn. In every table, a shaded row marks the primitive system under study, and differences between rates are in percentage points (pp).

\subsection{Experimental Setup}
\label{sec:experimental-setup}

\textbf{Tasks and Datasets.} The eight families cover the following tasks and benchmarks.
(i)~\textit{LLM tool use}: ToolBench, which evaluates the use of a large catalogue of real APIs~\citep{qin2023toolllm}; NESTFUL, whose call sequences feed one call's output into the next~\citep{basu2025nestful}; $\tau^2$-Bench, which extends the tool--agent--user setting of $\tau$-bench to dual control~\citep{barres2025tau,yao2024tau}; ACEBench, which targets robustness and fine-grained tool use across complex multi-tool workflows~\citep{chen2025acebench}; BFCLv4, which covers agentic web search and memory function calling~\citep{patil2025bfcl}; and 50 end-to-end real-world tasks in five domains~\citep{jin2026toolprimitives}.
(ii)~\textit{Repository construction}: the 522 frozen tasks of LEGO-REPO, across seven domains and five difficulty bands~\citep{jin2026codeprimitives}, and the external repository benchmarks RepoZero, RepoGenesis, and RepoCraft under their published protocols.
(iii)~\textit{Formal theorem proving}: 120 Putnam 2015--2025 problems and 60 IMO ProofBench problems in Lean~4 with Mathlib~\citep{demoura2021lean,mathlib2020}, held out from the Putnam 2000--2014 solutions from which the contract library was built~\citep{anonymous2026library}.
(iv)~\textit{Knowledge-intensive tasks}: MultiDoc2Dial, NarrativeQA, SWE-QA, APIBench, and AppWorld~\citep{feng2021multidoc2dial,kovcisky2018narrativeqa,peng2026swe,patil2023gorilla,trivedi2024appworld}.
(v)~\textit{Multi-agent systems}: AIME25, AIME24, MATH, GSM8K, HumanEval+, MBPP+, MedQA, and GPQA-Diamond~\citep{mathai_aime2025,maxwelljia_aime2024,hendrycks2021measuring,cobbe2021training,liu2023your,jin2021disease,rein2024gpqa}.
(vi)~\textit{Software engineering}: SWE-bench Verified~\citep{jimenez2023swebench}, Terminal-Bench~4.0~\citep{merrill2026terminal}, DevOps-Gym~\citep{tang2026devops}, and whole-repository migrations from SWE Refactor Bench~\citep{hong2026swe}.
(vii)~\textit{Large information spaces}: multi-document QA over HotpotQA~\citep{yang-etal-2018-hotpotqa}, 2WikiMultiHopQA~\citep{ho-etal-2020-constructing}, and MuSiQue~\citep{trivedi-etal-2022-musique}; RepoProbe-Python, SWE-QA-Pro, and GAIA-Text-103~\citep{repoprobe2026,cai-etal-2026-swe,ICLR2024_25ae35b5}; controlled spaces of 32K to 512K tokens built on LongAgent's multi-needle setting~\citep{zhao-etal-2024-longagent}; and a demand intervention in a fixed 512K space.
(viii)~\textit{Agent harnesses}: sampled subsets of 100 SWE-bench Verified and 45 Terminal-Bench~2.0 tasks, with two attempts per task and development tasks disjoint from evaluation~\citep{jin2026stitch}.
Additional details on each study's subsets, versions, and changes to the evaluation interface are in Appendix~\ref{app:ledger}.

\textbf{Baselines.} We compare each family with established methods of its task and with controls that remove or replace its primitives while the rest of the pipeline stays fixed.
(i)~\textit{LLM tool use}: ReAct~\citep{yao2023react}, retrieval-only selection, a planner--executor, and DFSDT, ToolLLM's depth-first search over a decision tree~\citep{qin2023toolllm}, all on the backbone and tool catalogue of the Tool Primitives, together with the reference systems of its benchmarks.
(ii)~\textit{Repository construction}: construction paradigms from single-shot generation through execution feedback, diagnosis, and retrieval to retrieval with a generic code editor, and the independently developed repository agents OpenHands, Claude Code, SWE-agent, and Agentless, each under its own harness~\citep{wang2025openhands,anthropic2025claudecode,yang2024swe,xia2024agentless}.
(iii)~\textit{Formal theorem proving}: for method selection, random order, BM25~\citep{robertson2009probabilistic}, TF-IDF~\citep{salton1988term}, and three embedding retrievers, two from OpenAI~\citep{neelakantan2022text} and Qwen3 Embedding~\citep{zhang2025qwen3}, with each embedding retriever used alone and with listwise reranking of its top ten by the selector backbone~\citep{sun2023chatgpt}; for proving, one attempt without retrieval (Base), pass@20 with twice the compilation budget, APOLLO~\citep{ospanov2026apollo}, and the same AxProverBase loop without a contract~\citep{requena2026minimalagentautomatedtheorem}.
(iv)~\textit{Knowledge-intensive tasks}: Hybrid RAG, RAPTOR~\citep{sarthi2024raptor}, HippoRAG~2~\citep{gutierrez2025rag}, and Agent Workflow Memory~\citep{wang2024awm}.
(v)~\textit{Multi-agent systems}: a single agent, TextMAS with natural-language communication, LatentMAS~\citep{zou2025latent} in its sequential configuration, and each single primitive applied to every query, the last three with four agents each; on Llama-3-70B-Instruct, eight existing MAS methods, among them LLM-Debate~\citep{du2023improving} and MAS-GPT~\citep{ye2025mas}, and the prompting baselines Chain-of-Thought~\citep{wei2022chain} and Self-Consistency~\citep{wang2022self}.
(vi)~\textit{Software engineering}: Codex~\citep{openai2025codex}, Claude Code~\citep{anthropic2025claudecode}, OpenHands~\citep{wang2025openhands}, mini-SWE-agent~\citep{yang2024swe}, Grok Build, SageAgent, and Aider with the same backbones.
(vii)~\textit{Large information spaces}: direct, iterative sparse, and dense retrieval, ChainRAG~\citep{zhu-etal-2025-mitigating}, S2G-RAG~\citep{li-etal-2026-s2g}, ReAct, OWL~\citep{hu2025owl}, the repository agents ReAct with RepoGraph~\citep{ICLR2025_4a4a3c19} and RepoDistill~\citep{yin-etal-2026-repodistill}, the benchmark-specific SWE-QA-Pro Agent~\citep{cai-etal-2026-swe}, and, for long contexts, LongAgent~\citep{zhao-etal-2024-longagent} and Chain of Agents~\citep{zhang2024chainagentslargelanguage}.
(viii)~\textit{Agent harnesses}: the mini-SWE-agent seed harness, SWE-agent~\citep{yang2024swe}, Codex, MemoHarness~\citep{huang2026memoharnessagentharnesseslearn}, and MetaHarness~\citep{lee2026metaharnessendtoendoptimizationmodel}.

\begin{table}[!htbp]
\centering
\researchtablesetup
\caption{\textbf{HEART and reference systems on the six ToolBench splits.} Pass and win rates (\%); reference rows use the backbone in the first column and HEART uses Qwen3-8B, whose matched-backbone controls are in Table~\ref{tab:tool-matched}~\citep{jin2026toolprimitives}. Green marks the best and amber the second-best value in each column.}
\label{tab:tool-toolbench}
\fitresearchtable{\begin{tabular}{llcccccccccccccc}
\toprule
\rowcolor{TabHead}
 &
& \multicolumn{2}{c}{\hdr{I1-Inst.}}
& \multicolumn{2}{c}{\hdr{I1-Tool}}
& \multicolumn{2}{c}{\hdr{I1-Cat.}}
& \multicolumn{2}{c}{\hdr{I2-Inst.}}
& \multicolumn{2}{c}{\hdr{I2-Cat.}}
& \multicolumn{2}{c}{\hdr{I3-Inst.}}
& \multicolumn{2}{c}{\hdr{Avg.}} \\
\hdrgapfill\cmidrule(lr){3-4} \cmidrule(lr){5-6} \cmidrule(lr){7-8}
\cmidrule(lr){9-10} \cmidrule(lr){11-12} \cmidrule(lr){13-14} \cmidrule(lr){15-16}
\rowcolor{TabHead}
\multirow{-2}{*}{\hdr{Model}} & \multirow{-2}{*}{\hdr{Method}} & \hdr{Pass} & \hdr{Win} & \hdr{Pass} & \hdr{Win} & \hdr{Pass} & \hdr{Win} & \hdr{Pass} & \hdr{Win} & \hdr{Pass} & \hdr{Win} & \hdr{Pass} & \hdr{Win} & \hdr{Pass} & \hdr{Win} \\
\midrule
Vicuna & ReACT \& DFSDT
& 0.0 & 0.0 & 0.0 & 0.0 & 0.0 & 0.0 & 0.0 & 0.0 & 0.0 & 0.0 & 0.0 & 0.0 & 0.0 & 0.0 \\
\midrule
Alpaca & ReACT \& DFSDT
& 0.0 & 0.0 & 0.0 & 0.0 & 0.0 & 0.0 & 0.0 & 0.0 & 0.0 & 0.0 & 0.0 & 0.0 & 0.0 & 0.0 \\
\midrule
 & ReACT
& 25.0 & 45.0 & 29.0 & 42.0 & 33.0 & 47.5 & 30.5 & 50.8 & 31.5 & 41.8 & 25.0 & 55.0 & 29.0 & 47.0 \\
 & DFSDT
& 57.0 & 55.0 & 61.0 & 55.3 & 62.0 & 54.5 & 77.0 & 68.5 & 77.0 & 58.0 & 66.0 & 69.0 & 66.7 & 60.0 \\
 \multirow{-3}{*}{ToolLLaMA} & DFSDT-Retriever
& 64.0 & 62.3 & 64.0 & 59.0 & 60.5 & 55.0 & \second{81.5} & 68.5 & 68.5 & 60.8 & 65.0 & 73.0 & 67.3 & 63.1 \\
\midrule
 & ReACT
& 41.5 & -- & 44.0 & -- & 44.5 & -- & 42.5 & -- & 46.5 & -- & 22.0 & -- & 40.2 & -- \\
 \multirow{-2}{*}{GPT-3.5-Turbo} & DFSDT
& 54.5 & 60.5 & 65.0 & 62.0 & 60.5 & 57.3 & 75.0 & 72.0 & 71.5 & 64.8 & 62.0 & 69.0 & 64.8 & 64.3 \\
\midrule
 & ReACT
& 5.5 & 31.0 & 3.5 & 27.8 & 5.5 & 33.8 & 6.0 & 35.0 & 6.0 & 31.5 & 14.0 & 47.5 & 6.8 & 34.4 \\
 \multirow{-2}{*}{Claude-2} & DFSDT
& 20.5 & 38.0 & 31.0 & 44.3 & 18.5 & 43.3 & 17.0 & 36.8 & 20.5 & 33.5 & 28.0 & 65.0 & 22.6 & 43.5 \\
\midrule
 & ReACT
& 12.0 & 28.5 & 20.0 & 35.3 & 20.0 & 31.0 & 8.5 & 29.8 & 14.5 & 29.8 & 24.0 & 45.0 & 16.5 & 33.2 \\
 \multirow{-2}{*}{Text-Davinci-003} & DFSDT
& 43.5 & 40.3 & 44.0 & 43.8 & 46.0 & 46.8 & 37.0 & 40.5 & 42.0 & 43.3 & 46.0 & 63.0 & 43.1 & 46.3 \\
\midrule
 & ReACT
& 53.5 & 60.0 & 50.0 & 58.8 & 53.5 & 63.5 & 67.0 & 65.8 & 72.0 & 60.3 & 47.0 & 78.0 & 57.2 & 64.4 \\
 \multirow{-2}{*}{GPT-4} & DFSDT
& 60.0 & 67.5 & \best{71.5} & 67.8 & 67.0 & 66.5 & 79.5 & 73.3 & 77.5 & 63.3 & 71.0 & 84.0 & 71.1 & 70.4 \\
\midrule
 & ReACT
& 54.5 & 62.0 & 52.8 & 58.5 & 54.2 & 63.5 & 65.4 & 65.8 & 71.5 & 62.0 & 46.5 & 76.5 & 57.5 & 64.7 \\
 \multirow{-2}{*}{GPT-5.4} & DFSDT
& 61.2 & 67.5 & 64.5 & 65.2 & 63.8 & 64.2 & 76.5 & 71.2 & 76.2 & 65.8 & 68.4 & 81.2 & 68.8 & 69.2 \\
\midrule
 & ReACT
& 57.4 & 65.2 & 55.2 & 61.5 & 57.8 & 67.5 & 70.2 & 68.4 & 75.4 & 64.5 & 51.2 & 81.5 & 61.2 & 68.1 \\
 \multirow{-2}{*}{Claude-4.6-Sonnet} & DFSDT
& \second{65.4} & \second{70.8} & \second{70.5} & \second{69.5} & \second{68.4} & \second{70.2} & \second{81.5} & \second{75.6} & \second{80.5} & \second{69.4} & \second{73.8} & \second{86.5} & \second{73.4} & \second{73.8} \\
\midrule
 & ReACT
& 56.2 & 63.8 & 53.8 & 60.5 & 56.5 & 65.4 & 68.4 & 67.5 & 73.2 & 63.2 & 49.5 & 79.5 & 59.6 & 66.6 \\
 \multirow{-2}{*}{Gemini-3.1-Pro} & DFSDT
& 63.8 & 69.2 & 67.5 & 66.8 & 66.2 & 67.5 & 79.5 & 73.4 & 78.4 & 67.8 & 71.2 & 84.5 & 71.1 & 71.5 \\
\midrule
\tabours HEART & --
& \best{67.5} & \best{72.8}
& \best{71.5} & \best{70.8}
& \best{69.5} & \best{72.1}
& \best{83.4} & \best{77.5}
& \best{81.6} & \best{71.5}
& \best{77.2} & \best{89.4}
& \best{75.1} & \best{75.7} \\
\bottomrule
\end{tabular}}
\end{table}

\begin{table}[!htbp]
\centering
\researchtablesetup
\caption{\textbf{HEART and reference models on NESTFUL nested call sequences.} F1 for function and parameter names, partial and full sequence-match accuracy, and win rate (every predicted API executable and leading to the gold answer) under one- and three-shot demonstrations; HEART runs Qwen3-8B in its three harness mechanisms and its Tool Primitives (8B\,$\times$\,4)~\citep{jin2026toolprimitives}. Green marks the best and amber the second-best value in each column.}
\label{tab:tool-nestful}
\fitresearchtable{\begin{tabular}{lccccccccccc}
\toprule
\rowcolor{TabHead}
 & & \multicolumn{5}{c}{\hdr{One-shot ICL}} & \multicolumn{5}{c}{\hdr{Three-shot ICL}} \\
\hdrgapfill\cmidrule(lr){3-7}\cmidrule(lr){8-12}
\rowcolor{TabHead}
\multirow{-2}{*}{\hdr{Model}} & \multirow{-2}{*}{\hdr{\#Params (B)}}
& \makecell[b]{\hdr{F1}\\\hdr{Func.}} & \makecell[b]{\hdr{F1}\\\hdr{Param.}} & \makecell[b]{\hdr{Part}\\\hdr{Acc.}} & \makecell[b]{\hdr{Full}\\\hdr{Acc.}} & \makecell[b]{\hdr{Win}\\\hdr{Rate}}
& \makecell[b]{\hdr{F1}\\\hdr{Func.}} & \makecell[b]{\hdr{F1}\\\hdr{Param.}} & \makecell[b]{\hdr{Part}\\\hdr{Acc.}} & \makecell[b]{\hdr{Full}\\\hdr{Acc.}} & \makecell[b]{\hdr{Win}\\\hdr{Rate}} \\
\midrule
xLAM-1b-fc-r & 1 & 0.19 & 0.08 & 0.09 & 0.00 & 0.01 & 0.22 & 0.09 & 0.09 & 0.03 & 0.02 \\
xLAM-2-1b-fc-r & 1 & 0.41 & 0.13 & 0.13 & 0.00 & 0.00 & 0.43 & 0.12 & 0.13 & 0.00 & 0.00 \\
xLAM-7b-fc-r & 7 & 0.49 & 0.17 & 0.15 & 0.00 & 0.03 & 0.55 & 0.23 & 0.23 & 0.15 & 0.14 \\
xLAM-2-8b-fc-r & 8 & 0.48 & 0.15 & 0.14 & 0.00 & 0.01 & 0.47 & 0.17 & 0.15 & 0.04 & 0.04 \\
Hammer2.0-7b & 7 & 0.56 & 0.24 & 0.21 & 0.07 & 0.16 & 0.61 & 0.30 & 0.29 & 0.22 & 0.25 \\
Hammer2.1-7b & 7 & 0.10 & 0.05 & 0.05 & 0.01 & 0.01 & 0.16 & 0.10 & 0.11 & 0.08 & 0.08 \\
Llama-3.1-8B-Instruct & 8 & 0.64 & 0.19 & 0.17 & 0.06 & 0.06 & 0.63 & 0.22 & 0.22 & 0.16 & 0.11 \\
ToolACE-8B & 8 & 0.43 & 0.13 & 0.13 & 0.00 & 0.00 & 0.50 & 0.15 & 0.13 & 0.00 & 0.00 \\
ToolACE-2-Llama-3.1-8B & 8 & 0.28 & 0.10 & 0.13 & 0.00 & 0.00 & 0.29 & 0.10 & 0.13 & 0.00 & 0.00 \\
Granite-20B-FunctionCalling & 20 & 0.64 & 0.20 & 0.17 & 0.02 & 0.05 & 0.61 & 0.25 & 0.26 & 0.21 & 0.20 \\
Mixtral-8x7B-Instruct-v0.1 & 46.7 & 0.22 & 0.07 & 0.05 & 0.00 & 0.01 & 0.32 & 0.13 & 0.14 & 0.09 & 0.09 \\
xLAM-8x7b-fc-r & 46.7 & 0.40 & 0.15 & 0.16 & 0.01 & 0.01 & 0.43 & 0.16 & 0.17 & 0.02 & 0.03 \\
Llama-3.1-70B-Instruct & 70 & 0.41 & 0.19 & 0.15 & 0.04 & 0.09 & 0.33 & 0.17 & 0.15 & 0.07 & 0.11 \\
Mixtral-8x22B-Instruct-v0.1 & 141 & 0.49 & 0.21 & 0.17 & 0.06 & 0.07 & 0.65 & 0.29 & 0.28 & 0.21 & 0.23 \\
xLAM-8x22b-fc-r & 141 & 0.53 & 0.21 & 0.22 & 0.12 & 0.03 & 0.50 & 0.23 & 0.25 & 0.17 & 0.06 \\
Llama-3.1-405B-Instruct-fp8 & 405 & 0.41 & 0.14 & 0.08 & 0.03 & 0.10 & 0.41 & 0.18 & 0.13 & 0.07 & 0.14 \\
DeepSeek-V3 & 685 & 0.69 & 0.36 & 0.27 & 0.09 & 0.43 & 0.69 & 0.42 & 0.37 & 0.29 & 0.60 \\
\midrule
GPT-4o (2024-08-06) & -- & 0.73 & 0.41 & 0.38 & 0.28 & 0.59 & 0.74 & 0.41 & 0.38 & 0.28 & 0.60 \\
GPT-5.4 & -- & \second{0.78} & \second{0.48} & \second{0.50} & \second{0.40} & \second{0.66} & \second{0.80} & \second{0.50} & \second{0.52} & \second{0.42} & \second{0.68} \\
Claude-4.6-Sonnet & -- & 0.77 & 0.47 & 0.49 & 0.39 & 0.65 & 0.79 & 0.49 & 0.51 & 0.41 & 0.67 \\
Gemini-3.1-Pro & -- & 0.76 & 0.46 & 0.48 & 0.38 & 0.63 & 0.78 & 0.48 & 0.50 & 0.40 & 0.65 \\
\midrule
\tabours HEART & 8\,$\times$\,4
& \best{0.82} & \best{0.52} & \best{0.55} & \best{0.44} & \best{0.75}
& \best{0.84} & \best{0.56} & \best{0.58} & \best{0.47} & \best{0.79} \\
\bottomrule
\end{tabular}}
\end{table}

\textbf{Models.} Each family keeps the backbones of its companion study, and its controls hold the backbone fixed across arms unless the control varies the model itself. Tool Primitives and HEART's request decomposition, value resolution, and result validation all run Qwen3-8B~\citep{qwen3technicalreport}. Code Primitives adapt with GPT-5.6 Terra, which also fills the backbone and diagnosis positions, and a model-allocation study varies all three. Math Primitives are selected by GPT-5.6 Luna~\citep{s06_openai2026gpt56}, GPT-5.6 Terra, or DeepSeek-V4.1-Flash~\citep{s06_xu2026deepseek}, and GPT-5.6 Luna drives the prover. SourceLearn learns and applies Knowledge Primitives with GPT-5.6 Luna, GPT-OSS-120B~\citep{agarwal2025gpt}, or DeepSeek-V4.1-Flash, one backend per configuration performing every LLM step of every method. Agent Primitives run Qwen3-8B and two DeepSeek-R1-Distill models~\citep{deepseekai2025deepseekr1incentivizingreasoningcapability} as worker backbones, three of the study's five, and Llama-3-70B-Instruct~\citep{grattafiori2024llama} in the comparison with existing MAS methods; they are composed per query by a GPT-5.2 Organizer, which Claude-4~\citep{anthropic2025claude} replaces in an ablation. The controls of Dev-Primitives run GPT-5.6 Sol, and their harness comparisons run each baseline's own backbone in both arms. Navigation Primitives and the rest of their system run on GPT-4.1, and ANTMAN-H keeps the GPT-4.1 orchestrator and runs every other component, the Navigation Primitives included, on Qwen3-8B. Harness Primitives are developed with GPT-5.6 Sol and composed for a GPT-5.6 Luna actor at medium effort, whose backbone the composer shares; GPT-5.6 Terra, DeepSeek-V4-Flash, and Claude-4.5-Haiku serve as actors not used in development.

\subsection{Tool Primitives for LLM Tool Use}
\label{sec:tool-experiments}

LLM tool use places the smallest demand on the last mile: each tool does what the task needs, and what remains is to call it correctly. In function calling, the model writes schema-compliant arguments for each call~\citep{schick2023toolformer,patil2023gorilla,patil2025bfcl} and maps one tool's output into the next tool's input schema, a step that nested and multi-turn calls make brittle, while a large catalogue in context degrades performance~\citep{liu-etal-2024-lost}. Tool Primitives reverse this burden: instead of requiring models to speak the language of APIs, they make tools speak the language of models~\citep{jin2026toolprimitives}. Each Tool Primitive resolves the schema of its own tool, and HEART, a harness engineering framework, decomposes each query into an invocation plan, grounds each step's arguments, and validates every returned result (Section~\ref{sec:var-tool}).

\textbf{Setup.} We evaluate HEART on the five benchmarks of its study and report three of them in full: the six single- and multi-tool splits of ToolBench~\citep{qin2023toolllm}, NESTFUL~\citep{basu2025nestful}, and the retail, airline, and telecom domains of $\tau^2$-Bench~\citep{barres2025tau}. The baselines are each benchmark's reference systems: on ToolBench, models from Vicuna~\citep{zheng2023judging} and Alpaca~\citep{alpaca} to ToolLLaMA~\citep{qin2023toolllm} and GPT-4~\citep{achiam2023gpt} under ReAct~\citep{yao2023react} and DFSDT; on NESTFUL, function-calling models such as xLAM~\citep{liu2024apigen}, Hammer~\citep{lin2024hammer}, ToolACE~\citep{liu2024toolace}, and Granite~\citep{abdelaziz2024granite} and general models from Llama-3.1~\citep{grattafiori2024llama} and Mixtral~\citep{jiang2024mixtral} to DeepSeek-V3~\citep{liu2024deepseek} and GPT-4o~\citep{hurst2024gpt}; and on $\tau^2$-Bench, GPT-4.1, GPT-4.1-mini, GPT-o4-mini, and Claude-3.7-Sonnet. GPT-5.4, Claude-4.6-Sonnet, and Gemini-3.1-Pro appear on all five. HEART runs Qwen3-8B~\citep{qwen3technicalreport} in request decomposition, value resolution, result validation, and the Tool Primitives, with revision budget $B=3$ and retrieval over the 25,519 tools of ToolFace; results average three runs, and eight on $\tau^2$-Bench.

\begin{wraptable}{R}{0.6\linewidth}
\wraplift
\researchpaneltrue
\begin{researchtable}
\centering
\researchtablesetup
\caption{\textbf{HEART and reference models on dual-control conversational tool use in $\tau^2$-Bench.} Pass$^k$ is the fraction of tasks whose $k$ independent trials all succeed, tokens, API cost (USD) and latency (s) are averages per task, and HEART runs its default Qwen3-8B backbone~\citep{jin2026toolprimitives}. Green marks the best and amber the second-best value in each Pass$^k$ column within each domain.}
\label{tab:tool-tau}
\fitresearchtable{\begin{tabular}{llccccccc}
\toprule
\rowcolor{TabHead}
\hdr{Method} & \hdr{Domain} & \hdr{Pass\textsuperscript{1}} & \hdr{Pass\textsuperscript{2}} & \hdr{Pass\textsuperscript{3}} & \hdr{Pass\textsuperscript{4}} & \hdr{Tokens} & \makecell[b]{\hdr{Cost}\\\hdr{(\$)}} & \makecell[b]{\hdr{Latency}\\\hdr{(s)}} \\
\midrule
 & Retail  & 0.75 & 0.64 & 0.58 & 0.53 & 7,401 & 0.0592 & 14.28 \\
        & Airline & 0.56 & 0.45 & 0.42 & 0.40 & 7,825 & 0.0626 & 15.62 \\
        \multirow{-3}{*}{GPT-4.1} & Telecom & 0.34 & 0.26 & 0.22 & 0.19 & 6,862 & 0.0549 & 34.41 \\
\midrule
 & Retail  & 0.71 & 0.59 & 0.52 & 0.46 & 5{,}127 & 0.0031 & 4.24 \\
            & Airline & 0.59 & 0.48 & 0.42 & 0.38 & 5{,}389 & 0.0032 & 4.68 \\
            \multirow{-3}{*}{GPT-o4-mini} & Telecom & 0.42 & 0.33 & 0.29 & 0.26 & 4{,}856 & 0.0029 & 26.05 \\
\midrule
 & Retail  & 0.66 & 0.52 & 0.44 & 0.39 & 5,214 & 0.0083 & 6.12 \\
             & Airline & 0.51 & 0.39 & 0.32 & 0.26 & 5,563 & 0.0089 & 6.81 \\
             \multirow{-3}{*}{GPT-4.1-mini} & Telecom & 0.44 & 0.30 & 0.22 & 0.18 & 5,021 & 0.0080 & 21.94 \\
\midrule
 & Retail  & 0.79 & 0.69 & 0.63 & 0.60 & 6,482 & 0.0972 & 11.45 \\
                  & Airline & 0.50 & 0.41 & 0.38 & 0.36 & 6,745 & 0.1012 & 12.32 \\
                  \multirow{-3}{*}{Claude-3.7-Sonnet} & Telecom & 0.49 & 0.37 & 0.30 & 0.25 & 6,038 & 0.0906 & 30.88 \\
\midrule
 & Retail  & 0.70 & 0.53 & 0.43 & 0.37 & 7,842 & 0.1176 & 16.24 \\
        & Airline & 0.64 & 0.55 & 0.50 & 0.48 & 8,126 & 0.1219 & 18.42 \\
        \multirow{-3}{*}{GPT-5.4} & Telecom & 0.56 & 0.40 & 0.34 & 0.29 & 7,319 & 0.1098 & 35.11 \\
\midrule
 & Retail  & \second{0.80} & \second{0.72} & \second{0.66} & \second{0.61} & 7,153 & 0.1073 & 14.12 \\
                  & Airline & \second{0.72} & \second{0.62} & \second{0.55} & \second{0.50} & 7,412 & 0.1112 & 16.52 \\
                  \multirow{-3}{*}{Claude-4.6-Sonnet} & Telecom & \second{0.58} & \second{0.47} & \second{0.39} & \second{0.33} & 6,728 & 0.1009 & 33.48 \\
\midrule
 & Retail  & 0.68 & 0.54 & 0.46 & 0.40 & 6,914 & 0.0830 & 9.15 \\
               & Airline & 0.54 & 0.41 & 0.33 & 0.29 & 7,258 & 0.0871 & 10.82 \\
               \multirow{-3}{*}{Gemini-3.1-Pro} & Telecom & 0.55 & 0.42 & 0.34 & 0.30 & 6,492 & 0.0779 & 28.46 \\
\midrule
\tabours  & Retail  & \best{0.82} & \best{0.78} & \best{0.76} & \best{0.75} & 21,684 & 0.0152 & 18.72 \\
\tabours       & Airline & \best{0.73} & \best{0.68} & \best{0.65} & \best{0.63} & 24,531 & 0.0172 & 21.34 \\
\tabours       \multirow{-3}{*}{HEART} & Telecom & \best{0.63} & \best{0.57} & \best{0.53} & \best{0.50} & 20,887 & 0.0146 & 37.95 \\
\bottomrule
\end{tabular}}
\end{researchtable}

\end{wraptable}
\textbf{Main Results.} On ToolBench (Table~\ref{tab:tool-toolbench}), HEART reaches an average pass rate of 75.1\% and win rate of 75.7\%. Without tree search, it leads Claude-4.6-Sonnet with DFSDT, the strongest commercial baseline, by 1.7 and 1.9~pp, and ToolLLaMA with DFSDT and its retriever, the strongest fine-tuned baseline, by 7.8 and 12.6~pp. The lead is largest on I3-Inst., the hardest split, whose multi-tool instructions require cross-API dependencies: 77.2\% pass and 89.4\% win rates, 3.4 and 2.9~pp above Claude-4.6-Sonnet with DFSDT. On the same Qwen3-8B backbone and ToolFace catalogue, HEART leads DFSDT, the strongest of four scaffolds on ToolBench, by 19.3~pp pass rate, 75.1\% against 55.8\%, and by 0.31 NESTFUL full accuracy (Table~\ref{tab:tool-matched}).\fillpar[0.22]

On NESTFUL (Table~\ref{tab:tool-nestful}), HEART is best on all five metrics under both one- and three-shot demonstrations. Its full-sequence accuracy is 0.44 and 0.47, against 0.40 and 0.42 for GPT-5.4, the strongest baseline, and its win rate 0.75 and 0.79, against 0.66 and 0.68. Fine-tuning on flat tool-use trajectories does not carry over to nested sequences: with one-shot demonstrations, the xLAM models up to 8B and both ToolACE models reach 0.00 full accuracy. Nor does scale close the gap, as DeepSeek-V3, with 685B parameters, reaches 0.09.

On $\tau^2$-Bench (Table~\ref{tab:tool-tau}), HEART has the highest Pass$^k$ in each domain for each $k$. Its Pass$^1$ margins over Claude-4.6-Sonnet, the strongest baseline model, are small, 0.82, 0.73, and 0.63 against 0.80, 0.72, and 0.58 on retail, airline, and telecom, and they widen with $k$: in the airline domain, whose tasks include booking a flight under the airline's policy, HEART succeeds in all four trials on 0.63 of the tasks against 0.50, and in telecom on 0.50 against 0.33. Although each invocation step adds a Tool Primitive's model call and HEART uses 20,887 to 24,531 tokens per task, it costs \$0.0146 to \$0.0172 per task on Qwen3-8B, against \$0.1098 to \$0.1219 for GPT-5.4, up to $7.7\times$ less.\fillpar[0.6][1]

% Wrapped table of sections/05a-tool.tex (formerly the right body of tables/pairs/tool-matched+tool-ablation.tex).
\begin{table}[!htbp]
\centering
\researchtablesetup
\caption{\textbf{Removing Tool Primitives or one of HEART's harness mechanisms.} Each row removes one component from the full configuration (Qwen3-8B, ToolFace): request decomposition, value resolution, result validation, or ToolFace with its Tool Primitives; metrics as in Table~\ref{tab:tool-matched}~\citep{jin2026toolprimitives}. Green marks the best and amber the second-best value in each column.}
\label{tab:tool-ablation}
\fitresearchtable{\begin{tabular}{lcccc}
\toprule
\rowcolor{TabHead}
 & \multicolumn{2}{c}{\hdr{ToolBench}} & \multicolumn{2}{c}{\hdr{NESTFUL}} \\
\hdrgapfill\cmidrule(lr){2-3}\cmidrule(lr){4-5}
\rowcolor{TabHead}
\multirow{-2}{*}{\hdr{Variant}} & \hdr{Pass} & \hdr{Win} & \hdr{Full Acc.} & \hdr{Win} \\
\midrule
w/o Decomposition               & 57.2 & 64.1 & 0.26 & 0.51 \\
w/o Resolution                  & \second{62.4} & \second{67.3} & \second{0.28} & \second{0.52} \\
w/o ToolFace \& Tool Primitives & 16.1 & 27.9 & 0.06 & 0.06 \\
w/o Validation                  & 47.6 & 49.6 & 0.15 & 0.33 \\
\midrule
\tabours HEART & \best{75.1} & \best{75.7} & \best{0.44} & \best{0.75} \\
\bottomrule
\end{tabular}}
\end{table}

\textbf{Component Ablation.} Removing one component at a time locates the gain in the natural-language interface (Table~\ref{tab:tool-ablation}). Without ToolFace and its Tool Primitives, value resolution resolves arguments against raw schemas and dispatches calls itself; ToolBench pass rate then falls from 75.1\% to 16.1\% and NESTFUL full accuracy from 0.44 to 0.06. Removing result validation, which drives revision, gives the second-largest drop, to 47.6\% and 0.15, and removing request decomposition or value resolution gives 57.2\% and 62.4\%. Removing the primitives also removes one model call per step; given that budget to self-refine its schema-compliant arguments at matched backbone, calls, and tokens, value resolution without primitives reaches 34.2\% (Table~\ref{tab:tool-budget}), so the interface carries most of the gain.\fillpar

On the other two benchmarks, HEART has the highest ACEBench Overall score, 86.9, and the highest BFCLv4 Web Search and Memory Overall scores, 86.0 and 69.38~\citep{chen2025acebench,patil2025bfcl}. On 50 end-to-end real-world tasks in five domains, it completes 84\%, against 20\% to 24\% for GPT-5.4, Claude-4.6-Sonnet, and Gemini-3.1-Pro in their product interfaces and 22\% to 32\% when they call the ToolFace tools through native function calling. We provide the full framework of HEART, its overview figure, the matched-backbone and budget-matched controls, and further diagnostics in Appendix~\ref{sec:tool-mechanism}. A Tool Primitive invokes its function as it is; a component that must itself change to fit its target calls for a primitive that adapts the resource it owns, the step a Code Primitive takes.

\wrapguard
\subsection{Code Primitives for Repository Construction}
\label{sec:code-experiments}

Repository construction asks an agent to deliver a complete project whose modules, interfaces, configuration, tests, and dependencies work together~\citep{ding2025nl2repo}. When the project reuses code, the last mile lies in making that code fit its target. A component reused from another repository becomes part of the delivered source, so its signatures, data representations, and dependencies must be reconciled with the target. Retrieval-augmented code generation finds related code~\citep{zhang2023repocoder,lu2022reacc,zhou2022docprompting} but leaves this reconciliation to the caller. Code Primitives assign it to the component itself: each Code Primitive adapts the implementation it owns to the target and checks the result against the tests it carries, while LEGO retrieves, coordinates, and verifies these primitives during construction~\citep{jin2026codeprimitives}.

\textbf{Setup.} LEGO-REPO poses the task as reconstruction. Each of its 522 frozen tasks removes the implementation of an installable Python package, keeps the package's interfaces, native tests, and environment, and grades the delivered repository with those tests. The tasks span seven domains and five difficulty bands, D1--D5, which combine module count with the number of native tests the original passes. Let $\Pi(X)$ denote the number of native tests passed using source set $X$. For task $T$, the original sources $O$ define the identity ceiling $c(T)=\Pi(O)$ and an empty reconstruction tree the floor $f(T)=\Pi(\varnothing)$, and generated sources $G$ score
\begin{equation}
s(G;T)=\operatorname{clamp}_{[0,1]}\left(\frac{\Pi(G)-f(T)}{c(T)-f(T)}\right),\qquad
S_{\mathrm{delivery}}=\frac{1}{522}\sum_{T=1}^{522}\widetilde{s}_T,\qquad
\widetilde{s}_T=\begin{cases}s(G_T;T),&\text{admissible},\\0,&\text{otherwise},\end{cases}
\label{eq:delivery-score}
\end{equation}
where a run is admissible when $c(T)-f(T)\geq5$, execution completes, and provenance confirms that the tests exercised the generated target code. The zero-score rate is the fraction of tasks assigned 0. The baselines are construction paradigms, from single-shot generation to retrieval with a generic code editor, and independently developed repository agents, among them OpenHands and Claude Code~\citep{wang2025openhands,anthropic2025claudecode}, each under its own harness with tasks, environment, and grading held fixed. Unless a configuration varies them, GPT-5.6 Terra fills the backbone, adaptation, and diagnosis positions, retrieval returns the top two of 1,424 CodeFace entries per requirement, and revision runs at most five rounds per module.

\begin{wraptable}{R}{0.56\linewidth}
\wraplift
\researchpaneltrue
\begin{researchtable}
\centering
\researchtablesetup
\caption{\textbf{LEGO against construction paradigms on 522 repositories.} Delivery score on the native $[0,1]$ scale for the smallest (D1) and largest (D5) difficulty bands and all tasks, zero-score rate and recorded-call cost per task (USD), with stages R retrieval, A adaptation, E generic editing, V execution feedback and D diagnosis; the two one-attempt rows are first-round prefixes of longer trajectories~\citep{jin2026codeprimitives}. Green marks the best and amber the second-best value in each of the D1, D5, All, and Zero columns. Lower is better for Zero.}
\label{tab:code-paradigms}
\fitresearchtable{\begin{tabular}{lcccccc}
\toprule
\rowcolor{TabHead}
\hdr{Configuration} & \hdr{Stages} & \hdr{D1} & \hdr{D5} & \hdr{All} & \hdr{Zero (\%)} & \hdr{\$/task} \\
\midrule
Single-shot             & \stg{-{}-{}-{}-} & --    & --    & 0.2172 & 57.1 & 0.86 \\
Retrieval, one attempt  & \stg{R-{}-{}-}   & --    & --    & 0.2314 & 58.6 & 0.87 \\
Feedback                & \stg{-{}-{}-V}   & 0.525 & 0.129 & 0.3180 & 41.0 & 1.55 \\
\midrule
Retrieval + Feedback    & \stg{R-{}-V}     & 0.544 & 0.141 & 0.3352 & 34.7 & 1.56 \\
Feedback + Diagnosis    & \stg{-{}-DV}     & 0.569 & 0.135 & 0.3469 & 30.1 & 1.79 \\
Retrieval + Diagnosis   & \stg{R-DV}       & 0.649 & 0.127 & 0.3980 & 31.6 & 1.80 \\
Retrieval + Adapt       & \stg{RA-V}       & 0.653 & \second{0.275} & \second{0.4795} & \best{17.8} & 1.83 \\
RAG + Edit              & \stg{RE\,DV}     & 0.637 & 0.136 & 0.4093 & 24.3 & 2.01 \\
\tabours LEGO           & \stg{RADV}       & \best{0.686} & \best{0.338} & \best{0.5134} & \second{18.8} & 2.08 \\
\quad same-repo donors excluded & \stg{RADV} & \second{0.683} & 0.221 & 0.4743 & 24.9 & 2.08 \\
\bottomrule
\end{tabular}}
\end{researchtable}

\end{wraptable}

\textbf{Main Results.} Construction from scratch weakens as repositories grow: under execution feedback, GPT-5.6 Terra is the strongest of the study's 13 backbones, and the mean delivery of the 13 falls from 0.393 on D1 to 0.038 on D5. Table~\ref{tab:code-paradigms} compares the construction paradigms on all 522 tasks with GPT-5.6 Terra. LEGO reaches a delivery score of 0.5134 against 0.3180 for execution feedback, a gain of 0.1954 or 61.4\%, and lowers the zero-score rate from 41.0\% to 18.8\%. Retrieval adds 0.0172 to execution feedback, whereas primitive adaptation adds 0.1154 to retrieval with diagnosis; RAG with a generic editor, the strongest paradigm without primitive adaptation, reaches 0.4093. The strongest independent repository agents stay near the feedback baseline: OpenHands reaches 0.3263 with GPT-5.6 Terra and Claude Code 0.3170 with Claude Sonnet 5, while SWE-agent and Agentless, run with GPT-5.6 Terra through an issue-to-patch adapter, reach 0.2750 and 0.2181~\citep{yang2024swe,xia2024agentless}. LEGO leads OpenHands and Claude Code on the same backbones by 57.3\% and 56.2\%, at 0.5134 and 0.4953. Across the 13 backbones of the study, with adaptation and diagnosis on GPT-5.6 Terra, LEGO improves all 13 and raises mean delivery from 0.1926 to 0.3400, a gain of 0.1474~\citep{jin2026codeprimitives}.

\begin{wraptable}{R}{0.48\linewidth}
\wraplift
\researchpaneltrue\researchpanelsize{9}
% Pair body (tables/pairs/code-consumption+code-fields.tex); a float of its own if input directly.
\begin{researchtable}
\centering
\researchtablesetup
\caption{\textbf{Access, invocation, and adaptation of the same reusable components.} The reference \stg{-{}-DV} uses no library; the other rows share LEGO's retrieval and diagnosis: \stg{RC\,DV} shows the retrieved component read-only, \stg{RI\,DV} vendors and calls it unchanged, and structured-schema adaptation sends the adaptation request as a typed field mapping instead of free text; delivery on the $[0,1]$ scale, $\Delta$ from \stg{-{}-DV}~\citep{jin2026codeprimitives}. Green marks the best and amber the second-best value in each of the D1, D5, and All columns.}
\label{tab:code-consumption}
\fitresearchtable{\begin{tabular}{lcccc}
\toprule
\rowcolor{TabHead}
\hdr{Configuration} & \hdr{D1} & \hdr{D5} & \hdr{All} & \hdr{$\Delta$ \stg{-{}-DV}} \\
\midrule
Feedback + Diagnosis (\stg{-{}-DV}) & 0.569 & 0.135 & 0.3469 & \textit{ref.} \\
Context-only (\stg{RC\,DV})         & 0.620 & 0.115 & 0.3544 & \sn{+0.0075} \\
Import + Call (\stg{RI\,DV})        & 0.627 & 0.060 & 0.3430 & \sn{-0.0039} \\
Structured-schema adaptation        & \second{0.679} & \second{0.327} & \second{0.4974} & \sn{+0.1505} \\
\tabours LEGO (\stg{RADV})          & \best{0.686} & \best{0.338} & \best{0.5134} & \sn{+0.1665} \\
\bottomrule
\end{tabular}}
\end{researchtable}

\end{wraptable}
\textbf{Access versus Adaptation.} Access to the retrieved components adds little; the gain comes from adapting them. The consumption controls of Table~\ref{tab:code-consumption} keep LEGO's retrieval and diagnosis and vary how the components are used. As read-only context with their contracts and tests, they reach 0.3544, and vendored and called unchanged they reach 0.3430, against 0.3469 for diagnosis without a library; on D5, unchanged invocation falls to 0.060 against 0.135 for that reference. Structured-schema adaptation keeps primitive adaptation but replaces the natural-language request with a typed field mapping, and it scores 0.4974 against 0.5134. Most of LEGO's 0.1041 margin over generic editing lies in the primitive's fields: given the carried tests, the interface contract and provenance, and the dependency closure, a generic editor over the same components recovers 0.0946 of it.

\textbf{Cost and Model Allocation.} Spending LEGO's recorded-call budget on more repair attempts does not close the gap: execution feedback with eight attempts per module costs \$2.07 per task and reaches 0.3437, against 0.5134 at \$2.08 for LEGO. The locally scoped models can also be small. With GPT-OSS-20B as the resident LLM $M_i$ of each primitive and as the diagnosis model, LEGO keeps 95.1\% of its delivery, 0.4880, at 24.0\% lower cost; replacing the backbone with DeepSeek-V3.2 keeps 48.1\%. Delivery thus depends more on the backbone, which plans and integrates the repository, than on these models.

The gain holds across repository sizes, library restrictions, and benchmarks. Over the five bands, the absolute gain stays between 0.161 and 0.209, so the relative gain rises from 30.7\% on D1 to 162.0\% on D5 as feedback falls. Excluding same-repository donors lowers delivery to 0.4743, and a library re-mined from a corpus disjoint from the benchmark repositories gains 0.1348 over feedback. Under the published protocols of three external repository benchmarks, LEGO with GPT-5.6 Terra raises the RepoZero (Python to Python) pass rate from 53.8\% to 62.9\%, the RepoGenesis resolved rate from 41.0\% to 48.2\%, and the RepoCraft normalized score from 0.436 to 0.501 over execution feedback, relative gains of 16.9\%, 17.6\%, and 14.9\%. With the CodeFace entries traced to these targets removed, the library covers 29.6--34.8\% of their target modules, against 61.4\% on LEGO-REPO.

We provide the full framework of LEGO, its overview figure, the field-by-field decomposition of the primitive, and the full model-allocation table in Appendix~\ref{sec:code-mechanism}, with breakdowns by band and domain, donor-provenance controls, attempt budgets, and seed uncertainty in Appendix~\ref{app:experiments-extra}. A Code Primitive checks its adaptation against the tests it carries and leaves the integrated result to the repository's build and tests. When an exact checker judges the final result, the open step moves earlier, to deciding which method to apply.

\wrapguard
\subsection{Math Primitives for Formal Theorem Proving}
\label{sec:method-experiments}

In formal theorem proving with Lean~4 and Mathlib~\citep{demoura2021lean,mathlib2020}, the kernel checks each completed proof, and LLM provers generate and revise proofs against this check~\citep{xin2024deepseek,first2023baldur,ospanov2026apollo}, aided by premise selection, Mathlib search, and lemma libraries~\citep{yang2023leandojo,zhu2026premise,gao2026leansearch,wang2024lego}. Here the last mile is deciding whether a method applies: which part of the goal it targets, what action it takes there, and which conditions that action requires. For $n\sum_i a_i^2\ge(\sum_i a_i)^2$ over the reals, similarity retrieval proposes AM--GM, which needs $a_i\ge0$, a condition the goal does not give, while Cauchy--Schwarz applies. Math Primitives let the library speak instead: each holds a proof method as a Method Contract, and under self-advertised method selection every contract proposes how it would be used on the goal before the contracts are ranked~\citep{anonymous2026library}.

\textbf{Setup.} We evaluate on 120 Putnam problems from 2015--2025 and 60 IMO ProofBench problems, held out from the Putnam 2000--2014 solutions behind the 82 contracts. Each problem is labeled with the contracts whose method appears in its official solution; GPT-5.6 Sol proposed the labels, and a human checked each one before any selector ran. The baselines are random ordering, BM25, TF-IDF cosine, and three embedding retrievers, each also paired with a listwise LLM reranker over its top ten candidates. The LLM-based selectors within a block of Table~\ref{tab:method-selection} share one backbone, GPT-5.6 Terra, GPT-5.6 Luna, or DeepSeek-V4.1-Flash.

\begin{table}[!htbp]
\centering
\researchtablesetup
\caption{\textbf{Selecting Method Contracts for held-out Putnam and IMO ProofBench problems.} hit@$k$ is the fraction of problems whose top-$k$ shortlist contains an annotated reference contract, R@$k$ the fraction of annotated contracts in the top $k$, and MRR the mean reciprocal rank; rerankers reorder the top ten candidates of an embedding retriever, and self-advertisement ranks all 82 contracts of the library from their goal-specific proposals~\citep{anonymous2026library}. Green marks the best and amber the second-best value in each column within each selector model.}
\label{tab:method-selection}
\fitresearchtable{\begin{tabular}{lcccccccccccc}
\toprule
\rowcolor{TabHead}
 & \multicolumn{6}{c}{\hdr{Putnam 2015--2025 ($n$\,=\,120)}} & \multicolumn{6}{c}{\hdr{IMO ProofBench ($n$\,=\,60)}} \\
\hdrgapfill\cmidrule(lr){2-7}\cmidrule(lr){8-13}
\rowcolor{TabHead}
\multirow{-2}{*}{\hdr{Selector}} & \hdr{hit@1} & \hdr{hit@3} & \hdr{hit@5} & \hdr{R@3} & \hdr{R@5} & \hdr{MRR} & \hdr{hit@1} & \hdr{hit@3} & \hdr{hit@5} & \hdr{R@3} & \hdr{R@5} & \hdr{MRR} \\
\midrule
\tabgroup{13}{No LLM} \\
Random             & 0.058 & 0.158 & 0.225 & 0.038 & 0.053 & 0.141 & 0.083 & 0.167 & 0.267 & 0.041 & 0.076 & 0.158 \\
BM25               & 0.208 & 0.400 & 0.508 & 0.104 & 0.140 & 0.337 & 0.183 & 0.333 & 0.400 & 0.103 & 0.147 & 0.295 \\
TF-IDF cosine      & 0.267 & 0.442 & 0.525 & 0.118 & 0.144 & 0.382 & 0.233 & 0.367 & 0.433 & 0.135 & 0.165 & 0.318 \\
OpenAI emb-3-small & 0.325 & 0.533 & 0.683 & 0.157 & 0.229 & 0.463 & 0.267 & 0.517 & 0.617 & 0.213 & 0.263 & 0.411 \\
OpenAI emb-3-large & 0.425 & 0.667 & 0.725 & 0.204 & 0.267 & 0.556 & 0.367 & 0.583 & 0.717 & 0.268 & 0.339 & 0.523 \\
Qwen3-Emb-0.6B     & 0.450 & 0.675 & 0.775 & 0.217 & 0.291 & 0.579 & 0.350 & 0.617 & 0.767 & 0.282 & 0.380 & 0.510 \\
\midrule
\tabgroup{13}{GPT-5.6 Terra} \\
OpenAI-small + rerank & 0.642 & 0.767 & 0.775 & 0.285 & 0.324 & 0.704 & 0.483 & 0.717 & 0.767 & 0.336 & 0.397 & 0.607 \\
OpenAI-large + rerank & 0.667 & 0.800 & 0.842 & 0.315 & 0.373 & 0.742 & \second{0.600} & 0.783 & \second{0.900} & 0.378 & 0.455 & 0.712 \\
Qwen3 + rerank        & \second{0.675} & 0.808 & 0.858 & 0.338 & 0.407 & 0.751 & \best{0.633} & 0.783 & \second{0.900} & 0.373 & 0.469 & \second{0.724} \\
\tabours Self-advertisement & 0.625 & \second{0.908} & \second{0.975} & \second{0.342} & \second{0.472} & \second{0.778} & 0.517 & \best{0.867} & 0.883 & \best{0.422} & \second{0.494} & 0.694 \\
\tabours \quad + runoff & \best{0.725} & \best{0.950} & \best{0.983} & \best{0.374} & \best{0.503} & \best{0.831} & \second{0.600} & \second{0.833} & \best{0.933} & \second{0.409} & \best{0.512} & \best{0.736} \\
\midrule
\tabgroup{13}{GPT-5.6 Luna} \\
OpenAI-small + rerank & 0.592 & 0.742 & 0.783 & 0.286 & 0.321 & 0.673 & 0.517 & 0.750 & 0.783 & 0.350 & 0.400 & 0.628 \\
OpenAI-large + rerank & \best{0.692} & 0.792 & 0.825 & 0.312 & 0.361 & 0.747 & \best{0.583} & \best{0.833} & 0.883 & 0.404 & 0.460 & \best{0.714} \\
Qwen3 + rerank        & \second{0.683} & 0.817 & 0.842 & 0.337 & 0.385 & 0.756 & \second{0.567} & \second{0.817} & 0.883 & 0.407 & 0.468 & 0.688 \\
\tabours Self-advertisement & 0.658 & \best{0.917} & \second{0.950} & \second{0.346} & \second{0.446} & \second{0.787} & 0.550 & \second{0.817} & \second{0.917} & \second{0.414} & \best{0.554} & 0.693 \\
\tabours \quad + runoff & \second{0.683} & \second{0.908} & \best{0.958} & \best{0.352} & \best{0.461} & \best{0.800} & \second{0.567} & \second{0.817} & \best{0.933} & \best{0.417} & \second{0.538} & \second{0.703} \\
\midrule
\tabgroup{13}{DeepSeek-V4.1-Flash} \\
OpenAI-small + rerank & 0.550 & 0.750 & 0.775 & 0.262 & 0.305 & 0.643 & \second{0.550} & 0.767 & 0.767 & 0.354 & 0.388 & 0.640 \\
OpenAI-large + rerank & 0.567 & 0.775 & 0.825 & 0.293 & 0.367 & 0.673 & \second{0.550} & \best{0.833} & 0.850 & 0.383 & 0.446 & \second{0.685} \\
Qwen3 + rerank        & 0.575 & 0.783 & 0.842 & \second{0.307} & 0.371 & 0.687 & \best{0.617} & 0.783 & 0.850 & 0.388 & 0.449 & \best{0.705} \\
\tabours Self-advertisement & \second{0.600} & \second{0.808} & \best{0.900} & 0.293 & \second{0.393} & \second{0.719} & 0.517 & \second{0.817} & \second{0.883} & \second{0.395} & \second{0.493} & 0.672 \\
\tabours \quad + runoff & \best{0.608} & \best{0.842} & \second{0.892} & \best{0.308} & \best{0.399} & \best{0.735} & 0.517 & \best{0.833} & \best{0.917} & \best{0.412} & \best{0.523} & 0.684 \\
\bottomrule
\end{tabular}}
\end{table}

\textbf{Main Results.} Self-advertisement gives the broadest shortlist coverage (Table~\ref{tab:method-selection}). Lexical matching reaches at most 0.525 Putnam hit@5, dense embeddings 0.775, and LLM reranking of an embedding pool 0.858. Self-advertisement attains the highest recall@5 in all six backbone--benchmark settings, and on Putnam it also leads at hit@3, hit@5, and MRR under each backbone. With GPT-5.6 Luna, the prover's backbone, it reaches 0.950 Putnam hit@5 against 0.842 for the strongest reranker, and 0.917 IMO ProofBench hit@5 against 0.883, a difference of two of the 60 problems. Its lead grows with the backbone while the rerankers plateau: from DeepSeek-V4.1-Flash to GPT-5.6 Luna to GPT-5.6 Terra, its Putnam hit@5 rises from 0.900 to 0.950 to 0.975, and the best reranker stays at 0.842, 0.842, and 0.858. The lead under DeepSeek-V4.1-Flash shows that it does not depend on the model family that drafted the contracts and labels. On IMO ProofBench, whose problems come from different competitions than the Putnam-derived library, self-advertisement keeps its recall@5 lead, at 0.494, 0.554, and 0.493 under GPT-5.6 Terra, GPT-5.6 Luna, and DeepSeek-V4.1-Flash, against at most 0.469, 0.468, and 0.449 for the rerankers.

Rerankers remain more precise at rank one in most settings; under GPT-5.6 Luna, the best reaches 0.692 Putnam hit@1 against 0.658 for self-advertisement. The runoff targets this gap: with the same backbone, it re-ranks self-advertisement's top ten by comparing the leading proposals. Under GPT-5.6 Terra, it raises Putnam hit@1 from 0.625 to 0.725, above the strongest reranker's 0.675, and IMO ProofBench hit@1 from 0.517 to 0.600, with the best MRR on both benchmarks (0.831 and 0.736). Under DeepSeek-V4.1-Flash, it gives the best Putnam hit@1 and MRR (0.608 and 0.735), and under GPT-5.6 Luna it changes each hit metric by at most three problems.

\begin{wraptable}{R}{0.553\linewidth}
\wraplift
\centering
\researchtablesetup
\caption{\textbf{Downstream proof completion, a separate stage from shortlist coverage (Table~\ref{tab:method-selection}).} Proof success (\%) with GPT-5.6 Luna under ten Lean compilations per problem, except pass@20, which draws 20 samples and so has twice the budget; Base is one attempt without retrieval, and the contract arm adds the top-ranked contract to AxProverBase's initial prompt~\citep{anonymous2026library}. Green marks the best and amber the second-best value in each column.}
\label{tab:method-proof}
\fitresearchtable{\begin{tabular}{lccccccc}
\toprule
\rowcolor{TabHead}
 & \multicolumn{4}{c}{\hdr{Putnam 2015--2025}} & \multicolumn{3}{c}{\hdr{IMO ProofBench}} \\
\hdrgapfill\cmidrule(lr){2-5}\cmidrule(lr){6-8}
\rowcolor{TabHead}
\multirow{-2}{*}{\hdr{Method}} & \hdr{ALG} & \hdr{ANA} & \hdr{DISC} & \hdr{All} & \hdr{Basic} & \hdr{Adv.} & \hdr{All} \\
\midrule
Base         & 0.0 & 0.0 & 0.0 & 0.0 & 3.3 & 0.0 & 1.7 \\
pass@20      & 2.3 & 0.0 & 0.0 & 0.8 & 3.3 & 0.0 & 1.7 \\
APOLLO       & \second{4.5} & 0.0 & 2.2 & 2.5 & 13.3 & 0.0 & 6.7 \\
AxProverBase & \best{9.1} & \second{3.3} & \second{4.3} & \second{5.8} & \second{20.0} & 0.0 & \second{10.0} \\
\midrule
\tabours + self-advertised contract & \best{9.1} & \best{10.0} & \best{17.4} & \best{12.5} & \best{26.7} & \best{3.3} & \best{15.0} \\
\bottomrule
\end{tabular}}
\end{wraptable}

\textbf{Downstream Proving.} The selected contract also helps the prover (Table~\ref{tab:method-proof}). AxProverBase is an Ax-Prover whole-proof repair loop driven by GPT-5.6 Luna with ten Lean compilations per problem~\citep{requena2026minimalagentautomatedtheorem}, and the contract arm adds the execution side of the top-ranked contract to its initial prompt. The contract more than doubles AxProverBase's success on Putnam, from 5.8\% to 12.5\%, and raises it from 10.0\% to 15.0\% on IMO ProofBench. On Putnam the gains come from analysis (3.3\% to 10.0\%) and discrete problems (4.3\% to 17.4\%), while algebra stays at 9.1\%; on IMO ProofBench the contract adds Basic problems (20.0\% to 26.7\%) and the one Advanced problem that any arm solves. Base and pass@20 solve almost no problems, although pass@20 has twice the budget, and APOLLO~\citep{ospanov2026apollo} trails both AxProverBase arms. Since the two arms share the prover, its backbone, and its budget, the gain reflects selection and contract content together.

\textbf{Applicability Beyond Similarity.} The coverage gains of Table~\ref{tab:method-selection} come largely from methods that similarity overlooks. We place each problem by the rank of its highest-ranked annotated method under self-advertisement with GPT-5.6 Luna and by that method's rank under BM25 and each embedding retriever (Figure~\ref{fig:method-ranks} in Appendix~\ref{app:experiments-extra}). Under all four retrievers, annotated methods that similarity ranks outside its top five enter the top five of self-advertisement, and many of them lie outside the top ten that the rerankers receive. This is the candidate-pool ceiling of Proposition~\ref{prop:coverage}(ii) at work: a reranker reorders the pool that similarity retrieved, so no reranking backbone recovers a method that the pool omits. Self-advertisement judges all 82 contracts, and each method it brings into the shortlist arrives with a target, an action, and required conditions that can be checked against the goal.

We provide the full method of self-advertised method selection, its overview figure, the theory behind the commitment gate and the runoff, and the experimental configuration in Appendix~\ref{sec:method-mechanism}. A Method Contract stays fixed once built, whereas a Knowledge Primitive's source model changes as the agent studies its source.

\wrapguard
\subsection{Knowledge Primitives for Knowledge-Intensive Tasks}
\label{sec:knowledge-experiments}

When many tasks draw on one persistent, authoritative source, such as a document collection, a code repository, or API documentation, the last mile is understanding that source: how its knowledge is structured, interpreted, and applied, beyond the passages that one task retrieves. Retrieval brings such passages into each task~\citep{lewis2020rag,guu2020retrieval}, structured indexes organize the source's content~\citep{sarthi2024raptor,edge2024local,gutierrez2025rag}, and agent memories keep experience from earlier tasks~\citep{shinn2023reflexion,zhao2024expel,wang2024awm}, yet repeated use of a source is still treated as repeated access. Knowledge Primitives turn this repeated use into \emph{source learning}: the resource of a Knowledge Primitive is a persistent \emph{source model} that holds reusable \emph{source-specific competence}, while the source itself remains the authority for factual content. SourceLearn~\citep{jin2026sourcelearn} develops this source model in two ways. Before future tasks are known, Self-Directed Source Learning studies what the source model does not yet explain well, and Task-Guided Source Learning uses guidance tasks to expose local gaps and recurring needs in how the source is represented. We provide the full framework of SourceLearn in Appendix~\ref{sec:knowledge-mechanism}, together with its overview figure, the benchmark and baseline settings, and the comparison at a matched context budget.

\textbf{Setup.}
We evaluate SourceLearn on question answering over document collections, stories, and code repositories with MultiDoc2Dial~\citep{feng2021multidoc2dial}, NarrativeQA~\citep{kovcisky2018narrativeqa}, and SWE-QA~\citep{peng2026swe}; tool use with APIBench~\citep{patil2023gorilla}; and interactive tasks with AppWorld~\citep{trivedi2024appworld}. Each document collection, story, repository, or API library is one source. Guidance and test tasks are disjoint, with three fixed 30\%/70\% splits and AppWorld's official training and test-normal splits. GPT-5.6 Luna judges QA answers against the references, and APIBench and AppWorld use their official evaluators. The baselines are task-local retrieval with Hybrid RAG, persistent source representations with RAPTOR~\citep{sarthi2024raptor} and HippoRAG~2~\citep{gutierrez2025rag}, and experience-based memory with AWM~\citep{wang2024awm}, which learns workflows from the same guidance tasks. Each configuration uses a single backbone, GPT-5.6 Luna~\citep{s06_openai2026gpt56}, GPT-OSS-120B~\citep{agarwal2025gpt}, or DeepSeek-V4.1-Flash~\citep{s06_xu2026deepseek}, for the LLM operations of all methods, which share the embedding model, the answer or agent prompt, and, on AppWorld, the ReAct-style code agent~\citep{yao2023react}. SourceLearn adds the activated source model to the Hybrid RAG evidence $R(D,q)$, which AWM also receives, so the comparison with Hybrid RAG isolates the contribution of the source model.

\begin{table}[!htbp]
\centering
\researchtablesetup
\caption{\textbf{SourceLearn against retrieval and memory baselines on five benchmarks and three backbones.} Accuracy or task success (\%) for document QA, code QA, tool use, and interactive environments, with the change from Hybrid RAG in pp in parentheses~\citep{jin2026sourcelearn}. Green marks the best and amber the second-best value in each column within each backbone.}
\label{tab:knowledge-main}
\fitresearchtable{\begin{tabular}{llccccc}
\toprule
\rowcolor{TabHead}
 & & \multicolumn{2}{c}{\hdr{Document QA}} & \hdr{Code QA} & \hdr{Tool Use} & \hdr{Interactive Env.} \\
\hdrgapfill\cmidrule(lr){3-4}\cmidrule(lr){5-5}\cmidrule(lr){6-6}\cmidrule(lr){7-7}
\rowcolor{TabHead}
\multirow{-2}{*}{\hdr{Model}} & \multirow{-2}{*}{\hdr{Method}} & \hdr{MultiDoc2Dial} & \hdr{NarrativeQA} & \hdr{SWE-QA} & \hdr{APIBench} & \hdr{AppWorld} \\
\midrule
 & Hybrid RAG & 64.9 & 66.0 & 56.9 & 47.8 & 72.0 \\
 & RAPTOR & \second{71.2}\dpp{+6.3} & \second{68.1}\dpp{+2.1} & 50.8\dpp{-6.1} & 61.9\dpp{+14.1} & 75.6\dpp{+3.6} \\
 & HippoRAG 2 & 66.3\dpp{+1.4} & 66.6\dpp{+0.6} & \second{62.7}\dpp{+5.8} & 55.2\dpp{+7.4} & \second{77.4}\dpp{+5.4} \\
 & AWM & 65.1\dpp{+0.2} & 62.1\dpp{-3.9} & 56.7\dpp{-0.2} & \second{62.8}\dpp{+15.0} & 66.7\dpp{-5.3} \\
\tabours \multirow{-5}{*}{GPT-5.6 Luna} & SourceLearn & \best{77.6}\dpp{+12.7} & \best{80.2}\dpp{+14.2} & \best{69.5}\dpp{+12.6} & \best{70.4}\dpp{+22.6} & \best{81.5}\dpp{+9.5} \\
\midrule
 & Hybrid RAG & 64.5 & 45.6 & 52.5 & 60.3 & \second{30.4} \\
 & RAPTOR & 61.5\dpp{-3.0} & 40.4\dpp{-5.2} & 32.8\dpp{-19.7} & 54.7\dpp{-5.6} & 26.8\dpp{-3.6} \\
 & HippoRAG 2 & \second{66.7}\dpp{+2.2} & \second{45.7}\dpp{+0.1} & \best{56.7}\dpp{+4.2} & 56.0\dpp{-4.3} & \best{33.9}\dpp{+3.5} \\
 & AWM & 64.1\dpp{-0.4} & 43.2\dpp{-2.4} & 51.5\dpp{-1.0} & \second{63.2}\dpp{+2.9} & 28.6\dpp{-1.8} \\
\tabours \multirow{-5}{*}{GPT-OSS-120B} & SourceLearn & \best{68.7}\dpp{+4.2} & \best{56.7}\dpp{+11.1} & \second{53.4}\dpp{+0.9} & \best{71.8}\dpp{+11.5} & 27.4\dpp{-3.0} \\
\midrule
 & Hybrid RAG & 63.3 & \second{64.2} & 45.9 & 67.1 & 83.9 \\
 & RAPTOR & \second{73.2}\dpp{+9.9} & 60.4\dpp{-3.8} & 36.8\dpp{-9.1} & 69.3\dpp{+2.2} & \second{88.7}\dpp{+4.8} \\
 & HippoRAG 2 & 67.3\dpp{+4.0} & 62.0\dpp{-2.2} & \second{55.7}\dpp{+9.8} & 69.6\dpp{+2.5} & 86.9\dpp{+3.0} \\
 & AWM & 63.7\dpp{+0.4} & 61.4\dpp{-2.8} & 47.4\dpp{+1.5} & \second{70.8}\dpp{+3.7} & 83.9\dpp{0.0} \\
\tabours \multirow{-5}{*}{DeepSeek-V4.1-Flash} & SourceLearn & \best{81.3}\dpp{+18.0} & \best{76.2}\dpp{+12.0} & \best{61.6}\dpp{+15.7} & \best{82.2}\dpp{+15.1} & \best{89.9}\dpp{+6.0} \\
\bottomrule
\end{tabular}}
\end{table}

\textbf{Main Results.}
Across the 15 settings of Table~\ref{tab:knowledge-main}, SourceLearn is the best method in 13 and the second best in one more. It improves on Hybrid RAG in 14 settings, and its average gain over the five benchmarks is 14.3~pp with GPT-5.6 Luna, 4.9~pp with GPT-OSS-120B, and 13.4~pp with DeepSeek-V4.1-Flash. The gains span document, code, and API sources and exceed 10~pp on four of the five benchmarks with GPT-5.6 Luna and with DeepSeek-V4.1-Flash; with GPT-5.6 Luna, MultiDoc2Dial rises from 64.9\% to 77.6\% and APIBench from 47.8\% to 70.4\%. The two settings in which it is not the best both use GPT-OSS-120B: HippoRAG~2 leads on SWE-QA (56.7\% against 53.4\%), and on AppWorld SourceLearn falls below Hybrid RAG (27.4\% against 30.4\%). The baselines help unevenly. RAPTOR's summary tree falls below Hybrid RAG on SWE-QA under all three backbones, where HippoRAG~2 is the strongest baseline, and AWM, the strongest baseline on APIBench, moves question-answering accuracy by at most 3.9~pp. Learning the source transfers across benchmarks more evenly than structuring its content or retaining task experience. SourceLearn also leads on three of the four sources tested with GPT-5.6 Luna at a matched context budget, where each retrieval baseline fills SourceLearn's per-question budget by reading further down its own ranking; HippoRAG~2 leads on Sphinx (73.5\% against 72.5\%; Table~\ref{tab:knowledge-budget}).

\textbf{Analysis.}
Both learning mechanisms contribute and complement each other. We remove each in turn with GPT-5.6 Luna on two sources of each QA benchmark, with the splits and task-time source access of the main results (Figure~\ref{fig:knowledge-ablation}). The full model $M_T$ is the most accurate on all three benchmarks, omitting Self-Directed or Task-Guided Source Learning lowers accuracy, and the initial model $M_0$ is the lowest. Within Task-Guided Source Learning, removing failure-guided local refinement or cross-task representation learning lowers accuracy on all three benchmarks, and removing both, which recovers the self-directed model $M_S$, lowers it further. The two pathways thus carry different information from the same guidance tasks: concrete failures repair local gaps, while recurring representation lessons change how the whole model represents the source. Learning also does more than compress the source (Figure~\ref{fig:knowledge-reconstruction}). From $M_0$ to $M_T$, the model shifts from isolated look-up content toward rules and procedures, and the share of units with an explicit applicability condition rises from roughly 30\% to 65--82\% on the four MultiDoc2Dial sources. The share of the claims required by test questions that the model represents rises from 23.2\% to 40.0\%, and questions with all of their required claims represented are answered correctly 86.6\% of the time, against 67.5\% when none are.

The Tool, Code, Math, and Knowledge Primitives each own a resource that a single model applies; Agent Primitives own a recurring multi-agent computation pattern, whose agents run on one LLM and exchange KV caches.

\wrapguard
\subsection{Agent Primitives for Multi-Agent Systems}
\label{sec:agent-experiments}

Where several agents share one problem, the last mile lies in how they combine their work. Multi-agent systems (MAS) let LLM agents critique, vote, and plan together, but they are usually designed task by task, with handcrafted roles and interaction prompts, so their complexity grows with the task, and no reusable abstraction amortizes it. Their agents also communicate in natural language, which degrades over long, multi-stage interactions. In two stress tests with Qwen3-8B, an instruction injected at the midpoint of a long context is followed in 15.6\% of cases through text and in 73.3\% through KV caches, and after ten unrelated sentences are injected into solved GSM8K traces, 47\% of the problems stay solved through text and 93\% through KV caches. Agent Primitives~\citep{jin2026agentprimitives} address both problems with three reusable latent building blocks, Review, Voting and Selection, and Planning and Execution, whose internal agents communicate through KV caches and which an LLM Organizer composes into a primitive-based MAS for each query.

\textbf{Setup.}
The evaluation covers mathematical problem solving on AIME24, AIME25, MATH, and GSM8K~\citep{maxwelljia_aime2024,mathai_aime2025,hendrycks2021measuring,cobbe2021training}, code generation on HumanEval+ and MBPP+, which extend the tests of HumanEval and MBPP~\citep{liu2023your,chen2021evaluatinglargelanguagemodels,austin2021program}, and question answering on MedQA and GPQA-Diamond~\citep{jin2021disease,rein2024gpqa}. The baselines are a single agent, a text-based MAS (TextMAS), LatentMAS~\citep{zou2025latent} in its sequential configuration, and each primitive applied alone to all queries; the last three use four agents each, whereas the Organizer, GPT-5.2 by default, sets the number of agents per query. KV cache communication assumes that the communicating agents share parameters and positional encoding, so all agents of a run use one worker backbone: Qwen3-8B~\citep{qwen3technicalreport}, DeepSeek-R1-Distill-Qwen-32B, or DeepSeek-R1-Distill-Llama-70B~\citep{deepseekai2025deepseekr1incentivizingreasoningcapability}, abbreviated R1/Qwen-32B and R1/Llama-70B.

\begin{table}[!htbp]
\centering
\researchtablesetup
\caption{\textbf{Primitives-based MAS against single agents, text and latent MAS, and each Agent Primitive used alone.} Accuracy (\%) on eight math, code and QA benchmarks for three worker backbones, where R1/ marks a DeepSeek-R1-Distill model; the Organizer sets the number of agents of a composed MAS per query, while a single primitive uses a fixed number~\citep{jin2026agentprimitives}. Green marks the best and amber the second-best value in each column within each backbone.}
\label{tab:agent-main}
\fitresearchtable{\begin{tabular}{llccccccccc}
\toprule
\rowcolor{TabHead}
 & & \multicolumn{4}{c}{\hdr{Math Problem Solving}} & \multicolumn{2}{c}{\hdr{Code Generation}} & \multicolumn{2}{c}{\hdr{Q\&A}} & \\
\hdrgapfill\cmidrule(lr){3-6}\cmidrule(lr){7-8}\cmidrule(lr){9-10}
\rowcolor{TabHead}
\multirow{-2}{*}{\hdr{Model}} & \multirow{-2}{*}{\hdr{Method}} & \hdr{AIME25} & \hdr{AIME24} & \hdr{MATH} & \hdr{GSM8K} & \hdr{HumanEval+} & \hdr{MBPP+} & \hdr{MedQA} & \makecell[b]{\hdr{GPQA-}\\\hdr{Diamond}} & \multirow{-2}{*}{\hdr{Avg.}} \\
\midrule
 & Single & 46.7 & 50.0 & 60.8 & 81.1 & 74.4 & 64.8 & 53.0 & 39.9 & 58.8 \\
 & TextMAS & 53.3 & 53.3 & 61.4 & 92.3 & 80.5 & 69.5 & 75.0 & 43.4 & 66.1 \\
 & LatentMAS & 53.3 & 56.7 & \second{62.6} & \second{93.8} & 80.5 & \second{74.6} & \second{75.3} & 45.5 & 67.8 \\
 & Review & 60.0 & 63.3 & 61.0 & 93.2 & 78.6 & 70.6 & 64.2 & 48.9 & 67.5 \\
 & Voting & \second{66.7} & \second{70.0} & 61.4 & 91.8 & \second{81.0} & 74.3 & 70.3 & \second{55.0} & \second{71.3} \\
 & Planning & \second{66.7} & 63.3 & 60.8 & 93.2 & 78.6 & \best{75.9} & 67.0 & 51.0 & 69.6 \\
\tabours \multirow{-7}{*}{Qwen3-8B} & Primitives-based MAS & \best{73.3} & \best{76.7} & \best{63.7} & \best{94.2} & \best{82.3} & \best{75.9} & \best{76.7} & \best{59.6} & \best{75.3} \\
\midrule
 & Single & 53.3 & \best{73.3} & 63.4 & 93.8 & 80.5 & 73.8 & 68.0 & 62.1 & 71.0 \\
 & TextMAS & 53.3 & \best{73.3} & 68.0 & 93.8 & 82.3 & \second{74.6} & 79.6 & 62.6 & 73.4 \\
 & LatentMAS & \second{56.7} & \best{73.3} & \second{78.2} & \best{95.2} & 83.5 & \best{75.7} & \second{81.2} & \second{63.6} & \second{75.9} \\
 & Review & 53.3 & \second{70.0} & 71.3 & 94.6 & 81.1 & 73.8 & 73.0 & 62.1 & 72.4 \\
 & Voting & \second{56.7} & \second{70.0} & 74.7 & \second{95.0} & \best{86.6} & \second{74.6} & 79.3 & 63.1 & 75.0 \\
 & Planning & 53.3 & 66.7 & 74.6 & 93.8 & \second{85.3} & 74.3 & 75.0 & 62.6 & 73.2 \\
\tabours \multirow{-7}{*}{R1/Qwen-32B} & Primitives-based MAS & \best{63.3} & \best{73.3} & \best{79.8} & \second{95.0} & \best{86.6} & \best{75.7} & \best{82.7} & \best{64.6} & \best{77.6} \\
\midrule
 & Single & 50.0 & 70.0 & 69.6 & 92.4 & 82.3 & 66.4 & 64.5 & 65.2 & 70.1 \\
 & TextMAS & \second{53.3} & 70.0 & 72.8 & \second{93.2} & 82.3 & \second{68.8} & \second{77.8} & \second{65.7} & 73.0 \\
 & LatentMAS & 40.0 & 43.3 & \second{78.6} & 78.6 & 73.2 & 55.6 & 68.6 & 41.9 & 60.0 \\
 & Review & 50.0 & 70.0 & 75.9 & 91.8 & 82.3 & 67.2 & 72.9 & 65.2 & 71.9 \\
 & Voting & \best{56.7} & \second{73.3} & 77.2 & \best{93.8} & \second{82.9} & \second{68.8} & \second{77.8} & \second{65.7} & \second{74.5} \\
 & Planning & 50.0 & 63.3 & 75.6 & 92.4 & 82.3 & 66.7 & 74.0 & 65.2 & 71.2 \\
\tabours \multirow{-7}{*}{R1/Llama-70B} & Primitives-based MAS & \best{56.7} & \best{76.7} & \best{79.3} & \best{93.8} & \best{85.3} & \best{70.6} & \best{81.9} & \best{66.7} & \best{76.4} \\
\bottomrule
\end{tabular}}
\end{table}

\textbf{Main Results.}
Composing primitives gives the highest mean accuracy with each backbone (Table~\ref{tab:agent-main}). With Qwen3-8B, the primitive-based MAS reaches 75.3\%, against 58.8\% for a single agent, 66.1\% for TextMAS, 67.8\% for LatentMAS, and 71.3\% for Voting and Selection, the strongest single primitive. With R1/Qwen-32B and R1/Llama-70B, it improves on a single agent by 6.6 and 6.3~pp. Across the three backbones, it is best or tied for best on each benchmark except GSM8K with R1/Qwen-32B, where LatentMAS reaches 95.2\% against 95.0\%. Each primitive used alone also improves on the mean accuracy of a single agent, and composing them adds 4.0, 2.6, and 1.9~pp on the three backbones over the strongest one, Voting and Selection. TextMAS improves on a single agent by 2.4 to 7.3~pp. LatentMAS is the strongest baseline with R1/Qwen-32B, at 75.9\%, but falls to 60.0\% with R1/Llama-70B, 10.1~pp below a single agent, which the study attributes to its aggressive chunking of latent reasoning steps; the more conservative KV cache communication of Agent Primitives improves on a single agent with each backbone.

\begin{researchtable}
\centering
\researchtablesetup
\caption{\textbf{Primitives-based MAS against independently designed multi-agent methods and two single-model prompting baselines.} Accuracy (\%) in the Agent Primitives study's unified Llama-3-70B-Instruct setting, whose backbone differs from those of Table~\ref{tab:agent-main}, with the change from its own single-agent reference in pp in parentheses; Chain-of-Thought and Self-Consistency are the prompting baselines~\citep{jin2026agentprimitives}. Green marks the best and amber the second-best value in each column.}
\label{tab:agent-mas}
\fitresearchtable{\begin{tabular}{lcccc}
\toprule
\rowcolor{TabHead}
\hdr{Method} & \hdr{MATH} & \hdr{GSM8K} & \hdr{HumanEval+} & \hdr{GPQA} \\
\midrule
Single & 50.6 & 92.4 & 75.8 & 36.7 \\
Chain-of-Thought & 53.2\dpp{+2.6} & 92.8\dpp{+0.4} & 77.0\dpp{+1.2} & 35.3\dpp{-1.4} \\
Self-Consistency & 61.6\dpp{+11.0} & \best{95.0}\dpp{+2.6} & 75.8\dpp{0.0} & 37.2\dpp{+0.5} \\
LLM-Debate & 61.4\dpp{+10.8} & 91.6\dpp{-0.8} & 74.5\dpp{-1.3} & 34.4\dpp{-2.3} \\
Self-Refine & 58.5\dpp{+7.9} & 90.8\dpp{-1.6} & 62.7\dpp{-13.1} & 38.3\dpp{+1.6} \\
Quality-Diversity & 60.5\dpp{+9.9} & 93.0\dpp{+0.6} & 70.2\dpp{-5.6} & 33.6\dpp{-3.1} \\
SPP & 51.7\dpp{+1.1} & 92.8\dpp{+0.4} & 73.3\dpp{-2.5} & 35.1\dpp{-1.6} \\
AgentVerse & 55.6\dpp{+5.0} & 93.4\dpp{+1.0} & 73.9\dpp{-1.9} & \second{40.2}\dpp{+3.5} \\
GPTSwarm & 55.4\dpp{+4.8} & 93.2\dpp{+0.8} & 73.9\dpp{-1.9} & 36.5\dpp{-0.2} \\
DyLAN & 59.6\dpp{+9.0} & 91.2\dpp{-1.2} & 75.8\dpp{0.0} & 36.0\dpp{-0.7} \\
MAS-GPT & \second{68.7}\dpp{+18.1} & 93.4\dpp{+1.0} & \second{78.9}\dpp{+3.1} & 37.6\dpp{+0.9} \\
\midrule
\tabours Primitives-based MAS & \best{72.4}\dpp{+21.8} & \second{93.8}\dpp{+1.4} & \best{82.3}\dpp{+6.5} & \best{53.2}\dpp{+16.5} \\
\bottomrule
\end{tabular}}
\end{researchtable}

\textbf{Existing MAS Methods.}
Table~\ref{tab:agent-mas} compares the primitive-based MAS on Llama-3-70B-Instruct~\citep{grattafiori2024llama} with eight MAS methods, LLM-Debate~\citep{du2023improving}, Self-Refine~\citep{madaan2023self}, Quality-Diversity~\citep{lu2024intelligent}, SPP~\citep{wang2024unleashing}, AgentVerse~\citep{chen2023agentverse}, GPTSwarm~\citep{zhuge2024gptswarm}, DyLAN~\citep{liu2024a}, and MAS-GPT~\citep{ye2025mas}, and with two prompting baselines, Chain-of-Thought~\citep{wei2022chain} and Self-Consistency~\citep{wang2022self}. It is the most accurate method on MATH, HumanEval+, and GPQA, with 72.4\%, 82.3\%, and 53.2\%. On MATH and HumanEval+, it leads MAS-GPT, the strongest prior method, by 3.7 and 3.4~pp, and on GPQA the other methods reach 33.6 to 40.2\%. Self-Consistency leads on GSM8K, with 95.0\% against 93.8\%. These gains come at moderate cost. Across the study's five worker backbones, the primitive-based MAS reduces token usage and inference latency by about 3--4$\times$ relative to TextMAS, with 1.3--1.6$\times$ overhead relative to single-agent inference, and on Llama-3-70B-Instruct it uses fewer output tokens than each MAS baseline, the Organizer's 200--400 tokens per query included.

\textbf{Analysis.}
The ablations attribute the gain to per-query composition and to KV cache communication. Replacing the Organizer's choice with a random structure from the Knowledge Pool lowers accuracy by 4.9 to 6.6~pp with Qwen3-8B and by up to 12.4~pp with R1/Llama-70B, and removing the pool costs up to 11.9~pp. Replacing GPT-5.2 with Claude-4~\citep{anthropic2025claude} as the Organizer keeps accuracy within 1.8~pp, so the main benefit comes from problem-aware structure selection and not from a particular Organizer model (Tables~\ref{tab:agent-organizer} and~\ref{tab:agent-pool}). KV cache communication in turn depends on re-indexing the sender's keys for RoPE. Without the re-indexing, the primitive-based MAS on R1/Llama-70B falls from 85.3\% to 31.1\% on HumanEval+ and from 81.9\% to 36.6\% on MedQA, and each primitive alone falls to between 16.7\% and 34.7\%, so the loss arises within the primitives; with Qwen3-8B, the largest drop is 13.3~pp (Table~\ref{tab:agent-rope}). With text communication, the three primitives reach 50.0, 56.7, and 53.3\% on Qwen3-8B AIME25, above the 46.7\% of a single agent, and KV caches raise them to 60.0, 66.7, and 66.7\%, so the primitive abstraction and KV cache communication each contribute.

We provide the full framework of Agent Primitives, its overview figure, and additional results in Appendix~\ref{sec:agent-mechanism}. The agents of an Agent Primitive share one LLM inside the primitive; when the units that must coordinate are the components of a repository, Dev-Primitives make each component a primitive of its own.

\wrapguard
\subsection{Dev-Primitives for Software Engineering}
\label{sec:dev-experiments}
\begin{table}[!htbp]
\centering
\researchtablesetup
\caption{\textbf{HERMES against same-backbone harnesses on SWE-bench Verified.} Resolved rate (\%, mean \textpm{} standard error over 500 instances), cost per instance and latency for each harness and backbone, with $\Delta$ the resolved-rate change of HERMES in pp. HERMES is run once per backbone, so a backbone compared with several harnesses repeats its HERMES results, and Qwen3-8B is run with HERMES only~\citep{jin2026devprimitives}. Green marks the best and amber the second-best value in each Resolved, Cost/Test, and Latency column. Lower is better for cost and latency.\fillcap}
\label{tab:dev-swe}
\fitresearchtable{\begin{tabular}{lccccccc}
\toprule
\rowcolor{TabHead}
 & \multicolumn{3}{c}{\hdr{Baseline Results}} & \multicolumn{3}{c}{\hdr{HERMES Results}} & \\
\hdrgapfill\cmidrule(lr){2-4}\cmidrule(lr){5-7}
\rowcolor{TabHead}
\multirow{-2}{*}{\hdr{Backbone}} & \hdr{Resolved (\%)} & \hdr{Cost/Test (\$)} & \hdr{Latency (s)} & \hdr{Resolved (\%)} & \hdr{Cost/Test (\$)} & \hdr{Latency (s)} & \multirow{-2}{*}{\hdr{$\Delta$ Res.}} \\
\midrule
\tabgroup{8}{Baseline harness: mini-SWE-agent} \\
GPT-5.6 Sol                & \second{96.20\,\textpm\,0.86} & 1.15      & 182.37 & \best{97.00\,\textpm\,0.76} & 1.80        & \second{205.34} & +0.80 \\
GPT-5.6 Terra              & 95.40\,\textpm\,0.94        & 0.40        & 180.00 & \second{96.20\,\textpm\,0.86} & 0.98      & \best{196.71} & +0.80 \\
GPT-5.6 Luna               & 93.00\,\textpm\,1.14        & \best{0.04} & 201.07 & 95.60\,\textpm\,0.92        & \best{0.10} & 218.46        & +2.60 \\
Claude Fable 5             & 95.00\,\textpm\,0.98        & 2.05        & 356.18 & 96.00\,\textpm\,0.88        & 4.50        & 381.27        & +1.00 \\
Claude Opus 5              & \best{97.00\,\textpm\,0.76} & 1.29        & 576.99 & \best{97.00\,\textpm\,0.76} & 2.25        & 472.18        & 0.00 \\
Claude Sonnet 5            & 79.60\,\textpm\,1.80        & 1.49        & 962.37 & 85.80\,\textpm\,1.56        & 2.10        & 824.63        & +6.20 \\
DeepSeek-V4                & 77.40\,\textpm\,1.87        & 0.44        & 634.61 & 82.80\,\textpm\,1.69        & 0.83        & 418.52        & +5.40 \\
GPT-5.5                    & 82.60\,\textpm\,1.70        & 1.36        & 426.43 & 85.60\,\textpm\,1.57        & 2.45        & 398.74        & +3.00 \\
GPT-5.4-mini               & 73.00\,\textpm\,1.99        & 0.51        & 326.46 & 82.40\,\textpm\,1.70        & 0.83        & 347.82        & +9.40 \\
GPT-5-mini                 & 60.80\,\textpm\,2.19        & \second{0.05} & 187.13 & 80.20\,\textpm\,1.78        & 0.21        & 213.55        & +19.40 \\
Claude Opus 4.8            & 88.60\,\textpm\,1.42        & 1.92        & 566.95 & 91.00\,\textpm\,1.28        & 2.65        & 482.73        & +2.40 \\
Claude Opus 4.7            & 82.00\,\textpm\,1.72        & 2.42        & 441.99 & 86.00\,\textpm\,1.55        & 3.10        & 396.52        & +4.00 \\
Gemini-3.5-Flash           & 78.80\,\textpm\,1.83        & 0.95        & 254.13 & 84.00\,\textpm\,1.64        & 1.42        & 276.84        & +5.20 \\
Gemini-3.1-Pro-Preview     & 78.80\,\textpm\,1.83        & 0.78        & 312.26 & 83.40\,\textpm\,1.67        & 1.26        & 328.75        & +4.60 \\
GPT-5.4 (xhigh)            & 78.20\,\textpm\,1.85        & 0.80        & 307.12 & 83.80\,\textpm\,1.65        & 1.23        & 322.68        & +5.60 \\
Claude Opus 4.6 (Thinking) & 78.20\,\textpm\,1.85        & 1.22        & 350.76 & 84.40\,\textpm\,1.62        & 1.86        & 341.28        & +6.20 \\
GPT-5.3-Codex              & 78.00\,\textpm\,1.85        & 0.46        & 246.53 & 83.00\,\textpm\,1.68        & 0.79        & 267.39        & +5.00 \\
Claude Sonnet 4.6          & 77.40\,\textpm\,1.87        & 1.30        & 511.79 & 82.80\,\textpm\,1.69        & 1.92        & 438.64        & +5.40 \\
Claude Opus 4.5 (Thinking) & 76.40\,\textpm\,1.90        & 1.05        & 320.54 & 81.80\,\textpm\,1.73        & 1.64        & 336.28        & +5.40 \\
Gemini-3-Pro               & 76.40\,\textpm\,1.90        & 0.77        & 394.49 & 82.20\,\textpm\,1.71        & 1.31        & 381.73        & +5.80 \\
\midrule
\tabgroup{8}{Baseline harness: Claude Code} \\
Claude Opus 4.8 & 85.80\,\textpm\,1.56 & 0.67 & \second{135.40} & 91.00\,\textpm\,1.28 & 2.65 & 482.73 & +5.20 \\
\midrule
\tabgroup{8}{Baseline harness: Codex} \\
GPT-5.5 & 76.40\,\textpm\,1.90 & 0.65 & \best{108.54} & 85.60\,\textpm\,1.57 & 2.45 & 398.74 & +9.20 \\
\midrule
\tabgroup{8}{No baseline harness} \\
Qwen3-8B & -- & -- & -- & 80.60\,\textpm\,1.77 & \second{0.13} & 229.81 & -- \\
\bottomrule
\end{tabular}}
\end{table}

The last mile of software engineering runs through the repository. The program state that a change must respect is spread over interdependent sources, configurations, and tests, and each edit changes the repository that the next step reads, so a single agent repeatedly reconstructs this distributed state within one trajectory, whose growing history leads to context explosion and semantic drift. Dev-Primitives~\citep{jin2026devprimitives} let individual software components reason about their own responsibilities and communicate relevant information directly to one another, and HERMES instantiates them through dynamic activation, localized collaboration, environment execution, and diagnosis-driven revision.

\textbf{Setup.}
We evaluate HERMES on four benchmarks: SWE-bench Verified~\citep{jimenez2023swebench}, 500 human-validated GitHub issues; SWE Refactor Bench~\citep{hong2026swe}, 20 whole-repository migrations scored by a composite; Terminal-Bench~4.0~\citep{merrill2026terminal}, 66 terminal tasks run five times each; and DevOps-Gym~\citep{tang2026devops}, scored by the unweighted mean over its build and configuration, monitoring, issue-resolving, and test-generation stages. The baselines are the harnesses of published configurations, among them mini-SWE-agent~\citep{yang2024swe}, Claude Code~\citep{anthropic2025claudecode}, Codex~\citep{openai2025codex}, and OpenHands~\citep{wang2025openhands}, and HERMES runs on the backbone and reasoning effort of each, so that the comparison isolates the harness. The backbones include GPT-5.6 Sol, Terra, and Luna, Claude Opus 5 and Sonnet 5, DeepSeek-V4, and Qwen3-8B~\citep{qwen3technicalreport}, with medium reasoning effort unless stated otherwise and $B=3$ revision rounds.

\begin{table}[!htbp]
\centering
\researchtablesetup
\caption{\textbf{HERMES against same-backbone harnesses on Terminal-Bench 4.0.} Resolution (\%, mean \textpm{} standard deviation over five runs of the 66 tasks) with tokens and cost totalled over each complete evaluation, per model and reasoning effort, and $\Delta$ the change of HERMES in pp. Qwen3-8B is run with HERMES only~\citep{jin2026devprimitives}. Green marks the best and amber the second-best value in each Resolution, Tokens, and Cost column. Lower is better for tokens and cost.}
\label{tab:dev-terminal}
\fitresearchtable{\begin{tabular}{llccccccc}
\toprule
\rowcolor{TabHead}
 & & \multicolumn{3}{c}{\hdr{Baseline Results}} & \multicolumn{3}{c}{\hdr{HERMES Results}} & \\
\hdrgapfill\cmidrule(lr){3-5}\cmidrule(lr){6-8}
\rowcolor{TabHead}
\multirow{-2}{*}{\hdr{Model}} & \multirow{-2}{*}{\hdr{Effort}} & \hdr{Resolution (\%)} & \hdr{Tokens (B)} & \hdr{Cost (\$k)} & \hdr{Resolution (\%)} & \hdr{Tokens (B)} & \hdr{Cost (\$k)} & \multirow{-2}{*}{\hdr{$\Delta$ Res.}} \\
\midrule
\tabgroup{9}{Baseline harness: Codex} \\
 & medium & 33.0\,\textpm\,3.4 & \second{3.6} & 1.9   & 51.8\,\textpm\,2.2        & 6.8  & 3.4        & +18.8 \\
              \multirow{-2}{*}{GPT-5.6 Sol} & max    & 37.3\,\textpm\,3.8 & 4.4  & 2.5        & \best{55.5\,\textpm\,3.1} & 8.9  & 4.7        & +18.2 \\
\midrule
 & medium & 18.2\,\textpm\,3.0 & 4.4  & 1.2        & 44.5\,\textpm\,3.3        & 7.2  & 2.2        & +26.3 \\
              \multirow{-2}{*}{GPT-5.6 Terra} & max    & 21.5\,\textpm\,3.3 & 5.7  & 1.7        & 48.2\,\textpm\,2.5        & 9.1  & 2.9        & +26.7 \\
\midrule
 & medium & 14.5\,\textpm\,2.6 & 9.1  & \best{0.2} & 29.1\,\textpm\,2.9        & 13.2 & \best{0.4} & +14.6 \\
              \multirow{-2}{*}{GPT-5.6 Luna} & max    & 17.3\,\textpm\,2.8 & 11.6 & \second{0.3} & 33.3\,\textpm\,3.2      & 15.8 & \second{0.5} & +16.0 \\
\midrule
\tabgroup{9}{Baseline harness: Claude Code} \\
 & medium & \second{47.0\,\textpm\,3.2} & 5.2  & 4.8 & 48.5\,\textpm\,3.6 & 7.3  & 6.7  & +1.5 \\
                \multirow{-2}{*}{Claude Opus 5} & max    & \best{51.8\,\textpm\,3.4} & 6.5  & 6.0 & \second{53.0\,\textpm\,2.1} & 9.4  & 8.6  & +1.2 \\
\midrule
Claude Fable 5  & max    & 44.5\,\textpm\,3.8        & 3.8  & 7.3 & 49.7\,\textpm\,3.3 & \second{6.2} & 10.1 & +5.2 \\
\midrule
GLM-5.3         & max    & 41.8\,\textpm\,3.2        & 8.7  & 2.7 & 46.4\,\textpm\,2.8 & 11.5 & 3.9  & +4.6 \\
\midrule
Claude Opus 4.8 & max    & 23.6\,\textpm\,3.6        & 6.4  & 6.5 & 34.2\,\textpm\,3.8 & 9.0  & 8.4  & +10.6 \\
\midrule
 & medium & 10.6\,\textpm\,2.7        & 15.9 & 7.1 & 35.8\,\textpm\,2.3 & 13.4 & 5.8  & +25.2 \\
                \multirow{-2}{*}{Claude Sonnet 5} & max    & 12.4\,\textpm\,3.1        & 21.6 & 9.6 & 40.0\,\textpm\,3.5 & 16.8 & 7.4  & +27.6 \\
\midrule
\tabgroup{9}{Baseline harness: mini-SWE-agent} \\
DeepSeek-V4      & max  & 28.8\,\textpm\,3.4 & 7.8  & 0.7 & 39.4\,\textpm\,3.1 & 10.1 & 1.1 & +10.6 \\
\midrule
Gemini-3.8-Flash & high & 19.1\,\textpm\,3.4 & 17.2 & 1.8 & 30.3\,\textpm\,3.7 & 20.4 & 2.6 & +11.2 \\
\midrule
Gemini-3.7-Flash & high & 11.2\,\textpm\,2.4 & 11.1 & 1.3 & 24.2\,\textpm\,3.4 & 14.8 & 1.9 & +13.0 \\
\midrule
\tabgroup{9}{Baseline harness: Grok Build} \\
Grok 4.6 & high & 20.3\,\textpm\,3.1 & 4.0        & 3.6 & 31.5\,\textpm\,2.5 & \second{6.2} & 5.2 & +11.2 \\
\midrule
Grok 4.5 & high & 12.4\,\textpm\,2.6 & \best{3.4} & 2.1 & 25.5\,\textpm\,2.2 & \best{5.0} & 3.0 & +13.1 \\
\midrule
\tabgroup{9}{No baseline harness} \\
Qwen3-8B & -- & -- & -- & -- & 27.6\,\textpm\,2.7 & 14.3 & 0.7 & -- \\
\bottomrule
\end{tabular}}
\end{table}

\begin{table}[!htbp]
\centering
\researchtablesetup
\caption{\textbf{HERMES against same-backbone harnesses on the four DevOps-Gym stages.} Stage success rates (\%), with Avg.\ the unweighted mean over stages and $\Delta$ the change of HERMES in pp. DeepSeek-V4 uses maximum reasoning effort. HERMES is run once per backbone, so a backbone compared with several harnesses repeats its HERMES results, and Qwen3-8B is run with HERMES only~\citep{jin2026devprimitives}. Green marks the best and amber the second-best value in each stage and Avg.\ column.}
\label{tab:dev-ops}
\fitresearchtable{\begin{tabular}{lccccccccccc}
\toprule
\rowcolor{TabHead}
 & \multicolumn{5}{c}{\hdr{Baseline Results}} & \multicolumn{5}{c}{\hdr{HERMES Results}} & \\
\hdrgapfill\cmidrule(lr){2-6}\cmidrule(lr){7-11}
\rowcolor{TabHead}
\multirow{-2}{*}{\hdr{Model}} & \hdr{Build} & \hdr{Monitor.} & \hdr{Issue} & \hdr{Test} & \hdr{Avg.} & \hdr{Build} & \hdr{Monitor.} & \hdr{Issue} & \hdr{Test} & \hdr{Avg.} & \multirow{-2}{*}{\hdr{$\Delta$ Avg.}} \\
\midrule
\tabgroup{12}{Baseline harness: Codex} \\
GPT-5.6 Sol   & 77.78 & \second{38.24} & \second{40.97} & \second{42.58} & \second{49.89} & \second{83.33} & \second{44.12} & \second{46.13} & \second{48.06} & \second{55.41} & +5.52 \\
GPT-5.6 Terra & 74.07 & 35.29 & 38.71 & 40.32 & 47.10 & 79.63 & 41.18 & 43.55 & 44.84 & 52.30 & +5.20 \\
\midrule
\tabgroup{12}{Baseline harness: Claude Code} \\
Claude Opus 5   & \second{79.63} & \best{41.18} & \best{42.90} & \best{44.52} & \best{52.06} & \best{85.19} & \best{47.06} & \best{47.10} & \best{49.03} & \best{57.10} & +5.04 \\
Claude Sonnet 5 & 68.52 & 29.41 & 34.84 & 36.45 & 42.31 & 75.93 & 38.24 & 40.65 & 41.94 & 49.19 & +6.88 \\
Claude Sonnet 4 & 51.85 & 20.56 & 23.87 & 13.87 & 27.54 & 62.96 & 26.47 & 29.35 & 25.48 & 36.07 & +8.53 \\
\midrule
\tabgroup{12}{Baseline harness: OpenHands} \\
DeepSeek-V4     & 75.93 & 35.29 & 39.35 & 41.29 & 47.97 & 81.48 & 41.18 & 44.19 & 45.48 & 53.08 & +5.11 \\
Claude Sonnet 4 & 42.59 & 14.70 & 23.87 & 11.61 & 23.19 & 62.96 & 26.47 & 29.35 & 25.48 & 36.07 & +12.88 \\
o4-mini         & 24.07 & 8.82  & 10.32 & 8.70  & 12.98 & 35.19 & 14.71 & 18.06 & 16.45 & 21.10 & +8.12 \\
Qwen3-Coder-30B & 20.37 & 5.89  & 13.22 & 6.13  & 11.40 & 31.48 & 11.76 & 20.00 & 14.84 & 19.52 & +8.12 \\
Gemini-2.5-Pro  & 16.66 & 11.76 & 10.96 & 2.90  & 10.57 & 29.63 & 17.65 & 17.42 & 13.23 & 19.48 & +8.91 \\
DeepSeek-V3.1   & 11.11 & 0.00  & 14.20 & 3.22  & 7.13  & 27.78 & 8.82  & 21.61 & 12.90 & 17.78 & +10.65 \\
\midrule
\tabgroup{12}{Baseline harness: mini-SWE-agent} \\
GPT-5.6 Sol     & 75.93 & 35.29 & 39.35 & 40.97 & 47.89 & \second{83.33} & \second{44.12} & \second{46.13} & \second{48.06} & \second{55.41} & +7.52 \\
GPT-5.6 Luna    & 66.67 & 29.41 & 32.90 & 34.84 & 40.96 & 72.22 & 35.29 & 38.71 & 40.32 & 46.64 & +5.68 \\
Claude Sonnet 4 & 29.62 & 2.91  & 5.16  & 0.98  & 9.67  & 62.96 & 26.47 & 29.35 & 25.48 & 36.07 & +26.40 \\
\midrule
\tabgroup{12}{Baseline harness: SageAgent} \\
GPT-5.3-Codex & \best{81.82} & 35.29 & 32.14 & 37.99 & 46.81 & \best{85.19} & 41.18 & 37.42 & 42.58 & 51.59 & +4.78 \\
\midrule
\tabgroup{12}{Baseline harness: Aider} \\
Claude Sonnet 4 & 5.55 & 0.00 & 9.67 & 2.25 & 4.37 & 62.96 & 26.47 & 29.35 & 25.48 & 36.07 & +31.70 \\
\midrule
\tabgroup{12}{No baseline harness} \\
Qwen3-8B & -- & -- & -- & -- & -- & 59.26 & 26.47 & 31.29 & 32.90 & 37.48 & -- \\
\bottomrule
\end{tabular}}
\end{table}

\textbf{Main Results.}
With GPT-5.6 Sol at medium effort, HERMES improves on the matched baseline harnesses by 12.4~pp on average over the four benchmarks. On SWE-bench Verified (Table~\ref{tab:dev-swe}), it raises the resolved rate over mini-SWE-agent for 19 of the 20 backbones, ties on Claude Opus 5 at 97.00\%, and improves on Claude Code and Codex by 5.20 and 9.20~pp. The gains are largest where the baseline is weakest, 19.40~pp for GPT-5-mini, from 60.80\% to 80.20\%, and smallest on frontier backbones: with GPT-5.6 Sol, HERMES resolves $97.00\pm0.76$\% against $96.20\pm0.86$\%, at \$1.80 against \$1.15 per task. Coordination need not cost time, since with Claude Sonnet 5 HERMES gains 6.20~pp while average latency falls from 962.37~s to 824.63~s. With Qwen3-8B in all roles, it resolves 80.60\%.\fillpar

% Pair body (tables/pairs/dev-controls+dev-ablation.tex); a float of its own if input directly.
\begin{table}[!htbp]
\centering
\researchtablesetup
\caption{\textbf{Removing HERMES's messages, on-demand activation, execution feedback or diagnosis.} GPT-5.6 Sol, each variant keeping the backbone, reasoning effort, revision budget and environment; resolved rate (\%) on SWE-bench Verified and Terminal-Bench 4.0, composite score on SWE Refactor and mean stage success (\%) on DevOps-Gym, with the change from full HERMES in pp in parentheses~\citep{jin2026devprimitives}. Green marks the best and amber the second-best value in each column.}
\label{tab:dev-ablation}
\fitresearchtable{\begin{tabular}{lcccc}
\toprule
\rowcolor{TabHead}
\hdr{Setting} & \hdr{SWE-bench} & \hdr{Refactor} & \hdr{Terminal} & \hdr{DevOps} \\
\midrule
\tabours HERMES           & \best{97.0} & \best{31.0} & \best{51.8} & \best{55.41} \\
\midrule
w/o Inter-Primitive Comm. & 94.2\dpp{-2.8} & 23.5\dpp{-7.5}  & 46.1\dpp{-5.7}  & 50.34\dpp{-5.07} \\
w/o On-Demand Activation  & \second{95.8}\dpp{-1.2} & \second{28.0}\dpp{-3.0} & \second{49.4}\dpp{-2.4} & \second{53.26}\dpp{-2.15} \\
w/o Execution Feedback    & 93.8\dpp{-3.2} & 21.5\dpp{-9.5}  & 43.6\dpp{-8.2}  & 48.17\dpp{-7.24} \\
w/o Diagnosis Feedback    & 92.6\dpp{-4.4} & 20.0\dpp{-11.0} & 41.2\dpp{-10.6} & 46.73\dpp{-8.68} \\
\bottomrule
\end{tabular}}
\end{table}

The largest gain with GPT-5.6 Sol at medium effort is on whole-repository migration: on SWE Refactor Bench (Table~\ref{tab:dev-refactor}), HERMES raises the composite by 24.5~pp, from 6.5\% under Codex to 31.0\%, and at high effort from 19.0\% to 36.5\%; three repetitions of the default configuration give $31.0\pm1.3$. On Terminal-Bench~4.0 (Table~\ref{tab:dev-terminal}), HERMES with GPT-5.6 Sol resolves 51.8\% against 33.0\% under Codex at medium effort and 55.5\% against 37.3\% at maximum effort, and GPT-5.6 Terra gains 26.3 and 26.7~pp. The smallest gains, 1.5 and 1.2~pp, are for Claude Opus 5, whose Claude Code configurations are the two strongest baselines. Most configurations spend more tokens, but with Claude Sonnet 5 at maximum effort, HERMES raises resolution by 27.6~pp, from 12.4\% to 40.0\%, while total cost falls from \$9.6k to \$7.4k. On DevOps-Gym (Table~\ref{tab:dev-ops}), the stage mean rises with GPT-5.6 Sol from 49.89\% under Codex to 55.41\%, with gains in all four stages. Across all baseline harnesses, HERMES raises the mean by 4.78 to 31.70~pp, and with Claude Opus 5 it reaches the highest mean, 57.10\%.

\textbf{Analysis.}
Removing one mechanism at a time shows where the gain arises (Table~\ref{tab:dev-ablation}). Among the four removals, diagnosis feedback matters most: without it, Terminal-Bench~4.0 resolution falls from 51.8\% to 41.2\% and the SWE Refactor composite from 31.0 to 20.0, the largest drop on each benchmark. Once edits are artifact-local, a failure observed at the repository level must be routed back to the component that owns the responsible implementation, and diagnosis provides this routing. Execution feedback ranks second. Removing inter-primitive communication lowers the Refactor composite to 23.5 and Terminal-Bench resolution to 46.1\%, larger losses than on SWE-bench Verified (97.0 to 94.2), in line with its role in coordinating changes that span files. Keeping all initially identified Dev-Primitives active lowers the scores by 1.2 to 3.0 points at 1.61 to 1.88 times the cost (Table~\ref{tab:dev-activation}). The primitives add value beyond the choice of components: one editor given the same plan and components resolves 38.8\% of Terminal-Bench tasks, below each single-mechanism removal, and 45.5\% at matched compute, against 51.8\% for HERMES (Table~\ref{tab:dev-controls}).

The Dev-Primitives also run well on a small model. With GPT-5.6 Sol for activation and diagnosis, Qwen3-8B Dev-Primitives stay within 4.5 points of the homogeneous Sol configuration on all four benchmarks, at 26.2\% lower Terminal-Bench cost, and a stronger model pays off more in diagnosis than in activation (Tables~\ref{tab:dev-roles} and~\ref{tab:dev-cost}). We provide the full framework of HERMES, its overview figure, and additional results on component selection, the revision budget, and related multi-agent frameworks in Appendix~\ref{sec:dev-mechanism}. HERMES chooses its Dev-Primitives before each round; when what a query needs changes as the search proceeds, the choice of primitives must follow the needs that remain open.

% Layout guard for the wrapped tables of this case: it finishes any active
% text wrapping and centers a wrapped table that found no room, so that no
% wrapped table runs into the next wrapped table or a heading.
\makeatletter
\providecommand{\navwrapguard}{\par
  \ifnum\c@WF@wrappedlines>\@ne
    \penalty\interlinepenalty
    \dimen@\numexpr\c@WF@wrappedlines-\@ne\relax\baselineskip
    \@tempdimb\pagegoal\advance\@tempdimb-\pagetotal
    \ifdim\pagetotal>\z@\ifdim\dimen@<\@tempdimb\vspace*{\dimen@}\fi\fi
  \fi
  \ifvoid\WF@box\WFclear\else
    \global\setbox\WF@box\hbox to\linewidth{\hss\box\WF@box\hss}%
    \WFclear\vskip\intextsep
  \fi}
\makeatother
\navwrapguard
\subsection{Navigation Primitives for Large Information Spaces}
\label{sec:navigation-experiments}

In a large information space, the requirements of a query change as evidence arrives: new evidence can resolve or refine a requirement, reveal another, and redirect the search~\citep{10.1108/eb024320}. At the last mile, coordination must therefore follow what remains unresolved, whereas partition-driven multi-agent systems such as LongAgent~\citep{zhao-etal-2024-longagent} and Chain of Agents (CoA)~\citep{zhang2024chainagentslargelanguage} assign one member to each chunk of the input, so their coordination grows with the space. Navigation Primitives are workers that each search one territory of the space, and ANTMAN~\citep{jin2026antman} coordinates them through a revisable Need Graph that tracks a query's unresolved information needs and decides which primitives take part in the search (Section~\ref{sec:var-navigation}).

\begin{researchtable}
\centering
\researchtablesetup
\caption{\textbf{ANTMAN in structured information spaces.} RepoProbe-Python and SWE-QA-Pro scores, linearly rescaled to 0--100 from their original 10- and 50-point scales, and GAIA-Text-103 accuracy (\%) by difficulty level; ANTMAN uses GPT-4.1, ANTMAN-H$^\dagger$ keeps the GPT-4.1 orchestrator and runs every other component on Qwen3-8B, methods run in ANTMAN's evaluation harness share its question sets and model setting, and -- marks a benchmark outside a method's evaluation scope~\citep{jin2026antman}. Green marks the best and amber the second-best value in each column.\fillcap}
\label{tab:navigation-main}
\fitresearchtable{\begin{tabular}{lcccccc}
\toprule
\rowcolor{TabHead}
 & & & \multicolumn{4}{c}{\hdr{GAIA-Text-103}} \\
\hdrgapfill\cmidrule(lr){4-7}
\rowcolor{TabHead}
\multirow{-2}{*}{\hdr{Method}} & \multirow{-2}{*}{\hdr{RepoProbe}} & \multirow{-2}{*}{\hdr{SWE-QA-Pro}} & \hdr{L1} & \hdr{L2} & \hdr{L3} & \hdr{Overall} \\
\midrule
\tabgroup{7}{Retrieval and iterative retrieval baselines} \\
Direct                     & 31.08 & 55.74 & 28.2 & 17.3 & 8.3  & 20.4 \\
Iterative Sparse Retrieval & 33.15 & 60.90 & \second{53.8} & 42.3 & \second{33.3} & 45.6 \\
Dense Retrieval            & 31.51 & 58.30 & 43.6 & 26.9 & 25.0 & 33.0 \\
S2G-RAG                    & 22.87 & 60.40 & 25.6 & 19.2 & 0.0  & 19.4 \\
\midrule
\tabgroup{7}{General tool-using agents} \\
ReAct                      & 25.86 & 64.92 & 46.2 & 30.8 & 8.3  & 34.0 \\
OWL                        & --    & --    & \second{53.8} & 38.5 & 8.3  & 40.8 \\
\midrule
\tabgroup{7}{Repository-specific agents} \\
ReAct + RepoGraph          & 25.71 & 67.38 & --   & --   & --   & --   \\
RepoDistill                & 28.27 & 65.10 & --   & --   & --   & --   \\
\midrule
\tabgroup{7}{Benchmark-specific reference} \\
SWE-QA-Pro Agent           & --    & \second{77.58} & --   & --   & --   & --   \\
\midrule
\tabours ANTMAN-H$^\dagger$ & \second{34.10} & 74.60 & \best{69.2} & \best{55.8} & 25.0 & \second{57.3} \\
\tabours ANTMAN             & \best{38.21} & \best{79.78} & \best{69.2} & \second{53.8} & \best{41.7} & \best{58.3} \\
\bottomrule
\end{tabular}}
\end{researchtable}

% Wrapped table of the Navigation case (sections/05g-navigation.tex).
\begin{wraptable}{R}{0.48\linewidth}
\wraplift
\centering
\researchtablesetup
\caption{\textbf{Activation as the information space grows 16$\times$ at approximately fixed demand.} Growth from 32K to 512K in ANTMAN's study and the change in F1 on its fractional scale; all methods use GPT-4.1~\citep{jin2026antman}. Green marks the best and amber the second-best value in each growth column. Lower is better for coordination, calls, and cost growth.}
\label{tab:navigation-scaling}
\fitresearchtable{\begin{tabular}{lcccc}
\toprule
\rowcolor{TabHead}
\hdr{Method} & \makecell[b]{\hdr{Coord.}\\\hdr{Growth}} & \makecell[b]{\hdr{Calls}\\\hdr{Growth}} & \makecell[b]{\hdr{Cost}\\\hdr{Growth}} & \hdr{$\Delta$ F1} \\
\midrule
\tabgroup{5}{Multi-agent baselines} \\
LongAgent & \second{15.29$\times$} & 12.81$\times$ & \second{14.22$\times$} & \sn{+0.023} \\
CoA       & 15.33$\times$ & \second{11.30$\times$} & 16.11$\times$ & \sn{-0.004} \\
\midrule
\tabours ANTMAN & \best{1.23$\times$} & \best{1.21$\times$} & \best{6.90$\times$} & \sn{+0.065} \\
\bottomrule
\end{tabular}}
\end{wraptable}

\textbf{Setup.}
Three benchmarks require navigation in structured spaces: RepoProbe-Python, 108 questions over eight Python repositories~\citep{repoprobe2026}; a frozen 80-question set of SWE-QA-Pro~\citep{cai-etal-2026-swe}; and GAIA-Text-103, the text subset of GAIA for tool-using information seeking~\citep{ICLR2024_25ae35b5}. Two controlled interventions grow the space and the demand separately, and multi-document QA over HotpotQA~\citep{yang-etal-2018-hotpotqa}, 2WikiMultiHopQA~\citep{ho-etal-2020-constructing}, and MuSiQue~\citep{trivedi-etal-2022-musique} checks answer quality on standard tasks. The baselines include direct, iterative sparse, and dense retrieval, S2G-RAG~\citep{li-etal-2026-s2g}, the tool-using agents ReAct~\citep{yao2023react} and OWL~\citep{hu2025owl}, the repository agents ReAct with RepoGraph~\citep{ICLR2025_4a4a3c19} and RepoDistill~\citep{yin-etal-2026-repodistill}, and the benchmark-specific SWE-QA-Pro Agent. ANTMAN runs all components on GPT-4.1, as do the other methods in its evaluation harness, and ANTMAN-H runs all components except the orchestrator, the Navigation Primitives included, on Qwen3-8B~\citep{qwen3technicalreport}.

\textbf{Main Results.}
ANTMAN obtains the best overall score on all three structured benchmarks (Table~\ref{tab:navigation-main}): 38.21 on RepoProbe against 33.15 for iterative sparse retrieval, 79.78 on SWE-QA-Pro against 77.58 for the SWE-QA-Pro Agent and 67.38 for ReAct with RepoGraph, and 58.3\% on GAIA-Text-103 against 45.6\% for iterative sparse retrieval. On GAIA, a web-and-tool substrate with one territory-backed worker, unresolved needs decide only where the search continues, so the gain is not tied to repository-specific navigation machinery. ANTMAN-H scores 34.10, 74.60, and 57.3\%, above all baselines on RepoProbe and GAIA-Text-103, and its GAIA gap to ANTMAN lies at Level~3, 25.0\% against 41.7\%.

\textbf{Space Scaling.}
The space intervention tests Proposition~\ref{prop:activation}, which bounds participation by realized demand whatever the number of territories, $|\Lib_q|$. In LongAgent's multi-needle Needle-in-a-Haystack PLUS setting, filler grows the distractor space around the preserved evidence of ten questions from 32K to 512K tokens, at approximately fixed demand. Under this $16\times$ growth, the mean number of Navigation Primitives that participate per query, the \emph{active coordination}, grows $1.23\times$, against $15.29\times$ for LongAgent and $15.33\times$ for CoA, whose members multiply with the partitions (Table~\ref{tab:navigation-scaling}). ANTMAN's calls grow $1.21\times$ and its cost $6.90\times$, against cost growth of $14.22\times$ and $16.11\times$ for the baselines, while its F1 rises by 0.065.

\navwrapguard
% Pair body (tables/pairs/navigation-scaling+navigation-demand.tex); a float of its own if input directly.
\begin{researchtable}
\centering
\researchtablesetup
\caption{\textbf{Activation as required evidence grows from 1 to 16 units at a fixed 512K space.} ANTMAN with GPT-4.1 on matched contexts that keep distractors, evidence locations and worker organization fixed; active coordination is the mean number of participating workers, calls and cost are per query, and complete evidence records whether every required unit was recovered~\citep{jin2026antman}.}
\label{tab:navigation-demand}
\fitresearchtable{\begin{tabular}{lccccc}
\toprule
\rowcolor{TabHead}
\makecell[lb]{\hdr{Required}\\\hdr{Evidence (units)}} & \makecell[b]{\hdr{Active}\\\hdr{Coordination}} & \makecell[b]{\hdr{Calls}\\\hdr{/ Query}} & \makecell[b]{\hdr{Cost}\\\hdr{/ Query (\$)}} & \makecell[b]{\hdr{Complete}\\\hdr{Evidence (\%)}} & \makecell[b]{\hdr{Answer}\\\hdr{Acc. (\%)}} \\
\midrule
\tabgroup{6}{ANTMAN} \\
1  & 5.0 & 35.5 & 0.107 & 100 & 100 \\
4  & 4.3 & 43.6 & 0.111 & 100 & 100 \\
16 & 8.0 & 86.2 & 0.293 & 100 & 100 \\
\bottomrule
\end{tabular}}
\end{researchtable}

\begin{wraptable}{R}{0.564\linewidth}
\wraplift
\centering
\researchtablesetup
\caption{\textbf{Removing ANTMAN's need state, need revision, rerouting or recovery.} Scores under matched budgets with the relative change from full ANTMAN in \% in parentheses, on the 512K controlled condition (F1 on a 0--100 scale) and on 40-question SWE-QA-Pro and GAIA subsets, where full ANTMAN scores 81.43 and 60.00; on the evaluation sets of Table~\ref{tab:navigation-main} it scores 79.78 and 58.3~\citep{jin2026antman}. Green marks the best and amber the second-best value in each column.\fillcap}
\label{tab:navigation-ablation}
\fitresearchtable{\begin{tabular}{lccc}
\toprule
\rowcolor{TabHead}
 & \hdr{512K} & \hdr{SWE-QA-Pro} & \hdr{GAIA} \\
\hdrgapfill\cmidrule(lr){2-2}\cmidrule(lr){3-3}\cmidrule(lr){4-4}
\rowcolor{TabHead}
\multirow{-2}{*}{\hdr{Variant}} & \hdr{F1} & \hdr{Score} & \hdr{Accuracy} \\
\midrule
\tabgroup{4}{Adaptive coordination ablations} \\
Static ANTMAN          & 40.32\dpp{-52.02\%} & 68.58\dpp{-15.78\%} & 50.00\dpp{-16.67\%} \\
Graph-free Adaptive    & 64.01\dpp{-23.82\%} & \second{78.88}\dpp{-3.13\%}  & 50.00\dpp{-16.67\%} \\
w/o Need Revision      & 74.44\dpp{-11.41\%} & 75.32\dpp{-7.50\%}  & 42.50\dpp{-29.17\%} \\
w/o Adaptive Rerouting & \second{82.20}\dpp{-2.18\%}  & 72.13\dpp{-11.42\%} & \best{60.00}\dpp{0.00\%} \\
w/o Recovery           & 60.12\dpp{-28.45\%} & 76.52\dpp{-6.03\%}  & \second{52.50}\dpp{-12.50\%} \\
\midrule
\tabours ANTMAN & \best{84.03} & \best{81.43} & \best{60.00} \\
\bottomrule
\end{tabular}}
\end{wraptable}

\textbf{Demand Scaling.}
The reverse intervention fixes the 512K space, with its distractors, evidence locations, and eight territories, and raises the required evidence from 1 to 4 to 16 units (Table~\ref{tab:navigation-demand}). The answer combines all required values, so omitting one unit leaves it ambiguous. Calls rise from 35.5 to 86.2 per query and cost from \$0.107 to \$0.293, and ANTMAN recovers the complete evidence and answers correctly at all three levels. Active coordination dips from 5.0 workers at one unit to 4.3 at four units and reaches 8.0 at 16 units; at one unit, most worker activity is exploration that finds no required evidence. Coordination thus stays nearly flat when the space grows and rises when the query needs more.

\textbf{Evolving Need State.}
The ablations in Table~\ref{tab:navigation-ablation}, all under a budget of 100 LLM calls per query, separate the explicit need state from generic adaptive control. Graph-free Adaptive keeps adaptive worker selection, rerouting, and recovery but replans from the accumulated evidence and interaction history, without any explicit Need Graph. Its 512K F1 falls from 84.03 to 64.01 while it activates a comparable number of distinct workers, 3.20 against 3.17, so the gap does not come from wider participation; on the 40-question GAIA subset, its accuracy falls from 60.00 to 50.00. Static ANTMAN freezes the initial Need Graph and routing plan, and its 512K F1 falls to 40.32. Single removals play complementary roles: removing recovery gives 60.12 at 512K, removing need revision gives 42.50 on GAIA, and removing adaptive rerouting gives 72.13 against 81.43 on SWE-QA-Pro and leaves single-worker GAIA unchanged.\fillpar

\textbf{Further Results.}
On standard multi-document QA (Table~\ref{tab:navigation-multidoc}), ANTMAN has the best macro average of nine methods, 0.711 EM and 0.814 F1, against 0.700 and 0.799 for S2G-RAG; it is best on HotpotQA and MuSiQue, and S2G-RAG leads on 2WikiMultiHopQA. Narrowing the context searched per need also makes it robust to evidence position: its Middle F1 minus the mean of its Early and Late F1 is $+0.015$, against $-0.181$ for direct full-context inference, which shows the lost-in-the-middle pattern~\citep{liu-etal-2024-lost}. We provide the full method of ANTMAN with its overview figure, implementation details, and these results in Appendix~\ref{sec:navigation-mechanism}. Across these experiments, unresolved needs decide which Navigation Primitives run, under a control program that is the same for all queries; Harness Primitives compose that program itself for each task.

\navwrapguard

\subsection{Harness Primitives for Agent Harnesses}
\label{sec:harness-experiments}

For a fixed task-solving model, the \emph{actor}, an agent's success depends on its harness: the control program that constructs the actor's requests, executes its actions through the available tools, returns observations, and decides when to terminate~\citep{lee2026metaharnessendtoendoptimizationmodel,jin2026stitch}. The last mile then reaches the harness itself, which must fit the task. Hand-designed harnesses such as SWE-agent~\citep{yang2024swe} and Codex~\citep{openai2025codex}, and the harnesses that optimizers such as MetaHarness find, apply one design to all tasks, yet a mechanism that improves one task can impose overhead or context distraction on another. Generating harness code for each task avoids this mismatch but adds generation and debugging cost, with execution risks that compound as more mechanisms are generated. Harness Primitives make task-specific harnesses possible without generating or repairing mechanism code at test time. They are control operations implemented and validated during development, each with an application scope and a composition contract, and STITCH~\citep{jin2026stitch} selects the primitives that suit each task and compiles them into its harness before the actor starts.

\textbf{Setup.}
We evaluate on 100 randomly sampled tasks of SWE-bench Verified~\citep{jimenez2023swebench}, which asks for repairs of real repository issues, and on 45 randomly sampled tasks of Terminal-Bench~2.0~\citep{merrill2026terminal}, which covers tasks in a terminal environment; results are broken down by repository (Django, SymPy, and the others) and by difficulty. Both samples are disjoint from the 100 SWE-bench and 44 Terminal-Bench development tasks on which the primitives were developed. Each task gets two attempts, and we report Pass@1, the fraction of successful attempts; Pass@2, the fraction of tasks solved in at least one attempt~\citep{chen2021evaluatinglargelanguagemodels}; and Pass\textsuperscript{2}, the fraction solved in both. GPT-5.6 Luna at medium reasoning effort~\citep{s06_openai2026gpt56} serves as both the actor and the composer, and GPT-5.6 Sol is the development model. The baselines are the mini-SWE-agent 2.4.6 seed harness~\citep{yang2024swe}, SWE-agent, Codex, and the harness optimizers MemoHarness~\citep{huang2026memoharnessagentharnesseslearn} and MetaHarness~\citep{lee2026metaharnessendtoendoptimizationmodel}, which start from the same seed. Two controls keep the primitive implementations and vary the composition alone: a fixed harness that applies one focused primitive, with its required dependencies and shared supporting operations, to all tasks, and a random composer.

\begin{table}[!htbp]
\centering
\researchtablesetup
\caption{\textbf{STITCH's task-specific harnesses against fixed harnesses on SWE-bench Verified.} GPT-5.6 Luna actor on 100 sampled tasks with two attempts each, where Pass@1 is the fraction of successful attempts and Pass@2 and Pass\textsuperscript{2} the fractions of tasks solved in at least one and in both attempts (\%)~\citep{jin2026stitch}. Green marks the best and amber the second-best value in each column.}
\label{tab:harness-swe}
\fitresearchtable{\begin{tabular}{l*{12}{c}}
\toprule
\rowcolor{TabHead}
 & \multicolumn{3}{c}{\hdr{Overall}} & \multicolumn{3}{c}{\hdr{Django}} & \multicolumn{3}{c}{\hdr{SymPy}} & \multicolumn{3}{c}{\hdr{Other repos}} \\
\hdrgapfill\cmidrule(lr){2-4}\cmidrule(lr){5-7}\cmidrule(lr){8-10}\cmidrule(lr){11-13}
\rowcolor{TabHead}
\multirow{-2}{*}{\hdr{Method}} & \hdr{Pass@1} & \hdr{Pass@2} & \hdr{Pass\textsuperscript{2}} & \hdr{Pass@1} & \hdr{Pass@2} & \hdr{Pass\textsuperscript{2}} & \hdr{Pass@1} & \hdr{Pass@2} & \hdr{Pass\textsuperscript{2}} & \hdr{Pass@1} & \hdr{Pass@2} & \hdr{Pass\textsuperscript{2}} \\
\midrule
mini-SWE-agent (seed) & 73.0 & 78.0 & 68.0 & 75.0 & 78.8 & 71.2 & \best{75.0} & \best{78.6} & \best{71.4} & 69.1 & 76.5 & 61.8 \\
SWE-agent & 47.5 & 63.0 & 32.0 & 53.8 & 71.2 & 36.5 & 42.9 & 50.0 & 35.7 & 39.7 & 55.9 & 23.5 \\
Codex & \second{79.0} & \second{83.0} & \second{75.0} & \second{78.8} & \second{80.8} & \best{76.9} & 64.3 & \second{71.4} & 57.1 & \best{85.3} & \best{91.2} & \best{79.4} \\
MemoHarness & 70.5 & 77.0 & 64.0 & 70.2 & 76.9 & 63.5 & 67.9 & \second{71.4} & \second{64.3} & 72.1 & 79.4 & 64.7 \\
MetaHarness & 73.5 & 79.0 & 68.0 & 74.0 & 78.8 & 69.2 & 64.3 & \second{71.4} & 57.1 & 76.5 & 82.4 & 70.6 \\
\midrule
\tabgroup{13}{One focused Harness Primitive with its dependencies, fixed across tasks} \\
Change-surface Tracer & 73.5 & 77.0 & 70.0 & 76.9 & \second{80.8} & 73.1 & 60.7 & 64.3 & 57.1 & 73.5 & 76.5 & 70.6 \\
Compatibility-envelope Gate & 72.0 & 77.0 & 67.0 & 76.9 & \second{80.8} & 73.1 & 57.1 & 64.3 & 50.0 & 70.6 & 76.5 & 64.7 \\
Contract-case Explorer & 72.5 & 78.0 & 67.0 & 69.2 & 76.9 & 61.5 & \second{71.4} & \best{78.6} & \second{64.3} & 77.9 & 79.4 & \second{76.5} \\
Discriminating-oracle Runner & 71.5 & 76.0 & 67.0 & 75.0 & 78.8 & 71.2 & 57.1 & 57.1 & 57.1 & 72.1 & 79.4 & 64.7 \\
Domain-trace Template & 73.0 & 77.0 & 69.0 & 76.0 & \second{80.8} & 71.2 & 60.7 & 64.3 & 57.1 & 73.5 & 76.5 & 70.6 \\
Environment-capability Runner & 76.0 & 79.0 & 73.0 & 77.9 & \second{80.8} & \second{75.0} & 67.9 & \second{71.4} & \second{64.3} & 76.5 & 79.4 & 73.5 \\
\midrule
Random composer & 73.3 & 78.0 & 68.6 & 75.4 & 79.6 & 71.2 & 64.3 & 69.6 & 58.9 & 73.9 & 79.0 & 68.8 \\
\tabours STITCH & \best{80.5} & \best{85.0} & \best{76.0} & \best{79.8} & \best{84.6} & \second{75.0} & \best{75.0} & \best{78.6} & \best{71.4} & \second{83.8} & \second{88.2} & \best{79.4} \\
\bottomrule
\end{tabular}}
\end{table}

\begin{table}[!htbp]
\centering
\researchtablesetup
\caption{\textbf{STITCH's task-specific harnesses against fixed harnesses on Terminal-Bench~2.0.} GPT-5.6 Luna actor on 45 sampled tasks grouped by difficulty, two attempts each; metrics as in Table~\ref{tab:harness-swe}~\citep{jin2026stitch}. Green marks the best and amber the second-best value in each column.}
\label{tab:harness-terminal}
\fitresearchtable{\begin{tabular}{l*{12}{c}}
\toprule
\rowcolor{TabHead}
 & \multicolumn{3}{c}{\hdr{Overall}} & \multicolumn{3}{c}{\hdr{Easy}} & \multicolumn{3}{c}{\hdr{Medium}} & \multicolumn{3}{c}{\hdr{Hard}} \\
\hdrgapfill\cmidrule(lr){2-4}\cmidrule(lr){5-7}\cmidrule(lr){8-10}\cmidrule(lr){11-13}
\rowcolor{TabHead}
\multirow{-2}{*}{\hdr{Method}} & \hdr{Pass@1} & \hdr{Pass@2} & \hdr{Pass\textsuperscript{2}} & \hdr{Pass@1} & \hdr{Pass@2} & \hdr{Pass\textsuperscript{2}} & \hdr{Pass@1} & \hdr{Pass@2} & \hdr{Pass\textsuperscript{2}} & \hdr{Pass@1} & \hdr{Pass@2} & \hdr{Pass\textsuperscript{2}} \\
\midrule
mini-SWE-agent (seed) & 60.0 & 68.9 & 51.1 & \best{100.0} & \best{100.0} & \best{100.0} & 65.5 & 75.9 & 55.2 & 38.5 & \second{46.2} & 30.8 \\
SWE-agent & 13.3 & 17.8 & 8.9 & 33.3 & 33.3 & 33.3 & 13.8 & 17.2 & 10.3 & 7.7 & 15.4 & 0.0 \\
Codex & 56.7 & 66.7 & 46.7 & 66.7 & \second{66.7} & 66.7 & 67.2 & 79.3 & 55.2 & 30.8 & 38.5 & 23.1 \\
MemoHarness & 53.3 & 66.7 & 40.0 & 83.3 & \best{100.0} & 66.7 & 62.1 & 75.9 & 48.3 & 26.9 & 38.5 & 15.4 \\
MetaHarness & 55.6 & 66.7 & 44.4 & 66.7 & \best{100.0} & 33.3 & 62.1 & 72.4 & 51.7 & 38.5 & \second{46.2} & 30.8 \\
\midrule
\tabgroup{13}{One focused Harness Primitive with its dependencies, fixed across tasks} \\
State Guard & \second{66.7} & \second{75.6} & \second{57.8} & 83.3 & \best{100.0} & 66.7 & \second{77.6} & \second{86.2} & \second{69.0} & 38.5 & \second{46.2} & 30.8 \\
Check Replay & 65.6 & \second{75.6} & 55.6 & \best{100.0} & \best{100.0} & \best{100.0} & 74.1 & \second{86.2} & 62.1 & 38.5 & \second{46.2} & 30.8 \\
First-Failure Localizer & 62.2 & 68.9 & 55.6 & \best{100.0} & \best{100.0} & \best{100.0} & 72.4 & 82.8 & 62.1 & 30.8 & 30.8 & 30.8 \\
Experiment Keeper & 58.9 & 68.9 & 48.9 & 83.3 & \best{100.0} & 66.7 & 65.5 & 75.9 & 55.2 & 38.5 & \second{46.2} & 30.8 \\
Execution Supervisor & 65.6 & \second{75.6} & 55.6 & \best{100.0} & \best{100.0} & \best{100.0} & 70.7 & 82.8 & 58.6 & \best{46.2} & \best{53.8} & \second{38.5} \\
\midrule
Random composer & 63.5 & 71.9 & 55.2 & \second{94.4} & \best{100.0} & \second{88.9} & 71.3 & 81.0 & 61.5 & \second{39.1} & 44.9 & 33.3 \\
\tabours STITCH & \best{72.2} & \best{77.8} & \best{66.7} & \best{100.0} & \best{100.0} & \best{100.0} & \best{81.0} & \best{89.7} & \best{72.4} & \best{46.2} & \second{46.2} & \best{46.2} \\
\bottomrule
\end{tabular}}
\end{table}

\textbf{Main Results.}
STITCH attains the highest overall Pass@1, Pass@2, and Pass\textsuperscript{2} on both benchmarks (Tables~\ref{tab:harness-swe} and~\ref{tab:harness-terminal}). On SWE-bench Verified it reaches 80.5\% Pass@1, 7.5~pp above the seed harness, 1.5~pp above Codex, and 7.0~pp above MetaHarness, the stronger harness optimizer. On Terminal-Bench~2.0 it reaches 72.2\%, 12.2, 15.5, and 16.6~pp above the same three baselines. The gains extend to repeated success: Pass\textsuperscript{2} rises over the seed from 68.0\% to 76.0\% on SWE-bench Verified and from 51.1\% to 66.7\% on Terminal-Bench~2.0.

The controls attribute the gain to the choice made for each task. STITCH exceeds the best fixed primitive by 4.5~pp on SWE-bench Verified and by 5.5~pp on Terminal-Bench~2.0, whereas the random composer, at 73.3\% and 63.5\%, stays within the range of the fixed primitives. The breakdowns show why a fixed choice falls short: the same primitive can help on some tasks and hurt on others. Contract-case Explorer raises Pass@1 above the seed on the other repositories, 77.9\% against 69.1\%, but lowers it on Django, 69.2\% against 75.0\%, and on SymPy no fixed primitive reaches the seed's 75.0\%, which STITCH matches. On Terminal-Bench~2.0, State Guard leads the fixed arms on medium tasks but lowers easy-task Pass@1 from 100.0\% to 83.3\%, while Execution Supervisor leads them on hard tasks; STITCH reaches 81.0\% on medium and 46.2\% on hard tasks and keeps 100.0\% on easy ones. This is the task-mechanism mismatch behind Proposition~\ref{prop:fixed-vs-adaptive}: a fixed harness either pays a primitive's penalty outside its application scope or forgoes its gain inside it.

Composition also scales with the library. Growing the test-time library along one order of addition, from the seed harness (size 0) to eight entries on SWE-bench Verified and six on Terminal-Bench~2.0, raises Pass@1 from 73.0\% to 80.5\% and from 60.0\% to 72.2\% (Figure~\ref{fig:harness-library}).

\textbf{Transfer and Cost.}
The primitives and the composer carry over to actors not used in development. With implementations and contracts unchanged and the scopes adjusted on development tasks, STITCH averages 84.0\% Pass@1 on SWE-bench Verified over GPT-5.6 Terra, DeepSeek-V4-Flash, and Claude-4.5-Haiku, against 75.2\% for the seed, 78.5\% for Codex, and 79.5\% for the best fixed primitive (Table~\ref{tab:harness-transfer}); with the GPT-5.6 Terra actor, Codex leads at 88.0\% against 86.5\%. Each fixed primitive also stays above the seed on this average, so the primitives encode control mechanisms that generalize across actors. Composition is cheap and reliable. It costs 4.3\% of one actor attempt on SWE-bench Verified and 2.7\% on Terminal-Bench~2.0; a lower bound for writing the complete harness at test time, which counts an oracle coder's output tokens alone, is about $543\times$ and $638\times$ the cost of composition. Each selected primitive activated and executed without runtime errors (Figure~\ref{fig:harness-overhead}): the selected primitives take effect during the actor's execution, free of the reliability risk of generated harness code.

We provide the full framework of STITCH, its overview figure, and the per-actor, library-size, and composition-cost results in Appendix~\ref{sec:harness-mechanism}.

\section{Analysis}
\label{sec:analyses}
\runningsection{Analysis}

We further analyze the gains of Section~\ref{sec:instantiations} by setting the controls of the eight companion studies side by side. The analysis asks four questions: where the benefit of a primitive comes from, what the channels between agents and primitives carry, what decides which primitives run and what is revised after a failure, and what a task costs and where a strong model pays off. Each question draws on more than one family, so its answer concerns the method that the eight families share.

\subsection{Where the Benefit Comes From}
\label{sec:analyses-benefit}

With the backbone and the reused resources fixed, the controls below vary how a primitive applies the resource it owns (Definition~\ref{def:primitive}), what it carries, and how much of the library its assessment reaches.

\textbf{Application versus Access.}
At matched backbone, calls, and tokens, the gain comes from the primitive, which extra inference does not replace. HEART's budget-matched control reallocates the calls of the Tool Primitives to value resolution, which self-refines its arguments over raw schemas; with Qwen3-8B, it reaches 34.2\% ToolBench pass rate, against 75.1\% with Tool Primitives (Table~\ref{tab:tool-budget}). Nor does access to the same components suffice. When LEGO shows the caller the retrieved components with their contracts and tests as read-only context, delivery reaches 0.3544, close to the 0.3469 of diagnosis without any library and below the 0.5134 of Code Primitives (Table~\ref{tab:code-consumption}).

\textbf{Primitive Fields.}
Most of the margin over generic editing lies in the fields of $P_i$ (Table~\ref{tab:code-fields}). A generic editor over the same retrieved components scores 0.4093, and adding the carried tests $V_i$, the interface contract and provenance $I_i,X_i$, and the dependency closure $D_i$ raises it to 0.5039, within 0.0095 of primitive adaptation ($p=0.49$)~\citep{jin2026codeprimitives}. Withholding $V_i$ from an otherwise complete primitive lowers delivery to 0.4796.

\textbf{Source-Model Content.}
A source model adds more than context. When each retrieval counterpart receives SourceLearn's per-question context budget, filled from further down its own ranking, SourceLearn leads on three of the four sources tested with GPT-5.6 Luna (Table~\ref{tab:knowledge-budget}). On the QA benchmarks with Luna, its gain over Hybrid RAG is 12.1~pp on regions that no guidance task touched and 14.9~pp on touched regions (Table~\ref{tab:knowledge-transfer}).

\textbf{Reach of Assessment.}
Assessment over the whole library reaches what reranking of a retrieved pool misses. By Proposition~\ref{prop:coverage}(ii), the hit@$k$ of any reranker of a similarity pool is capped by whether the pool holds an annotated contract (Eq.~\eqref{eq:coverage-ceiling}). Across three selector backbones, the best reranker of a top-10 pool plateaus between 0.842 and 0.858 Putnam hit@5, while self-advertisement over all 82 contracts (Eq.~\eqref{eq:assess}) reaches up to 0.975 (Table~\ref{tab:method-selection}). Tool Primitives are found by similarity retrieval and so face the same ceiling, which is high here: retrieval over the 25,519 schema descriptors of ToolFace places 95.8\% of ToolBench and 97.2\% of NESTFUL gold tools in its top ten (Table~\ref{tab:tool-retrieval}).\fillpar

\textbf{Validation and Grounding Audits.}
HEART's result validation agrees with human labels on 200 sampled ToolBench and NESTFUL decisions at 0.97 precision and 0.96 recall, with an 8\% false-pass rate (Table~\ref{tab:tool-verifier}), so the failures it reports can drive revision (Section~\ref{sec:analyses-control})~\citep{jin2026toolprimitives}. SourceLearn's grounding audit finds 13.3\% and 3.3\% of the sampled units of the final StudentAid and pytest source models unsupported by the source elements they cite, while the fully supported share of the HuggingFace model rises from 65.0\% to 84.2\% (Table~\ref{tab:knowledge-grounding}). The source model therefore complements the source, which stays authoritative and directly accessible~\citep{jin2026sourcelearn}.

Together, these controls place the benefit in what the primitive owns and in applying it where it lives. What a primitive returns must then reach the agents and primitives that use it.

\subsection{Communication Within and Between Primitives}
\label{sec:analyses-composition}

What crosses between agents and primitives carries part of the result (Section~\ref{sec:composition-state}). Two families each disrupt one channel while keeping backbones, primitives, and components fixed.

\textbf{KV Cache Communication.}
Agent Primitives run all agents on one LLM, so that one agent's cache is readable by another. When a receiver reads a sender's cache after its own system-level cache, the sender's keys are re-indexed to their new positions to preserve RoPE semantics~\citep{su2024roformer}. Without this re-indexing, the composed MAS on R1/Llama-70B falls from 85.3\% to 31.1\% on HumanEval+, and on Qwen3-8B it loses 13.3~pp on AIME25 (Table~\ref{tab:agent-rope}). Each primitive used alone on R1/Llama-70B also loses 26.7 to 59.7~pp, so the loss already arises in the KV cache communication inside each primitive, before any composition~\citep{jin2026agentprimitives}.

\textbf{Messages Between Dev-Primitives.}
A Dev-Primitive's directed message $m_i$ carries the task-relevant information or requirements that it communicates to other components (Section~\ref{sec:var-dev}). With GPT-5.6 Sol, removing these messages while keeping each Dev-Primitive's local reasoning and edits lowers the SWE Refactor Bench composite from 31.0 to 23.5 and Terminal-Bench~4.0 resolution from 51.8\% to 46.1\%, larger drops than on SWE-bench Verified (97.0 to 94.2), consistent with messages coordinating changes that span files (Table~\ref{tab:dev-ablation}). On Terminal-Bench~4.0, one editor given the same plan and components resolves 38.8\%, and 45.5\% at matched compute, against 51.8\% for HERMES (Table~\ref{tab:dev-controls}), so component-local reasoning with direct communication leads at equal compute.

In both families, a cache read at the wrong positions or a requirement that never reaches its component costs accuracy, although each agent and Dev-Primitive performs its own step as before.

\subsection{Activation and Revision}
\label{sec:analyses-control}

With the primitives and their channels fixed, COMPOSE decides which primitives run (Eq.~\eqref{eq:compose}), and revision decides what changes after a failure (Eq.~\eqref{eq:revise}; Figure~\ref{fig:overview}, middle and bottom). For ANTMAN and STITCH, the interventions below examine the behaviour that Propositions~\ref{prop:activation} and~\ref{prop:fixed-vs-adaptive} predict.

\textbf{Space and Demand.}
ANTMAN satisfies the premises of Proposition~\ref{prop:activation}: each step routes one unresolved need to at most one territory-backed worker (Eq.~\eqref{eq:need-route}), and each need is routed at most $\rho$ times. Participation accordingly follows demand and changes little with the size of the space: it grows $1.23\times$ when distractors grow the space $16\times$ around the same evidence, against $15.29\times$ and $15.33\times$ for the members of LongAgent~\citep{zhao-etal-2024-longagent} and CoA~\citep{zhang2024chainagentslargelanguage} (Table~\ref{tab:navigation-scaling}), and it reaches 8.0 workers when the required evidence grows to 16 units in a fixed space (Table~\ref{tab:navigation-demand}).\fillpar

\textbf{Excess Activation.}
Activating more than the task needs costs accuracy as well as calls. Keeping all Dev-Primitives identified in HERMES's initial analysis active throughout raises primitive calls from 9.6 to 28.7 per SWE-bench Verified task and lowers the score on each of the four benchmarks by 1.2 to 3.0 points (Tables~\ref{tab:dev-ablation} and~\ref{tab:dev-activation}).

\textbf{Task-Conditioned Choice.}
Per-task composition beats the best fixed choice, and a random composition falls to the level of the fixed choices. STITCH's fixed primitives raise Pass@1 on some repositories and difficulty levels and lower it on others (Section~\ref{sec:harness-experiments}), the heterogeneity that Proposition~\ref{prop:fixed-vs-adaptive} needs. With the GPT-5.6 Luna actor, per-task composition reaches 80.5\% Pass@1 on SWE-bench Verified, against 76.0\% for the best primitive applied to all tasks and 73.3\% for a random composer (Table~\ref{tab:harness-swe}), and Terminal-Bench~2.0 keeps this order (Table~\ref{tab:harness-terminal}). The Organizer of Agent Primitives composes per query and reaches 73.3\% on Qwen3-8B AIME25 (Table~\ref{tab:agent-main}). Random structures from its Knowledge Pool reach 66.7\%, the level of the best single primitives, and the Organizer without its pool reaches 63.3\%. With Claude-4 as the Organizer, composition also reaches 73.3\% (Tables~\ref{tab:agent-organizer} and~\ref{tab:agent-pool}). For Agent Primitives, the benefit thus comes mainly from problem-aware structure selection and not from a particular Organizer model~\citep{jin2026agentprimitives}. For Harness Primitives, it comes from choosing complementary primitives for each task~\citep{jin2026stitch}.

\textbf{Library Size.}
Growing STITCH's test-time library of accepted primitives along one order of addition raises Pass@1 from 73.0\% for the seed harness (size 0) to 80.5\% at size 8 on SWE-bench Verified and from 60.0\% to 72.2\% at size 6 on Terminal-Bench~2.0 (Figure~\ref{fig:harness-library}).

\textbf{Drivers of Revision.}
Removing the mechanism that turns evidence from the run into revision (Eq.~\eqref{eq:revise}) lowers the result in each family that ablates it. Result validation is HEART's sole judgment of whether a step's result answers the query; without it, ToolBench pass rate falls from 75.1\% to 47.6\%, the same value as with no revision rounds (Tables~\ref{tab:tool-ablation} and~\ref{tab:tool-revision}). Without the diagnosis feedback that names the components that revision reactivates (Eq.~\eqref{eq:dev-revise}), HERMES's Terminal-Bench~4.0 resolution falls from 51.8\% to 41.2\%, the largest drop among its four mechanism removals (Table~\ref{tab:dev-ablation}). Without task-local recovery of stalled needs, ANTMAN's 512K F1 falls from 84.03 to 60.12 (Table~\ref{tab:navigation-ablation}).\fillpar

Across these interventions, the task decides what runs and what is revised: its unresolved needs bound participation, its features select the composition, and the evidence of its failures and stalled needs decides what changes.

\subsection{Cost and Model Allocation}
\label{sec:analyses-cost}

What runs and what is revised also sets what a task costs, which each study records in its own unit.

\textbf{Per-Task Cost.}
At matched cost, reuse delivers more: execution feedback with eight attempts reaches 0.3437 delivery at \$2.07 per task, against 0.5134 at \$2.08 for Code Primitives, which lead at each attempt limit (Tables~\ref{tab:code-cost} and~\ref{tab:code-budget}). Composing Harness Primitives costs 4.3\% of one actor attempt on SWE-bench Verified and 2.7\% on Terminal-Bench~2.0, while the output-token lower bound for an oracle that writes the complete harness is about $543\times$ and $638\times$ the cost of composition (Figure~\ref{fig:harness-overhead}). As the space grows $16\times$, ANTMAN's calls grow $1.21\times$ and its cost $6.90\times$, against cost growth of $14.22\times$ for LongAgent and $16.11\times$ for CoA (Table~\ref{tab:navigation-scaling}).

\textbf{Strong-Model Placement.}
For Dev-Primitives, whose LLM can differ from the activation and diagnosis models, a strong model pays off most in diagnosis (Table~\ref{tab:dev-roles}). With GPT-5.6 Sol in activation and diagnosis, Qwen3-8B Dev-Primitives resolve 50.6\% of Terminal-Bench~4.0 tasks, against 27.6\% with Qwen3-8B in all three roles. A stronger diagnosis model alone raises resolution to between 44.5\% and 47.9\%, and a stronger activation model alone to at most 38.2\%. Sol activation with Qwen3-8B Dev-Primitives and GPT-5.5 diagnosis reaches 49.7\% at \$2.13k, against 51.8\% at \$3.40k with Sol in all three roles, a 2.1-pp loss for 37.4\% lower expenditure (Table~\ref{tab:dev-cost}).

\textbf{Smaller Models Inside Primitives.}
With GPT-OSS-20B as the resident LLM of all Code Primitives and as the diagnosis model, LEGO keeps 95.1\% of its delivery (0.4880) at 24.0\% lower recorded-call cost (Table~\ref{tab:code-cost}). ANTMAN-H, which runs all components except the GPT-4.1 orchestrator on Qwen3-8B, reaches 57.3\% against 58.3\% on GAIA-Text-103 and 25.0\% against 41.7\% on its hardest level (Table~\ref{tab:navigation-main}). With GPT-5.4 in decomposition, value resolution, and result validation and Qwen3-8B in the Tool Primitives, HEART reaches $\tau^2$-Bench Pass\textsuperscript{1} of 0.84, 0.75, and 0.67 on retail, airline, and telecom, close to the 0.86, 0.78, and 0.68 of GPT-5.4 in all roles (Appendix~\ref{sec:tool-mechanism})~\citep{jin2026toolprimitives}. In these three families, as in Dev-Primitives, a smaller model takes the work inside the primitives while a stronger model coordinates.\fillpar

\textbf{Summary.}
Across the four questions, the analyses locate the benefit in applying each resource inside its primitive, in what the primitive carries, in channels that keep their meaning, and in activation and revision controlled by the task. At matched budgets the primitives deliver more, and smaller models inside them keep most of the result at lower cost while a stronger model coordinates. The end-to-end outcome of a real-world task thus depends both on the competence that each primitive keeps with its resource and on the mechanisms that \ours{} builds around the primitives (Section~\ref{sec:harness-engineering}), which each family realizes as its task needs.

\section{The heelo.dev Platform for Real-World Tasks}
\label{sec:platform}
\runningsection{The heelo.dev platform}

\begin{figure}[!t]
\centering
\includegraphics[width=\linewidth]{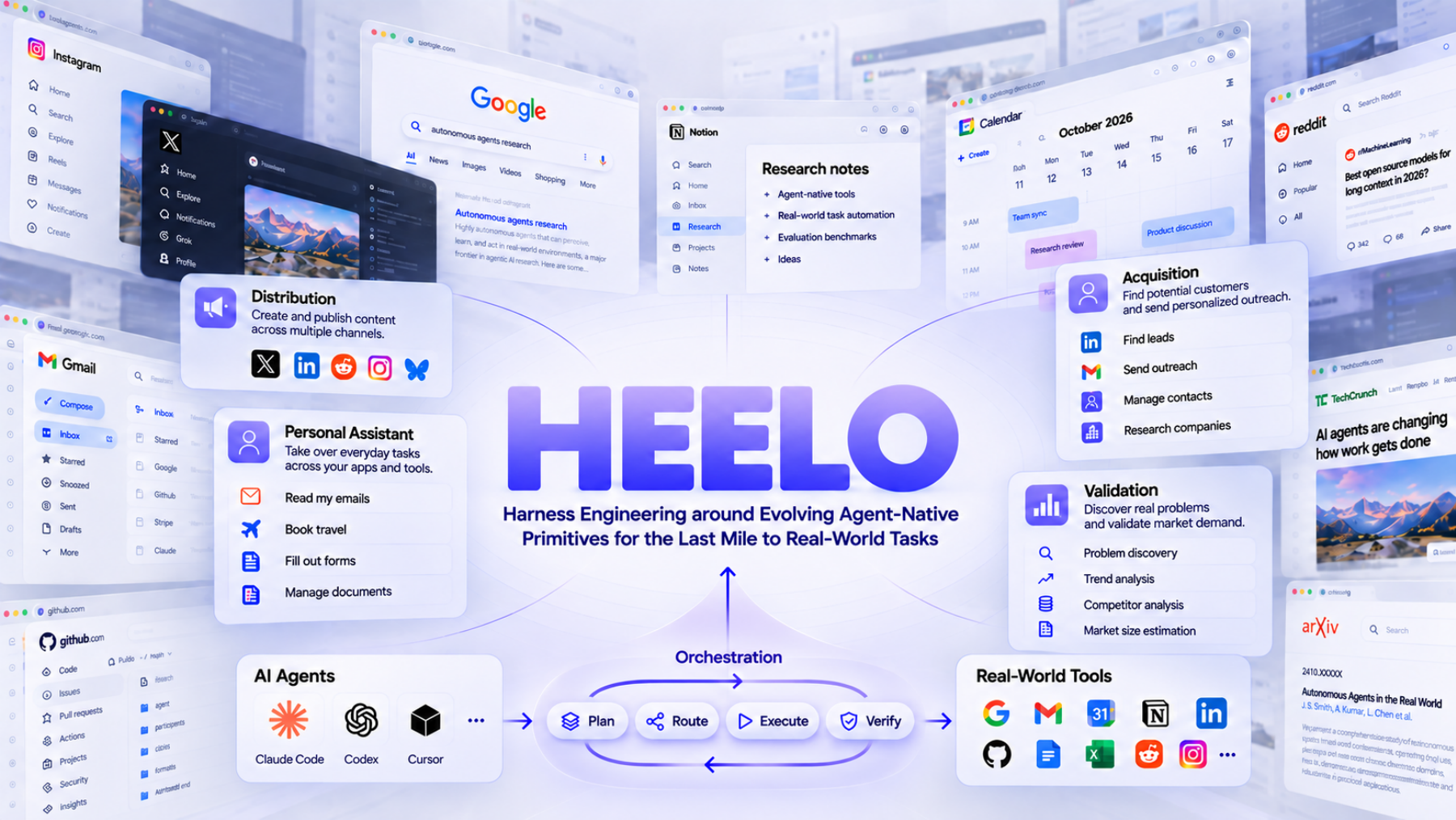}
\caption{\textbf{The heelo.dev platform.} Agents such as Claude Code, Codex, or Cursor reach real-world tools through \ours{}, which orchestrates agent-native primitives for each task: it plans the task, routes each need to a primitive, executes the primitives, and verifies the result, revising the plan when a check fails. Four loops test the platform end to end: distribution creates and publishes content across channels, acquisition finds potential customers and sends personalized outreach, the personal assistant takes over everyday tasks across apps and tools, and validation discovers real problems and validates market demand.}
\label{fig:banner}
\end{figure}

On the research benchmarks of Section~\ref{sec:instantiations}, as in the cross-family protocol of Appendix~\ref{app:evaluation}, the environment is frozen and the grader is fixed in advance. Real-world tasks also run against live services, accounts, and people. To test \ours{} under these conditions, we built heelo.dev (\url{https://heelo.dev}), a platform that exposes real-world actions as agent-native primitives (Figure~\ref{fig:banner}), and we compared agents with and without \ours{} on 46 such tasks.

\suppressfloats[t]

\subsection{Platform Overview}
\label{sec:platform-overview}

heelo.dev is a tool execution backend for AI agents. It exposes third-party actions, such as flights, calendars, email, social posting, company research, and browser automation, as tools behind one invocation endpoint, and each tool follows the agent-native design of Section~\ref{sec:primitive-definition}. A tool accepts the fields of its input schema or a natural-language request and encapsulates its own execution strategy, from LLM calls to browser automation and third-party APIs, and it keeps its state in a session. When a request lacks a detail, the tool returns a clarification question with the session's identifier, and the caller continues the session until the tool returns a result or an error. Callers reach the tools in four ways: direct invocation; a natural-language router that selects a tool from the descriptions and examples of all registered tools; a ReAct loop that chains tools step by step, each result informing the next action; and an MCP server through which external agents, such as Claude Code, connect to \ours{}. Tool calls are recorded in a shared log.\fillpar

\subsection{Four Real-World Loops}
\label{sec:platform-tasks}

\begin{table}[!t]
\centering
\researchtablesetup
% Category and Tools of one loop form a nested two-column block, so that the
% Loop and Task description cells are centred on the whole block. The widths
% are fixed here, since \linewidth inside a cell is the width of that cell. The
% block sits in a p column: its centred tabular then lies on the row's axis.
\edef\loopcatw{\the\dimexpr0.18\linewidth\relax}%
\edef\looptoolw{\the\dimexpr0.49\linewidth-8\tabcolsep\relax}%
\def\looptools#1{\begin{tabular}{@{}>{\raggedright\arraybackslash}p{\loopcatw}>{\raggedright\arraybackslash}p{\looptoolw}@{}}#1\end{tabular}}%
\caption{\textbf{The four real-world loops of heelo.dev.} The end-to-end task of each loop and the tools that it composes, grouped by category. A name that ends in * stands for all tools of a family, such as \texttt{notion.*} for reading, creating, updating, and deleting Notion pages.}
\label{tab:platform-loops}
\begin{tabular}{>{\raggedright\arraybackslash}m{0.095\linewidth}>{\raggedright\arraybackslash}m{0.235\linewidth}p{\dimexpr0.67\linewidth-6\tabcolsep\relax}}
\toprule
\rowcolor{TabHead}
\hdr{Loop} & \hdr{Task description} & \makebox[\dimexpr\loopcatw+2\tabcolsep\relax][l]{\hdr{Category}}\hdr{Tools} \\
\midrule
\addlinespace[2pt]
Distribution & Publish a campaign's landing page, post one text adapted to each platform's rules, render its share cards, and attribute each click to the contact who clicked &
\looptools{
Page publishing & \texttt{inbound.page.publish}, \texttt{inbound.page.audit} \\[2.5pt]
Click attribution & \texttt{inbound.link.wrap}, \texttt{inbound.stats} \\[2.5pt]
Cross-platform posting & \texttt{channel.expand} \\[2.5pt]
Single-platform posting & \texttt{twitter.post}, \texttt{linkedin.post}, \texttt{instagram.post}, \texttt{bluesky.post\_browser} \\[2.5pt]
Share cards & \texttt{media.render.card}} \\
\addlinespace[2pt]
\midrule
\addlinespace[2pt]
Acquisition & Find people or companies with a matching need, verify their contacts, and send them a multi-step sequence from an authenticated domain, checking consent and suppression before every send &
\looptools{
Find leads & \texttt{demand.radar} \\[2.5pt]
Research companies & \texttt{icp.search}, \texttt{icp.export\_results} \\[2.5pt]
Manage contacts & \texttt{contact.resolve}, \texttt{outbound.mail.contact\_*}, \texttt{outbound.mail.consent\_record} \\[2.5pt]
Send outreach & \texttt{outbound.mail.sender\_*}, \texttt{outbound.mail.sequence\_*}, \texttt{outbound.mail.suppress\_*}} \\
\addlinespace[2pt]
\midrule
\addlinespace[2pt]
Assistant & Read an invoice email, extract and check its fields, and, after a person approves, add the calendar event and the Notion page and render a PDF receipt &
\looptools{
Read emails & \texttt{mail.read\_list}, \texttt{mail.read\_get}, \texttt{mail.read\_attachment}, \texttt{mail.read\_watch} \\[2.5pt]
Manage documents & \texttt{doc.extract\_text}, \texttt{doc.extract\_fields}, \texttt{doc.review\_decide}, \texttt{doc.render\_pdf}, \texttt{doc.render\_docx}, \texttt{doc.render\_xlsx} \\[2.5pt]
Calendars and notes & \texttt{google.calendar\_*}, \texttt{outlook\_calendar.*}, \texttt{apple\_calendar.*}, \texttt{notion.*} \\[2.5pt]
Travel and forms & \texttt{kiwi.book\_flight}, \texttt{booking\_com.reserve\_stay}, \texttt{web.fill\_form}} \\
\addlinespace[2pt]
\midrule
\addlinespace[2pt]
Validation & Mine the problems that people report in a domain, size the market for one of them, and test its demand with an experiment that reaches real people through acquisition &
\looptools{
Scraping & \texttt{scrape.structured.extract} \\[2.5pt]
Problem discovery & \texttt{problem.mine} \\[2.5pt]
Trend analysis & \texttt{trends.signals} \\[2.5pt]
Market and competitors & \texttt{market.size.assess} \\[2.5pt]
Demand experiment & \texttt{validation.run.*}} \\
\addlinespace[2pt]
\bottomrule
\end{tabular}
\end{table}

The real-world tasks of the platform are organized as four loops, each with one end-to-end task (Table~\ref{tab:platform-loops}). Each loop adds three tool families to the platform, such as \texttt{mail.read}, \texttt{doc.extract}, and \texttt{doc.render} for the assistant, and draws on tools that the platform already has, such as calendars, Notion, and the posting tools of single platforms. Distribution publishes a campaign's content to several channels, each under its own platform rules, and hosts the pages that capture visitors. Acquisition finds people and companies with a matching need and sends them multi-step outreach from an authenticated domain. The assistant turns an invoice email into checked fields, approved writes, and a receipt, and validation mines the problems that people report and tests the demand for one of them. The loops also feed one another: distribution attributes each click from acquisition's outreach to the contact who clicked and stops that contact's sequence once the contact is captured, and validation reaches real people through acquisition. A loop's task is therefore solved only when the tools compose within a loop and across loops, each tool's result reaching the tool that consumes it. The four tasks need 26 capabilities: distribution, for instance, needs page publishing, click attribution, safety and input checks, multi-platform posting, budget control, share-card rendering, and the launch pipeline that chains them.

\subsection{Evaluation Protocol}
\label{sec:platform-protocol}

The acceptance criterion $R_q$ of a loop's task (Section~\ref{sec:formulation}) is met only when the whole chain runs live: the assistant loop's live runner, for instance, checks the chain against real accounts link by link, each link reported as passed, failed, or not run, and a link that has not run never counts as passed. To measure what the primitives add, we pose the 26 capabilities as 46 tasks that name no heelo.dev tool. The pass criteria of each task are fixed before its first run as facts that a person can check, such as an invoice total, the pages of a PDF, or a refused channel, and the tasks include failure paths, such as a wrong contact identifier that must return a clean error or a page that robots.txt disallows. The tasks also serve as acceptance tests of the primitives: when a run exposes a defect in a primitive, the primitive is fixed and the task runs again, and we report the runs after the fix.

We run two agents, Claude Opus 5.5 with Claude Code and GPT-6.1 Sol with Codex, both at medium reasoning effort, in two arms each: on their own, and connected to \ours{} through the MCP server of heelo.dev, which gives the agent \ours{}'s primitives as tools; in five validation tasks, Codex with \ours{} is also prompted to use them. Each agent keeps the built-in tools that a task allows, from web search and web fetch to file editing and a sandboxed shell. On its own, an agent receives no credentials, and each run starts in an empty directory outside the heelo.dev repository, in a new session unless it follows up an earlier task; Claude Code cannot read this repository, and Codex, whose shell can read outside its directory, runs with its user configuration and connected apps turned off. A task with side effects uses test accounts and allowlisted recipients only, and an agent on its own, which cannot act on any account, is scored by the pass criteria that its work meets, from a ready draft to a script that needs only a credential. A connected agent calls the staging deployment of heelo.dev, or a local build where a task needs what staging lacks, such as a sandbox channel, a second tenant, a CMS site, or a browser for Lighthouse, and the validation tasks run on a local build with the loop's latest fixes. Most tasks run three times per arm under a fixed time limit.

In distribution and the assistant loop, Claude Sonnet 5.5, which is not one of the tested agents, grades the answers blind, with the arm names and heelo.dev's tool names removed, and a person reviews its grades against the run records; acquisition and validation runs are checked against their criteria from the answers or tool results, the artifacts they link, and the request logs. Each run records whether it passes, its wall-clock time and tokens, and its cost at API prices, which for each model call is its tokens multiplied by the model's price per token, with cached input tokens at the cache price. A run scores 1 when it passes. A run that does not pass scores the share of its pass criteria that it meets, or of its target that it reaches, where a task measures one, as in distribution, lead discovery, problem mining, and the six-stage end-to-end validation, and 0 otherwise; a task's score averages its runs. A capability's score averages its tasks, a loop's score the loop's tasks, and the overall score all 46 tasks. A task is done when all its runs pass. An arm's time and cost per completed task are its time and cost summed over 43 tasks, which leave out three tasks whose runs another task already counts, divided by the number of tasks it completes out of 46. In the six-stage end-to-end validation, only the time of the agent's sessions counts, not that of the launch, which runs on without the agent.\fillpar

\subsection{Results}
\label{sec:platform-results}

\begin{researchtable}
\centering
\researchtablesetup
\caption{\textbf{Results on the 46 real-world tasks of heelo.dev.} Task score (\%) over the tasks of each loop (14, 15, 7, and 10) and over all 46 tasks, tasks completed out of 46, and time and cost per completed task, for Claude Opus 5.5 with Claude Code and GPT-6.1 Sol with Codex, both at medium reasoning effort, each on its own and connected to \ours{} through MCP (+\,\ours{}). Time and cost per completed task divide an arm's totals over 43 tasks, which leave out three tasks whose runs another task already counts, by the number of tasks it completes out of 46. Green marks the best and amber the second-best value in each column; lower is better for time and cost.\fillcap}
\label{tab:platform-results}
\fitresearchtable{\begin{tabular}{lcccccccc}
\toprule
\rowcolor{TabHead}
 & \multicolumn{5}{c}{\hdr{Task score (\%)}} & & \multicolumn{2}{c}{\hdr{Per completed task}} \\
\hdrgapfill\cmidrule(lr){2-6}\cmidrule(lr){8-9}
\rowcolor{TabHead}
\multirow{-2}{*}{\hdr{Arm}} & \hdr{Distribution} & \hdr{Acquisition} & \hdr{Assistant} & \hdr{Validation} & \hdr{All} & \multirow{-2}{*}{\hdr{Completed}} & \hdr{Time (min)} & \hdr{Cost (\$)} \\
\midrule
Claude Code & 23 & \second{41} & 0 & 35 & 28 & 9 & 6.3 & 1.24 \\
\tabours \quad + \ours{} & \best{98} & \best{84} & \best{100} & \best{81} & \best{90} & \best{39} & \best{3.4} & 0.80 \\
\midrule
Codex & 14 & 34 & 0 & 46 & 25 & 9 & 8.0 & \second{0.65} \\
\tabours \quad + \ours{} & \second{90} & \best{84} & \second{83} & \second{73} & \second{83} & \second{35} & \second{3.5} & \best{0.42} \\
\bottomrule
\end{tabular}}
\end{researchtable}

\begin{researchtable}
\centering
\researchtablesetup
\caption{\textbf{Results of the 26 capabilities of heelo.dev.} Task score (\%), time per task, and cost per task of Claude Opus 5.5 with Claude Code (CC) and GPT-6.1 Sol with Codex, each on its own and connected to \ours{} through MCP (+H), for each capability of Figure~\ref{fig:heelo-results}(c). Time and cost are means over the capability's tasks, with costs at API prices: each model call's tokens multiplied by the model's price per token, with cached input tokens at the cache price. Shaded columns are the agents connected to \ours{}.\fillcap}
\label{tab:platform-capabilities}
\fitresearchtable{\begin{tabular}{ll*{6}{c>{\columncolor{TabOurs}}c}}
\toprule
\rowcolor{TabHead}
 & & \multicolumn{4}{c}{\hdr{Task score (\%)}} & \multicolumn{4}{c}{\hdr{Time per task (s)}} & \multicolumn{4}{c}{\hdr{Cost per task (\$)}} \\
\hdrgapfill\cmidrule(lr){3-6}\cmidrule(lr){7-10}\cmidrule(lr){11-14}
\rowcolor{TabHead}
\multirow{-2}{*}{\hdr{Loop}} & \multirow{-2}{*}{\hdr{Capability}} & \hdr{CC} & \hdr{+H} & \hdr{Codex} & \hdr{+H} & \hdr{CC} & \hdr{+H} & \hdr{Codex} & \hdr{+H} & \hdr{CC} & \hdr{+H} & \hdr{Codex} & \hdr{+H} \\
\midrule
\multirow{7}{*}{Distribution} & Page publishing & \phantom{00}0 & 100 & \phantom{00}0 & 100 & \phantom{0}16.3 & \phantom{0}156.4 & \phantom{0}58.2 & \phantom{0}92.1 & 0.050 & 0.734 & 0.053 & 0.152 \\
 & Click attribution & \phantom{00}0 & 100 & \phantom{0}17 & 100 & \phantom{0}18.8 & \phantom{0}241.8 & \phantom{0}92.5 & 218.3 & 0.053 & 0.590 & 0.066 & 0.204 \\
 & Safety and input checks & \phantom{00}0 & 100 & \phantom{00}0 & 100 & \phantom{0}12.1 & \phantom{0}104.6 & \phantom{0}54.6 & \phantom{0}97.7 & 0.046 & 0.530 & 0.056 & 0.152 \\
 & Multi-platform posting & \phantom{0}56 & 100 & \phantom{0}39 & \phantom{0}56 & \phantom{0}88.7 & \phantom{0}107.0 & 112.4 & 167.5 & 0.281 & 0.390 & 0.091 & 0.152 \\
 & Budget control & \phantom{00}0 & 100 & \phantom{00}0 & 100 & \phantom{0}55.4 & \phantom{00}98.8 & \phantom{0}51.8 & 108.4 & 0.169 & 0.467 & 0.039 & 0.200 \\
 & Share-card rendering & \phantom{0}77 & \phantom{0}83 & \phantom{00}0 & 100 & 353.6 & \phantom{0}422.7 & \phantom{0}65.1 & 181.6 & 0.535 & 1.145 & 0.057 & 0.195 \\
 & Launch pipeline & \phantom{00}0 & 100 & \phantom{0}33 & 100 & \phantom{0}61.0 & \phantom{0}217.3 & 102.7 & 346.8 & 0.197 & 0.584 & 0.096 & 0.322 \\
\midrule
\multirow{7}{*}{Acquisition} & Outreach pipeline & \phantom{00}0 & 100 & \phantom{00}0 & 100 & \phantom{0}95.0 & \phantom{0}293.0 & 192.8 & 754.2 & 0.287 & 1.733 & 0.148 & 1.776 \\
 & Sequenced sending & \phantom{00}0 & 100 & \phantom{00}0 & 100 & 124.0 & \phantom{00}57.8 & 116.3 & \phantom{0}85.5 & 0.414 & 0.380 & 0.053 & 0.109 \\
 & Deliverability & \phantom{0}75 & 100 & \phantom{0}50 & 100 & 110.0 & \phantom{00}56.3 & \phantom{0}90.8 & 112.4 & 0.363 & 0.407 & 0.075 & 0.158 \\
 & Consent and compliance & \phantom{0}50 & 100 & \phantom{0}50 & 100 & \phantom{0}67.5 & \phantom{00}53.1 & 110.5 & 115.5 & 0.193 & 0.375 & 0.107 & 0.139 \\
 & Contact finding & \phantom{00}0 & 100 & \phantom{00}0 & 100 & 201.0 & \phantom{0}456.1 & 137.5 & 596.5 & 1.125 & 3.045 & 0.381 & 2.843 \\
 & Source coverage and robots.txt & 100 & 100 & 100 & 100 & \phantom{0}97.0 & \phantom{0}134.5 & \phantom{0}59.4 & 227.7 & 0.299 & 0.825 & 0.139 & 0.592 \\
 & Lead discovery & \phantom{00}6 & \phantom{0}20 & \phantom{00}3 & \phantom{0}20 & \phantom{0}88.5 & \phantom{0}199.1 & 211.7 & 244.9 & 0.424 & 1.094 & 0.450 & 0.831 \\
\midrule
\multirow{5}{*}{Assistant} & Mail and attachments & \phantom{00}0 & 100 & \phantom{00}0 & 100 & \phantom{0}10.7 & \phantom{00}25.8 & \phantom{0}16.7 & \phantom{0}24.6 & 0.054 & 0.116 & 0.029 & 0.053 \\
 & Cost routing & \phantom{00}0 & 100 & \phantom{00}0 & 100 & \phantom{0}10.3 & \phantom{00}26.1 & \phantom{0}17.6 & \phantom{0}25.6 & 0.054 & 0.116 & 0.032 & 0.053 \\
 & Invoice fields and checks & \phantom{00}0 & 100 & \phantom{00}0 & 100 & \phantom{0}10.8 & \phantom{00}26.4 & \phantom{0}16.7 & \phantom{0}29.3 & 0.055 & 0.116 & 0.033 & 0.053 \\
 & Receipt rendering & \phantom{00}0 & 100 & \phantom{00}0 & \phantom{00}0 & \phantom{00}7.7 & \phantom{00}18.3 & \phantom{0}13.7 & \phantom{0}20.9 & 0.049 & 0.147 & 0.023 & 0.080 \\
 & Review and approval & \phantom{00}0 & 100 & \phantom{00}0 & \phantom{0}80 & \phantom{00}9.4 & \phantom{00}23.4 & \phantom{0}15.0 & \phantom{0}21.1 & 0.053 & 0.116 & 0.024 & 0.055 \\
\midrule
\multirow{7}{*}{Validation} & Scraping compliance & \phantom{0}50 & 100 & \phantom{0}33 & 100 & \phantom{0}14.6 & \phantom{00}17.7 & \phantom{0}25.3 & \phantom{0}14.5 & 0.049 & 0.072 & 0.029 & 0.024 \\
 & Crawling and extraction & \phantom{0}33 & 100 & \phantom{0}67 & 100 & \phantom{0}43.8 & \phantom{00}44.0 & \phantom{0}41.4 & \phantom{0}31.0 & 0.123 & 0.178 & 0.066 & 0.061 \\
 & Request audit log & \phantom{00}0 & 100 & \phantom{00}0 & 100 & \phantom{0}10.1 & \phantom{00}12.9 & \phantom{0}10.6 & \phantom{00}8.5 & 0.212 & 0.301 & 0.007 & 0.008 \\
 & Problem mining & \phantom{00}0 & \phantom{0}57 & 100 & 100 & 340.9 & \phantom{0}287.0 & 431.6 & 603.3 & 1.218 & 0.560 & 0.822 & 1.272 \\
 & Lead hand-off & \phantom{00}0 & \phantom{00}0 & 100 & \phantom{0}67 & \phantom{0}97.1 & 1200.0 & 677.5 & 662.3 & 0.286 & 2.240 & 1.212 & 1.365 \\
 & Funnel math & 100 & 100 & \phantom{00}0 & \phantom{00}0 & \phantom{0}22.8 & \phantom{00}38.5 & \phantom{0}26.2 & \phantom{0}29.8 & 0.115 & 0.207 & 0.034 & 0.053 \\
 & End-to-end validation & \phantom{0}42 & \phantom{0}75 & \phantom{0}29 & \phantom{0}29 & \phantom{0}47.7 & \phantom{0}768.2 & \phantom{0}67.1 & 105.9 & 0.259 & 3.223 & 0.105 & 0.177 \\
\bottomrule
\end{tabular}}
\end{researchtable}

\textbf{Overall Results.} Connected to \ours{}, Claude Code completes 39 of the 46 tasks instead of 9, and Codex 35 instead of 9, and their overall scores rise from 28\% to 90\% and from 25\% to 83\%; Figure~\ref{fig:heelo-results}(c) and Table~\ref{tab:platform-capabilities} give the score of every capability, and Table~\ref{tab:platform-results} the scores by loop. Of the 26 capabilities, Claude Code completes every task in 2 on its own and in 21 with \ours{}, and Codex in 3 and in 19. On their own, both agents score zero in 12 capabilities, most of which act on an account or a service: publishing a page, sending a sequence, finding a contact, reading a mailbox, or holding an invoice's writes for a person's approval.

\textbf{Distribution and Acquisition.} In distribution, Claude Code scores 98\% with \ours{} and Codex 90\%, against 23\% and 14\% on their own. Claude Code with \ours{} completes every task except the card check over four templates and four sizes, in which one of three runs stops at the time limit, and Codex with \ours{} every task except two of the three multi-platform posts. In acquisition, both agents score 84\% with \ours{}, against 41\% and 34\% on their own, and with \ours{} they complete every task in all capabilities but lead discovery. On their own, the agents score mainly where the open web and their own code suffice, as in multi-platform posting (56\% and 39\%) and source coverage within robots.txt (100\%). Lead discovery, which asks for people who state a matching need on public platforms, is the only capability in which no arm scores above 20\%: both connected agents reach 20\%, and Claude Code and Codex on their own 6\% and 3\%.

\textbf{Assistant.} On their own, both agents score 0\% in the assistant loop, because an agent on its own cannot read the mailbox, the calendar, or Notion. With \ours{}, Claude Code completes every assistant task and passes all 15 runs behind them, from an invoice email to its planned writes, through an invoice whose total does not add up and is held for correction, to the receipt of an approved invoice. Codex with \ours{} completes five of the seven tasks and scores 83\%: in the three runs that end with the receipt of an approved invoice, heelo.dev returns the receipt, but Codex's answer leaves it out. Every link of the loop's chain also passes live against real accounts: the invoice email is read, its eight fields are extracted by the rule-based tier, the arithmetic checks pass, nothing is written until a person approves, and the approval then produces the calendar event, the Notion page, and the PDF receipt.

\textbf{Validation.} In validation, Claude Code scores 81\% with \ours{} and Codex 73\%, against 35\% and 46\% on their own, and they complete 7 and 6 of the 10 tasks, against 2 and 3. Only the connected agents record the status code of every request in a scrape. On a self-contained version of the mining task that asks for problems ranked with their trends, Codex completes the task with and without \ours{}; for Claude Code with \ours{}, heelo.dev's problem miner gives a trend for 8 of the 14 mined problems. On its own, Codex also completes the hand-off of 100 people with a draft of 100 candidate contacts; with \ours{}, it passes two of its three runs, and Claude Code with \ours{} stops at the time limit, because the agent resolves the contacts one at a time, with a call for each. Codex scores zero on the funnel math with and without \ours{}: its answers compute the rate but leave out the caveat of a small sample. Of the two end-to-end tasks, Claude Code with \ours{} passes all 12 runs of the shorter one and Codex with \ours{} 7. In the six-stage end-to-end validation, Claude Code with \ours{} passes stages two to four, up to a launch that finds 110 people at 77 companies, and Codex with \ours{} stops before the assessment, because its fresh scans return no assessable problem.

\textbf{Time and Cost.} In most tasks, a connected run costs more than a run on its own, because it reads more tokens, from the tool menu to the structured results of the primitives: the median assistant run of Claude Code with \ours{} reads 160,701 tokens and takes 23.4 seconds, against 19,456 tokens and 9.4 seconds on its own. Because the connected agents complete most of their tasks, they need less time and lower cost per completed task: Claude Code with \ours{} needs 3.4 minutes and \$0.80, against 6.3 minutes and \$1.24 on its own, and Codex with \ours{} 3.5 minutes and \$0.42, against 8.0 minutes and \$0.65, which is $1.9\times$ and $2.3\times$ less time and $1.55\times$ lower cost. Per completed task, Codex with \ours{} is the cheapest of the four arms. Repeated work also costs less with \ours{}: when a task revisits a JavaScript page whose extraction recipe heelo.dev stored on the first visit, the second visit costs less than half the first in every run of both connected agents, against one of three runs of Codex on its own.\fillpar

Across the four loops, the gap between an agent on its own and the same agent connected to \ours{} opens where a task must act on an account or a service: an agent on its own can find a page, draft a message, or write a script, but a published page, a sent sequence, or an invoice read from a mailbox takes a primitive that holds the account, applies its resource in the task's context, and verifies the outcome before it reports success.

\section{Conclusion}
\label{sec:conclusion}
\runningsection{Conclusion}

How can an agent handle a real-world task end to end? Our answer is to engineer each task's harness around agent-native primitives, which address the last mile to the real world by keeping the competence to apply a resource with the resource itself. An agent-native primitive equips a resource with a resident LLM that applies it to a natural-language request in context and sends other primitives the requirements that its result places on them; where the domain offers an exact procedure, a program takes the place of the resident LLM. \ours{} builds each task's harness from a library of such primitives through three mechanisms. Library selection and composition retrieves and assesses candidates and composes the few that the task needs into a task-specific graph. Evidence-based execution and improvement runs the graph, verifies its result against evidence from the task's environment, and after a failure revises only the primitives that the evidence implicates, within a budget of revision rounds. Validation and publishing admits new or improved primitives into the library only after validation, on a path separate from any task.\fillpar[0.6][1]

Each of the eight primitive families comes from the solution we built for one kind of real-world task, realizes the mechanisms that its task needs, and improves on the reference baseline of its companion study. On the tasks where frontier harnesses fall short, Dev-Primitives lead Codex by 18.2 points on Terminal-Bench~4.0 with GPT-5.6 Sol at maximum effort, and composing Harness Primitives per task raises Terminal-Bench~2.0 Pass@1 from 60.0\% for the seed harness to 72.2\%, above the 56.7\% of Codex. The analyses show that binding inside a Tool Primitive beats budget-matched self-refinement over raw schemas and that the fields a Code Primitive carries account for most of its margin over generic editing. What crosses between agents and primitives must keep its meaning: exchanged KV caches need positional re-encoding, and requirements must reach the components they concern. The task, in turn, decides which primitives run, as active Navigation Primitives grow $1.23\times$ while distractors grow the information space $16\times$ around the same evidence.

The families share one method, so a new kind of task can join \ours{} as a new primitive family that realizes the same definition and mechanisms. Beyond the research benchmarks, heelo.dev exposes real-world actions to agents as agent-native primitives. On its 46 real-world tasks, Claude Code and Codex complete 9 tasks each on their own and 39 and 35 when connected to \ours{} through MCP, with $1.9\times$ and $2.3\times$ less time and $1.55\times$ lower cost per completed task. Overall, primitives let the competence developed for one task outlast that task: once validated and published, the primitive that carries this competence becomes a building block that later tasks compose.\fillpar

\clearpage
\phantomsection
{\small
\runningsection{References}
\bibliographystyle{unsrtnat}
\bibliography{reference}
}
\clearpage
\appendix
\section{\texorpdfstring{Proofs for Section~\ref{sec:analysis}}{Proofs for the Theoretical Guarantees}}
\label{app:proofs}
\runningsection{Proofs}

Section~\ref{sec:analysis} states three results that the companion studies prove about selection: demand-bounded activation in ANTMAN, the comparison of fixed and task-conditioned composition in STITCH, and the coverage and retrieval ceiling in self-advertised method selection~\citep{jin2026antman,jin2026stitch,anonymous2026library}. This appendix proves them in that order. Each proof follows the proof of its study and uses the study's notation, which holds only within the subsection of that result, and a closing paragraph records what each premise contributes. Section~\ref{sec:analysis} states the same results in the notation of the model, with the following correspondence. For demand-bounded activation, ANTMAN's territory-backed workers are the primitives, its information needs are the needs that COMPOSE routes, its realized need count $H(q)$ is $D(q)$, and its number of territories $M$ is $|\Lib_q|$. For task-conditioned composition, STITCH's task $x$, task distribution $\mathcal{D}$, primitive $p_i$, and library size $M$ are $q$, $\mathcal{Q}$, $P_i$, and $N$; its utility $u_i$ and penalty $c_i$ are $g_i$ and $\kappa_i$; its activation vector $\mathbf{a}$ is $\mathbf{z}$; and its validity indicator $V(H)$ is $\mathbf{1}[H\text{ executes validly}]$. For the coverage results, the study's Lean goal $g$ with its local context, its contract descriptions, and its contract library $\mathcal{C}$ are the task $q$ with its context, the interfaces, and $\Lib$; its set $A(g)$ of annotated reference contracts is $\mathcal{U}_q$; its information variable $F$ is $Z$; and its top-$m$ similarity pool $P_m(g)$ is $\mathcal{S}_m(q)$.\fillpar

\subsection{Demand-Bounded Activation}

The proof follows ANTMAN's proof of its Proposition~1~\citep{jin2026antman}, in ANTMAN's notation (Appendix~\ref{sec:navigation-mechanism}). At coordination step $t$, ANTMAN selects one unresolved need $n_t$ of the Need Graph $G_t=(N_t,\Delta_t)$ and routes it to the territory-backed worker $i_t$ (Eq.~\eqref{eq:need-route}). The realized need set $\mathcal{H}(q)=\bigcup_{t=0}^{T_q}N_t$ of Eq.~\eqref{eq:navigation-realized}, with $H(q)=|\mathcal{H}(q)|$, collects every need that the graph holds during execution, and a reframed or newly introduced need counts as a new member; it is distinct from the attempt history $\mathcal{H}_t(n)$ that each need carries in Eq.~\eqref{eq:need-state}. For a realized need $n$, $a(n)$ counts its routing attempts, its initial routing and every rerouting caused by recovery, with $a(n)\le\rho$ under the execution budget, and $\mathcal{A}(q)=\bigcup_{t=0}^{T_q-1}\{i_t\}$ is the set of distinct workers activated (Eq.~\eqref{eq:navigation-dispatch}).

\begin{proof}[Proof of Proposition~\ref{prop:activation}]
Every worker activation in Algorithm~\ref{alg:navigation} occurs through a routing decision at Line~7, and each such decision is associated with the one need $n_t$ selected at that step, which belongs to $N_t\subseteq\mathcal{H}(q)$. Summing the routing events of all needs encountered during execution therefore counts every worker-dispatch event once:
\[
K(q)=\sum_{n\in\mathcal{H}(q)}a(n).
\]
A single routing event activates at most one territory-backed worker. Mapping each activated worker to the first routing event that activated it therefore sends distinct workers to distinct events, so the number of distinct workers activated cannot exceed the number of routing events, $|\mathcal{A}(q)|\le K(q)$. Each need is routed at most $\rho$ times, so
\[
K(q)=\sum_{n\in\mathcal{H}(q)}a(n)\ \le\ \sum_{n\in\mathcal{H}(q)}\rho\ =\ \rho\,|\mathcal{H}(q)|\ =\ \rho\,H(q).
\]
Combining the two inequalities gives $|\mathcal{A}(q)|\le K(q)\le\rho H(q)$. ANTMAN contains only $M$ territory-backed workers, so $|\mathcal{A}(q)|\le M$, and together $|\mathcal{A}(q)|\le\min\{M,\rho H(q)\}$, which is Eq.~\eqref{eq:activation}.
\end{proof}

\textbf{Role of the Premises.}
$M$ enters only as the size of the available candidate pool and not as a factor of the demand-dependent bound $\rho H(q)$, so enlarging the information space and increasing $M$ does not by itself force additional workers to become active. If realized demand stays bounded, $H(q)\le\bar H$, and $\rho$ does not depend on $M$, then $|\mathcal{A}(q)|\le\rho\bar H=O(1)$ with respect to $M$, which is ANTMAN's Corollary~1. A harder query can still produce more needs, revisions, or recovery attempts and so raise $H(q)$ or $K(q)$; the proposition attributes this growth to task demand. The argument uses three properties of ANTMAN's coordination: every activation goes through a routing decision on one selected need, a routing event activates at most one worker, and each need is routed at most $\rho$ times, with a reframed need counted as a new need. These are the properties that Section~\ref{sec:analysis-activation} requires of a COMPOSE that activates a primitive only by routing an unresolved need to it. The bound concerns active coordination, that is, whether worker participation is structurally coupled to the number of available territories; the retrieval stage that ranks WorkerCards for each routing decision is a separate cost, which can grow with the information space.

\subsection{Fixed versus Task-Conditioned Composition}

The proof follows STITCH's proofs of its Proposition~1 and of part~1 of its Proposition~2~\citep{jin2026stitch}, in STITCH's notation (Section~\ref{sec:analysis-adaptive} and Appendix~\ref{sec:harness-mechanism}). Expectations include task sampling and any randomness in primitive selection. Because the additive expression lies in $[0,1]$ for every task and activation vector, $\Delta_i(x)=R(x,\mathbf{e}_i)-R(x,\mathbf{0})$, where $\mathbf{e}_i$ activates $p_i$ alone and $\mathbf{0}$ activates no primitive, lies in $[-1,1]$, and every expectation below exists.

\begin{proof}[Proof of Proposition~\ref{prop:fixed-vs-adaptive}]
\emph{Part~(i).} Under the pointwise sign assumptions, $\Delta_i(x)=u_i(x)>0$ for $x\in\Omega_i$ and $\Delta_i(x)=-c_i(x)<0$ otherwise. Since $R(x,\mathbf{a})$ is additive and every combination is admissible, one oracle activation that maximizes $R(x,\mathbf{a})$ on each task is
\[
a_i^*(x)=\mathbf{1}(x\in\Omega_i).
\]
The expected performance of the oracle task-adaptive harness is
\[
\mathbb{E}_{x\sim\mathcal{D}}\bigl[R(x,\mathbf{a}^*(x))\bigr]=\mathbb{E}[R_0(x)]+\sum_{i=1}^{M}\mathbb{E}_x\bigl[\mathbf{1}(x\in\Omega_i)\,u_i(x)\bigr]=\mathbb{E}[R_0(x)]+\sum_{i=1}^{M}\mu_i\bar u_i .
\]
For a fixed harness $\mathbf{a}\in\{0,1\}^M$, $\mathbb{E}_{x\sim\mathcal{D}}[R(x,\mathbf{a})]=\mathbb{E}[R_0(x)]+\sum_{i=1}^{M}a_i\,\mathbb{E}_x[\Delta_i(x)]$, and the law of total expectation gives, for each $i$,
\[
\mathbb{E}_x[\Delta_i(x)]=\Pr(x\in\Omega_i)\,\mathbb{E}[u_i(x)\mid x\in\Omega_i]-\Pr(x\notin\Omega_i)\,\mathbb{E}[c_i(x)\mid x\notin\Omega_i]=\mu_i\bar u_i-(1-\mu_i)\bar c_i .
\]
The sum separates over $i$, so the optimal fixed harness sets $a^*_{i,\mathrm{fixed}}=1$ if $\mu_i\bar u_i>(1-\mu_i)\bar c_i$ and $a^*_{i,\mathrm{fixed}}=0$ otherwise, and
\begin{filldisp}[.6][1]\[
\mathbb{E}_{x\sim\mathcal{D}}\bigl[R(x,\mathbf{a}^*_{\mathrm{fixed}})\bigr]=\mathbb{E}[R_0(x)]+\sum_{i=1}^{M}\max\bigl(0,\ \mu_i\bar u_i-(1-\mu_i)\bar c_i\bigr).
\]\end{filldisp}
Subtracting this from the expected performance of the oracle gives $\Gamma_{\mathrm{fixed}}=\sum_{i=1}^{M}\bigl(\mu_i\bar u_i-\max(0,\ \mu_i\bar u_i-(1-\mu_i)\bar c_i)\bigr)$, and the identity $y-\max(0,y-z)=\min(y,z)$ with $y=\mu_i\bar u_i$ and $z=(1-\mu_i)\bar c_i$ gives the first line of Eq.~\eqref{eq:fixed-gap}. Because $\mu_i\in(0,1)$, $\bar u_i>0$, and $\bar c_i>0$, each term of the sum is strictly positive, so $\Gamma_{\mathrm{fixed}}>0$.

\emph{Part~(ii).} Write $A=\mathbb{E}_x[R(x,\mathbf{a}^*(x))]$ and $B=\mathbb{E}[J(x,H_{\mathrm{comp}})]$. The final activation $a_i$ of $p_i$ in $H_{\mathrm{comp}}$ depends on the task and on the composer's randomness. Conditioning on the application scope and using the definitions of $\alpha_i$ and $\beta_i$,
\begin{filldisp}\[
\mathbb{E}\bigl[a_i\Delta_i(x)\bigr]=\mu_i\,\mathbb{E}\bigl[a_iu_i(x)\mid x\in\Omega_i\bigr]-(1-\mu_i)\,\mathbb{E}\bigl[a_ic_i(x)\mid x\notin\Omega_i\bigr]=\mu_i(1-\beta_i)\,\bar u_i-(1-\mu_i)\,\alpha_i\,\bar c_i ,
\]\end{filldisp}
because $\mathbb{E}[a_iu_i(x)\mid x\in\Omega_i]=\bar u_i-\mathbb{E}[(1-a_i)u_i(x)\mid x\in\Omega_i]=(1-\beta_i)\,\bar u_i$. The compositions execute validly, $V(H_{\mathrm{comp}})=1$, so $J(x,H_{\mathrm{comp}})=R(x,\mathbf{a})$ at the final activations, and summing over $i$ under the additive model gives
\begin{filldisp}\[
B=A-\sum_{i=1}^{M}\bigl[\mu_i\beta_i\bar u_i+(1-\mu_i)\alpha_i\bar c_i\bigr].
\]\end{filldisp}
As in STITCH, the best static harness $H^*_{\mathrm{fixed}}$, which activates $\mathbf{a}^*_{\mathrm{fixed}}$ on every task, executes validly, so part~(i) gives $\mathbb{E}[J(x,H^*_{\mathrm{fixed}})]=A-\Gamma_{\mathrm{fixed}}$. Comparing the two expressions, $B>A-\Gamma_{\mathrm{fixed}}$ if and only if the utility-weighted selection loss is smaller than $\Gamma_{\mathrm{fixed}}$, which is the second line of Eq.~\eqref{eq:fixed-gap}.
\end{proof}

\textbf{Role of the Premises.}
Additivity is the modeling assumption: it abstracts away interactions between primitives, so it excludes primitives that help only together or that interfere. Admissibility of every combination lets the maximum over fixed harnesses separate over primitives, and the range $[0,1]$ keeps every expectation finite. Part~(i) compares two oracle values. Part~(ii) adds valid execution, $V(H_{\mathrm{comp}})=1$, which STITCH assumes for compositions of validated primitives built by its deterministic compiler; every selected primitive activated and executed without runtime errors, an activation rate of 100\% that STITCH reads as support for part~2 of its Proposition~2 (Figure~\ref{fig:harness-overhead}). The composer need not be perfect, since its selection loss need only stay below $\Gamma_{\mathrm{fixed}}$. Each $a_i$ is the final activation of mechanism $p_i$, after selection, bounded correction, and fallback to the seed harness, and the dependencies that the compiler supplies belong to the mechanism that requires them, as in STITCH's fixed arms. Part~2 of STITCH's Proposition~2, the comparison with raw code generation under independent implementation failures, is Eq.~\eqref{eq:harness-generation}, proved in Appendix~\ref{sec:harness-mechanism}. Each fixed arm of Section~\ref{sec:harness-experiments} applies one focused primitive with its required dependencies and shared supporting operations to every task. It is one of the fixed activations over which the maximum in Eq.~\eqref{eq:fixed-gap} ranges, so the best fixed arm bounds the value of the best fixed harness from below.

\subsection{Coverage and Retrieval Ceiling}

The three parts of Proposition~\ref{prop:coverage} restate the information bound (Proposition~2), the candidate-pool ceiling (Corollary~1), and the multi-method coverage result (Proposition~4) of the study of self-advertised method selection~\citep{anonymous2026library}. The proofs follow the study's appendix, in its notation (Section~\ref{sec:analysis-coverage} and Appendix~\ref{sec:method-mechanism}). A Method Contract makes its own guarantee conditional on explicit premises, as program contracts do~\citep{hoare1969axiomatic,meyer1992contract}: once $\Pre_c(\theta)$ and $\Obl_c(\theta)$ are proved for a binding $\theta$, its action yields $G_c(\theta)$ (Proposition~\ref{prop:math-scaffold}). The three parts proved here concern selection instead, that is, which contracts a selector can place in its shortlist. Throughout, $X=(g,\Gamma,\text{contract descriptions})$ is the selector's input, the nonempty annotated set $A=A(g)$ is random given $X$, and $T=\phi(X)$ holds the features available to a similarity-only selector.

\begin{proof}[Proof of Proposition~\ref{prop:coverage}]
\emph{Part~(i).} For each value $t$ of $T$, choose a set $S(t)$ with $|S(t)|=k$ that attains the maximum in the definition of $V_k(T)$; a maximizer exists because $\mathcal{C}$ is finite. Then
\[
V_k(T)=\mathbb{E}\bigl[\Pr(S(T)\cap A\neq\varnothing\mid T)\bigr]=\Pr\bigl(S(T)\cap A\neq\varnothing\bigr)=\mathbb{E}\bigl[\Pr(S(\phi(X))\cap A\neq\varnothing\mid X)\bigr],
\]
where the middle equality is the tower property and the last one uses that $S(\phi(X))$ is a function of $X$. For every $x$, the set $S(\phi(x))$ is one of the sets over which the inner maximum in $V_k(X)$ is taken, so $\Pr(S(\phi(x))\cap A\neq\varnothing\mid X=x)\le\max_{|S|=k}\Pr(S\cap A\neq\varnothing\mid X=x)$. Taking expectations over $X$ gives $V_k(T)\le V_k(X)$. For strictness, fix $n$ with $k<n\le|\mathcal{C}|$, and let $X$ be uniform over $n$ distinct values $x_1,\dots,x_n$ with $\phi(x_i)=t$ for every $i$. Let $c_1,\dots,c_n$ be distinct contracts and set $A=\{c_i\}$ whenever $X=x_i$. Given $X=x_i$, any shortlist containing $c_i$ hits with probability one, so $V_k(X)=1$. Given $T=t$, which carries no information about $i$, a shortlist $S$ hits with probability $\frac{1}{n}|S\cap\{c_1,\dots,c_n\}|\le k/n$, so $V_k(T)=k/n<1$.

\emph{Part~(ii).} By construction, the first $m$ positions of the final list are a permutation of $P_m(g)$, and the remaining positions hold $\mathcal{C}\setminus P_m(g)$. For $k\le m$, therefore, $\pi_k(g)\subseteq P_m(g)$, and hence $\pi_k(g)\cap A(g)\subseteq P_m(g)\cap A(g)$. Taking the indicator of nonemptiness of both sides, and their cardinalities divided by $|A(g)|$, gives the two inequalities of Eq.~\eqref{eq:coverage-ceiling}. Their right-hand sides are the first-stage selector's own hit@$m$ and recall@$m$ on $g$, since $P_m(g)$ is its top-$m$ list, and averaging over problems preserves both inequalities. The same containment bounds the reciprocal rank: if $P_m(g)\cap A(g)=\varnothing$, the first annotated contract appears at position $m+1$ or later, so $1/\operatorname{rank}(g)\le1/(m+1)$.

\emph{Part~(iii).} Fix a realization of $A$ and let $h_A(S)=\mathbf{1}[S\cap A\neq\varnothing]$. If $S\subseteq S'$, then $S\cap A\subseteq S'\cap A$, so $h_A(S)\le h_A(S')$, and $h_A$ is monotone. For $c\notin S'$, the marginal gain of $c$ at $S$ is
\[
h_A(S\cup\{c\})-h_A(S)=\mathbf{1}[c\in A]\cdot\mathbf{1}[S\cap A=\varnothing],
\]
which is nonincreasing in $S$ because $\mathbf{1}[S'\cap A=\varnothing]\le\mathbf{1}[S\cap A=\varnothing]$ when $S\subseteq S'$, so $h_A$ is submodular. Since $H_x(S)=\mathbb{E}[h_A(S)\mid X=x]$ is an average of the functions $h_A$ with nonnegative weights, and both properties are preserved under such averages, $H_x$ is monotone and submodular, with $H_x(\varnothing)=0$. The greedy guarantee is the classical bound for monotone submodular maximization under a cardinality constraint~\citep{nemhauser1978analysis}. Let $S^{\mathrm{opt}}$ maximize $H_x$ over sets of size $k$, and let $S_i$ be the greedy set after $i$ steps, with $S_0=\varnothing$. For each $i<k$, monotonicity and submodularity give
\begin{filldisp}\begin{align*}
H_x(S^{\mathrm{opt}})
&\le H_x\bigl(S_i\cup S^{\mathrm{opt}}\bigr)\\
&\le H_x(S_i)+\sum_{c\in S^{\mathrm{opt}}\setminus S_i}\bigl[H_x(S_i\cup\{c\})-H_x(S_i)\bigr]\\
&\le H_x(S_i)+k\bigl[H_x(S_{i+1})-H_x(S_i)\bigr],
\end{align*}\end{filldisp}
where the last step uses $|S^{\mathrm{opt}}\setminus S_i|\le k$ and the fact that the greedy step picks the largest marginal gain. Writing $\delta_i=H_x(S^{\mathrm{opt}})-H_x(S_i)$, this rearranges to $\delta_{i+1}\le(1-1/k)\,\delta_i$, so $\delta_k\le(1-1/k)^k\,\delta_0=(1-1/k)^k\,H_x(S^{\mathrm{opt}})$. Hence $H_x(S_k)\ge\bigl(1-(1-1/k)^k\bigr)H_x(S^{\mathrm{opt}})\ge(1-1/e)\,H_x(S^{\mathrm{opt}})$, using $(1-1/k)^k\le e^{-1}$.
\end{proof}

\textbf{Role of the Premises.}
Part~(i) is a statement about information and leaves open whether a particular model extracts it. An LLM reranker reads full descriptions, but only of the contracts that a similarity stage passes to it, so similarity still determines what the reranker can consider. Part~(ii) requires only that the second stage reorder the pool and place the remaining contracts after it, so it holds for every reranking backbone. The study's rerankers follow this design with $m=10$, and every reported $k$ satisfies $k\le m$, so an annotated contract that similarity ranks low cannot be recovered by any of them (Figure~\ref{fig:method-ranks}). Self-advertisement judges all 82 contracts from their proposals, and its coverage is limited instead by the information in $X$ and by how accurately the model judges applicability. Part~(iii) concerns expected hit@$k$. Expected recall@$k$ is additive over contracts, $R_x(S)=\sum_{c\in S}w_c(x)$ with $w_c(x)=\mathbb{E}\bigl[\mathbf{1}[c\in A]/|A|\bigm| X=x\bigr]$, so sorting by $w_c(x)$ maximizes it, and the two objectives coincide when a goal has a single reference contract. When several methods apply, a near-duplicate of a contract that is likely to apply adds little to expected hit@$k$, so sorting per-contract scores need not maximize it. The study sorts per-contract scores, which is optimal for expected recall@$k$, and part~(iii) identifies the objective that a diversity-aware selector would target.\fillpar[0.6][1]

\section{Framework Details of the Eight Primitive Families}
\label{app:families}
\runningsection{Framework details}

This appendix gives the full framework of each primitive family, one subsection per family and in the order of Section~\ref{sec:instantiations}: Tool Primitives and HEART~\citep{jin2026toolprimitives}, Code Primitives and LEGO~\citep{jin2026codeprimitives}, Math Primitives and self-advertised method selection in Lean~\citep{anonymous2026library}, Knowledge Primitives and SourceLearn~\citep{jin2026sourcelearn}, Agent Primitives~\citep{jin2026agentprimitives}, Dev-Primitives and HERMES~\citep{jin2026devprimitives}, Navigation Primitives and ANTMAN~\citep{jin2026antman}, and Harness Primitives and STITCH~\citep{jin2026stitch}. Each subsection gives one family's framework in the notation of its own paper, with the paper's equations, its algorithm where the paper has one, and its implementation specifics, such as models, budgets, and retrieval sizes. The symbols of a subsection hold only within that subsection, and each subsection ends with a short mapping of its framework to the method of Section~\ref{sec:method}.

\subsection{Tool Primitives: Natural-Language Invocation and Harness Engineering}
\label{sec:tool-mechanism}

\begin{figure}[!htbp]
\centering
\includegraphics[width=\linewidth]{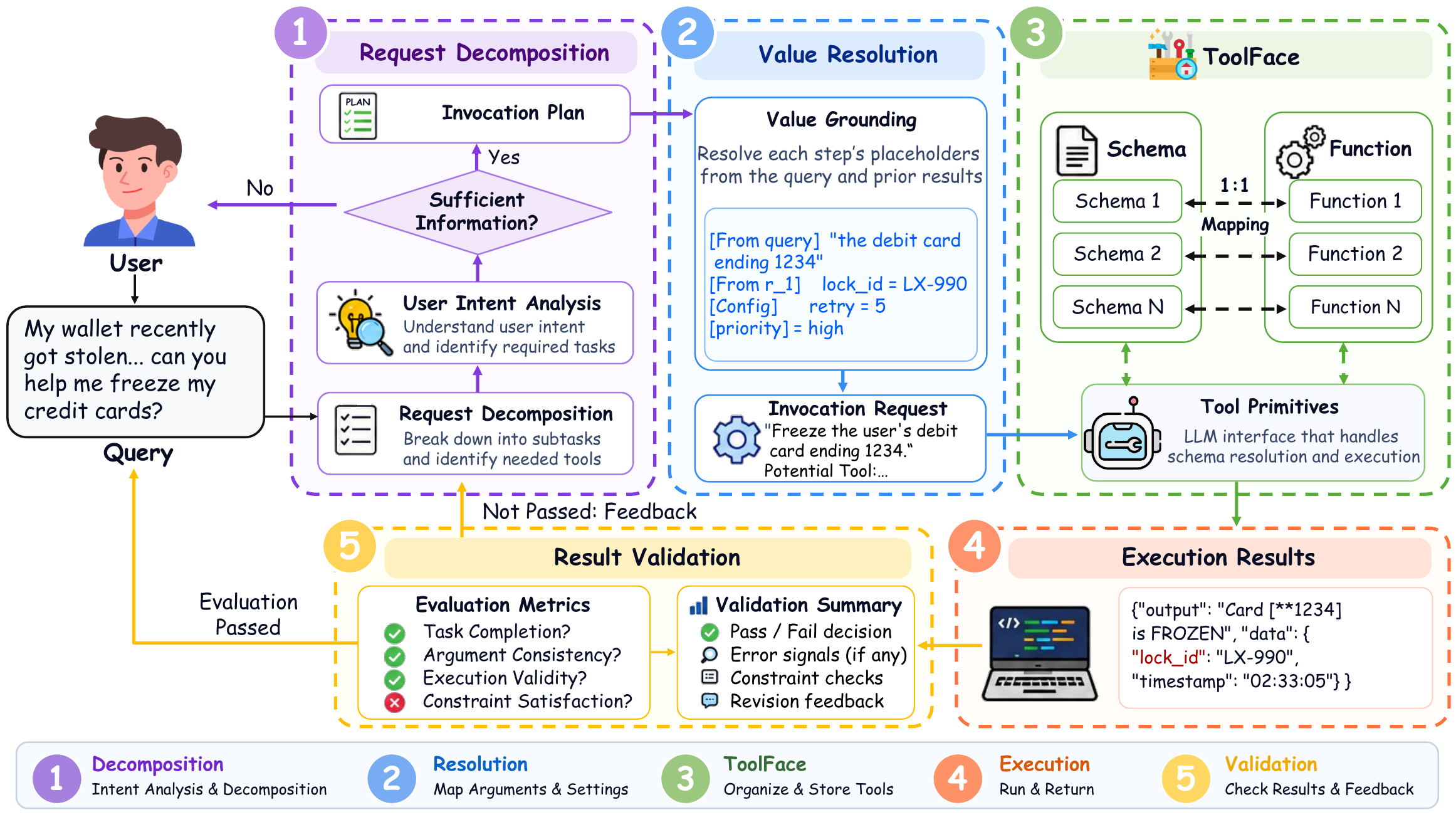}
\caption{\textbf{Overview of HEART.} \textbf{(1)~Request decomposition} analyses the user's intent and checks whether the context supports a plan, soliciting missing details if not. \textbf{(2)~Value resolution} grounds each step's placeholders from the query and from prior results into a natural-language invocation request. \textbf{(3)~ToolFace} stores tools as schema--function pairs, each wrapped as a Tool Primitive that resolves the schema internally. \textbf{(4)~Execution} returns the structured result, which \textbf{(5)~result validation} checks against task completion, argument consistency, execution validity, and constraint satisfaction; on failure it returns feedback that triggers a targeted revision~\citep{jin2026toolprimitives}.}
\label{fig:tool-overview}
\end{figure}

Tool Primitives give each tool a natural-language interface~\citep{jin2026toolprimitives}, and HEART, a \textbf{H}arness \textbf{E}ngineering framework via \textbf{A}gent-native, \textbf{R}eusable \textbf{T}ool Primitives, makes these primitives practical at repository scale. With ToolFace storing the tools, the primitives representing them, and the harness supplying what this abstraction leaves open, HEART decouples tool storage, representation, and invocation, which prior methods couple by exposing raw schemas in the model's context. In Figure~\ref{fig:tool-overview}, an illustrative request to freeze the cards in a stolen wallet passes through these parts, from the clarification that request decomposition asks for to the validation of each step.

\textbf{ToolFace.} ToolFace is a centralized registry $\mathcal{T}=\{(s_i,f_i)\}_{i=1}^{N}$ of schema--function pairs. Each manually authored schema $s_i$ gives the parameter types, constraints, and return specifications of tool $i$, and $f_i$ is its executable implementation. The 25,519 tools comprise 16,464 live APIs from ToolBench~\citep{qin2023toolllm}, 40 mathematical reasoning and 4,348 coding tools from NESTFUL~\citep{basu2025nestful}, 68 tools from $\tau^2$-Bench~\citep{barres2025tau}, 4,538 bilingual APIs from ACEBench~\citep{chen2025acebench}, 2 web-search and 5 memory tools from BFCLv4~\citep{patil2025bfcl}, and 54 manually crafted interaction tools such as form filling. Semantic search over schema descriptors retrieves tools on demand, so the model need not know the catalogue upfront and context pressure follows the number of candidates retrieved per step. A Tool Primitive is defined by the tool it wraps, so the set of primitives is induced by ToolFace and grows with it, and invocations chain along the data dependencies between calls.

\textbf{Tool Primitives.} A Tool Primitive is an LLM-based wrapper around one ToolFace tool, and all Tool Primitives share one base LLM $\mathcal{M}$, conditioned on different schemas:
\begin{equation}
\mathcal{P}_i(x;\,c)=\mathcal{M}\bigl([\,s_i;\,c;\,x\,]\bigr),
\label{eq:tool-primitive}\csname ltx@label\endcsname{eq:tool-exec}
\end{equation}
where $x$ is the natural-language request of the upstream caller, $c$ is optional context such as a prior primitive's result, empty when there is none, and $[\cdot]$ denotes prompt concatenation. $\mathcal{P}_i$ maps each value that the request describes to the argument space of $s_i$, with type coercion, enum matching, unit conversion, required-field, range, and format checks, schema defaults, and binding of fields referenced in the upstream context; it then invokes $f_i$ and returns the execution result $r_i$. On success, $r_i$ holds the tool name, the bound arguments, the function's result, and a one-sentence summary that lets it serve as the context $c$ of a downstream primitive. A failed schema resolution or a runtime error is not retried silently but returned as a structured failure of type \texttt{SCHEMA\_RESOLUTION}, \texttt{CONSTRAINT\_VIOLATION}, or \texttt{RUNTIME\_ERROR}, with the parameter or constraint that failed. The resulting properties are \emph{natural-language invocation}, \emph{inter-tool communication}, and \emph{execution isolation}; under the last, a primitive never calls another tool, modifies the plan, or recovers from a failure on its own.

\textbf{Harness Engineering.} HEART supplies what the abstraction leaves undetermined with three mechanisms.

\emph{Request decomposition} is needed because a primitive answers the request it is handed and cannot ask what the user wants. Intent analysis identifies the target task and its objective, the relevant tool categories, and the user's explicit constraints or preferences. Information sufficiency checking then decides whether the context $\mathcal{C}_t$ of query, acquired information, and execution history supports a fully specified plan:
\begin{equation}
\delta_t=\textsc{Decompose}(q,\,\mathcal{C}_t)\in\{\texttt{sufficient},\,\texttt{insufficient}\}.
\label{eq:tool-compose}\csname ltx@label\endcsname{eq:tool-harness}
\end{equation}
While $\delta_t=\texttt{insufficient}$, decomposition returns to the user a targeted clarification request $c_t$ for the missing information, and the reply enters the context. It then writes the invocation plan $\Pi=(\pi_1,\dots,\pi_K)$, whose steps give a tool category, a tool hint, the step's objective, and the steps it depends on.

\emph{Value resolution} is needed because a step's values come from the user query and from earlier results, which the primitive does not see. It resolves the values of step $\pi_k$ from $\mathcal{C}_t$ into a natural-language invocation request $x_k$ that names the target Tool Primitive and embeds the values inline, together with execution hyperparameters:
\begin{equation}
\beta_k=\textsc{Resolve}(\pi_k,\,\mathcal{C}_t)=\bigl(x_k,\,\texttt{H-Params}_k\bigr).
\label{eq:tool-resolve}
\end{equation}
Values may come from the query, earlier clarifications, or the result of any upstream step, referenced as \texttt{\$step\_j.\allowbreak<field>}. Value resolution never invents an ungrounded value, never overrides the primitive chosen by decomposition, and resolves a value shared by several steps once. $\texttt{H-Params}_k$ sets \texttt{retry} in $[1,5]$, \texttt{priority} to \texttt{low}, \texttt{normal}, or \texttt{high}, and \texttt{timeout\_s}, 30 seconds by default. The step's primitive then runs with the previous result as its context, $r_k\leftarrow\mathcal{P}_k(\beta_k\mid r_{k-1})$.

\emph{Result validation} is needed because a primitive reports whether its own call executed, not whether the result answers the query. It decides
\begin{equation}
v_k=\textsc{Validate}(r_k,\,\pi_k,\,s_k)\in\{\texttt{pass},\,\texttt{fail}\}
\label{eq:tool-validation}
\end{equation}
on four criteria, each assessed independently: \emph{task completion} of the step's objective, \emph{argument consistency} with the user's intent and the constraints of $s_k$, \emph{execution validity}, without runtime errors and with output of the expected schema, and \emph{constraint satisfaction} of task-level or domain-specific constraints such as rate limits, access permissions, or business rules. A step passes if all four hold and fails otherwise; execution then proceeds to $\pi_{k+1}$ or, after $\pi_K$, returns the final result. On a failure, $\phi_k\leftarrow\textsc{Validate}.\textsc{Feedback}(r_k,\pi_k,s_k)$ names each failed criterion and why, with the root cause first.

\emph{Targeted revision} returns this feedback to decomposition, which re-plans from the enriched context:
\begin{equation}
\mathcal{C}_{t+1}=\mathcal{C}_t\cup\{\phi_k\},\qquad\Pi'=\textsc{Decompose}(q,\,\mathcal{C}_{t+1}).
\label{eq:tool-replan}\csname ltx@label\endcsname{eq:tool-validate}
\end{equation}
The revised plan addresses the diagnosed failure, so a failed call is re-planned instead of terminating the trajectory, and recovery stays confined to the step that failed validation. Revision repeats for at most $B$ rounds, after which the current result is returned (Algorithm~\ref{alg:heart}).

\begin{algorithm}[!htbp]
\caption{End-to-end execution flow of HEART~\citep{jin2026toolprimitives}}
\label{alg:heart}
\begin{algorithmic}[1]
\Require user query $q$, ToolFace $\mathcal{T}$, maximum revision budget $B$
\Ensure final execution result or structured failure report
\State $\mathcal{C}\leftarrow\{q\}$, $b\leftarrow 0$
\While{$\textsc{Decompose}(q,\mathcal{C})=\texttt{insufficient}$}
  \State $c\leftarrow\textsc{Decompose}.\textsc{Clarify}(q,\mathcal{C})$;\quad $\mathcal{C}\leftarrow\mathcal{C}\cup\{\textsc{User}(c)\}$
\EndWhile
\State $\Pi\leftarrow\textsc{Decompose}.\textsc{Plan}(q,\mathcal{C})$
\While{$b\le B$}
  \For{each step $\pi_k\in\Pi$}
    \State $\beta_k\leftarrow\textsc{Resolve}(\pi_k,\mathcal{C})$;\quad $r_k\leftarrow\mathcal{P}_k(\beta_k\mid r_{k-1})$
    \If{$\textsc{Validate}(r_k,\pi_k,s_k)=\texttt{fail}$}
      \State $\phi_k\leftarrow\textsc{Validate}.\textsc{Feedback}(r_k,\pi_k,s_k)$;\quad $\mathcal{C}\leftarrow\mathcal{C}\cup\{\phi_k\}$
      \State $\Pi\leftarrow\textsc{Decompose}.\textsc{Plan}(q,\mathcal{C})$;\quad $b\leftarrow b+1$;\quad \textbf{break}
    \EndIf
  \EndFor
  \State \textbf{if} all $K$ steps passed validation \textbf{then return} $r_K$
\EndWhile
\State \Return $\textsc{FailureReport}(\mathcal{C})$
\end{algorithmic}
\end{algorithm}

\textbf{Implementation.} HEART is training-free, unlike tool-calling models fine-tuned on curated trajectories~\citep{lin2024hammer,liu2024toolace,liu2024apigen}. Qwen3-8B~\citep{qwen3technicalreport} runs decomposition, value resolution, result validation, and the Tool Primitives, which share its weights and differ in their system prompts. All experiments use $B=3$ and retrieval over the full repository, with results averaged over three runs, and over eight on $\tau^2$-Bench. A backbone study on $\tau^2$-Bench separates the orchestration layer of the three mechanisms from the execution layer of the Tool Primitives: with GPT-5.4 in the first and Qwen3-8B in the second, HEART reaches Pass\textsuperscript{1} of 0.84, 0.75, and 0.67 on retail, airline, and telecom, close to the 0.86, 0.78, and 0.68 of GPT-5.4 in all roles, so the two layers can be scaled independently. The default Qwen3-8B configuration reaches 0.82, 0.73, and 0.63 at more than $12\times$ lower cost than GPT-5.4 in all roles.

\textbf{Matched Backbone and Budget.} HEART pairs its harness with a small backbone, so its gains could come from the scaffold, the backbone, or the tool catalogue. Holding the backbone at Qwen3-8B and the catalogue at ToolFace, we compare HEART with ReAct, retrieval-only tool selection, a planner--executor pipeline, and DFSDT (Table~\ref{tab:tool-matched}). HEART reaches 75.1\% ToolBench pass rate against 55.8\% for DFSDT, the strongest of the four there, and 0.44 NESTFUL full accuracy against 0.19 for the planner--executor, the strongest on that metric.

\begin{table}[!htbp]
\centering\researchpaneltrue\researchpanelsize{9}
% Pair body (tables/pairs/tool-matched+tool-ablation.tex); a float of its own if input directly.
\begin{researchtable}
\centering
\researchtablesetup
\caption{\textbf{HEART against scaffolds with the same backbone and tool catalogue.} Every row uses Qwen3-8B and ToolFace; ToolBench reports average pass and win rates (\%), NESTFUL full-sequence accuracy and win rate (fractions)~\citep{jin2026toolprimitives}. Green marks the best and amber the second-best value in each column.}
\label{tab:tool-matched}
\fitresearchtable{\begin{tabular}{lcccc}
\toprule
\rowcolor{TabHead}
 & \multicolumn{2}{c}{\hdr{ToolBench}} & \multicolumn{2}{c}{\hdr{NESTFUL}} \\
\hdrgapfill\cmidrule(lr){2-3}\cmidrule(lr){4-5}
\rowcolor{TabHead}
\multirow{-2}{*}{\hdr{Method}} & \hdr{Pass} & \hdr{Win} & \hdr{Full Acc.} & \hdr{Win} \\
\midrule
ReAct                    & 33.5 & 46.2 & 0.08 & 0.10 \\
Retrieval-only selection & 41.0 & 49.5 & 0.10 & 0.12 \\
Planner--executor        & 52.6 & 55.8 & \second{0.19} & \second{0.28} \\
DFSDT                    & \second{55.8} & \second{57.4} & 0.13 & 0.16 \\
\midrule
\tabours HEART & \best{75.1} & \best{75.7} & \best{0.44} & \best{0.75} \\
\bottomrule
\end{tabular}}
\end{researchtable}

\end{table}

Removing ToolFace and the Tool Primitives also removes one model call per invocation step. The budget-matched control of Table~\ref{tab:tool-budget} gives these calls to value resolution, which self-refines its schema-compliant arguments with backbone, total call count, and token budget matched to full HEART. It raises the 16.1\% pass rate and 0.06 full accuracy of the plain ablation to 34.2\% and 0.15, against 75.1\% and 0.44 with Tool Primitives.

\begin{table}[!htbp]
\centering
\researchtablesetup
\caption{\textbf{Removing Tool Primitives at a matched call budget.} The equal-budget row reassigns the removed primitive calls to value-resolution self-refinement over raw schemas, with backbone, total call count and token budget matched to full HEART; the second row is the ablation of Table~\ref{tab:tool-ablation} without reallocation~\citep{jin2026toolprimitives}. Green marks the best and amber the second-best value in each column.}
\label{tab:tool-budget}
\fitresearchtable{\begin{tabular}{lcc}
\toprule
\rowcolor{TabHead}
\hdr{Variant} & \hdr{ToolBench Avg. Pass (\%)} & \hdr{NESTFUL Full Acc.} \\
\midrule
Equal-budget value-resolution self-refinement (no Tool Primitives) & \second{34.2} & \second{0.15} \\
w/o ToolFace \& Tool Primitives, no budget reallocation & 16.1 & 0.06 \\
\midrule
\tabours HEART & \best{75.1} & \best{0.44} \\
\bottomrule
\end{tabular}}
\end{table}

\textbf{Further Benchmarks.} HEART reaches the highest ACEBench Overall score, 86.9, and the highest BFCLv4 Web Search and Memory Overall scores, 86.0 and 69.38; outside HEART, the same Qwen3-8B reaches 12.0 on BFCLv4 Web Search. The 50 real-world tasks, 10 in each of travel planning, healthcare, e-commerce, finance, and local services, are grounded in user personas, range from a single-tool lookup to a multi-tool workflow of search, comparison, and conditional booking, and are scored by human annotators on whether the system completes the task. Through their product interfaces and built-in tools, GPT-5.4, Claude-4.6-Sonnet, and Gemini-3.1-Pro complete 20\% to 24\% of these tasks; given the ToolFace tools through native function calling, they complete 22\% to 32\%, a gain of at most 8~pp, against 84\% for HEART.

\textbf{Validation and Revision.} Because the recovery loop is conditioned on result validation, HEART's study compares 200 sampled validation decisions from ToolBench and NESTFUL trajectories with human pass/fail labels (Table~\ref{tab:tool-verifier}): precision is 0.97, recall 0.96, and the false-pass and false-fail rates are 8\% and 4\%. Most false passes fall on the task-completion criterion for partially satisfied objectives, and fewer than 3\% of trajectories end on a false pass; false fails trigger at least one redundant revision round in 10--15\% of trajectories without changing outcomes. ToolBench pass rate grows from 47.6\% without revision to 51.4\%, 65.9\%, and 75.1\% with one to three rounds and saturates at 75.3\% with five (Table~\ref{tab:tool-revision}), a plateau indicating that the loop terminates because validation converges.

\textbf{Retrieval and Robustness.} Over the full repository, Recall@5 is 92.4 on ToolBench and 94.7 on NESTFUL, and the top ten hold 95.8\% and 97.2\% of the gold tools (Table~\ref{tab:tool-retrieval}). Retrieval misses account for 15--20\% of HEART's residual end-task failures, and validation-triggered revision recovers part of the first-pass misses by reformulating the tool query. On subsampled repositories, ToolBench Recall@5 is 94.6 at 5k tools and 93.7 at 10k, against 92.4 at 25,519. Because tool schemas stay in each primitive's own prompt and never enter the caller's, HEART's attack success rate on ToolBench is 0.0\% when a malicious tool document is injected into the retrieved candidates following ToolHijacker~\citep{shi2025prompt}, under both gradient-free and gradient-based attacks, against 66.0--82.2\% for GPT-5.4, Claude-4.6-Sonnet, and Gemini-3.1-Pro.

\textbf{Mapping to the Method.} Schema resolution with its structured failure is the carried validation of a Tool Primitive (Section~\ref{sec:primitive-definition}), and running $\mathcal{P}_k(\beta_k\mid r_{k-1})$ is the act step of Eq.~\eqref{eq:act}, in which the summary in $r_{k-1}$ carries the earlier result along a dependency of the plan.
\FloatBarrier

\subsection{Code Primitives: Adaptation with Carried Tests}
\label{sec:code-mechanism}

\begin{figure}[!htbp]
\centering
\includegraphics[width=\linewidth]{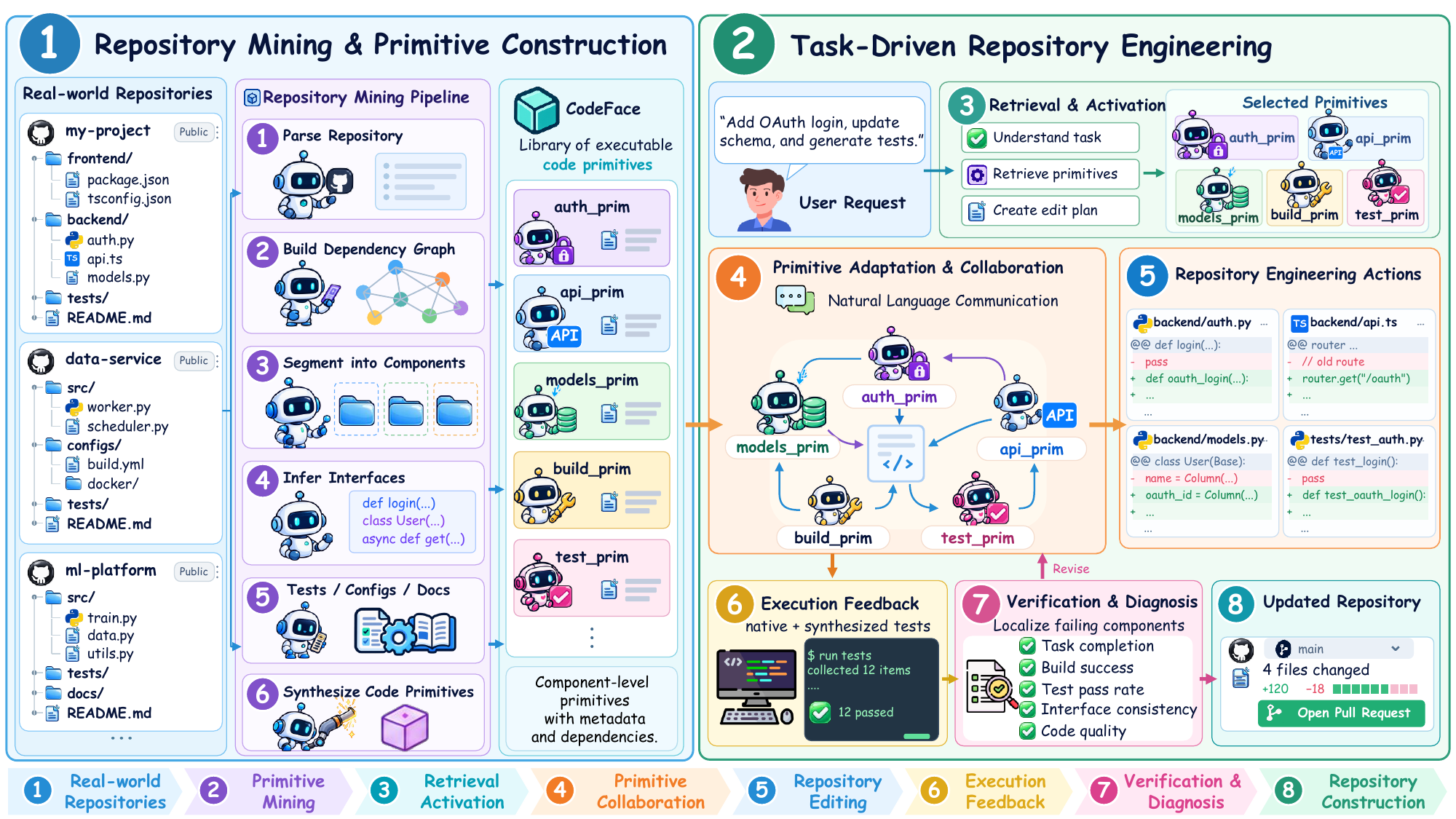}
\caption{\textbf{Overview of LEGO.} \emph{Left:} real-world repositories are mined into component-level Code Primitives, each carrying its interface, dependencies, tests, configuration, and documentation, and organized in CodeFace. \emph{Right:} a user request activates the required primitives, which adapt themselves and exchange the requirements their changes impose, before the adapted implementations are edited into the repository and revised under execution feedback and diagnosis~\citep{jin2026codeprimitives}.}
\label{fig:code-overview}
\end{figure}

Code Primitives are reusable software components that adapt their implementations, interfaces, and dependencies to target repositories~\citep{jin2026codeprimitives}. Tool Primitives and Agent Primitives expose tools and reusable behaviours through interfaces that support invocation; a Code Primitive instead provides an interface for adaptation, because it becomes part of the delivered repository. LEGO, \textbf{L}arge-scale repository \textbf{E}ngineering via a\textbf{G}ent-native reusable c\textbf{O}de primitives, builds repositories with Code Primitives drawn from CodeFace, a library mined once from real repositories. Adaptation is performed by the component that owns the implementation, while constraints that cross components are resolved at the repository level. In Figure~\ref{fig:code-overview}, CodeFace is mined offline, and a single illustrative OAuth request then passes through adaptation, execution, and diagnosis.

\textbf{Code Primitives.} A Code Primitive is a reusable executable software component equipped with a resident LLM, $P_i=(C_i,I_i,D_i,V_i,X_i,M_i)$, and the resident LLM is its natural-language interface. Its boundary follows functional coherence and the dependencies required for independent execution, so its granularity is induced by mining: entries have a median of 214 implementation lines (p25 96, p75 468) across 2 files (p75 4), and they export 4 symbols and carry 9 tests at the median. Given a requirement $r$ and construction context $\mathcal{C}$, $M_i$ runs in two modes:
\begin{equation}
(s_i,u_i)=M_i^{\mathrm{assess}}(P_i,r,\mathcal{C}),\qquad
(\widetilde{P}_i,\mu_i)=M_i^{\mathrm{adapt}}(P_i,r,\mathcal{C}),
\label{eq:code-primitive}\csname ltx@label\endcsname{eq:code-assess}
\end{equation}
where $s_i$ is the relevance assessment, $u_i$ the usage proposal, and $\mu_i=\{\mu_{i\to j}\}_j$ is empty for locally contained changes. ASSESS never modifies the primitive, and it answers UNSUITABLE when the capability does not support the request or when meeting the request would require replacing the capability itself. ADAPT may change the implementation, exported signatures, data representations, dependencies, and configuration while preserving the selected capability. The primitive then updates its interface and dependency records and reruns $V_i$, and this carried validation distinguishes primitive adaptation from generic editing.

\textbf{Prompt Contract.} A primitive receives a fixed four-part input: its own state $(C_i,I_i,D_i,V_i,X_i)$ verbatim; the requirement $r$ as the backbone wrote it; a context slice naming the target module path, the symbols the module must export, and the signatures of the sibling components it binds to; and the diagnostics $e$ of a previous attempt, if any, with the incoming requirements $\mu_{\to i}$ in a construction round. It sees no native test, no other module's source, and neither the repository-level plan nor its diagnosis. ADAPT returns a complete replacement implementation, the updated exports, behavioural contract, and dependencies, the outgoing requirements $\mu_{i\to j}$, a mapping that marks each carried test \textsf{kept}, \textsf{rewritten}, or \textsf{dropped}, and the execution log of $V_i$ on the rewritten component. A test may be dropped when it checks donor-specific behaviour, never because the adapted implementation fails it. A return that cannot be parsed, misses a field, or leaves a retained test failing is re-requested with the failure attached up to 3 times and then returns to repository-level handling as an unsuccessful adaptation.

\textbf{CodeFace and Mining.} CodeFace organizes the validated primitives as a library $\mathcal{F}=\{P_i\}_{i=1}^{N}$. The evaluated snapshot holds 1,424 entries: 871 mined, 343 harvested from construction runs whose repository-level suite verified the component in context, and 210 web-sourced for capabilities that the mined corpus lacked, all admitted under the same requirements. Mining parses a source repository $\mathcal{R}$ into program entities and relations and proposes component boundaries by weighted community discovery,
\begin{equation}
G_{\mathcal{R}}=(V,E_{\mathrm{call}}\cup E_{\mathrm{data}}\cup E_{\mathrm{struct}}),\qquad
\mathcal{S}^{(0)}_{\mathcal{R}}=\operatorname{COMMUNITY}(G_{\mathcal{R}})=\{S^{(0)}_1,\ldots,S^{(0)}_m\},
\label{eq:code-mining}
\end{equation}
whose edges capture calls, data and type dependencies, and structural relations such as imports, configuration references, and implementation--test associations. Semantic refinement then splits separable functions or merges entities that serve one capability into a component $S_i$. Across its dependency boundary $\partial S_i$, local helpers and configuration state are internalized into $C_i$, external requirements are recorded in $D_i$, and the exposed interactions, associated tests, and documentation with provenance supply $I_i$, $V_i$, and $X_i$. Each candidate is validated in isolation, and its build, runtime, and test diagnostics $e_i^t$ refine it,
\begin{equation}
\partial S_i=\{v\notin S_i\mid \exists u\in S_i,\;(u,v)\in E\},\qquad
(C_i^{t+1},D_i^{t+1},I_i^{t+1})=\operatorname{REFINE}(C_i^t,D_i^t,I_i^t,e_i^t),
\label{eq:code-refine}
\end{equation}
until it is admitted, with executable closure, a passing component suite, and a recorded source repository, commit, and license, or excluded. A security hold set is applied when the library is loaded and saved, and because each entry records its source repository, same-repository donors can be excluded per target (Table~\ref{tab:code-provenance}).

\textbf{Retrieval and Activation.} The construction context $\mathcal{C}_t$ holds the repository state, the constructed components with their recorded interfaces and dependencies, the decisions of earlier rounds, and the accumulated diagnostics. LEGO decomposes the request and activates primitives:
\begin{equation}
\Pi_t=(r_1,\ldots,r_K)=\operatorname{DECOMPOSE}(q,\mathcal{C}_t),\qquad
\mathcal{A}_t=\operatorname{MATCH}\bigl(\Pi_t,\{\mathcal{F}_k\}_{k=1}^{K},\{(s_i,u_i)\}\bigr)\subseteq\textstyle\bigcup_k\{i:P_i\in\mathcal{F}_k\}.
\label{eq:code-match}
\end{equation}
Each requirement $r_k$ names a capability with the interface it must expose and the dependencies it may assume. Retrieval runs outside the LLM, by embedding and reranking a query built from these constraints, and returns $|\mathcal{F}_k|=m$ candidates, with $m=2$. MATCH weighs coverage of $\Pi_t$ and compatibility with the target interfaces, the other selected primitives, and the existing repository. It activates the smallest mutually compatible set for the round, and a requirement that no suitable primitive covers goes to task-specific code $g_t$.

\textbf{Primitive Adaptation and Collaboration.} Round $t$ instantiates ADAPT as
\begin{equation}
(\widetilde{P}_i,\;\mu_i)=M_i^{\mathrm{adapt}}\bigl(P_i,\;r_{k(i)},\;\mathcal{C}_t\cup\mu_{\to i}\bigr),\qquad i\in\mathcal{A}_t,
\label{eq:code-adapt}
\end{equation}
with $k(i)$ the capability that $P_i$ was selected to cover and $\mu_{\to i}=\{\mu_{j\to i}:j\in\mathcal{A}_t\}$ the requirements addressed to it. A requirement $\mu_{i\to j}$, such as a changed signature, representation, dependency, or shared behavioural assumption, reaches its consumer in $\mathcal{A}_t$ within the same round, along the dependency edges that $\Pi_t$ records, and the consumer adapts its own implementation; no primitive edits another's component.

\textbf{Editing, Execution, and Diagnosis.} Repository Editing combines the adapted primitives with integration code into $\mathcal{S}_t=\operatorname{INTEGRATE}(\{\widetilde{P}_i\}_{i\in\mathcal{A}_t},\,g_t)$ and resolves the composition constraints across primitive boundaries,
\begin{equation}
\Gamma(\{\widetilde{P}_i\}_{i\in\mathcal{A}_t})=\{\Gamma_{\mathrm{api}},\Gamma_{\mathrm{type}},\Gamma_{\mathrm{dep}},\Gamma_{\mathrm{config}}\},
\label{eq:code-constraints}
\end{equation}
namely API bindings, shared data representations and types, package dependencies, and configuration; the harness, not diagnosis, evaluates $\Gamma$. Execution Feedback runs the target's build checks and tests together with unit and integration tests synthesized from $\Pi_t$ and records $o_t=\operatorname{EXECUTE}(\mathcal{S}_t)$: build status, test outcomes and failing cases, runtime errors, and traces. Each task runs in a separate environment; suites share a 900-second timeout, and each test has a 180-second limit, tightened to 45 seconds inside repair after a timed-out round. Verification and Diagnosis computes
\begin{filldisp}\begin{equation}
(v_t,\phi_t)=\operatorname{DIAGNOSE}(q,\Pi_t,\mathcal{S}_t,o_t),\qquad v_t\in\{\pass,\fail\},\quad \phi_t=(e_t,c_t,u_t),
\label{eq:code-diagnose}
\end{equation}\end{filldisp}
on task completion, build success, test pass rate, interface consistency, and code quality. Build success and test pass rate come from execution, so $v_t=\fail$ whenever the build or a required test fails. Diagnosis ties each observed failure in $e_t$ to execution evidence, finds the smallest set of components $c_t\subseteq\mathcal{A}_t$ or interactions that explains it, classifies it among seven types, from an incorrect implementation or a missing capability to an integration failure, and gives each implicated component a revision objective in $u_t$. It modifies no file and uses no held-out grading outcomes, original implementations, or floor and ceiling counts.

\textbf{Revision.} On failure, $\mathcal{C}_{t+1}=\mathcal{C}_t\cup\{\phi_t\}$, and the primitives in $c_t$ are reactivated for another round of Eq.~\eqref{eq:code-adapt}, while components that the new evidence does not contradict are preserved. The context accumulates diagnostics and is never truncated within a task. $\Pi_t$ is recomputed each round, so a requirement unmatched in round $t$ can be matched once neighbouring interfaces settle, and additional primitives are reserved for capabilities or dependencies that the evidence shows to be missing. A run stops when $v_t=\pass$, after $b=5$ rounds per module, after two consecutive rounds without an additional passing test, or after two timeout rounds, and it returns the updated repository and its patch.\fillpar

\textbf{Controls.} Configurations are named by their active stages: retrieval (\stg{R}), primitive adaptation (\stg{A}), execution feedback (\stg{V}), and diagnosis (\stg{D}). Each stage ablation is one substitution. Without \stg{D}, revision runs on the raw execution evidence $o_t$ in place of $\phi_t$; without \stg{A}, the retrieved $P_i$ is integrated in place of $\widetilde{P}_i$; and without \stg{R}, $\mathcal{A}_t$ is empty. Scoped generic editing (\stg{E}) replaces primitive adaptation over the same retrieved components and differs from a primitive in the prompt contract above: the editor receives none of the primitive's carried tests, metadata, or execution state. Vendor-and-invoke (\stg{I}) copies the selected primitive unchanged into a vendored package and calls it, and read-only context (\stg{C}) places its implementation, contract, and carried tests in the repository-level prompt without delivering its files. The independently developed agents run their own released harnesses with default scaffolding, tools, and stopping rules. A task adapter presents each LEGO-REPO task in the input format that the agent expects, and no agent receives the frozen test identifiers, the floor and ceiling counts, or any CodeFace entry.

\textbf{Implementation.} GPT-5.6 Terra is the default backbone, adaptation model, and diagnosis model, and one adaptation model instantiates all $M_i$; the study sweeps 13 backbones. The harness runs 10 concurrent model calls with at most 9,000 interface characters per module, top-2 retrieval, and $b=5$. A task uses on average 856k input and 48k output tokens for repository-level construction, 11k input and 2.3k output tokens per adaptation over 7.4 successful adaptations, and 18k input and 2.6k output tokens per diagnosis round, and embedding and reranking cost a fixed \$0.011. Priced at the hosted rates of the evaluated deployments, recorded calls cost \$1.55 per task for execution feedback and \$2.08 for LEGO with GPT-5.6 Terra.

\textbf{Model Allocation.} Table~\ref{tab:code-cost} varies the models in the backbone, adaptation, and diagnosis positions. Under a GPT-5.6 Terra backbone, GPT-OSS-20B in adaptation and diagnosis yields 0.4880 at \$1.58 per task, above feedback with diagnosis at 0.3469 and \$1.79, whereas a DeepSeek-V3.2 backbone with GPT-5.6 Terra in the other two positions reaches 0.2471 at \$0.79.

\begin{table}[!htbp]
\centering
\researchtablesetup
\caption{\textbf{Model allocation across LEGO's backbone, adaptation and diagnosis.} Configurations name the backbone / adaptation / diagnosis models (Terra is GPT-5.6 Terra, OSS-20B is GPT-OSS-20B, Sonnet 5 is Claude Sonnet 5), of which only adaptation runs inside the Code Primitives; cost is recorded-call USD per task, counting backbone and diagnosis calls, successful adaptation calls only, and a fixed retrieval charge, with open-weight models priced as hosted deployments; \% LEGO is relative to the full LEGO score, and the last two rows are library-free references~\citep{jin2026codeprimitives}. Green marks the best and amber the second-best value in the All and \$/task columns. Lower is better for \$/task.\fillcap[0.6][1]}
\label{tab:code-cost}
\fitresearchtable{\begin{tabular}{lccc}
\toprule
\rowcolor{TabHead}
\hdr{Configuration (c/a/d)} & \hdr{All} & \hdr{\$/task} & \hdr{\% LEGO} \\
\midrule
\tabours Terra / Terra / Terra     & \best{0.5134} & 2.08        & 100.0 \\
Terra / OSS-20B / OSS-20B          & \second{0.4880} & \second{1.58} & 95.1 \\
Sonnet 5 / OSS-20B / OSS-20B       & 0.4661        & 3.31        & 90.8 \\
DeepSeek-V3.2 / Terra / Terra      & 0.2471        & \best{0.79} & 48.1 \\
\midrule
Feedback + Diagnosis, $b$\,=\,5    & 0.3469        & 1.79        & 67.6 \\
Feedback, $b$\,=\,8                & 0.3437        & 2.07        & 66.9 \\
\bottomrule
\end{tabular}}
\end{table}

\textbf{Worked Adaptation.} CodeFace entry \stg{cf-0412/schema-coercion}, mined from \texttt{marshmallow}, spans 261 lines across 2 files, exports a field-coercion dispatcher and its error type, and carries 14 tests. Reconstructing \texttt{pandera}, LEGO required it to coerce untyped input to declared field types and to collect per-field errors, each with a location path, into one report. The adaptation replaced the donor's exception-per-field control flow with an accumulating error collector, renamed the exported error type to the target's contract, and moved the location path from a string to the target's tuple representation. It dropped 3 tests that asserted on \texttt{marshmallow}'s message formatting, rewrote 5, and kept 6, and the 11 retained tests passed before repository-level composition bound the component to 4 sibling modules. A generic editor can produce the same rewrite but performs no carried-test check.

\begin{figure}[!htbp]
\centering
\begin{minipage}[t]{0.49\linewidth}\vspace{0pt}\includegraphics[width=\linewidth]{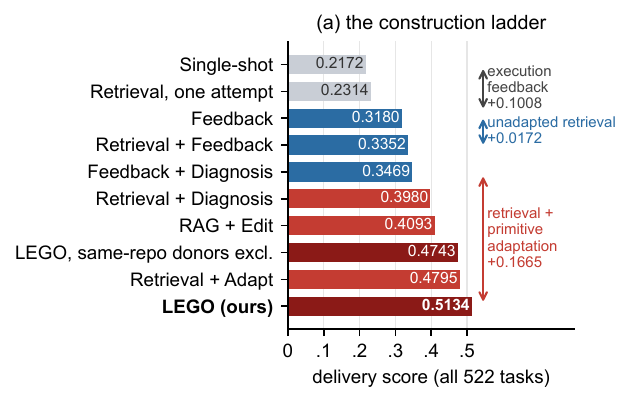}\end{minipage}\hfill
\begin{minipage}[t]{0.49\linewidth}\vspace{0pt}\includegraphics[width=\linewidth]{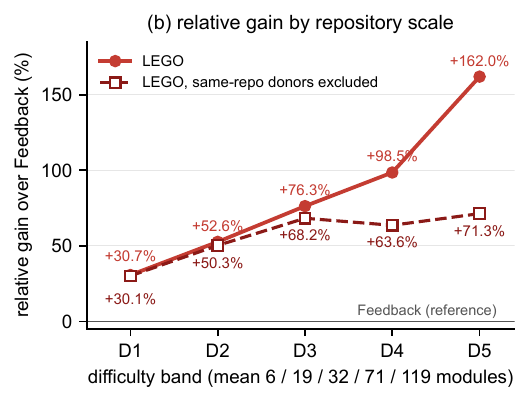}\end{minipage}
\caption{\textbf{Where Code Primitives add delivery, and how the gain scales with the repository.} (a) Delivery of the construction paradigms of Table~\ref{tab:code-paradigms} on all 522 tasks, with the increments of execution feedback, unadapted retrieval, and retrieval with primitive adaptation. (b) Relative gain over execution feedback by difficulty band, with and without same-repository donors. The absolute gain is nearly flat across bands (Table~\ref{tab:code-band-domain})~\citep{jin2026codeprimitives}.}
\label{fig:code-analysis}
\end{figure}

\textbf{Construction Ladder.} Figure~\ref{fig:code-analysis}(a) plots the ten configurations of Table~\ref{tab:code-paradigms} as a ladder of delivery: execution feedback adds 0.1008 to single-shot generation, unadapted retrieval 0.0172 to feedback, and retrieval with primitive adaptation 0.1665 to feedback with diagnosis. Panel (b) gives the relative gain over execution feedback by band, and the gap between the curves with and without same-repository donors at D4 and D5 is the part of the gain that those donors account for.

\textbf{Primitive Fields.} Table~\ref{tab:code-fields} locates LEGO's 0.1041 margin over generic editing in the fields of $P_i$. A scoped generic editor over the retrieved components scores 0.4093, and adding the carried tests $V_i$, then the interface contract and provenance $I_i,X_i$, then the dependency closure $D_i$ raises it to 0.4431, 0.4818, and 0.5039. Added in this order, the fields recover 0.0946 of the margin, and the residual 0.0095 has a 95\% interval of $[-0.0071,+0.0148]$ with Wilcoxon $p=0.49$. Withholding the carried tests from an otherwise complete primitive lowers delivery to 0.4796, a loss of 0.0338 that equals their forward increment~\citep{jin2026codeprimitives}.

\begin{table}[!htbp]
\centering\researchpaneltrue\researchpanelsize{9}
% Pair body (tables/pairs/code-consumption+code-fields.tex); a float of its own if input directly.
\begin{researchtable}
\centering
\researchtablesetup
\caption{\textbf{The fields of $P_i$ that make a Code Primitive useful.} Rows add carried tests $V_i$, interface contract and provenance $I_i,X_i$, and dependency closure $D_i$ cumulatively to RAG with generic editing (\stg{RE\,DV}), with retrieval, diagnosis and feedback fixed, and the fifth row withholds $V_i$ from a complete primitive; delivery on the $[0,1]$ scale, with $\Delta$ to \stg{RE\,DV} and to LEGO~\citep{jin2026codeprimitives}. Green marks the best and amber the second-best value in the All column.}
\label{tab:code-fields}
\fitresearchtable{\begin{tabular}{lccc}
\toprule
\rowcolor{TabHead}
\hdr{Configuration} & \hdr{All} & \hdr{$\Delta$ \stg{RE\,DV}} & \hdr{$\Delta$ LEGO} \\
\midrule
RAG + Edit (\stg{RE\,DV})                 & 0.4093 & \textit{ref.} & \sn{-0.1041} \\
\quad + carried tests $V_i$               & 0.4431 & \sn{+0.0338} & \sn{-0.0703} \\
\quad + contract and provenance $I_i,X_i$ & 0.4818 & \sn{+0.0725} & \sn{-0.0316} \\
\quad + dependency closure $D_i$          & \second{0.5039} & \sn{+0.0946} & \sn{-0.0095} \\
primitive without $V_i$                   & 0.4796 & \sn{+0.0703} & \sn{-0.0338} \\
\tabours LEGO (\stg{RADV})                & \best{0.5134} & \sn{+0.1041} & \textit{ref.} \\
\bottomrule
\end{tabular}}
\end{researchtable}

\end{table}

\textbf{Mapping to the Method.} Within the method of Section~\ref{sec:method}, ADAPT returns $\widetilde{P}_i$ as the result of Eq.~\eqref{eq:primitive} and $\mu_i$ as its messages, Repository Editing and Execution Feedback together form EXECUTE, DIAGNOSE is VERIFY, and admission into CodeFace is LEGO's validation and publishing step, Eq.~\eqref{eq:publish}.
\FloatBarrier

\subsection{Math Primitives: Self-Advertised Method Selection}
\label{sec:method-mechanism}
\begin{figure}[!htbp]
\centering
\includegraphics[width=\linewidth]{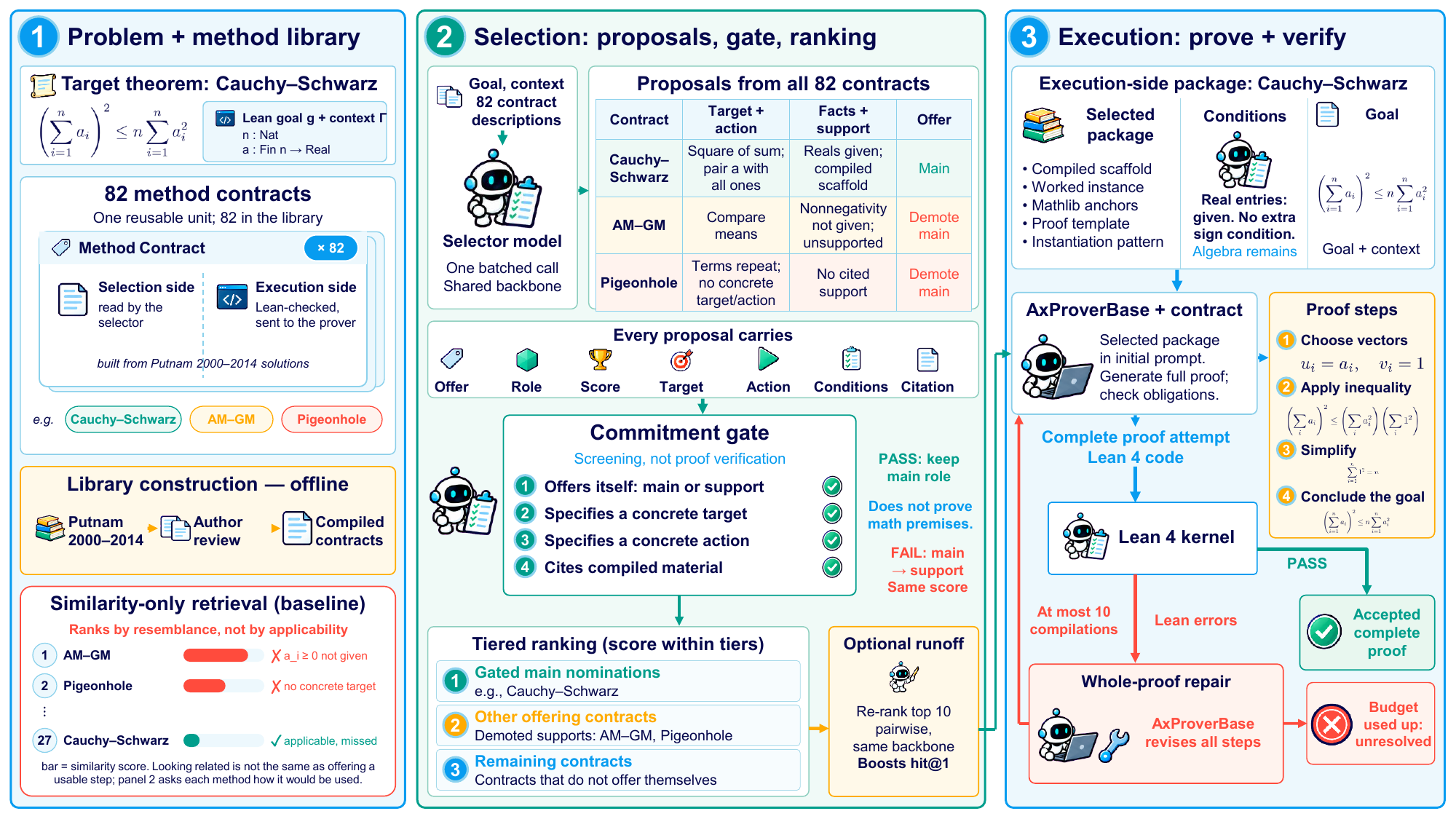}
\caption{\textbf{Math Primitives: self-advertised method selection on an illustrative Cauchy--Schwarz goal.} (1)~The task is a Lean goal with its local context. The library holds 82 Method Contracts, built offline from Putnam 2000--2014 solutions with every extracted method reviewed by the authors, and a selector reads only their selection sides: the method, its typical target, when it applies and when not, and a manifest of its verified material. Similarity retrieval alone proposes AM--GM, whose condition $a_i\ge0$ the goal does not give. (2)~In one batched call of a shared backbone, every contract proposes a target, an action, and the facts its use requires, cites verified material, and states its offer, role, and score. The commitment gate screens these proposals and demotes a vague (Pigeonhole) or unsupported (AM--GM) main nomination to a supporting role; the tiered ranking and an optional runoff over the top ten then order the contracts. (3)~The execution side of the top-ranked contract enters the initial prompt of AxProverBase, Lean compiles each complete proof attempt, and its errors drive the prover's whole-proof repair within ten compilations~\citep{anonymous2026library}.}
\label{fig:method-overview}
\end{figure}

A Math Primitive stores a proof method as a Method Contract, with the method's required conditions, intended action, and residual obligations, and self-advertised method selection lets each contract propose its own use on a goal~\citep{anonymous2026library}. We give the full method and its experimental settings in that study's notation. Figure~\ref{fig:method-overview} follows an illustrative Cauchy--Schwarz goal through the three stages of the method: the library whose selection sides the selector reads, self-advertisement with its commitment gate and tiered ranking, and proving under Lean compilation.

\textbf{Problem and Contracts.}
Let $g$ be a Lean goal with local context $\Gamma$, and let $\mathcal{C}$ be a finite library of Method Contracts. A contract $c$ has parameters $x_c$, a selection-facing description, required conditions $\Pre_c(x_c)$, a target statement $G_c(x_c)$ with an intended action, an execution scaffold $S_c$, and a list of residual obligations $\Obl_c(x_c)$. The action reduces the target to the required conditions and the obligations: once $\Pre_c(\theta)$ and $\Obl_c(\theta)$ are proved for a binding $\theta$ of $x_c$, the action yields $G_c(\theta)$. A selector receives $g$, $\Gamma$, and the contract descriptions and returns an ordered list $\pi(g)=(c_1,\dots,c_{|\mathcal{C}|})$ of the whole library. A contract is \emph{applicable} to $g$ if some correct proof of $g$ contains a step that instantiates it. The experiments measure against a nonempty set $A(g)$ of annotated reference contracts. Correct annotations are applicable, and other proofs may use other methods, so agreement with $A(g)$ understates how often a shortlist contains a usable method.

\textbf{Selection and Execution Sides.}
The \emph{selection side}, which is what a selector reads, states the method, its typical target, when it applies and when not, and a manifest naming the verified material of the execution side without its Lean code. The \emph{execution side}, which is supplied to the prover, holds a Lean scaffold over $x_c$, a compiled worked instance, the Mathlib declarations it relies on, $\Pre_c$ as Lean hypotheses, $\Obl_c$, $G_c$, and a verification record.

\textbf{Self-Advertisement.}
One batched model call reads $g$, $\Gamma$, and the selection sides of all contracts and returns a proposal for each, so similarity plays no part in ranking or filtering the candidates. The proposal of $c$ makes a commitment that can be checked against the goal: a target $t_c$, which instantiates $G_c(\hat\theta)$ in the goal's own objects at a proposed binding $\hat\theta$; an action $a_c$; and the required facts $\widehat{\Pre}_c$, which instantiate $\Pre_c(\hat\theta)$. It also cites verified material $\sigma_c$ from the contract and records whether the contract offers itself ($o_c$), its role $\rho_c\in\{\text{main},\text{support}\}$, and an integer applicability score $\hat s_c$. A main nomination must pass the commitment gate
\begin{equation}
\operatorname{Gate}(c)=o_c\wedge\operatorname{Spec}(t_c)\wedge\operatorname{Spec}(a_c)\wedge(\sigma_c\neq\varnothing),
\label{eq:method-gate}
\end{equation}
where $\operatorname{Spec}$ holds when a target or action names a concrete object of the goal; a main nomination that fails the gate is demoted to a supporting role with its score unchanged. The tiered ranking orders contracts lexicographically by $(\tau_c,-\hat s_c)$, where $\tau_c=1$ for main nominations that pass the gate, $\tau_c=2$ for other contracts that offer themselves, and $\tau_c=3$ otherwise. Ties are broken by contract identifier, so the ranking is deterministic given the model outputs.

\textbf{Runoff.}
A prover that receives one contract depends on hit@1, and integer scores often leave the leading proposals close or tied. The optional runoff re-ranks the top $q$ contracts in one top-down pass of $q-1$ calls: each call sees a contract, the contract above it, and that contract's earlier proposal, writes a proposal for the lower contract, and swaps the two when that proposal fits the proof better. Contracts outside the window keep their order. The runoff uses the selector's own backbone, so any gain it brings comes from direct comparison, and $q=10$ matches the rerankers' candidate count.\fillpar

\textbf{Conditional Scaffold Soundness.}
The study's first result states what a checked scaffold establishes.

\begin{proposition}[{Conditional scaffold soundness~\citep{anonymous2026library}}]
\label{prop:math-scaffold}
Suppose a contract provides a Lean-checked term with no \texttt{sorry} or admitted proof holes, of type
\begin{equation}
S_c:\ \forall x_c,\ \Pre_c(x_c)\to\Obl_{c,1}(x_c)\to\cdots\to\Obl_{c,r}(x_c)\to G_c(x_c).
\label{eq:method-action}\csname ltx@label\endcsname{eq:math-primitive}
\end{equation}
For any binding $\theta$ whose required conditions and obligations have checked proofs, applying $S_c$ yields a checked proof of the target $G_c(\theta)$.
\end{proposition}

\begin{proof}[Proof sketch]
Instantiate the checked term at $\theta$ and apply it to the checked proofs of its premises. Lean checks that the resulting term proves $G_c(\theta)$.
\end{proof}

The proposition applies to scaffolds checked in this parameterized form; a compiled example certifies that one example, not arbitrary new bindings. It concerns reuse, not discovery: a proposal identifies the remaining work for the new goal, $\Pre_c(\hat\theta)$ and $\Obl_c(\hat\theta)$, without discharging it.

\textbf{Similarity and Coverage.}
Let $X=(g,\Gamma,\text{contract descriptions})$ be the selector's input, with $A=A(g)$ random given $X$, and let $T=\phi(X)$ be the features available to a similarity-only selector. With information $F$, the best achievable expected hit@$k$ is $V_k(F)=\mathbb{E}\bigl[\max_{S\subseteq\mathcal{C},\,|S|=k}\Pr(S\cap A\neq\varnothing\mid F)\bigr]$. Proposition~\ref{prop:coverage} restates the study's information bound, $V_k(X)\ge V_k(T)$; its candidate-pool ceiling for a selector that reorders the similarity top-$m$ pool $P_m(g)$ and places the other contracts after it; and its multi-method coverage result, under which $H_x(S)=\Pr(S\cap A\neq\varnothing\mid X=x)$ is monotone and submodular, so greedy selection with the true marginal gains attains at least $1-1/e$ of the best size-$k$ coverage~\citep{nemhauser1978analysis}. The proofs of all three are in Appendix~\ref{app:proofs}.

\textbf{Ranking Regret.}
With $w_c(x)=\mathbb{E}\bigl[\mathbf{1}[c\in A]/|A|\bigm|X=x\bigr]$ for a fixed input $X=x$, a shortlist $S$ has expected recall@$k$ equal to $R_x(S)=\sum_{c\in S}w_c(x)$, and self-advertisement can be read as estimating, from each contract's proposal, the ordering by $w_c(x)$ that maximizes it.

\begin{proposition}[{Top-$k$ ranking regret~\citep{anonymous2026library}}]
\label{prop:math-regret}
Let $S_k^*$ contain the $k$ contracts with the largest $w_c(x)$, let $\widehat S_k$ contain the $k$ contracts with the largest scores $\hat s_c$, and suppose there is a strictly increasing map $f$ such that $|f(\hat s_c)-w_c(x)|\le\epsilon$ for every $c\in S_k^*\cup\widehat S_k$. Then
\[
R_x(S_k^*)-R_x(\widehat S_k)\le 2k\epsilon .
\]
\end{proposition}

Estimation errors affect the shortlist when they fall on contracts that are, or appear to be, among the top $k$. The gate moves contracts that appear to be among the top $k$ but cannot state a checkable use below all supported nominations, and the runoff compares the leading proposals directly, since absolute scores from different contracts need not lie on the common scale that the premise requires.

\textbf{Selection and Proof Success.}
For the event $E_k$ that the first $k$ positions contain an annotated contract and the event $Z_B$ that the prover completes a Lean proof within $B$ compile calls, any prover satisfies $\Pr(Z_B)=\Pr(Z_B\mid\neg E_k)+\Pr(E_k)\,\Delta_k$ with $\Delta_k=\Pr(Z_B\mid E_k)-\Pr(Z_B\mid\neg E_k)$. Selection controls $\Pr(E_k)$, while $\Delta_k$ depends on the contract's execution guidance and residual obligations; the contract arm of Table~\ref{tab:method-proof} is the case $k=1$, $B=10$.

\textbf{Contract Library.}
The 82 contracts are built once, offline, in three steps from solutions to Putnam 2000--2014 problems, none of them an evaluation problem. GPT-5.6 Sol extracts the methods used in the reference solutions and proposes merges between methods that play the same role in different problems, and the authors review each extracted method and each proposed merge. The model then drafts both sides of each contract from its source solutions, and each execution-side Lean artifact is compiled with Lean v4.27.0 and Mathlib at commit \texttt{a3a10db}. All 82 contracts pass, each with a scaffold stated for arbitrary parameters and a compiled worked instance. Construction and proving share this environment, so a scaffold that compiled at construction stays valid for the prover, as Proposition~\ref{prop:math-scaffold} requires, and the runner exits at startup if the library's recorded Mathlib revision differs from the environment's Lake manifest.

\textbf{Experimental Configuration.}
The selection baselines are random ordering, BM25~\citep{robertson2009probabilistic}, TF-IDF cosine~\citep{salton1988term}, OpenAI's text-embedding-3-small and text-embedding-3-large~\citep{neelakantan2022text}, Qwen3-Embedding-0.6B~\citep{zhang2025qwen3}, and a listwise LLM reranker over the top 10 candidates of each embedding model~\citep{sun2023chatgpt}. All LLM-based selectors run with DeepSeek-V4.1-Flash~\citep{s06_xu2026deepseek}, GPT-5.6 Luna~\citep{s06_openai2026gpt56}, and GPT-5.6 Terra, one backbone per block of Table~\ref{tab:method-selection}. With $\pi_k(g)=\{c_1,\dots,c_k\}$ and $\operatorname{rank}(g)=\min\{i:c_i\in A(g)\}$, the metrics over the $N$ problems of a benchmark are
\begin{gather*}
\text{hit@}k=\frac{1}{N}\sum_{g}\mathbf{1}\bigl[\pi_k(g)\cap A(g)\neq\varnothing\bigr],\qquad
\text{recall@}k=\frac{1}{N}\sum_{g}\frac{|\pi_k(g)\cap A(g)|}{|A(g)|},\\
\text{MRR}=\frac{1}{N}\sum_{g}\frac{1}{\operatorname{rank}(g)}.
\end{gather*}
For proving, GPT-5.6 Luna runs a whole-proof repair loop with a budget of $B=10$ Lean compilations per problem. Base is one attempt without retrieval; pass@20 draws 20 independent samples and so has twice the budget; APOLLO alternates generation with Lean-guided repair~\citep{ospanov2026apollo}; AxProverBase, Ax-Prover run without contracts under the same budget, is the direct control~\citep{requena2026minimalagentautomatedtheorem}; and the contract arm adds the top-ranked self-advertised contract to AxProverBase's initial prompt.

\textbf{Mapping to the Method.}
Within the method of Section~\ref{sec:method}, the selection side of a contract is its interface $d_i$, and a proposal is the relevance assessment of Eq.~\eqref{eq:assess}: the score $\hat s_c$ plays the role of $s_i$, and the target, action, and required facts form the proposed use $u_i$.

\subsection{Knowledge Primitives: Source Learning}
\label{sec:knowledge-mechanism}
\begin{figure}[!htbp]
\centering
\includegraphics[width=\linewidth]{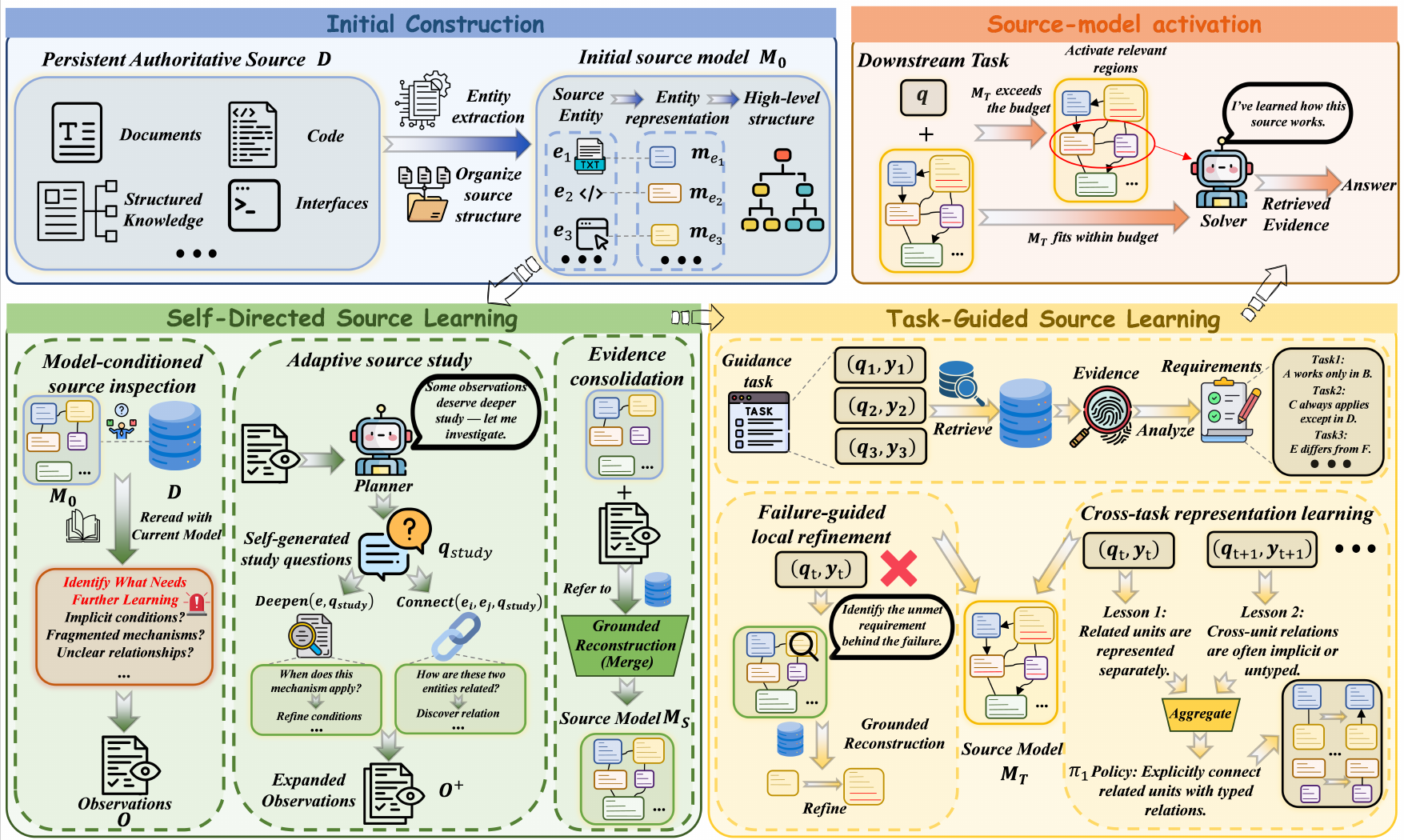}
\caption{\textbf{Overview of SourceLearn.} \textbf{Initial Construction} builds a provisional source model $M_0$ from the authoritative source $D$. \textbf{\textit{Self-Directed Source Learning}} deepens and connects source understanding through adaptive study, producing $M_S$. \textbf{\textit{Task-Guided Source Learning}} uses guidance-task failures and recurring representation needs to further refine the model into $M_T$. At inference time, \textbf{Source-Model Activation} combines relevant source-model context with directly retrieved evidence for downstream tasks~\citep{jin2026sourcelearn}.}
\label{fig:knowledge-overview}
\end{figure}

Each Knowledge Primitive owns a source model, an explicit and persistent state that holds the understanding of one authoritative source while remaining grounded in it. SourceLearn~\citep{jin2026sourcelearn} is the framework that develops this model through repeated study and use of the source, and we describe it in its own notation: $M$ is a source model, $F$ the task solver, $B$ a token budget, $R(D,q)$ the source context of a task $q$ and $R$ alone a source region, and $\Pi_0$ and $\Pi_1$ are instructions on how to represent a source. Initial construction builds a provisional source model $M_0$ from the source $D$, Self-Directed and Task-Guided Source Learning refine it into $M_S$ and then $M_T$ through grounded source reconstruction, and at test time the solver receives the activated $M_T$ beside the retrieved evidence (Figure~\ref{fig:knowledge-overview}).

\textbf{Problem.}
A persistent authoritative source $D$ grounds recurring tasks $(q,y)\sim\mathcal{P}_D$, where $y$ is the desired outcome of $q$. Knowledge access, source learning, and the use of the learned understanding are
\begin{equation}
\hat{y}=F\big(q,R(D,q)\big),\qquad
M_t=\operatorname{Learn}(D,H_t),\qquad
\hat{y}=F\big(q,R(D,q),M_t\big),
\label{eq:knowledge-access}
\end{equation}
where $R$ may be retrieval, long-context access, or a structured source representation, and $H_t$ is the accumulated study and task experience. Repeated access alone can leave reusable knowledge fragmented or in need of reconstruction for each task. Source learning therefore faces two questions: what source knowledge to learn before future tasks are known, and how task feedback can improve source-specific competence beyond the task that produced it, since retaining an observed answer or correction risks learning the task rather than the source. Self-Directed Source Learning addresses the first question and Task-Guided Source Learning the second:
\begin{equation}
M_0=\operatorname{Construct}(D),\qquad M_S=\operatorname{SelfLearn}(D,M_0),\qquad M_T=\operatorname{TaskLearn}(D,M_S;\mathcal{G}),
\label{eq:knowledge-stages}\csname ltx@label\endcsname{eq:knowledge-learning}
\end{equation}
where $\mathcal{G}=\{(q_t,y_t)\}_{t=1}^{T}$ holds guidance tasks grounded in the same source, with reference outcomes $y_t$.

\textbf{Source Model.}
$M$ is organized around source entities $e$, each with a reusable representation $m_e$ of its properties, mechanisms, governing conditions, procedures, exceptions, distinctions, and relations to other entities, and higher-level structure organizes dependencies across entities. Each unit makes one reusable semantic commitment and cites the source elements it rests on; facts that a reader can look up stay in $D$. $\operatorname{Construct}$ reads the source entity by entity within coherent regions, such as code modules or document sections, under a generic modeling instruction $\Pi_0$ that favors major entities, relations, procedures, and governing conditions over redundant low-level content. The resulting $M_0$ is deliberately provisional.

\textbf{Grounded Source Reconstruction.}
All later updates follow one principle: learning signals determine what should be reconsidered, while persistent knowledge is reconstructed from the authoritative source. For an affected source region $R$ and a learning signal $\xi$,
\begin{equation}
O_R=\operatorname{Inspect}(D_R,M_R;\xi),\qquad
M'=\operatorname{GroundApply}\big(M,\operatorname{Reconstruct}(M_R,O_R);D_R\big),
\label{eq:knowledge-grounded}
\end{equation}
where the temporary observations $O_R$, gathered by rereading the region's source content against the current model, are never copied into $M$. Each proposed unit passes through the grounding gate, which judges a statement supported when its cited excerpts entail it and a negative claim supported when an excerpt states the denial; any other unit counts as unsupported. The region rewrite is committed if the preservation check finds each previously supported meaning still stated, entailed, or covered by a broader unit, and withheld otherwise.

\textbf{Source-Model Activation.}
At task time, the solver receives at most $B=24$k tokens of $M$ in addition to the evidence retrieved for the query:
\begin{equation}
A_B(M,q)=
\begin{cases}
M, & \operatorname{tok}(M)\leq B,\\
\operatorname{ActivateRegions}(M,q;B), & \operatorname{tok}(M)>B,
\end{cases}
\qquad
\hat{y}=F\big(q,R(D,q),A_B(M,q)\big),
\label{eq:knowledge-query}\csname ltx@label\endcsname{eq:knowledge-primitive}
\end{equation}
where activation selects task-relevant regions among those established during construction and preserves the internal structure of each region it selects.

\textbf{Self-Directed Source Learning.}
Before future task requirements are known, one Inspect--Study--Consolidate cycle decides what deserves deeper study:
\begin{equation}
O=\operatorname{Inspect}(D,M),\qquad O^{+}=\operatorname{Study}(D,M,O),\qquad M'=\operatorname{Consolidate}(D,M,O^{+}).
\label{eq:knowledge-self}
\end{equation}
\emph{Model-conditioned source inspection} rereads the source content $S_e$ of each entity against its representation, $O_e=\operatorname{Observe}(S_e,m_e)$, and records what the representation still fails to explain well; observations do not modify $M$. \emph{Adaptive source study} then selects, at planning round $b$, study actions from the current model, the accumulated observations $\overline{O}_b$, and the study history $H_{<b}$:
\begin{equation}
\Gamma_b=\operatorname{PlanStudy}(M,\overline{O}_b,H_{<b}),\qquad
\Gamma_b\subseteq\bigl\{\textsc{Deepen}(e,q^{\mathrm{study}}),\ \textsc{Connect}(e_i,e_j,q^{\mathrm{study}})\bigr\}.
\label{eq:knowledge-plan}
\end{equation}
\textsc{Deepen} investigates an unresolved aspect of one entity, and \textsc{Connect} jointly studies two entities whose dependency remains insufficiently understood. The temporary study question $q^{\mathrm{study}}$ of each action guides retrieval and renewed reading and is not stored. \emph{Evidence consolidation} then applies Eq.~\eqref{eq:knowledge-grounded} once, a \emph{read-many, write-once} pattern, and turns $M_0$ into $M_S$.

\textbf{Task-Guided Source Learning.}
In one pass over $\mathcal{G}$, SourceLearn first returns to the source for each guidance task:
\begin{equation}
\widetilde{E}_t=\operatorname{RetrieveGuide}(D,q_t,y_t),\qquad (E_t^{*},C_t)=\operatorname{Diagnose}(q_t,y_t,\widetilde{E}_t),
\label{eq:knowledge-diagnose}
\end{equation}
where $E_t^{*}$ is the source evidence that supports the reference outcome and $C_t$ the source requirements of the task. No other step of source learning directly observes $y_t$, and neither $y_t$ nor $E_t^{*}$ is stored as persistent knowledge. The task is then answered through Eq.~\eqref{eq:knowledge-query}; when the answer $\hat{y}_t$ is wrong, \emph{failure-guided local refinement} rereads from $D$ the region behind the requirements that the model does not represent and revises it through Eq.~\eqref{eq:knowledge-grounded}:
\begin{equation}
C_t^{-}=\{c\in C_t:c\text{ is not represented in }M_t\},\qquad M_{t+1}=\operatorname{Refine}(D,M_t;C_t^{-}).
\label{eq:knowledge-refine}
\end{equation}
Requirements that $M_t$ represents trigger no writing, which separates a representational deficiency from an error made although the required understanding is available. The pass yields the locally refined model $M_{\mathrm{local}}$. \emph{Cross-task representation learning} draws from each guidance task at most one lesson about how, not what, to represent, and aggregates the recurring lessons with $\Pi_0$ into a representation policy $\Pi_1$, which introduces no source knowledge:
\begin{equation}
\ell_t=\operatorname{Lesson}(q_t,C_t,M_t,\hat{y}_t),\qquad \Pi_1=\operatorname{Aggregate}\big(\Pi_0,\{\ell_t\}_{t=1}^{T}\big).
\label{eq:knowledge-policy}
\end{equation}
Recalibration then revisits the source under $\Pi_1$ and incorporates structure that is missing or represented at an inappropriate granularity through Eq.~\eqref{eq:knowledge-grounded}:
\begin{equation}
M_T=\operatorname{Recalibrate}(D,M_{\mathrm{local}};\Pi_1).
\label{eq:knowledge-recalibrate}
\end{equation}
$M_T$ stays fixed for the test tasks.

\begin{figure}[!htbp]
\centering
\includegraphics[width=0.92\linewidth]{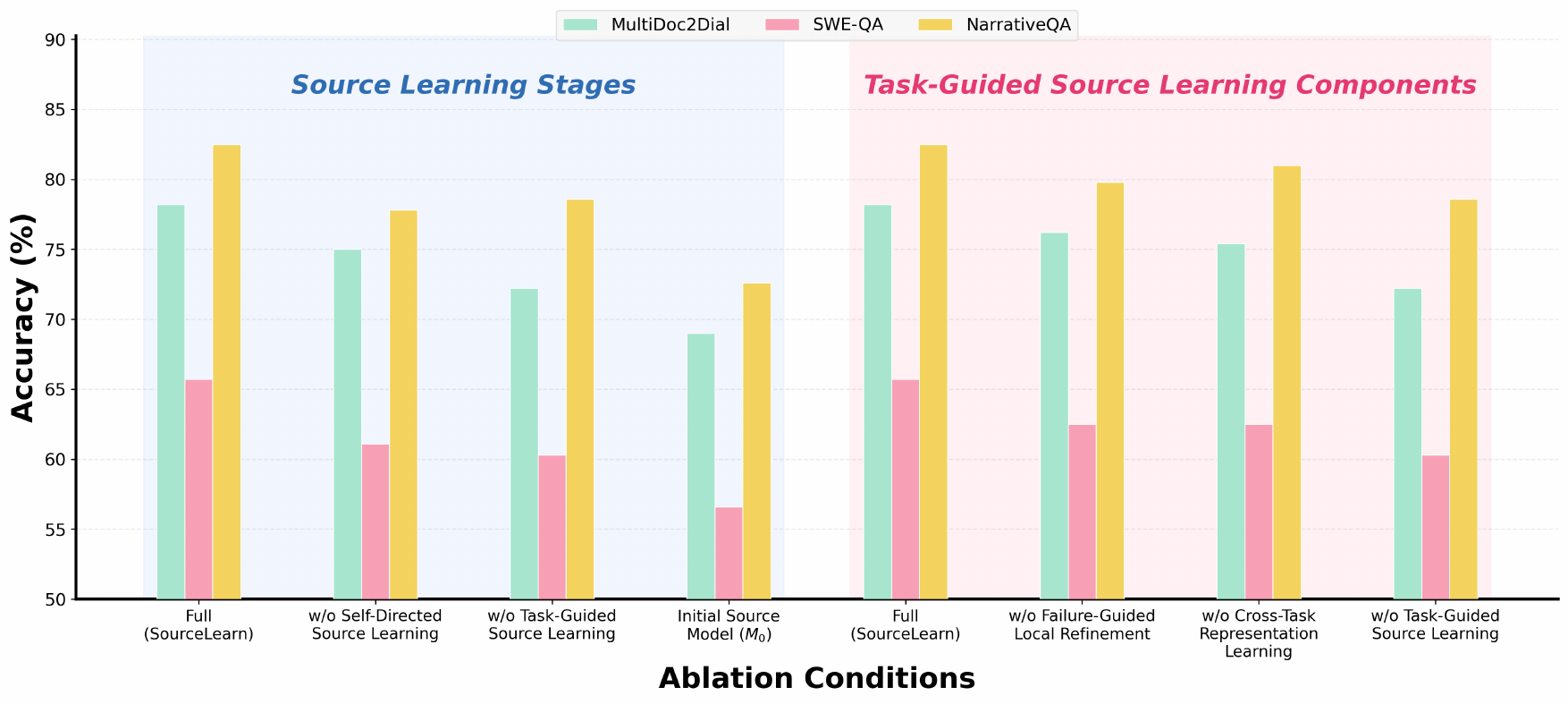}
\caption{\textbf{Both learning stages and both task-guided pathways add accuracy.} Accuracy (\%) with GPT-5.6 Luna on two sources per QA benchmark. Left: the full source model $M_T$ against removing self-directed or task-guided learning and against the initial model $M_0$. Right: removing failure-guided local refinement, cross-task representation learning, or both, which recovers $M_S$, under shared task-time access~\citep{jin2026sourcelearn}.\fillcap[0.6][1]}
\label{fig:knowledge-ablation}
\end{figure}

\textbf{Sources and Counterparts.}
MultiDoc2Dial contributes four government domains, DMV, SSA, StudentAid, and VA, each a source of 92--149 documents with 60 questions. NarrativeQA contributes 12 stories of its official test split, six movie scripts and six books with 27--30 questions each, read in full. SWE-QA contributes six of its fifteen repositories, streamlink, sphinx, xarray, pytest, flask, and conan, each a source at its pinned commit with 48 questions. APIBench has one source per model hub, TorchHub with 186 instructions and HuggingFace and TensorFlow Hub with 100 each, whose accuracies are averaged; a prompt that exceeds a backend's context window counts as a failure. AppWorld's API documentation, 457 API specifications across nine simulated apps, is a single source. RAPTOR~\citep{sarthi2024raptor} runs its official implementation with default settings (100-token chunks, recursive clustering, and collapsed-tree retrieval of the top 10 nodes within 3,500 tokens). HippoRAG~2~\citep{gutierrez2025rag} runs its official package in its reproduction setting and passes its top 8 passages, matching the evidence budget of Hybrid RAG. AWM~\citep{wang2024awm} induces workflows from the successful guidance-task experiences of each source and split and places them in the prompt of each test task of that split, beside the same Hybrid RAG evidence that SourceLearn receives.

\textbf{Implementation.}
GPT-5.6 Luna~\citep{s06_openai2026gpt56} and GPT-OSS-120B~\citep{agarwal2025gpt} run at medium reasoning effort, and DeepSeek-V4.1-Flash~\citep{s06_xu2026deepseek} at high effort on the QA benchmarks and medium effort on APIBench and AppWorld; all methods embed with text-embedding-3-large. $M_0$ is constructed once per source and backend. Self-Directed Source Learning runs one cycle with $K=20$ study actions, and its $M_S$ is shared by the three splits. Hybrid RAG, which supplies $R(D,q)$, splits documents into sections and code into files and symbol-level units, fuses the top 30 elements by BM25~\citep{robertson2009probabilistic} and by dense retrieval~\citep{karpukhin-etal-2020-dense} by reciprocal rank fusion, and passes the top 8 to the solver. On AppWorld, the ReAct-style code agent~\citep{yao2023react} runs for at most 50 steps, its documentation calls answered by a channel that returns each method's own source context, with the built-in documentation app removed. The three 30\%/70\% splits of the QA benchmarks and APIBench keep related tasks on the same side, and AppWorld's 90 training and 168 test-normal tasks serve as guidance and test tasks. The QA judge, GPT-5.6 Luna at medium reasoning effort, accepts an answer that makes the key claims of the reference without contradicting it, and accuracy is macro-averaged over sources and then averaged over splits; APIBench uses its AST-based evaluator, and AppWorld reports task goal completion. The ablations in Figure~\ref{fig:knowledge-ablation} use two sources per QA benchmark, SSA and VA, pytest and conan, and the scripts of \emph{All About Eve} and \emph{The Breakfast Club}, with the guidance and test splits of the main results. Removing Self-Directed Source Learning starts task-guided learning from $M_0$, removing Task-Guided Source Learning leaves $M_S$, removing cross-task representation learning leaves $M_{\mathrm{local}}$, and removing failure-guided local refinement recalibrates $M_S$ under $\Pi_1$.\fillpar[0.6][1]

\begin{wraptable}{R}{0.48\linewidth}
\wraplift
\centering
\researchtablesetup
\caption{\textbf{SourceLearn against retrieval given the same per-question context budget.} Accuracy (\%) with GPT-5.6 Luna on four sources, each retrieval counterpart reading further down its own ranking up to SourceLearn's evidence tokens (last row)~\citep{jin2026sourcelearn}. Green marks the best and amber the second-best value in each column.\fillcap}
\label{tab:knowledge-budget}
\fitresearchtable{\begin{tabular}{lcccc}
\toprule
\rowcolor{TabHead}
 & \multicolumn{2}{c}{\hdr{MultiDoc2Dial}} & \hdr{SWE-QA} & \hdr{APIBench} \\
\hdrgapfill\cmidrule(lr){2-3}\cmidrule(lr){4-4}\cmidrule(lr){5-5}
\rowcolor{TabHead}
\multirow{-2}{*}{\hdr{Method}} & \hdr{DMV} & \hdr{VA} & \hdr{Sphinx} & \hdr{HuggingFace} \\
\midrule
Hybrid RAG & \second{76.2} & 70.2 & 62.7 & 48.6 \\
RAPTOR & 70.2 & \second{73.8} & 61.8 & \second{52.9} \\
HippoRAG 2 & 71.4 & 71.5 & \best{73.5} & 51.4 \\
\tabours SourceLearn & \best{77.0} & \best{77.8} & \second{72.5} & \best{56.7} \\
\midrule
Evidence tokens & 16.5k & 18.2k & 23.4k & 17.1k \\
\bottomrule
\end{tabular}}
\end{wraptable}

\textbf{Additional Results.}
The activated source model gives the solver more task-time context than the native retrieval baselines receive. We therefore give each retrieval baseline the per-question context budget of SourceLearn, which it fills by reading further down its own ranking without changing its ranking or index (Table~\ref{tab:knowledge-budget}). SourceLearn remains the best on three of the four sources and second on Sphinx, so task-time context alone does not explain its performance. The remaining analyses also use GPT-5.6 Luna with the sources, splits, and judge of the main results. On the QA benchmarks, the gain over Hybrid RAG is 14.9~pp on source regions that guidance tasks touched and 12.1~pp on regions that none touched (Table~\ref{tab:knowledge-transfer}), so the learned understanding carries over to parts of the source that no guidance task exposed. On MultiDoc2Dial, the source model alone, with no retrieved evidence, performs comparably to Hybrid RAG with all eight retrieved elements, and removing the gold document costs Hybrid RAG about 25~pp but SourceLearn about 8~pp (Figure~\ref{fig:knowledge-degradation}). The audits of Tables~\ref{tab:knowledge-grounding} and~\ref{tab:knowledge-preservation} cover StudentAid, pytest, and HuggingFace: at most 1\% of the sampled units of $M_0$, $M_S$, and $M_T$ are contradicted by their cited source elements, and 93.3\% to 100\% of the claims that a learning stage removed remain represented. Some Task-Guided units on the document and code sources are partially supported or unsupported; on StudentAid, the fully supported share falls from 79.2\% in $M_0$ to 63.3\% in $M_T$, with 13.3\% unsupported. The learned model therefore complements the source with reusable understanding, and the source stays authoritative and directly accessible through $R(D,q)$.

\textbf{Mapping to the Method.}
As in Section~\ref{sec:var-knowledge}, each source model is the resource of one Knowledge Primitive, source-model activation specializes COMPOSE, and grounded source reconstruction realizes VALIDATE on the publishing path of Eq.~\eqref{eq:publish}, along which the two learning stages improve each source model apart from the test tasks.

\FloatBarrier

\subsection{Agent Primitives: Reusable Latent Building Blocks}
\label{sec:agent-mechanism}
\begin{figure}[!htbp]
\centering
\includegraphics[width=\linewidth]{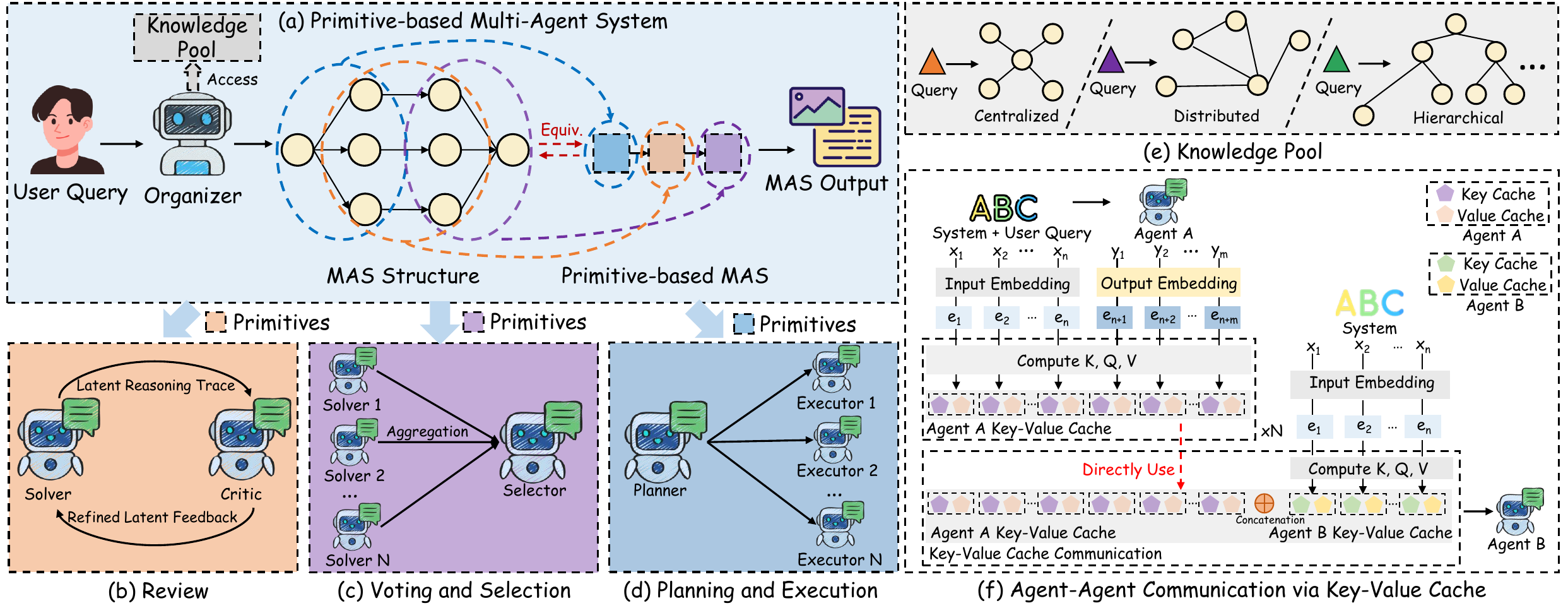}
\caption{\textbf{Overview of Agent Primitives.} (a)~Agent Primitives enable system-level functionality to be constructed by composing reusable latent operators, which replace manually designed agent roles and natural-language interaction protocols. (b)--(d)~Three representative primitives: Review, Voting and Selection, and Planning and Execution, each encapsulating a recurring multi-agent computation pattern as a reusable latent operator. (e)~A Knowledge Pool stores previously observed queries and effective system structures, providing structural guidance for system construction. (f)~Internal coordination within and across primitives is carried out via latent KV cache communication~\citep{jin2026agentprimitives}.}
\label{fig:agent-overview}
\end{figure}

Agent Primitives~\citep{jin2026agentprimitives} address two limitations of existing multi-agent systems (MAS): they lack a reusable, task-agnostic abstraction for organizing multi-agent computation, so their complexity scales with task complexity, and they communicate in natural language, which degrades in long contexts and under accumulated noise. Drawing on neural network design, where complex models are built from reusable blocks such as residual blocks and attention heads~\citep{vaswani2017attention}, the study decomposes existing MAS architectures into a few recurring, minimal computation patterns, such as iterative review, multi-agent consensus, and plan--execute decomposition, and instantiates three representative primitives that communicate via key--value (KV) cache. Latent communication methods mainly improve how information passes between agents~\citep{zheng2025thought,fu2025cache,zou2025latent}; Agent Primitives use KV cache communication to implement the computation structure of each primitive. For each query, an LLM Organizer selects and composes the primitives into a primitive-based MAS, guided by a lightweight Knowledge Pool (Figure~\ref{fig:agent-overview}).

\textbf{Latent Communication.}
For a Transformer decoder of $L$ layers~\citep{vaswani2017attention}, the KV cache after $n$ positions is $\mathcal{Z}_n=\{(K_n^\ell,V_n^\ell)\}_{\ell=1}^L$, and decoding step $t$ appends its key and value, $K_t^\ell=[K_{t-1}^\ell;\mathbf{k}_t^\ell]$ and $V_t^\ell=[V_{t-1}^\ell;\mathbf{v}_t^\ell]$. Latent communication is stated for two agents $A$ and $B$ for simplicity; the Selector of Voting and Selection reads the caches of all its Solvers. The input prompt $x^{A}_{1:n_A}$ of $A$ holds its system prompt and the user query. After processing its full sequence $s^{A}=[x^{A}_{1:n_A};y^{A}_{n_A+1:n_A+m_A}]$ of length $T_A=n_A+m_A$, $A$ holds the final KV cache $\mathcal{Z}^{A}_{T_A}$, a latent representation of $s^A$, and the fixed system prompt $x^{B}_{1:n_B}$ of $B$, processed alone, yields the system-level cache $\mathcal{Z}^{B}_{n_B}$:
\begin{equation}
\mathcal{Z}^{A}_{T_A}=\{(K^{A,\ell}_{T_A},V^{A,\ell}_{T_A})\}_{\ell=1}^L,\qquad
\mathcal{Z}^{B}_{n_B}=\{(K^{B,\ell}_{n_B},V^{B,\ell}_{n_B})\}_{\ell=1}^L.
\label{eq:agent-cache}
\end{equation}
A combined KV cache concatenates the latent states of $A$ and the system prompt of $B$ at each layer:
\begin{equation}
\tilde{K}^{B,\ell}=[K^{B,\ell}_{n_B};K^{A,\ell}_{T_A}],\qquad
\tilde{V}^{B,\ell}=[V^{B,\ell}_{n_B};V^{A,\ell}_{T_A}].
\label{eq:agent-splice}
\end{equation}
The resulting cache $\tilde{\mathcal{Z}}_{B}=\{(\tilde{K}^{B,\ell},\tilde{V}^{B,\ell})\}_{\ell=1}^L$ replaces the KV cache of $B$, which then decodes autoregressively without natural-language communication from $A$.

\textbf{Input-Output Alignment Assumption.}
KV cache communication assumes that the decoder conditions future tokens on its accumulated key--value states alone, so that conditioning on $\mathcal{Z}^A$ is distributionally equivalent to conditioning on the token sequence $s^A$:
\begin{equation}
p(y\mid\mathcal{Z}^A,x^{B}_{1:n_B})\approx p(y\mid s^A,x^{B}_{1:n_B}).
\label{eq:agent-alignment}
\end{equation}
The approximation holds when both agents share the same model parameters and positional encoding scheme, so all agents use the same LLM, and all evaluations stay within one model architecture.

\textbf{KV Cache Positional Re-encoding.}
Rotary positional encoding (RoPE)~\citep{su2024roformer} rotates the key at position $t$ as $\mathrm{RoPE}(K_t^\ell)=R(t)K_t^\ell$, where $R(t)$ is a deterministic block-diagonal rotation operator whose 2D rotation angles are linear in $t$. The concatenated cache is treated as one continuous sequence, in which the system prompt of $B$ occupies positions $1$ to $n_B$ and the cache of $A$ positions $1$ to $T_A$. To preserve RoPE semantics, the KV states of $A$ are re-indexed by the offset $n_B$, for each layer $\ell$ and $t\in\{1,\dots,T_A\}$,
\begin{equation}
\mathrm{RoPE}(K_t^{A,\ell})\rightarrow R(t+n_B)K_t^{A,\ell},
\label{eq:agent-rope}
\end{equation}
so that the concatenated cache can be used directly in standard decoding.

\textbf{Communication Bottleneck.}
Two stress tests with Qwen3-8B~\citep{qwen3technicalreport} motivate KV cache as the medium. In long-context task injection, an auxiliary instruction is injected at the beginning, midpoint, or end of the long contexts of 351 En.QA examples from InfiniteBench. Without injection, natural-language communication reaches 51.6\% accuracy and KV cache communication 61.3\%. At the midpoint, injection compliance falls to 15.6\% for natural language, against 73.3\% for KV cache, while accuracy stays stable across positions, so the degradation stems from communication failure. In communication noise, 100 correctly solved GSM8K~\citep{cobbe2021training} traces, truncated to 6{,}000 tokens, are continued by another Qwen3-8B within 5{,}000 tokens while 0 to 25 sentences of unrelated reasoning are injected. With one and three noise sentences, natural language solves 91\% and 73\% of the problems and KV cache all of them; with 10 and 25, 47\% and 40\% against 93\% and 77\%.

\textbf{Primitive Design.}
Internal coordination runs exclusively through latent KV cache exchange, so a primitive's behavior is independent of text-level message passing, and each primitive can serve as a drop-in replacement for a single agent. The role system prompts fix role boundaries independently of task-specific content and encode no explicit problem-solving strategy. \emph{Review}, an iterative self-critique operator, connects a Solver $A$ and a Critic $B$ through a latent feedback channel: $B$ consumes the KV cache of $A$ to generate corrective feedback, and the modified latent representation returns to $A$ for refinement until a stopping condition derived from the intermediate latent states holds, which gives adaptive depth of refinement. The Critic provides evaluative signals and neither revises the solution nor answers. In \emph{Voting and Selection}, a consensus operator, parallel Solvers produce latent candidates independently, and a Selector votes and selects over these candidates directly in latent space, computing an aggregated latent state that produces the output; it judges correctness, consistency, and overall solution quality without introducing reasoning beyond the candidates. In \emph{Planning and Execution}, a latent decomposition operator, a Planner $P$ produces a latent plan that decomposes the task into intermediate steps or subgoals without solving it, and an Executor $E$ follows the plan without modifying it to generate the output.

\textbf{Primitive-Based MAS.}
A primitive-based MAS is a composition graph of primitives, which an LLM Organizer, GPT-5.2 by default, builds per query. The Organizer's prompt supplies the query, the three primitives, and the Knowledge Pool, and instructs it to analyze the task requirements and complexity, to select primitives, and to determine their execution order and composition structure. The Organizer reasons at the level of primitives, designs no task-specific roles or natural-language protocols, and does not solve the task. Its primitive composition plan specifies (i)~the primitives to instantiate and (ii)~their execution structure in code, at 200--400 output tokens per query. The lightweight Knowledge Pool holds 45 previously observed queries paired with effective MAS structures from frameworks including Multi-Agent Debate~\citep{du2023improving}, DyLAN~\citep{liu2024a}, Self-Refine~\citep{madaan2023self}, AFlow~\citep{zhang2024aflow}, and MAS-GPT~\citep{ye2025mas}. The Organizer retrieves entries by the input query as structural guidance and abstracts them into primitive compositions, replacing task-specific agents with Agent Primitives. The system then executes the plan, coordinating within and across primitives through KV cache, and externally behaves like a standard MAS.

\textbf{Implementation.}
The study runs five worker backbones, Qwen3-4B, Qwen3-8B, Qwen3-14B, DeepSeek-R1-Distill-Qwen-32B, and DeepSeek-R1-Distill-Llama-70B, of which Table~\ref{tab:agent-main} reports three. With Qwen3-4B and Qwen3-14B, the primitive-based MAS reaches 69.6\% and 77.6\% average accuracy, against 55.7\% and 65.5\% for a single agent. LatentMAS~\citep{zou2025latent} runs in its sequential configuration with $m=40$, its best setting, and four agents, so TextMAS and the single-primitive baselines use four as well: two rounds for Review, three Solvers and one Selector for Voting and Selection, and one Planner with three Executors for Planning and Execution. In the primitive-based MAS, the Organizer sets the number of agents per query. The metrics are accuracy, token usage, and average inference time per query.

\textbf{RoPE Re-indexing.}
Table~\ref{tab:agent-rope} disables the re-indexing of Eq.~\eqref{eq:agent-rope} while the backbones and primitives stay unchanged. With Qwen3-8B, the composed MAS loses 13.3~pp on AIME25 and at most 2.7~pp on HumanEval+ and MedQA. With R1/Llama-70B, each primitive and the composed MAS lose between 26.7 and 59.7~pp, so positional re-encoding is critical for stable KV cache communication, most of all on the Llama-based backbone.

\begin{table}[!htbp]
\centering
\researchtablesetup
\caption{\textbf{RoPE re-indexing in the KV cache communication of Agent Primitives.} Accuracy (\%) with and without positional re-encoding for each primitive used alone and for the composed MAS, all communicating agents using the same LLM (R1/Llama-70B is DeepSeek-R1-Distill-Llama-70B), with the change from the re-indexed run in pp in parentheses~\citep{jin2026agentprimitives}.}
\label{tab:agent-rope}
\fitresearchtable{\begin{tabular}{lllccc}
\toprule
\rowcolor{TabHead}
\hdr{Model} & \hdr{RoPE} & \hdr{Method} & \hdr{AIME25} & \hdr{HumanEval+} & \hdr{MedQA} \\
\midrule
 & & Review & 60.0 & 78.6 & 64.2 \\
 & & Voting   & 66.7 & 81.0 & 70.3 \\
 & & Planning & 66.7 & 78.6 & 67.0 \\
\tabours & \multirow{-4}{*}{w/} & Primitives-based MAS & 73.3 & 82.3 & 76.7 \\
\cmidrule(lr){2-6}
 & & Review & 56.7\dpp{-3.3} & 73.8\dpp{-4.8} & 61.5\dpp{-2.7} \\
 & & Voting   & 60.0\dpp{-6.7} & 79.9\dpp{-1.1} & 66.8\dpp{-3.5} \\
 & & Planning & 56.7\dpp{-10.0} & 78.0\dpp{-0.6} & 65.2\dpp{-1.8} \\
 \multirow{-8}{*}[0.4ex]{Qwen3-8B} & \multirow{-4}{*}{w/o} & Primitives-based MAS & 60.0\dpp{-13.3} & 81.1\dpp{-1.2} & 74.0\dpp{-2.7} \\
\midrule
 & & Review & 50.0 & 82.3 & 72.9 \\
 & & Voting   & 56.7 & 82.9 & 77.8 \\
 & & Planning & 50.0 & 82.3 & 74.0 \\
\tabours & \multirow{-4}{*}{w/} & Primitives-based MAS & 56.7 & 85.3 & 81.9 \\
\cmidrule(lr){2-6}
 & & Review & 16.7\dpp{-33.3} & 22.6\dpp{-59.7} & 31.4\dpp{-41.5} \\
 & & Voting   & 26.7\dpp{-30.0} & 29.9\dpp{-53.0} & 34.7\dpp{-43.1} \\
 & & Planning & 23.3\dpp{-26.7} & 24.4\dpp{-57.9} & 30.5\dpp{-43.5} \\
 \multirow{-8}{*}[0.4ex]{R1/Llama-70B} & \multirow{-4}{*}{w/o} & Primitives-based MAS & 26.7\dpp{-30.0} & 31.1\dpp{-54.2} & 36.6\dpp{-45.3} \\
\bottomrule
\end{tabular}}
\end{table}

\textbf{Further Results.}
Table~\ref{tab:agent-cost} prices the methods of Table~\ref{tab:agent-mas} per query at market output-token prices of \$1.75 per million tokens for Llama-3-70B-Instruct and \$14 per million for GPT-5.2, and divides accuracy by cost. With the same Llama-3-70B-Instruct workers, Claude-4 as Organizer reaches 72.8\%, 93.2\%, 81.1\%, and 53.6\% on MATH, GSM8K, HumanEval+, and GPQA, Qwen3-32B 71.3\%, 89.8\%, 78.6\%, and 49.1\%, and Llama-3-70B-Instruct itself, the iso-model setting, 72.0\%, 91.8\%, 81.0\%, and 52.2\%, so the framework does not rely solely on a proprietary Organizer. Over 100 examples per task category, Review accounts for 15\%, 22\%, and 48\% of the selected primitives on MATH, HumanEval, and MedQA, Voting and Selection for 52\%, 18\%, and 35\%, and Planning and Execution for 33\%, 60\%, and 17\%, with 3.4, 2.8, and 2.5 primitives per query, so the Organizer adapts its composition to the task. On ARC-Challenge, which has no entry in the Knowledge Pool, the Organizer without the pool outperforms a single agent by 2.5 and 2.3~pp with Qwen3-4B and Qwen3-8B and random structure selection by 1.9 and 1.1~pp, and the pool adds a further 1.2~pp on average.

\textbf{Mapping to the Method.}
The Organizer realizes COMPOSE (Eq.~\eqref{eq:compose}), with the Knowledge Pool as the structural guidance it reads, and its primitive composition plan is the plan of that equation. Along the plan's dependencies, agents that share one LLM pass the combined cache of Eq.~\eqref{eq:agent-splice}. The Critic and the Selector judge results inside their primitives, the Review loop is the family's revision, and the primitives and the pool stay fixed across queries.
\FloatBarrier

\subsection{Dev-Primitives: Activation, Collaboration, Execution, and Diagnosis}
\label{sec:dev-mechanism}
\begin{figure}[!htbp]
\centering
\includegraphics[width=\linewidth]{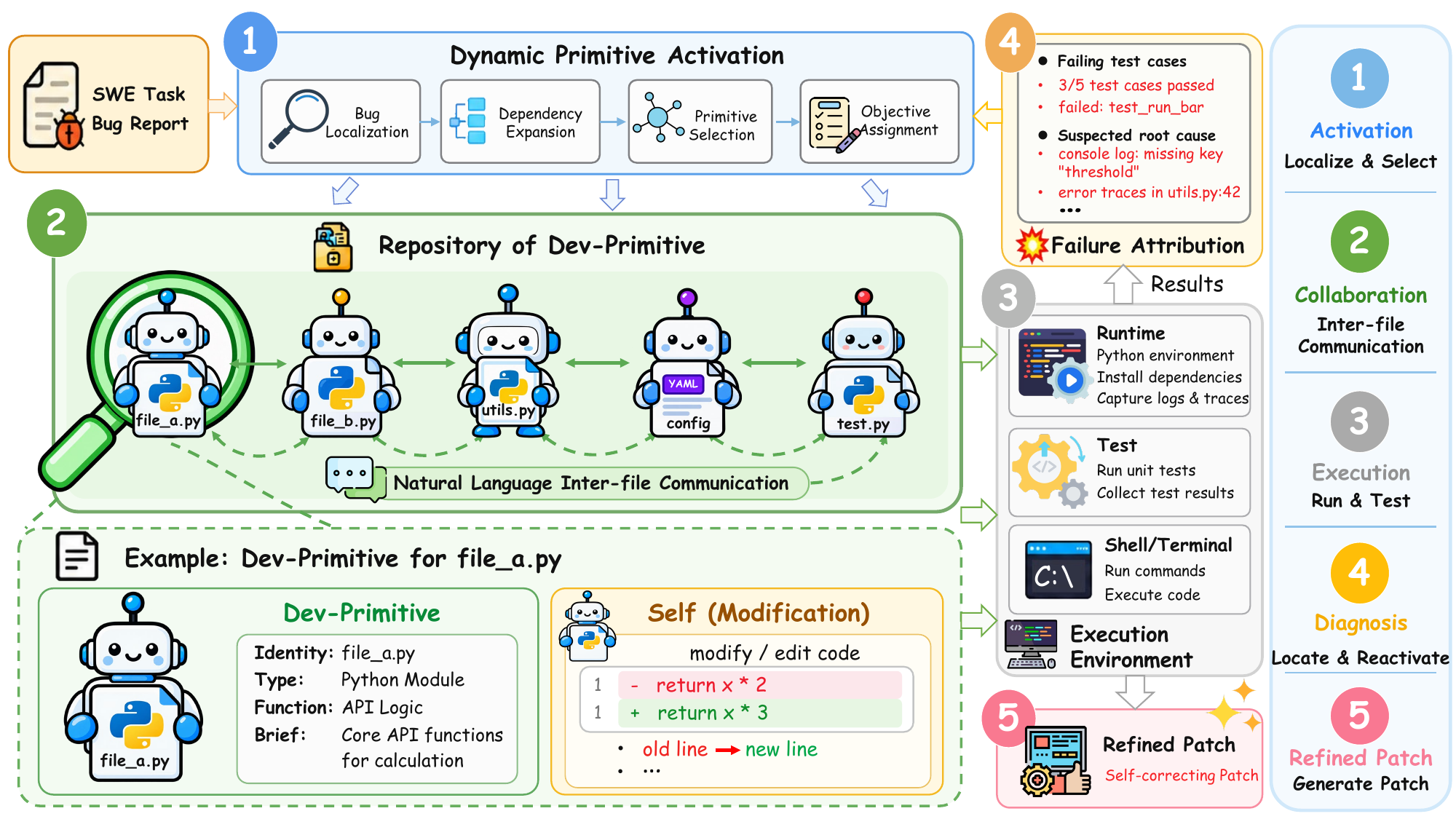}
\caption{\textbf{Overview of HERMES.} Dynamic activation selects the repository components relevant to the task and instantiates Dev-Primitives, which modify their own artifacts and exchange requirements in natural language. The modified repository is executed, and bug diagnosis maps the observed failure back to the components that must be revised, reactivating only those for the next round~\citep{jin2026devprimitives}.}
\label{fig:development-overview}
\end{figure}

Dev-Primitives are modular executable interfaces that turn repository components into active participants in software engineering~\citep{jin2026devprimitives}. HERMES, the harness engineering framework that instantiates them, consists of an activation module, a dynamically activated set of Dev-Primitives, an executable environment, and a diagnosis module. Dynamic activation and bug diagnosis, both implemented by prompting an LLM, are the two mechanisms that it supplies beyond the Dev-Primitive abstraction. Figure~\ref{fig:development-overview} gives the general workflow, in which bug diagnosis maps the observed failure back to the components that must be revised and the next round reactivates them. In one SWE-bench Verified trajectory with Qwen3-8B, six activated Dev-Primitives exchange 15 directed messages, and the issue is resolved in the first round.

\textbf{Dev-Primitives.}
A repository consists of $N$ software components $\mathcal{R}=\{a_i\}_{i=1}^{N}$, and each component $a_i$ is associated with a Dev-Primitive $P_i$ instantiated with an LLM $\mathcal{M}$:
\begin{equation}
(a_i',\,m_i)=P_i(a_i,\,x_i,\,\mathcal{C}_i)=\mathcal{M}\bigl([\,a_i;\ x_i;\ \mathcal{C}_i\,]\bigr),
\label{eq:dev-primitive}\csname ltx@label\endcsname{eq:dev-local}
\end{equation}
where $x_i$ is the local objective assigned to $P_i$, $\mathcal{C}_i$ the communication context of messages received from other Dev-Primitives, $a_i'$ the optionally modified artifact ($a_i'=a_i$ when no modification is required), and $m_i$ a natural-language message with task-relevant information or requirements for other components. In \emph{local modification}, the primitive reasons over and edits its own artifact. In \emph{inter-primitive communication}, it sends requirements such as interface changes, dependency updates, or testing requirements to the components that its local reasoning implicates, in practice its callers, callees, configurations, and tests. Each message is placed verbatim into its target's communication context, and no central agent interprets, filters, or rewrites it. Participation does not imply modification: a primitive may contribute information while leaving its artifact unchanged.

\textbf{State and Granularity.}
A Dev-Primitive persists across the revision rounds of a task with its current artifact and the communication accumulated during the trajectory; model hidden states are not preserved across calls. A new persistent file that an active Dev-Primitive creates is registered as a Dev-Primitive, while temporary outputs, caches, and logs are environment observations. Files are the default granularity, and on Terminal-Bench the components are the persistent filesystem artifacts, such as shell scripts, configuration files, and service definitions.

\textbf{Dynamic Primitive Activation.}
The activation module converts the issue $q$ into a plan over repository components,
\begin{equation}
\Pi=\operatorname{ACTIVATE}(q,\mathcal{R})=\{(i,x_i)\}_{i\in\mathcal{A}},\qquad
\mathcal{P}_{\mathcal{A}}=\{P_i\mid i\in\mathcal{A}\},
\label{eq:dev-activate}
\end{equation}
where $\mathcal{A}\subseteq\{1,\ldots,N\}$ indexes the selected components; HERMES instantiates $\mathcal{P}_{\mathcal{A}}$, and inactive components receive no model calls. Activation is a single LLM call chain of issue analysis, component localization, dependency analysis, task decomposition, and edit planning. Here $\mathcal{R}$ denotes access to the repository, not its concatenation into one context: the module retrieves candidate files and expands along dependencies by reading them, without a precomputed call graph. For each activated component, the plan gives an objective, dependencies, communication targets, and an expected action (\texttt{inspect}, \texttt{modify}, or \texttt{validate}), followed by an execution plan.

\textbf{Collaboration and Execution.}
The activated primitives operate as one round, and HERMES applies their edits:
\begin{equation}
\{a_i'\}_{i\in\mathcal{A}}\leftarrow\textsc{Collaborate}\bigl(\{P_i\}_{i\in\mathcal{A}},\,\{x_i\}_{i\in\mathcal{A}}\bigr),\qquad
\mathcal{R}'\leftarrow\textsc{UpdateRepository}\bigl(\mathcal{R},\,\{a_i'\}_{i\in\mathcal{A}}\bigr).
\label{eq:dev-messages}
\end{equation}
Communication is interleaved with local modification and follows the repository dependencies encountered during execution, so a recipient may revise its modification after receiving a requirement within the round. The modified repository is executed in the benchmark-provided environment,
\begin{equation}
o=\operatorname{EXECUTE}(\mathcal{R}')=\bigl(o_{\mathrm{shell}},\,o_{\mathrm{test}},\,o_{\mathrm{runtime}},\,o_{\mathrm{trace}}\bigr),
\label{eq:dev-execute}
\end{equation}
which records shell commands, outputs, exit codes, and build status; test results; runtime behaviour and exceptions; and logs and stack traces. During solving, all modules use task-visible checks, such as repository tests, build commands, linters, and reproduction scripts written while solving. Held-out evaluation tests, such as the \texttt{FAIL\_TO\_PASS} and \texttt{PASS\_TO\_PASS} tests of SWE-bench Verified, are withheld until HERMES terminates.

\textbf{Bug Diagnosis and Revision.}
Given the issue, the plan, the modified repository, and the execution observation, one LLM call returns the verdict
\begin{equation}
v=\operatorname{DIAGNOSE}(q,\Pi,\mathcal{R}',o)\in\{\pass,\fail\},
\label{eq:dev-diagnose}\csname ltx@label\endcsname{eq:dev-check}
\end{equation}
and, on \fail{}, structured feedback $\phi=(e,c,u)$: the observed failure $e$, the suspected root cause $c$ together with the components implicated by the evidence, including components that activation may have missed, and revision guidance $u$ for each component. Diagnosis does not modify artifacts and cannot override deterministic execution failures: a trajectory with failing task-visible tests or unsuccessful compilation cannot be marked \pass{} on model judgment alone. Activation then revises the plan,
\begin{equation}
\Pi'=\operatorname{ACTIVATE}(q,\mathcal{R}',\Pi,\phi),
\label{eq:dev-revise}
\end{equation}
retaining the modifications that new execution evidence does not contradict, updating objectives, removing components that are no longer relevant, and activating newly implicated Dev-Primitives. HERMES returns $\mathcal{R}'$ on \pass{} and a structured failure report $\textsc{FailureReport}(q,\mathcal{R}',o)$ once the budget $B$ is spent. $B$ counts the revision rounds after the initial plan--collaborate--execute--diagnose round, so $B=0$ performs that round without revision.

\textbf{Implementation.}
Each Dev-Primitive is instantiated with its own LLM, which may differ from the activation and diagnosis models (Table~\ref{tab:dev-roles}). The primary backbones are GPT-5.6 Sol, Terra, and Luna, Claude Sonnet 5, Claude Opus 5, DeepSeek-V4, and Qwen3-8B~\citep{qwen3technicalreport}. Reasoning effort is medium where configurable unless a table states otherwise, and $B=3$. Qwen3-8B is served locally with Ollama on NVIDIA A40 GPUs, with its default sampling parameters (temperature 0.6, top-$p$ 0.95, top-$k$ 20) and Ollama's 32K context window; all other models are accessed through their official APIs. SWE-bench Verified uses the official snapshots, SWE Refactor Bench its composite score, Terminal-Bench~4.0 five runs per task, and DevOps-Gym the unweighted average of its build and configuration, monitoring, issue-resolving, and test-generation categories. Token counts and costs include all activation, Dev-Primitive, and diagnosis calls. Baseline results come from the benchmark leaderboard or published evaluation where the model, harness, reasoning effort, and benchmark version are reported, and are otherwise run under the same snapshot and protocol as HERMES. On SWE Refactor Bench, each configuration runs on all 20 migration tasks, and the Claude and GPT-5.6 models are evaluated at medium and high effort (Table~\ref{tab:dev-refactor}).

\begin{table}[!htbp]
\centering
\researchtablesetup
\caption{\textbf{HERMES against same-backbone harnesses on SWE Refactor Bench.} Composite score (\%) and cost per task (USD) on the 20 whole-repository migration tasks, per model and reasoning effort, with $\Delta$ the composite change of HERMES in pp. Qwen3-8B is run with HERMES only~\citep{jin2026devprimitives}. Green marks the best and amber the second-best value in each Comp.\ and Cost column. Lower is better for cost.}
\label{tab:dev-refactor}
\fitresearchtable{\begin{tabular}{llccccc}
\toprule
\rowcolor{TabHead}
 & & \multicolumn{2}{c}{\hdr{Baseline Results}} & \multicolumn{2}{c}{\hdr{HERMES Results}} & \\
\hdrgapfill\cmidrule(lr){3-4}\cmidrule(lr){5-6}
\rowcolor{TabHead}
\multirow{-2}{*}{\hdr{Model}} & \multirow{-2}{*}{\hdr{Effort}} & \hdr{Comp. (\%)} & \hdr{Cost (\$)} & \hdr{Comp. (\%)} & \hdr{Cost (\$)} & \multirow{-2}{*}{\hdr{$\Delta$ Comp.}} \\
\midrule
\tabgroup{7}{Baseline harness: Claude Code} \\
 & medium & \second{28.5} & 38.4 & \second{36.5} & 58.6 & +8.0 \\
                \multirow{-2}{*}{Claude Opus 5} & high   & \best{34.5}   & 55.7 & \best{41.5}   & 82.4 & +7.0 \\
\midrule
 & medium & 15.0 & 11.9 & 22.0 & 22.7 & +7.0 \\
                \multirow{-2}{*}{Claude Sonnet 5} & high   & 6.0  & 24.6 & 25.5 & 34.8 & +19.5 \\
\midrule
Kimi-K3         & max    & 19.5 & 28.9 & 27.0 & 43.8 & +7.5 \\
\midrule
Qwen3.8-Max     & max    & 10.0 & 14.5 & 18.5 & 22.6 & +8.5 \\
\midrule
DeepSeek-V4     & max    & 7.0  & 4.3  & 15.0 & 11.6 & +8.0 \\
\midrule
GLM-5.2         & max    & 6.5  & 17.5 & 14.5 & 27.2 & +8.0 \\
\midrule
\tabgroup{7}{Baseline harness: Codex} \\
 & medium & 6.5  & 6.0  & 31.0 & 24.8 & +24.5 \\
                \multirow{-2}{*}{GPT-5.6 Sol} & high   & 19.0 & 7.7  & \second{36.5} & 39.6 & +17.5 \\
\midrule
 & medium & 4.5  & 3.4  & 26.0 & 14.6 & +21.5 \\
                \multirow{-2}{*}{GPT-5.6 Terra} & high   & 11.5 & 4.8  & 30.0 & 23.9 & +18.5 \\
\midrule
 & medium & 0.0  & \best{1.7}   & 16.0 & \best{4.9}   & +16.0 \\
                \multirow{-2}{*}{GPT-5.6 Luna} & high   & 4.0  & \second{1.8} & 20.0 & \second{7.2} & +16.0 \\
\midrule
\tabgroup{7}{No baseline harness} \\
Qwen3-8B & -- & -- & -- & 10.5 & 8.3 & -- \\
\bottomrule
\end{tabular}}
\end{table}

\textbf{Component Selection.}
Selection is measured against a target component set $\mathcal{G}$, with $\mathcal{A}$ the Dev-Primitives activated across all rounds: recall is $|\mathcal{A}\cap\mathcal{G}|/|\mathcal{G}|$ and precision $|\mathcal{A}\cap\mathcal{G}|/|\mathcal{A}|$, micro-averaged within each benchmark. Where a reference patch exists, $\mathcal{G}$ holds the files that it modifies, and otherwise the components that benchmark-specific supervision implicates; DevOps-Gym's monitoring tasks are excluded. With GPT-5.6 Sol, recall is 95.4\% on SWE-bench Verified, 83.3\% on SWE Refactor Bench, 84.5\% on Terminal-Bench~4.0, and 85.7\% on DevOps-Gym, precision ranges from 64.9\% to 72.6\%, and 4.7 to 12.6 Dev-Primitives are activated per task. Solved tasks exceed failed tasks in recall by 16.5 to 24.7 points, where a SWE Refactor Bench task counts as solved when its composite score is non-zero.

\textbf{Revision Budget.}
Raising $B$ from 0 to 3 adds 12.5 points on SWE Refactor Bench, 12.1 on Terminal-Bench, 9.50 on DevOps-Gym, and 4.8 on SWE-bench Verified, while raising it from 3 to 5 adds at most 0.9, which makes $B=3$ the default.\fillpar

% Pair body (tables/pairs/dev-controls+dev-ablation.tex); a float of its own if input directly.
\begin{wraptable}{R}{0.48\linewidth}
\wraplift
\centering
\researchtablesetup
\caption{\textbf{A centralized editor over the same components.} Terminal-Bench 4.0 with GPT-5.6 Sol at medium effort; all settings share the selected components, environment, diagnosis module and revision budget, and tokens (B) and cost (\$k) are evaluation totals~\citep{jin2026devprimitives}. Green marks the best and amber the second-best value in the Resolution column.}
\label{tab:dev-controls}
\fitresearchtable{\begin{tabular}{lccc}
\toprule
\rowcolor{TabHead}
\hdr{Setting} & \makecell[b]{\hdr{Resolution}\\\hdr{(\%)}} & \makecell[b]{\hdr{Tokens}\\\hdr{(B)}} & \makecell[b]{\hdr{Cost}\\\hdr{(\$k)}} \\
\midrule
\tabgroup{4}{GPT-5.6 Sol} \\
Single editor            & 38.8 & 5.9 & 2.90 \\
\quad + compute matching & \second{45.5} & 6.7 & 3.35 \\
\tabours HERMES          & \best{51.8} & 6.8 & 3.40 \\
\bottomrule
\end{tabular}}
\end{wraptable}

\textbf{Controls and Ablations.}
Table~\ref{tab:dev-controls} separates the Dev-Primitives from the choice of components on Terminal-Bench~4.0. The single editor keeps GPT-5.6 Sol, the activation-selected components, the environment, the diagnosis module, and the revision budget, and replaces the Dev-Primitives and their messages by one editor that receives the whole plan with the contents of all selected components and modifies them jointly. Across the four benchmarks, it uses 10--14\% fewer tokens and 11--16\% lower cost than HERMES. On Terminal-Bench~4.0, it resolves 38.8\% of the tasks against 51.8\% for HERMES. Its compute-matched variant generates several candidate revisions per round, evaluated with the same environment and diagnosis, until tokens and cost approximately match, 6.7B against 6.8B tokens, and reaches 45.5\%. Over the same components, the editor also loses 3.6~pp on SWE-bench Verified, 14.2~pp on SWE Refactor Bench, and 11.20~pp on DevOps-Gym, so on all benchmarks except SWE-bench Verified the replacement costs more than removing any single mechanism. Among the single-mechanism removals of Table~\ref{tab:dev-ablation}, removing inter-primitive communication keeps each Dev-Primitive's local reasoning and editing but exchanges no messages, and lowers SWE-bench Verified from 97.0\% to 94.2\%. Without on-demand activation, all Dev-Primitives identified in the initial task analysis stay active through all rounds, and primitive calls per SWE-bench Verified task rise from 9.6 to 28.7 (Table~\ref{tab:dev-activation}). Without execution feedback, diagnosis judges the repository by static inspection alone, and Terminal-Bench resolution falls to 43.6\%. Without diagnosis feedback, the raw execution outputs remain available and later rounds proceed, but no structured diagnosis or guidance is produced. Three repetitions of the default configuration on SWE Refactor Bench give composite scores of 29.5\%, 32.0\%, and 31.5\%, that is, $31.0\pm1.3$.

\textbf{Related Frameworks.}
A Dev-Primitive is defined by the artifact it owns, so the set of primitives is induced by the repository, whereas the libraries of Agent and Tool Primitives are built ahead of their tasks~\citep{jin2026agentprimitives,jin2026toolprimitives}. MAGIS and CodeTeam instead assign work to designer-specified roles or centrally coordinated file-level developers~\citep{tao2024magis,wang2026codeteam}, skill-based methods encapsulate reusable, task-agnostic behaviours~\citep{li2025codeskill}, and context-management methods reorganize information within one centralized trajectory~\citep{liu2026context,wang2026improving}. In their original evaluation settings, MAGIS reports 13.9\% resolution on SWE-bench against 36.6\% for HERMES with GPT-5.6 Luna, and CodeTeam reports average test pass rates of 34.6\% and 42.3\% on NL2Repo-Bench~\citep{ding2025nl2repo} for its prompting and supervised fine-tuning variants, against 44.9\% for HERMES.

\textbf{Relation to the Method.}
Dynamic Primitive Activation (Eq.~\eqref{eq:dev-activate}) realizes COMPOSE (Eq.~\eqref{eq:compose}), and Dev-Primitive Collaboration (Eq.~\eqref{eq:dev-messages}) the act step of Eq.~\eqref{eq:act}. Execution and Bug Diagnosis (Eqs.~\eqref{eq:dev-execute} and~\eqref{eq:dev-diagnose}) realize EXECUTE and VERIFY, with $\phi=(e,c,u)$ as the feedback $\phi_t$, and the re-planned activation of Eq.~\eqref{eq:dev-revise} is the revision of Eq.~\eqref{eq:revise}. The repository induces the library, which grows by the persistent files that active Dev-Primitives create.
\FloatBarrier

\subsection{Navigation Primitives: Need-Conditioned Coordination}
\label{sec:navigation-mechanism}
\begin{figure}[!htbp]
\centering
\includegraphics[width=\linewidth]{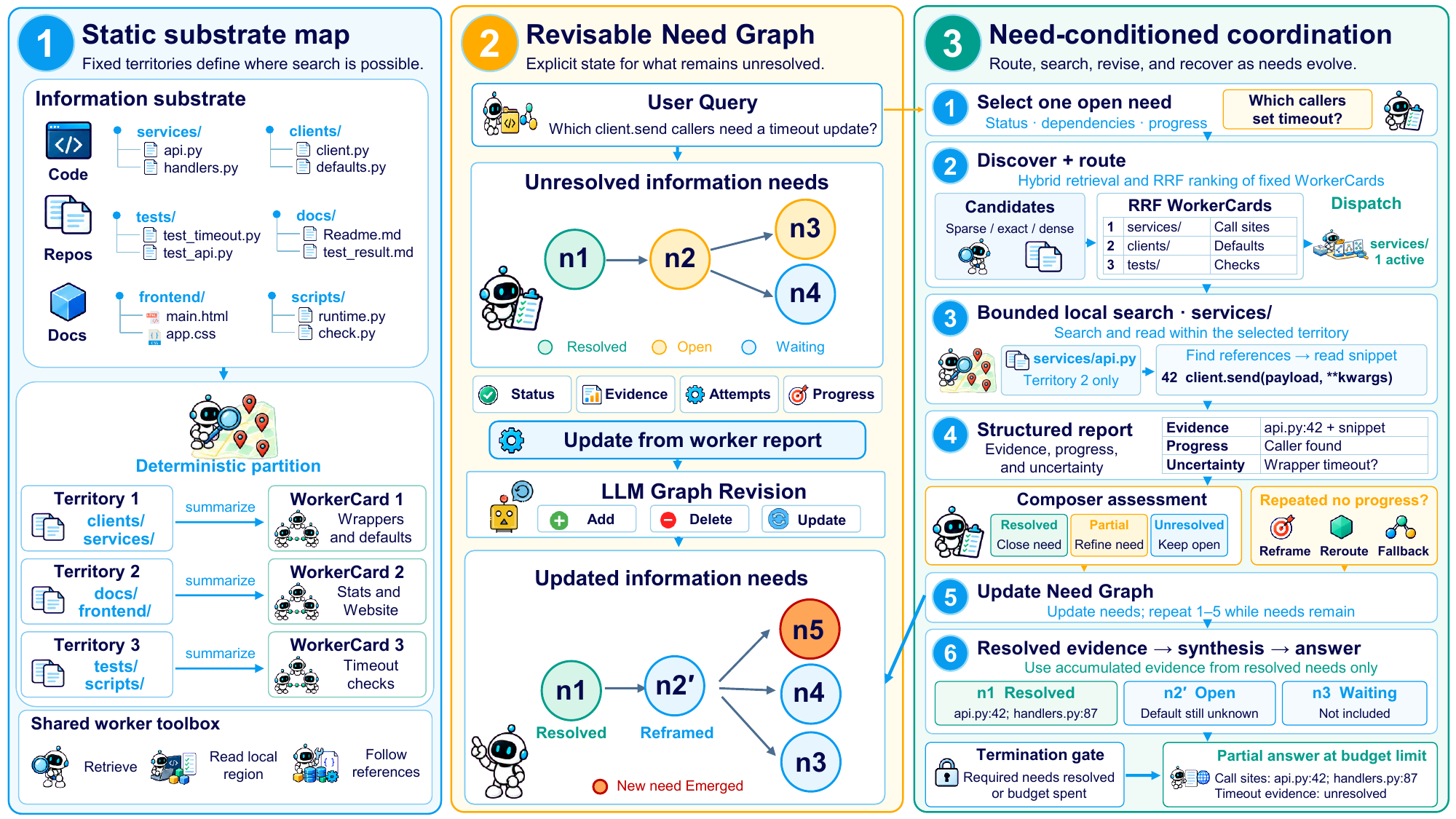}
\caption{\textbf{Overview of ANTMAN.} \textbf{(1)~Static substrate map:} a deterministic partition splits the information substrate, such as code, repositories, and documents, into fixed territories, each summarized by a WorkerCard, and every worker uses one shared toolbox. \textbf{(2)~Revisable Need Graph:} the user query becomes unresolved information needs whose status, evidence, attempts, and progress are updated from each worker report; an LLM graph revision then adds, deletes, or updates needs, so that a need can be resolved or reframed and a new need can emerge. \textbf{(3)~Need-conditioned coordination:} one open need is selected and routed through WorkerCard retrieval to one territory worker, which searches only its territory and returns evidence, progress, and uncertainty; the assessment closes, refines, or keeps the need open, repeated lack of progress triggers reframing, rerouting, or fallback, and the answer is synthesized from resolved evidence once the required needs are resolved or the budget is spent. The example follows the query ``Which client.send callers need a timeout update?'' over a small repository~\citep{jin2026antman}.\fillcap}
\label{fig:navigation-overview}
\end{figure}

Navigation Primitives are scope-specialized workers, each searching one territory of an information space, and ANTMAN~\citep{jin2026antman}, an adaptive coordination framework, decides which of them search for a query. ANTMAN answers a query $q$ from an information space $\mathcal{I}$ whose available information can grow large while the information that $q$ requires stays comparatively small. Newly encountered information can change the searcher's understanding of the problem and redirect the search~\citep{10.1108/eb024320}, so ANTMAN makes evolving unresolved information needs the runtime control state for coordination, which lets worker activation and search decisions adapt as those needs change. The method has three parts, shown in Figure~\ref{fig:navigation-overview}: a static substrate map, a revisable Need Graph, and need-conditioned coordination and recovery.

\textbf{Static Substrate Map.}
The map describes where ANTMAN can search. A substrate-specific partitioning procedure $P$ deterministically splits $\mathcal{I}$ into $M$ territories, bounded searchable regions, and each territory receives a WorkerCard, lightweight metadata that summarizes its scope and content:
\begin{equation}
\mathcal{T}=P(\mathcal{I})=\{\tau_1,\ldots,\tau_M\},\qquad
c_i=\mathrm{Card}(\tau_i),\qquad
\mathcal{C}=\{c_1,\ldots,c_M\}.
\label{eq:navigation-primitive}
\end{equation}
The worker $w_i$ of territory $\tau_i$ is scope-specialized: ANTMAN requires no distinct model architectures, tool sets, or personas, and within a substrate all workers share one interface for searching, inspecting, and returning evidence. Eq.~\eqref{eq:navigation-primitive} defines the workers that are available, not those that must participate in a query, and the map stays fixed during execution.\fillpar

\textbf{Revisable Need Graph.}
The Need Graph describes what remains unresolved. The coordinator initializes $G_0=\mathrm{InitializeNeedGraph}(q)$ and maintains the runtime state
\begin{equation}
G_t=(N_t,\Delta_t),\qquad
z_t(n)=\bigl(s_t(n),\,\mathcal{E}_t(n),\,\mathcal{H}_t(n),\,p_t(n)\bigr),
\label{eq:need-state}
\end{equation}
where $N_t$ is the set of information needs at step $t$, $\Delta_t$ holds their dependencies, and each need $n$ carries its resolution status $s_t(n)$, accumulated evidence $\mathcal{E}_t(n)$, attempt history $\mathcal{H}_t(n)$, and progress state $p_t(n)$. The unresolved needs form $\mathcal{U}_t=\{n\in N_t\mid s_t(n)\neq\mathrm{resolved}\}$; resolved needs stay in $N_t$, and final synthesis reads their evidence. After each report $r_t$, the task-local update $G_{t+1}=U(G_t,r_t)$ may resolve a need, keep it unresolved, reframe it into $n'$, or add requirements and dependencies that new evidence reveals, so needs may branch during execution. Within $U$, \emph{evidence selection} keeps the evidence most relevant to the query and discards redundant or unsupported material, and \emph{need-resolution assessment} judges the need resolved, partial, or unresolved: a resolved need is closed, a partial need is replaced or refined into a more specific unresolved requirement, and an unresolved need stays active. \emph{Need Graph consolidation} adds each proposed need as a new requirement, attaches it to an existing dependency, merges it with an overlapping need, subsumes it under an existing one, or discards it if redundant.

\textbf{Need-Conditioned Coordination and Recovery.}
At step $t$, ANTMAN selects an unresolved need, routes it with the Need Graph and the WorkerCards, and invokes the corresponding territory-backed worker:
\begin{equation}
n_t=\mathrm{Select}(G_t),\qquad
i_t=\mathrm{Route}(n_t,G_t,\mathcal{C}),\qquad
r_t=\mathrm{Exec}(w_{i_t},n_t,\tau_{i_t}),
\label{eq:need-route}
\end{equation}
where $i_t\in\{1,\ldots,M\}$ identifies the selected worker. Route has two stages: a non-LLM retrieval stage ranks the WorkerCards by sparse lexical, exact-term, and dense matching, combined by reciprocal-rank fusion, into a bounded candidate set, and the orchestrator then decides from $G_t$ whether and where to dispatch the need, respecting the dependencies between needs. Retrieval bounds the WorkerCard information within one routing decision, while the Need Graph governs the sequence of decisions and when search ends. The dispatched worker plans a bounded sequence of actions with its substrate's tools, searches its own territory, and returns a report $r_t$ of evidence, progress, and remaining uncertainty; global state remains in $G_t$. If repeated attempts make no progress, $\mathrm{Stuck}(n_t,G_{t+1})$ holds and $\mathrm{Recover}$ returns $\mathrm{Reframe}$, $\mathrm{Reroute}$ to another territory, or $\mathrm{Fallback}$ to an alternative resolution path, leaving the workers and the map unchanged. When the required needs are resolved or the execution budget $B$ is spent, ANTMAN synthesizes the answer from the evidence of resolved needs, which completes the execution procedure of Algorithm~\ref{alg:navigation}.

\begin{algorithm}[!htbp]
\caption{ANTMAN execution procedure~\citep{jin2026antman}. The substrate map remains fixed, while the Need Graph is revised during execution to select, route, and recover unresolved information needs.}
\label{alg:navigation}
\begin{algorithmic}[1]
\Require query $q$, information space $\mathcal{I}$, execution budget $B$
\Ensure final answer $y$
\State $\mathcal{T}\gets P(\mathcal{I})$ \Comment{fixed territories}
\State $\mathcal{C}\gets\{\mathrm{Card}(\tau_i)\}_{i=1}^{|\mathcal{T}|}$ \Comment{static WorkerCards}
\State $G_0\gets\mathrm{InitializeNeedGraph}(q)$
\State $t\gets 0$
\While{\textbf{not} $\mathrm{Terminate}(G_t,B)$}
  \State $n_t\gets\mathrm{Select}(G_t)$ \Comment{choose an unresolved need}
  \State $i_t\gets\mathrm{Route}(n_t,G_t,\mathcal{C})$ \Comment{select territory-backed worker}
  \State $r_t\gets\mathrm{Exec}(w_{i_t},n_t,\tau_{i_t})$ \Comment{bounded local information seeking}
  \State $G_{t+1}\gets U(G_t,r_t)$ \Comment{update evidence, status, and dependencies}
  \If{$\mathrm{Stuck}(n_t,G_{t+1})$}
    \State $a_t\gets\mathrm{Recover}(n_t,G_{t+1},\mathcal{C})$
    \If{$a_t=\mathrm{Reframe}$}
      \State $G_{t+1}\gets\mathrm{ReframeNeed}(n_t,G_{t+1})$
    \ElsIf{$a_t=\mathrm{Reroute}$}
      \State $\mathrm{MarkForRerouting}(n_t,G_{t+1})$
    \ElsIf{$a_t=\mathrm{Fallback}$}
      \State $G_{t+1}\gets\mathrm{Fallback}(n_t,G_{t+1})$
    \EndIf
  \EndIf
  \State $t\gets t+1$
\EndWhile
\State $\mathcal{E}^{*}\gets\mathrm{SelectResolvedEvidence}(G_t)$
\State $y\gets\mathrm{Synthesize}(q,\mathcal{E}^{*})$
\State \Return $y$
\end{algorithmic}
\end{algorithm}

\textbf{Runtime Coordination Demand.}
Because the graph is revised, the needs realized over a query can outnumber the initial set $N_0$:
\begin{filldisp}\begin{equation}
\mathcal{H}(q)=\bigcup_{t=0}^{T_q}N_t,\qquad H(q)=|\mathcal{H}(q)|,
\label{eq:navigation-realized}
\end{equation}\end{filldisp}
where $T_q$ is the final coordination step and a reframed or newly introduced need counts as new. A need $n$ receives $a(n)\le\rho$ routing attempts, its initial routing and any reroutes, and the worker-dispatch events and distinct activated workers are
\begin{filldisp}\begin{equation}
K(q)=\sum_{n\in\mathcal{H}(q)}a(n),\qquad
\mathcal{A}(q)=\bigcup_{t=0}^{T_q-1}\{i_t\}.
\label{eq:navigation-dispatch}
\end{equation}\end{filldisp}
Each routing decision at Line~7 of Algorithm~\ref{alg:navigation} concerns one need and activates at most one worker, so $|\mathcal{A}(q)|\le K(q)\le\rho H(q)$, and with $|\mathcal{A}(q)|\le M$ this gives ANTMAN's Proposition~1, $|\mathcal{A}(q)|\le\min\{M,\rho H(q)\}$, which Proposition~\ref{prop:activation} restates in the notation of Section~\ref{sec:method} and Appendix~\ref{app:proofs} proves together with its corollary. A partition-driven method that activates a constant fraction of its $M$ partitions instead has $|\mathcal{A}_{\mathrm{partition}}(q)|=\Theta(M)$. A larger space enlarges the candidate pool, which stays passive until the Need Graph creates demand for it.

\textbf{Implementation.}
ANTMAN runs all components on GPT-4.1, and ANTMAN-H keeps the GPT-4.1 orchestrator and runs all non-orchestrator components on Qwen3-8B~\citep{qwen3technicalreport}. The information-access layer adapts to each substrate, while the Need Graph, the worker organization, and the coordination procedure stay the same. In the controlled long-context setting, workers use lexical and semantic retrieval with bounded reading of selected document regions; repository navigation adds structural navigation over symbols, imports, references, inheritance, and call relations. GAIA-Text-103 uses a single territory-backed worker with web search, backed by Tavily, webpage text extraction, and isolated Python execution, an interface that all methods on GAIA share. Agentic methods on RepoProbe-Python and all ablation variants run under at most 100 LLM calls per query, and the iterative sparse retrieval baseline allows up to three lexical retrieval rounds, in which the LLM judges whether the accumulated evidence suffices and reformulates the query.

\textbf{Evaluation Settings.}
Multi-document QA scores 30 questions from each of HotpotQA~\citep{yang-etal-2018-hotpotqa}, 2WikiMultiHopQA~\citep{ho-etal-2020-constructing}, and MuSiQue~\citep{trivedi-etal-2022-musique} for nine methods, 810 predictions in all, with one frozen answer-extraction and evaluation pipeline, and its baselines also run on GPT-4.1. In the space intervention, each per-length entry averages 30 evaluations, ten questions with evidence at early, middle, and late positions. The demand study uses ten matched 512K base contexts, each with 16 planted evidence records over a fixed pool of eight territories; documents, distractors, evidence locations, territory assignment, and WorkerCards stay fixed across demand levels, while the query and its gold answer change with the level. Each record holds an independently generated value $e_i\in\{0,\ldots,7\}$, and at level $k\in\{1,4,16\}$ the answer is $y_k=\bigl(\sum_{e_i\in D_k}e_i\bigr)\bmod 8$ over nested required-evidence sets $D_1\subset D_4\subset D_{16}$, so omitting any required unit leaves the answer ambiguous over all eight classes. Runs are capped at ten coordination rounds, answers are scored by exact match, and Complete Evidence counts a unit as recovered when its content is extracted, rather than when its document is accessed. The ablations use the 512K condition, a 40-question SWE-QA-Pro subset sampled with seed 42, and a fixed 40-question GAIA subset. Their variants keep the territories, WorkerCards, models, tools, execution budget, and synthesis procedure, and the Need-Graph variants also share the initial decomposition and Need Graph. Graph-free Adaptive receives the same query, WorkerCards, accumulated evidence, and interaction history but no explicit Need Graph; w/o Need Revision prevents changes to the graph's structure or formulation, w/o Adaptive Rerouting fixes each need's worker after its initial routing, and w/o Recovery disables stuck detection, reframing, and fallback.

\begin{figure}[!htbp]
\centering
% Three panels in one row at one common scale (equal heights, about 0.32\linewidth each).
\includegraphics[height=0.223\linewidth]{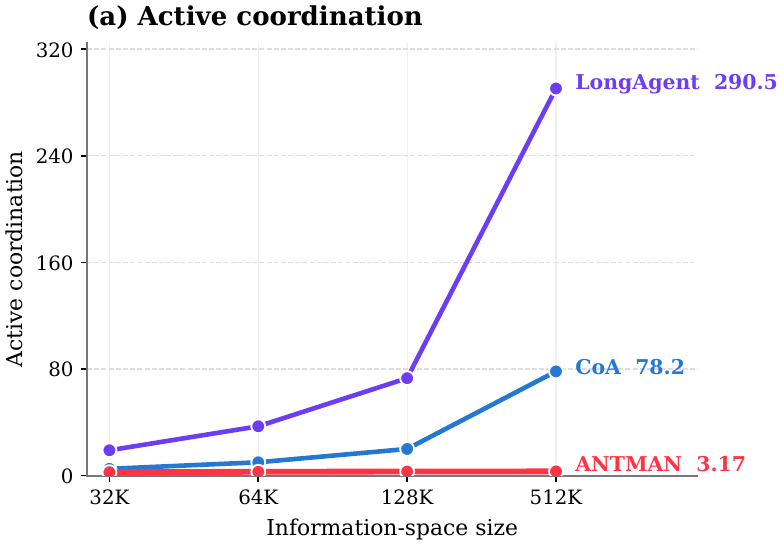}\hfill
\includegraphics[height=0.223\linewidth]{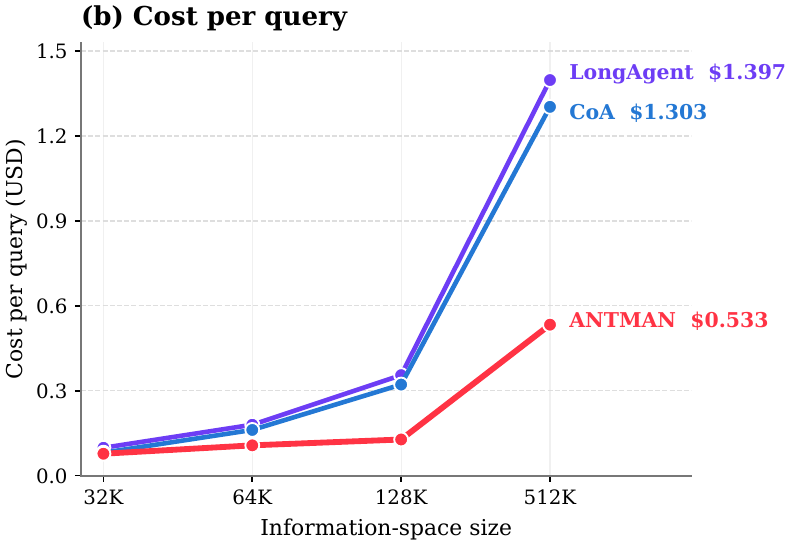}\hfill
\includegraphics[height=0.223\linewidth]{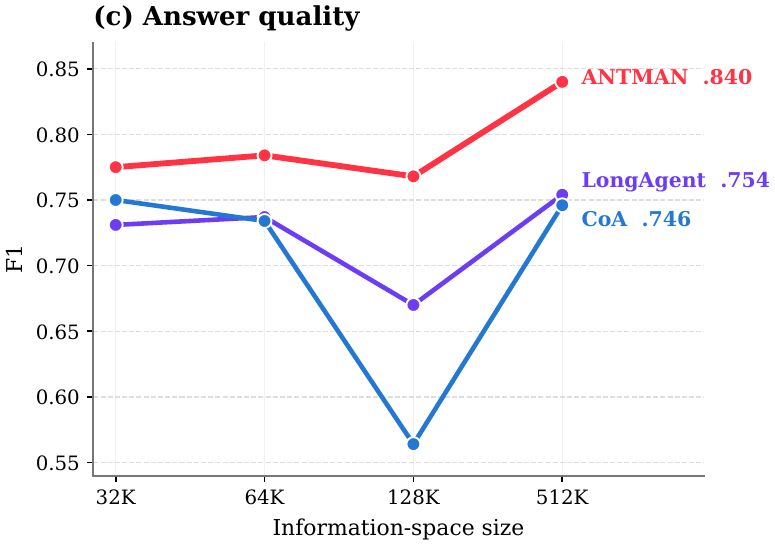}
\caption{\textbf{ANTMAN's active coordination stays nearly flat as the information space grows.} Active coordination (a), cost per query in USD (b), and F1 (c) of ANTMAN, LongAgent, and CoA as distractor filler grows the context from 32K to 512K tokens around preserved evidence at approximately fixed demand. The study uses ten questions with evidence at early, middle, and late positions and GPT-4.1 for all components. Per-length values are in Table~\ref{tab:navigation-full-scaling}~\citep{jin2026antman}.}
\label{fig:navigation-scaling}
\end{figure}

\textbf{Additional Results.}
On multi-document QA (Table~\ref{tab:navigation-multidoc}), ANTMAN's macro average of 0.711 EM and 0.814 F1 leads the specialized S2G-RAG~\citep{li-etal-2026-s2g}, at 0.700 and 0.799, so need-driven coordination preserves answer quality on standard tasks, and ANTMAN-H, at 0.622 and 0.763, is above all baselines except S2G-RAG. Across the lengths of the space intervention, ANTMAN keeps its active coordination nearly constant and its F1 stable, while the active coordination of LongAgent and CoA grows with the space (Figure~\ref{fig:navigation-scaling} and Table~\ref{tab:navigation-full-scaling}). Because it narrows the context searched for each unresolved need, ANTMAN is also robust to evidence position: averaged over the 32K to 128K contexts, its Middle F1 minus the mean of its Early and Late F1 is $+0.015$, against $-0.181$ for direct full-context inference, which shows the lost-in-the-middle pattern~\citep{liu-etal-2024-lost}. Under growing demand, the trajectory-level split of worker activity traces the dip in active coordination to exploration at low demand: at one unit, 4.1 of 5.0 worker activities are unrelated to the required evidence (Table~\ref{tab:navigation-demand-diagnostics}).

\begin{table}[!htbp]
\centering
\researchtablesetup
\caption{\textbf{ANTMAN on standard multi-document QA.} EM and token-level F1 on their native $[0,1]$ scales over 30 questions per benchmark, 810 predictions in all, scored with one frozen answer-extraction and evaluation pipeline, with macro averages over the three benchmarks; all methods use GPT-4.1, and ANTMAN-H$^\dagger$ keeps the GPT-4.1 orchestrator and runs all other components on Qwen3-8B~\citep{jin2026antman}. Green marks the best and amber the second-best value in each column.}
\label{tab:navigation-multidoc}
\fitresearchtable{\begin{tabular}{lcccccccc}
\toprule
\rowcolor{TabHead}
 & \multicolumn{2}{c}{\hdr{HotpotQA}} & \multicolumn{2}{c}{\hdr{2WikiMultiHopQA}} & \multicolumn{2}{c}{\hdr{MuSiQue}} & \multicolumn{2}{c}{\hdr{Avg.}} \\
\hdrgapfill\cmidrule(lr){2-3}\cmidrule(lr){4-5}\cmidrule(lr){6-7}\cmidrule(lr){8-9}
\rowcolor{TabHead}
\multirow{-2}{*}{\hdr{Method}} & \hdr{EM} & \hdr{F1} & \hdr{EM} & \hdr{F1} & \hdr{EM} & \hdr{F1} & \hdr{EM} & \hdr{F1} \\
\midrule
\tabgroup{9}{Retrieval and RAG baselines} \\
Iterative Sparse Retrieval (BM25) & 0.533 & 0.684 & 0.367 & 0.515 & 0.433 & 0.582 & 0.444 & 0.594 \\
Dense Retrieval (DPR)             & 0.533 & 0.696 & 0.267 & 0.379 & 0.200 & 0.292 & 0.333 & 0.456 \\
ChainRAG                          & 0.567 & 0.711 & 0.733 & 0.791 & 0.500 & 0.654 & 0.600 & 0.719 \\
S2G-RAG                           & \second{0.667} & \second{0.810} & \best{0.800} & \best{0.858} & \second{0.633} & \second{0.730} & \second{0.700} & \second{0.799} \\
\midrule
\tabgroup{9}{Agentic and multi-agent baselines} \\
ReAct     & 0.533 & 0.734 & 0.567 & 0.781 & 0.467 & 0.626 & 0.522 & 0.714 \\
LongAgent & 0.433 & 0.655 & 0.467 & 0.697 & 0.400 & 0.494 & 0.433 & 0.616 \\
CoA       & 0.567 & 0.769 & 0.600 & 0.778 & 0.533 & 0.682 & 0.567 & 0.743 \\
\midrule
\tabours ANTMAN-H$^\dagger$ & \best{0.700} & \best{0.842} & 0.700 & 0.796 & 0.467 & 0.651 & 0.622 & 0.763 \\
\tabours ANTMAN             & \best{0.700} & \best{0.842} & \second{0.767} & \second{0.823} & \best{0.667} & \best{0.777} & \best{0.711} & \best{0.814} \\
\bottomrule
\end{tabular}}
\end{table}

\textbf{Mapping to the Method.}
Within the method of Section~\ref{sec:method}, a territory $\tau_i$ is the resource $a_i$ of a Navigation Primitive, and its WorkerCard $c_i$ is the interface $d_i$. Select and the two-stage Route realize COMPOSE one need at a time, the need-resolution assessment inside $U$ plays the role of VERIFY, and recovery is the revision step of Eq.~\eqref{eq:revise}.
\FloatBarrier

\subsection{Harness Primitives: Composing Task-Specific Harnesses}
\label{sec:harness-mechanism}
\begin{figure}[!htbp]
\centering
\includegraphics[width=\linewidth]{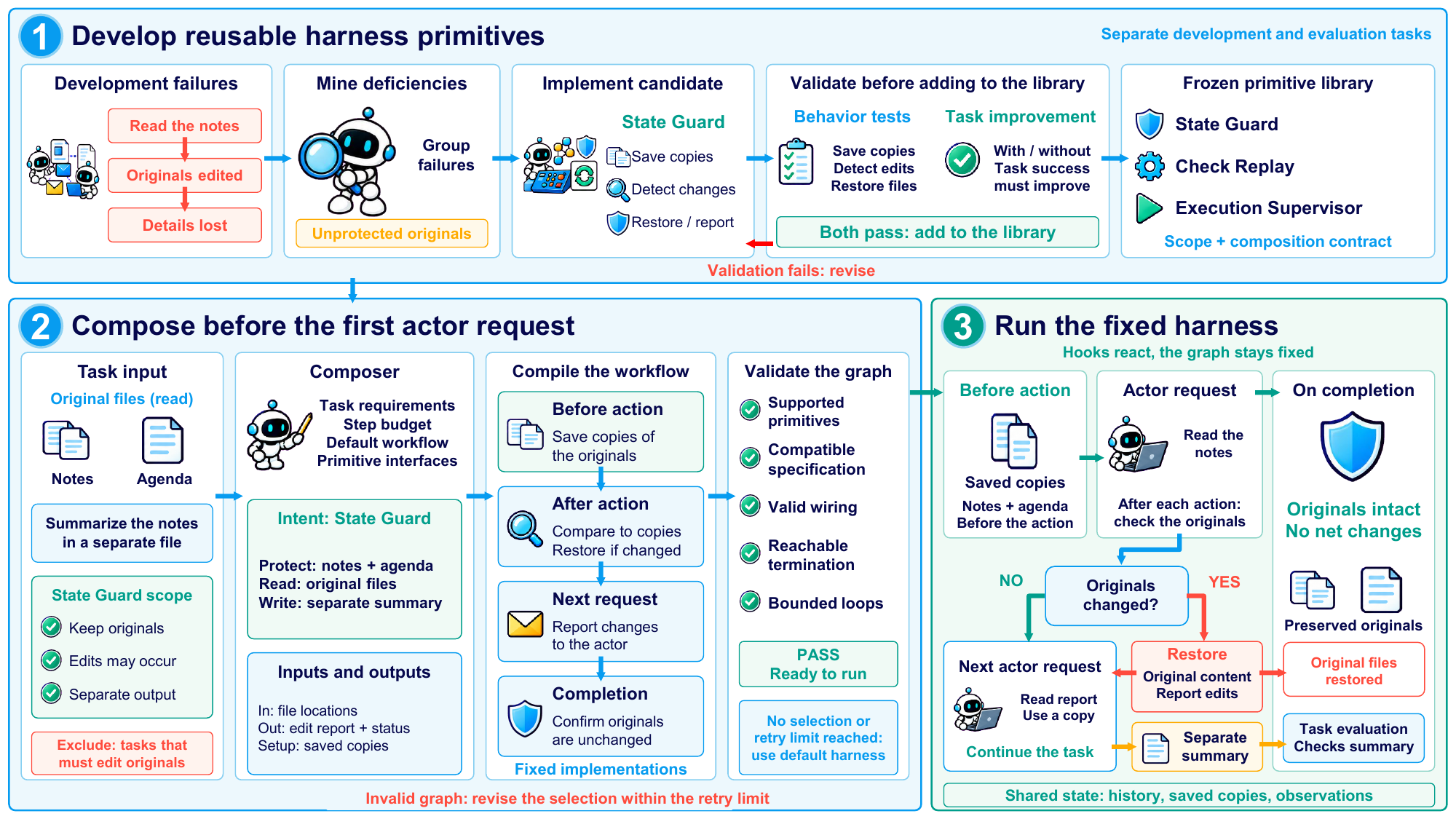}
\caption{\textbf{Overview of STITCH, shown with State Guard.} (1)~On development tasks kept apart from evaluation, recurring failures, such as an actor editing the original files whose details it later needs, are grouped into deficiencies; a candidate primitive is implemented and enters the frozen library only after it passes behavior tests and improves task success over the same harness without it, and each primitive carries an application scope and a composition contract. (2)~Before the first actor request, the harness composer reads the task requirements, step budget, default workflow, and primitive interfaces and states a composition intent, which the deterministic compiler wires into a workflow; the graph is validated for supported primitives, compatible specifications, valid wiring, reachable termination, and bounded loops, an invalid graph is revised within a retry limit, and the default harness is used otherwise. (3)~The compiled harness then runs fixed: its hooks save copies before each action, check the originals after it, and restore and report any change to the actor~\citep{jin2026stitch}.}
\label{fig:harness-overview}
\end{figure}

Harness Primitives are reusable harness mechanisms with a clear application scope and composition contract, mined from failed task trajectories~\citep{jin2026stitch}. Based on them, STITCH \textbf{S}elects suitable primitives given \textbf{T}ask \textbf{I}nformation and compiles them into \textbf{T}ask-spe\textbf{C}ific \textbf{H}arnesses at test time, keeping the development of Harness Primitives separate from their composition into each task's harness. In Figure~\ref{fig:harness-overview}, development turns a recurring failure, checks that fail without triggering repair, into a validated repair primitive, and test-time composition selects and wires different primitives from the same library for each of three tasks. We state the framework below in STITCH's own notation.

\textbf{Problem.}
Let $x$ be a task instruction and $c_x$ an execution specification that describes the task-solving language model, the \emph{actor}, its interaction interface, and the task environment setup. A harness $H$ is the control program that mediates the actor's interaction with the environment, and for the resulting trajectory $\tau$, $R_x(\tau)\in\{0,1\}$ indicates task success. With the actor model and the environment fixed, a construction procedure $F$ maps the task information to a harness $H_x=F(x,c_x)$, and the goal is
\begin{equation}
H_x^\star\in\underset{H\in\mathcal{H}(c_x)}{\arg\max}\;\mathbb{E}_{\tau\sim p(\cdot\mid x,c_x,H)}\bigl[R_x(\tau)\bigr],
\label{eq:harness-objective}
\end{equation}
where $\mathcal{H}(c_x)$ is the set of compatible harnesses and $p(\cdot\mid x,c_x,H)$ the induced trajectory distribution.

\textbf{Primitives and Harness Graphs.}
The library $\mathcal{P}=\{p_1,\dots,p_M\}$ holds implemented, reusable control operations. A primitive's application scope describes when it is expected to facilitate the task and when its overhead or behavior may be undesirable, according to its success on the evaluated tasks; its composition contract specifies its inputs, outputs, required environment setup, and dependencies. Scopes guide the harness composer, and contracts guide the implementation of the harness compiler. A harness is a directed graph $G=(V,E)$ whose nodes are control operations, each invoking a primitive, and whose edges specify execution order and conditional branches; the operations communicate through shared execution state, which holds the interaction history and intermediate results. The seed harness $H_{\mathrm{seed}}$ is abstracted as a graph $G_{\mathrm{seed}}$, and the task-specific graph $G_x$ represents $H_x$.

\textbf{State Guard.}
State Guard, one of the five Terminal-Bench primitives, addresses a recurring failure in which actor actions destroy inputs needed later. It blocks recognized direct writes, restores files that an action changes, and reports the affected paths in a failed-action observation. Its scope selects it when local inputs are irreplaceable, actions plausibly risk modifying them, and outputs are separate, and it excludes files that the task requires editing in place. Its contract has the source paths to protect and the declared output paths as inputs, mutation reports and integrity-check results as outputs, file access and storage for recoverable copies as setup, and as dependencies a snapshot before the first actor action, a check after each action, violation feedback to the next actor request, and the final integrity check for completion logic.\fillpar

\textbf{Primitive Development.}
A development agent, distinct in role from the actor, runs the actor under $H_{\mathrm{seed}}$ on a development set $\mathcal{D}_{\mathrm{dev}}$ and analyzes the failed attempts ($R_x(\tau)=0$) for recurring deficiencies. For each deficiency, it proposes a mechanism, specifies when it should activate and which subsequent actor decision it should influence, and implements it. Mechanism tests check that the operation activates and that its output changes the relevant request, action, or control-flow decision, and matched development runs with and without the candidate measure its benefit to $R_x(\tau)$. This evidence drives iterative revision and library acceptance: only primitives with a positive effect are accepted, development ends once the number of valid primitives passes a threshold $\theta$, and each accepted primitive then receives its scope and contract. The seed harness is mini-SWE-agent 2.4.6~\citep{yang2024swe}, the development model GPT-5.6 Sol, and the actor during development GPT-5.6 Luna; the 100 SWE-bench and 44 Terminal-Bench development tasks are sampled at random and disjoint from the evaluation samples. For an actor model or a task domain not used in development, implementations and contracts stay unchanged, and the scopes are adjusted on development tasks.\fillpar[0.6][1]

\textbf{Test-Time Composition.}
Before the first actor request, the composer $\pi$, which shares the actor's backbone (GPT-5.6 Luna at medium reasoning effort), reads the task, $c_x$, the budget, a compact description of $G_{\mathrm{seed}}$, and the whole primitive catalog in one \texttt{initial\_single\_pass} prompt. It applies each primitive's \texttt{choose\_only\_if} criteria literally and returns \texttt{should\_mutate}, a rationale, and an intent. With \texttt{should\_mutate=false} and \texttt{intent=null}, it keeps the seed graph when no other primitive fully qualifies; otherwise the composition intent $I_x$ names primitive instances with their parameters at a design level, without generating their implementation code or handling their integration. The deterministic compiler $K$ then builds the candidate graph
\begin{equation}
I_x=\pi\bigl(x,c_x;\,G_{\mathrm{seed}},\mathcal{P}\bigr),\qquad
\widetilde{G}_x=K\bigl(G_{\mathrm{seed}},I_x;\,\mathcal{P},c_x\bigr),
\label{eq:harness-compile}
\end{equation}
connecting the selected primitives to the seed graph and supplying the dependencies that their contracts require; selecting a repository map, for instance, also inserts the map into the actor's request. The candidate is checked for supported primitives, compatibility with $c_x$, valid connections, reachable termination, and bounded loops. The diagnostics of an invalid candidate return to the composer for bounded correction, the seed harness is retained if correction fails, and $K$ builds the valid graph into $H_x=F(x,c_x)$, as summarized in Algorithm~\ref{alg:harness}. During the run, the structure of $G_x$ stays fixed while its conditional branches respond to observations, and selected verification and repair primitives return a failed check to the actor for repair. We record whether each selected primitive is activated and executed without runtime errors, and the benchmark scores $R_x(\tau)$ after the run.

\begin{algorithm}[!htbp]
\caption{Test-time harness composition in STITCH. The procedure and the composer prompt follow the companion study~\citep{jin2026stitch}.}
\label{alg:harness}
\begin{algorithmic}[1]
\Require task $x$, execution specification $c_x$, budget, seed graph $G_{\mathrm{seed}}$, fixed library $\mathcal{P}$ with application scopes and composition contracts
\State $(\texttt{should\_mutate},\ \text{rationale},\ I_x)\gets\pi(x,c_x;G_{\mathrm{seed}},\mathcal{P})$ \Comment{before the first actor request}
\If{\texttt{should\_mutate} is false}
  \State \Return $G_{\mathrm{seed}}$ \Comment{\texttt{intent=null} retains the seed graph}
\EndIf
\State $\widetilde{G}_x\gets K(G_{\mathrm{seed}},I_x;\mathcal{P},c_x)$ \Comment{connect primitives, supply contract dependencies}
\While{the candidate fails the check for supported primitives, compatibility with $c_x$, valid connections, reachable termination, or bounded loops}
  \If{the correction bound is reached}
    \State \Return $G_{\mathrm{seed}}$ \Comment{correction failed; the seed harness is retained}
  \EndIf
  \State send the diagnostics to $\pi$, which corrects $I_x$
  \State $\widetilde{G}_x\gets K(G_{\mathrm{seed}},I_x;\mathcal{P},c_x)$
\EndWhile
\State \Return the harness $H_x=F(x,c_x)$ that $K$ builds from the valid graph $G_x=\widetilde{G}_x$
\end{algorithmic}
\end{algorithm}

\textbf{Comparison with Code Generation.}
STITCH's analysis assigns zero utility to invalid harnesses: with $V(H)\in\{0,1\}$ indicating the absence of the modeled implementation failures and $R(x,H)$ the success probability given valid execution, a harness earns $J(x,H)=V(H)R(x,H)$. Proposition~\ref{prop:fixed-vs-adaptive} restates STITCH's Proposition~1 and part~1 of its Proposition~2, and part~2 compares composition with raw code generation. With $\Omega_i$ the application scope of $p_i$ and $\mu_i$ its task prevalence (Section~\ref{sec:analysis-adaptive}), let $k(x)=\sum_{i=1}^M\mathbf{1}(x\in\Omega_i)$ be the number of oracle-selected mechanisms, generated in one attempt without repair. If, conditional on $x$, each mechanism fails independently with probability $\epsilon\in(0,1)$ and the harness is valid if and only if none of these failures occurs, then $\mathbb{P}(V(H_{\mathrm{gen}})=1\mid x)=(1-\epsilon)^{k(x)}$, while compositions of validated primitives by the deterministic compiler are assumed to execute validly, $V(H_{\mathrm{comp}})=1$. Let $\bar k=\mathbb{E}_x[k(x)]=\sum_{i=1}^M\mu_i$, assume $k(x)=\bar k$ for all tasks, although the selected mechanisms may differ across tasks, and suppose that valid generated harnesses attain the performance $R(x,\mathbf{a}^*(x))$ of the oracle activation $a_i^*(x)=\mathbf{1}(x\in\Omega_i)$. If $\mathbb{E}[J(x,H_{\mathrm{comp}})]>0$, composition outperforms this oracle generator if and only if
\begin{filldisp}\begin{equation}
\bar{k}>k^*=\frac{\ln\bigl(\mathbb{E}[R(x,\mathbf{a}^*(x))]\,/\,\mathbb{E}[J(x,H_{\mathrm{comp}})]\bigr)}{-\ln(1-\epsilon)}\approx\frac{1}{\epsilon}\ln\frac{\mathbb{E}[R(x,\mathbf{a}^*(x))]}{\mathbb{E}[J(x,H_{\mathrm{comp}})]},
\label{eq:harness-generation}
\end{equation}\end{filldisp}
where the approximation holds for small $\epsilon$. With $A=\mathbb{E}_x[R(x,\mathbf{a}^*(x))]$ and $B=\mathbb{E}[J(x,H_{\mathrm{comp}})]$, the failure model gives $\mathbb{E}[J(x,H_{\mathrm{gen}})]=\mathbb{E}_x[(1-\epsilon)^{k(x)}R(x,\mathbf{a}^*(x))]=(1-\epsilon)^{\bar k}A$. Since $0<B\le A$, the inequality $B>(1-\epsilon)^{\bar k}A$ holds if and only if $\bar k>\ln(A/B)/(-\ln(1-\epsilon))$, and the approximation follows from $-\ln(1-\epsilon)=\epsilon+O(\epsilon^2)$. Part~2 thus characterizes a reliability trade-off under a constant mechanism count and independent implementation failures.\fillpar[0.6][1]

\textbf{Baselines and Settings.}
Each fixed-primitive baseline evaluates a fixed harness configuration that contains one designated focused control mechanism, its required dependencies, and shared supporting operations, which together form a minimal compatible set of operations handled by the compiler. The harness optimizers MemoHarness~\citep{huang2026memoharnessagentharnesseslearn} and MetaHarness~\citep{lee2026metaharnessendtoendoptimizationmodel} start from the same mini-SWE-agent seed and use GPT-5.6 Sol as their development model. The actors not used in development are GPT-5.6 Terra with medium reasoning, DeepSeek-V4-Flash without reasoning, and Claude-4.5-Haiku. Appendix~\ref{app:experiments-extra} reports the transfer of primitives across task domains and the composer's selection patterns: the composer keeps the seed harness for about half of the tasks and otherwise mostly selects a single primitive.\fillpar

\textbf{Additional Results.}
On SWE-bench Verified with the three actors not used in development (Table~\ref{tab:harness-transfer}), STITCH has the best average on all three metrics and the best scores with DeepSeek-V4-Flash and Claude-4.5-Haiku, while Codex leads with GPT-5.6 Terra, at 88.0\% against 86.5\% Pass@1; each fixed primitive exceeds the seed on the average.\fillpar[.6][1]

\begin{table}[!htbp]
\centering
\researchtablesetup
\caption{\textbf{Harness Primitives with actor models not used to develop the primitives, on SWE-bench Verified.} Implementations and composition contracts are unchanged, while each application scope is adjusted according to the new actor's execution on development tasks; Overall averages the three actors, and metrics are as in Table~\ref{tab:harness-swe}~\citep{jin2026stitch}. Green marks the best and amber the second-best value in each column.}
\label{tab:harness-transfer}
\fitresearchtable{\begin{tabular}{l*{12}{c}}
\toprule
\rowcolor{TabHead}
 & \multicolumn{3}{c}{\hdr{Overall}} & \multicolumn{3}{c}{\hdr{GPT-5.6 Terra}} & \multicolumn{3}{c}{\hdr{DeepSeek-V4-Flash}} & \multicolumn{3}{c}{\hdr{Claude-4.5-Haiku}} \\
\hdrgapfill\cmidrule(lr){2-4}\cmidrule(lr){5-7}\cmidrule(lr){8-10}\cmidrule(lr){11-13}
\rowcolor{TabHead}
\multirow{-2}{*}{\hdr{Method}} & \hdr{Pass@1} & \hdr{Pass@2} & \hdr{Pass\textsuperscript{2}} & \hdr{Pass@1} & \hdr{Pass@2} & \hdr{Pass\textsuperscript{2}} & \hdr{Pass@1} & \hdr{Pass@2} & \hdr{Pass\textsuperscript{2}} & \hdr{Pass@1} & \hdr{Pass@2} & \hdr{Pass\textsuperscript{2}} \\
\midrule
mini-SWE-agent (seed) & 75.2 & 80.3 & 70.0 & 82.0 & 84.0 & 80.0 & 82.5 & 89.0 & 76.0 & 61.0 & 68.0 & 54.0 \\
SWE-agent & 73.3 & 79.7 & 67.0 & 74.5 & 83.0 & 66.0 & \second{89.5} & \second{93.0} & \second{86.0} & 56.0 & 63.0 & 49.0 \\
Codex & 78.5 & 84.7 & 72.3 & \best{88.0} & \best{92.0} & \best{84.0} & 82.0 & 90.0 & 74.0 & 65.5 & 72.0 & 59.0 \\
\midrule
\tabgroup{13}{One focused Harness Primitive with its dependencies, fixed across tasks} \\
Change-surface Tracer & 77.7 & 83.7 & 71.7 & 79.5 & 84.0 & 75.0 & 87.5 & 92.0 & 83.0 & 66.0 & 75.0 & 57.0 \\
Compatibility-envelope Gate & 77.3 & 83.3 & 71.3 & 79.5 & 85.0 & 74.0 & 85.5 & 90.0 & 81.0 & 67.0 & 75.0 & 59.0 \\
Contract-case Explorer & 78.7 & 83.7 & 73.7 & 80.0 & 83.0 & 77.0 & 88.5 & \second{93.0} & 84.0 & 67.5 & 75.0 & 60.0 \\
Discriminating-oracle Runner & 76.3 & 81.0 & 71.7 & 77.5 & 82.0 & 73.0 & 86.5 & 90.0 & 83.0 & 65.0 & 71.0 & 59.0 \\
Domain-trace Template & 78.7 & \second{85.0} & 72.3 & 81.5 & 86.0 & 77.0 & 84.5 & 91.0 & 78.0 & \second{70.0} & \second{78.0} & \second{62.0} \\
Environment-capability Runner & \second{79.5} & 84.3 & \second{74.7} & 83.5 & 86.0 & 81.0 & 89.0 & \second{93.0} & 85.0 & 66.0 & 74.0 & 58.0 \\
\midrule
Random composer & 77.4 & 82.8 & 71.9 & 80.6 & 84.4 & 76.9 & 85.9 & 90.9 & 80.9 & 65.6 & 73.3 & 58.0 \\
\tabours STITCH & \best{84.0} & \best{88.3} & \best{79.7} & \second{86.5} & \second{90.0} & \second{83.0} & \best{91.5} & \best{95.0} & \best{88.0} & \best{74.0} & \best{80.0} & \best{68.0} \\
\bottomrule
\end{tabular}}
\end{table}

\begin{figure}[!htbp]
\centering
\begin{minipage}[t]{0.49\linewidth}\vspace{0pt}\includegraphics[width=\linewidth]{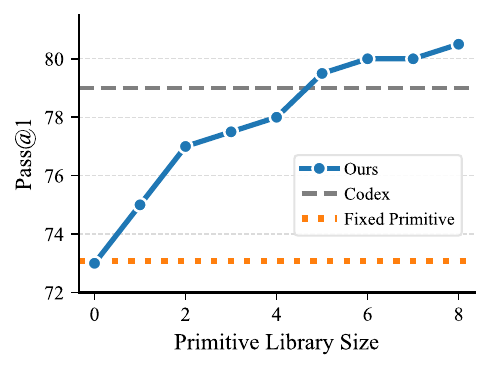}\end{minipage}\hfill
\begin{minipage}[t]{0.49\linewidth}\vspace{0pt}\includegraphics[width=\linewidth]{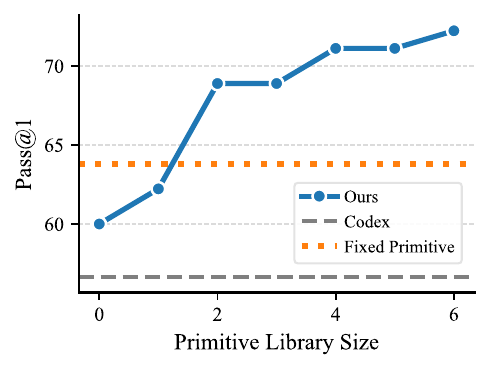}\end{minipage}
\caption{\textbf{Pass@1 of STITCH grows with the primitive library.} SWE-bench Verified (left) and Terminal-Bench~2.0 (right) as the test-time library grows from the seed harness (size 0) to 8 and 6 entries, along STITCH's single order of addition. The curve labelled Ours is STITCH. The dashed line marks Codex, and the dotted line marks the fixed-primitive arms at their mean Pass@1, whose best members reach 76.0\% and 66.7\% (Tables~\ref{tab:harness-swe} and~\ref{tab:harness-terminal})~\citep{jin2026stitch}.}
\label{fig:harness-library}
\end{figure}

As the test-time library grows along a single order of addition (Figure~\ref{fig:harness-library}), Pass@1 rises from 73.0\% for the seed harness (size 0) to 80.5\% at size 8 on SWE-bench Verified and from 60.0\% to 72.2\% at size 6 on Terminal-Bench~2.0, ending above Codex on both benchmarks.

\begin{figure}[!htbp]
\centering
\begin{minipage}[t]{0.49\linewidth}\vspace{0pt}\includegraphics[width=\linewidth]{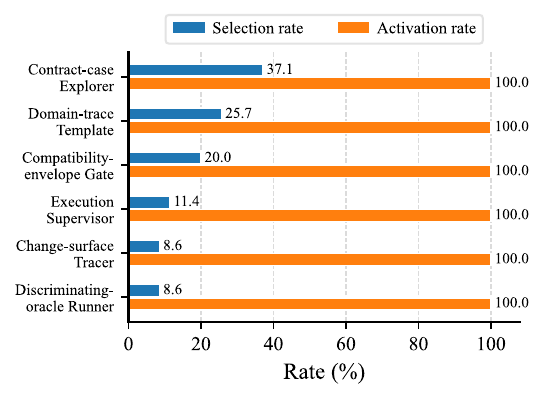}\end{minipage}\hfill
\begin{minipage}[t]{0.49\linewidth}\vspace{0pt}\includegraphics[width=\linewidth]{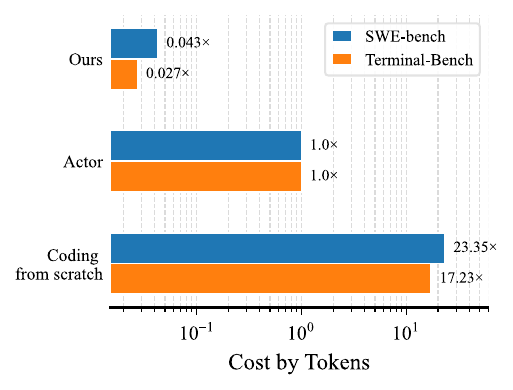}\end{minipage}
\caption{\textbf{Each selected Harness Primitive activates, and composition costs a few percent of one actor attempt.} Left: selection and activation rates (\%) per selected primitive, where activation means that the selected primitive was activated and executed without runtime errors. Right: token cost of composition (Ours) relative to one actor attempt on SWE-bench Verified and Terminal-Bench~2.0, against the output-token lower bound for an oracle that writes the complete harness with no input, reasoning, or debugging cost, on a logarithmic axis~\citep{jin2026stitch}.\fillcap[0.6][1]}
\label{fig:harness-overhead}
\end{figure}

The activation rate is 100\% for each selected primitive (Figure~\ref{fig:harness-overhead}), which supports part~2, the comparison with raw code generation: free of the reliability risk of generated harness code, the selected primitives take effect during the actor's execution. Composition costs 4.3\% of one actor attempt on SWE-bench Verified and 2.7\% on Terminal-Bench~2.0, whereas the output-token lower bound for an oracle that writes the complete harness, with no input, reasoning, or debugging cost, is 23.35 and 17.23 actor attempts, about $543\times$ and $638\times$ the cost of composition.

\textbf{Relation to the Method.}
The compiled graph $G_x$ is the plan of Eq.~\eqref{eq:compose}: the composition intent $I_x$ names its primitive instances and their parameters, and the compiler adds the dependencies along which results flow; the shared execution state is the context of Eq.~\eqref{eq:act}, and primitive development is STITCH's form of the validation and publishing of Section~\ref{sec:harness-engineering} (Eq.~\eqref{eq:publish}).
\FloatBarrier

\section{Additional Experimental Results}
\label{app:experiments-extra}
\runningsection{Additional results}

This appendix completes the experimental record of the eight primitive families, in the order of Section~\ref{sec:instantiations}. Each subsection extends the family's subsection there with complete benchmark breakdowns, retrieval and verification diagnostics, model allocations, provenance and budget controls, grounding audits, scaling trajectories, or selection patterns, and it names the mechanism of the family that its material bears on. Appendix~\ref{app:families} states each family's full method in its own notation, and all values are copied from the eight companion studies, whose settings and evidence Appendix~\ref{app:ledger} records. The tables of this appendix follow the table convention of Section~\ref{sec:experimental-setup}.

\subsection{Tool Primitives: Breakdowns, Retrieval, and Validation}

Section~\ref{sec:tool-experiments} compares HEART with the reference systems of ToolBench, NESTFUL, and $\tau^{2}$-Bench and removes one component at a time, a harness mechanism or ToolFace with its Tool Primitives (Tables~\ref{tab:tool-toolbench}, \ref{tab:tool-nestful}, \ref{tab:tool-tau}, and~\ref{tab:tool-ablation}); the controls at a matched backbone and tool catalogue and at a matched call budget accompany the framework in Appendix~\ref{sec:tool-mechanism} (Tables~\ref{tab:tool-matched} and~\ref{tab:tool-budget}). This subsection adds the diagnostics of retrieval, validation, and revision. HEART is evaluated on five benchmarks~\citep{jin2026toolprimitives}. ToolBench covers large-scale API invocation~\citep{qin2023toolllm}, NESTFUL nested call sequences, and $\tau^{2}$-Bench interactive dual-control execution~\citep{basu2025nestful,barres2025tau}. ACEBench targets fine-grained multi-tool workflows, and BFCLv4 agentic web search and memory function calling~\citep{chen2025acebench,patil2025bfcl}. The ACEBench and BFCLv4 results, which Section~\ref{sec:tool-experiments} summarizes, appear in full in the companion study's own tables. Three further suites are related benchmarks: StableToolBench, which replaces ToolBench's live APIs with deterministic virtual servers~\citep{guo2024stabletoolbench}, HammerBench~\citep{wang2025hammerbench}, and ToolSandbox~\citep{lu2025toolsandbox}. Each benchmark table follows its benchmark's own protocol.

The ToolBench reference rows include Vicuna~\citep{zheng2023judging}, Alpaca~\citep{alpaca}, and GPT-4~\citep{achiam2023gpt}. The NESTFUL comparison also lists Granite's function-calling model~\citep{abdelaziz2024granite}, Llama~3.1~\citep{grattafiori2024llama}, Mixtral~\citep{jiang2024mixtral}, DeepSeek-V3~\citep{liu2024deepseek}, and GPT-4o~\citep{hurst2024gpt}. Their configurations and scores follow the companion study. NESTFUL's own metrics are F1 for function and parameter names, partial and full sequence accuracy, and win rate.

The last three tables concern how HEART's harness finds Tool Primitives and judges their results. Table~\ref{tab:tool-retrieval} measures how often semantic retrieval over the 25,519 schema descriptors returns the gold tools. A plan step can invoke only a tool that retrieval returns, the containment behind the candidate-pool ceiling of Proposition~\ref{prop:coverage}(ii), and validation-triggered revision recovers part of the first-pass misses by reformulating the tool query. Table~\ref{tab:tool-verifier} compares result validation, which judges every step's result on four criteria, with human labels. HEART audits validation because its recovery loop is conditioned on it: most false passes fall on the task-completion criterion for partially satisfied objectives, fewer than 3\% of trajectories end on a false pass, false fails cost redundant revision rounds without changing outcomes, and the plateau at $B=3$ indicates that the loop terminates because validation converges (Appendix~\ref{sec:tool-mechanism}). Table~\ref{tab:tool-revision} varies HEART's maximum revision budget $B$; $B=0$ disables revision.
% Pair float: Table~\ref{tab:tool-retrieval} (left) and Table~\ref{tab:tool-verifier} (right), 9pt panels.
% Replaces \input{tables/restored/tool-retrieval} and \input{tables/restored/tool-verifier} (appendix).
% Each body keeps its own \caption and \label, so numbering and \ref's are unchanged.
\begin{table}[!htbp]
\centering\researchpaneltrue\researchpanelsize{9}
\begin{minipage}[t]{0.489\linewidth}% Pair body (tables/pairs/tool-retrieval+tool-verifier.tex); a float of its own if input directly.
\begin{researchtable}
\centering
\researchtablesetup
\caption{\textbf{Retrieval recall over the 25,519 tools of ToolFace.} Recall@$k$ (\%) of semantic retrieval over schema descriptors, with each task's gold tools as ground truth, measures candidate access separately from invocation success~\citep{jin2026toolprimitives}.}
\label{tab:tool-retrieval}
\fitresearchtable{\begin{tabular}{lcccc}
\toprule
\rowcolor{TabHead}
\hdr{Benchmark} & \hdr{Recall@1} & \hdr{Recall@3} & \hdr{Recall@5} & \hdr{Recall@10} \\
\midrule
ToolBench & 76.5 & 88.2 & 92.4 & 95.8 \\
NESTFUL   & 81.3 & 91.6 & 94.7 & 97.2 \\
\bottomrule
\end{tabular}}
\end{researchtable}
\end{minipage}\hfill
\begin{minipage}[t]{0.486\linewidth}% Pair body (tables/pairs/tool-retrieval+tool-verifier.tex); a float of its own if input directly.
\begin{researchtable}
\centering
\researchtablesetup
\caption{\textbf{Decision quality of HEART's result validation.} Result validation shares the Qwen3-8B backbone of the other roles; HEART's study compares 200 sampled validation decisions from ToolBench and NESTFUL trajectories with human pass/fail labels~\citep{jin2026toolprimitives}.}
\label{tab:tool-verifier}
\fitresearchtable{\begin{tabular}{lcccc}
\toprule
\rowcolor{TabHead}
\hdr{Metric} & \hdr{Precision} & \hdr{Recall} & \makecell[b]{\hdr{False-pass rate}\\\hdr{(\%)}} & \makecell[b]{\hdr{False-fail rate}\\\hdr{(\%)}} \\
\midrule
Value & 0.97 & 0.96 & 8 & 4 \\
\bottomrule
\end{tabular}}
\end{researchtable}
\end{minipage}
\end{table}

\begin{table}[!htbp]
\centering
\researchtablesetup
\caption{\textbf{HEART under different revision budgets.} The maximum revision budget $B$ bounds the revision rounds after a failed validation; $B$\,=\,0 disables revision, and the shaded $B$\,=\,3 is HEART's default; metrics as in Table~\ref{tab:tool-matched}~\citep{jin2026toolprimitives}. Green marks the best and amber the second-best value in each column.}
\label{tab:tool-revision}
\fitresearchtable{\begin{tabular}{lcccc}
\toprule
\rowcolor{TabHead}
 & \multicolumn{2}{c}{\hdr{ToolBench}} & \multicolumn{2}{c}{\hdr{NESTFUL}} \\
\hdrgapfill\cmidrule(lr){2-3}\cmidrule(lr){4-5}
\rowcolor{TabHead}
\multirow{-2}{*}{\hdr{Revision budget}} & \hdr{Pass} & \hdr{Win} & \hdr{Full Acc.} & \hdr{Win} \\
\midrule
$B$\,=\,0 & 47.6 & 49.6 & 0.15 & 0.33 \\
$B$\,=\,1 & 51.4 & 57.9 & 0.27 & 0.49 \\
$B$\,=\,2 & 65.9 & 66.8 & \second{0.42} & \second{0.68} \\
\tabours $B$\,=\,3 & \second{75.1} & \second{75.7} & \best{0.44} & \best{0.75} \\
$B$\,=\,5 & \best{75.3} & \best{76.2} & \best{0.44} & \best{0.75} \\
\bottomrule
\end{tabular}}
\end{table}

\FloatBarrier

\subsection{Code Primitives: Breakdowns, Controls, and Uncertainty}

Section~\ref{sec:code-experiments} compares LEGO with construction paradigms, consumption and field controls, and model allocations (Tables~\ref{tab:code-paradigms}, \ref{tab:code-consumption}, \ref{tab:code-fields}, and~\ref{tab:code-cost}). The tables below break delivery down by band and domain, restrict donor provenance, fix the attempt budget, give the uncertainty over decoding seeds and paired tasks, and report the independently developed agents by band with their cost. Table~\ref{tab:code-band-domain} breaks LEGO-REPO delivery down by difficulty band and domain. Across the five bands, the absolute gain over feedback stays between +0.161 and +0.209, so the larger relative gains on harder bands come mostly from a falling feedback score. Its coverage column measures offline reachability in the library. Table~\ref{tab:code-provenance} restricts library eligibility step by step, from excluding same-repository donors to a library re-mined from a disjoint corpus. These are measured library variants, and the gain over feedback stays positive under every restriction. The restrictions are possible because every CodeFace entry is admitted with a recorded source repository, commit, and license (Section~\ref{sec:building}). Table~\ref{tab:code-budget} fixes the same integer attempt limit for both configurations; at every reported limit from 1 to 8, the library configuration stays above feedback alone~\citep{jin2026codeprimitives}.
\begin{table}[!htbp]
\centering
\researchtablesetup
\caption{\textbf{Feedback and LEGO by difficulty band and software domain.} The two configurations of Table~\ref{tab:code-paradigms} partitioned two ways, with $\Delta$ their difference in mean delivery over each partition's tasks; coverage is the offline module-level reachability of the library~\citep{jin2026codeprimitives}.}
\label{tab:code-band-domain}
\fitresearchtable{\begin{tabular}{lcccccc}
\toprule
\rowcolor{TabHead}
\hdr{Partition} & \hdr{Tasks} & \hdr{Modules} & \hdr{Feedback} & \hdr{LEGO} & \hdr{$\Delta$} & \hdr{Coverage (\%)} \\
\midrule
D1 (mean 6 modules)   & 117 & 721 & 0.525 & 0.686 & \sn{+0.161} & 75.3 \\
D2 (mean 19 modules)  & 111 & 2{,}064 & 0.382 & 0.583 & \sn{+0.201} & 71.2 \\
D3 (mean 32 modules)  & 116 & 3{,}654 & 0.274 & 0.483 & \sn{+0.209} & 65.8 \\
D4 (mean 71 modules)  &  96 & 6{,}823 & 0.206 & 0.409 & \sn{+0.203} & 61.2 \\
D5 (mean 119 modules) &  82 & 9{,}723 & 0.129 & 0.338 & \sn{+0.209} & 56.7 \\
\midrule
machine-learning      & 158 & 9{,}649 & 0.2498 & 0.4368 & \sn{+0.1870} & 56.3 \\
databases-storage     & 149 & 5{,}537 & 0.3390 & 0.4946 & \sn{+0.1556} & 66.5 \\
compilers-languages   & 131 & 4{,}166 & 0.3640 & 0.6104 & \sn{+0.2464} & 68.5 \\
graphics-vision       &  32 & 1{,}420 & 0.3195 & 0.5055 & \sn{+0.1860} & 60.4 \\
scientific-computing  &  26 & 1{,}322 & 0.3179 & 0.5213 & \sn{+0.2034} & 58.3 \\
quant-finance         &  14 & 465 & 0.3711 & 0.5877 & \sn{+0.2166} & 62.4 \\
crypto-security       &  12 & 426 & 0.3880 & 0.6139 & \sn{+0.2259} & 52.1 \\
\midrule
all tasks             & 522 & 22{,}985 & 0.3180 & 0.5134 & \sn{+0.1954} & 61.4 \\
\bottomrule
\end{tabular}}
\end{table}

% Pair float: Table~\ref{tab:code-provenance} (left) and Table~\ref{tab:code-budget} (right), 8.5pt panels.
% Replaces \input{tables/restored/code-provenance} and \input{tables/restored/code-budget} (appendix).
% Each body keeps its own \caption and \label, so numbering and \ref's are unchanged.
\begin{table}[!htbp]
\centering\researchpaneltrue\researchpanelsize{8.5}
\begin{minipage}[t]{0.535\linewidth}\captionsetup{skip=17.4pt}% Pair body (tables/pairs/code-provenance+code-budget.tex); a float of its own if input directly.
\begin{researchtable}
\centering
\researchtablesetup
\caption{\textbf{Reuse under progressively stricter donor-provenance controls.} Every row is a measured library variant, from donor filters applied per target to a library re-mined from a corpus disjoint from the benchmark repositories; coverage as in Table~\ref{tab:code-band-domain}~\citep{jin2026codeprimitives}. Green marks the best and amber the second-best value in the All column.}
\label{tab:code-provenance}
\fitresearchtable{\begin{tabular}{lcccc}
\toprule
\rowcolor{TabHead}
\hdr{Library eligibility} & \hdr{Entries} & \makecell[b]{\hdr{Coverage}\\\hdr{(\%)}} & \hdr{All} & \makecell[b]{\hdr{$\Delta$ vs}\\\hdr{Feedback}} \\
\midrule
\tabours all entries (default)          & 1{,}424 & 61.4 & \best{0.5134} & \sn{+0.1954} \\
\textminus{} same-repository donors     & 1{,}424 & 41.2 & \second{0.4743} & \sn{+0.1563} \\
\textminus{} same-org donors            & 1{,}424 & 38.6 & 0.4597 & \sn{+0.1417} \\
\textminus{} any fork or near-duplicate & 1{,}365 & 33.1 & 0.4455 & \sn{+0.1275} \\
re-mined from disjoint corpus           & 518     & 44.7 & 0.4528 & \sn{+0.1348} \\
\bottomrule
\end{tabular}}
\end{researchtable}
\end{minipage}\hfill
\begin{minipage}[t]{0.440\linewidth}% Pair body (tables/pairs/code-provenance+code-budget.tex); a float of its own if input directly.
\begin{researchtable}
\centering
\researchtablesetup
\caption{\textbf{Executed attempt budgets with and without the library.} Both configurations use the same integer attempt limit $b$ per module on the same 522 tasks; $b$\,=\,5 is the default, $b$\,=\,1 rows are first-round prefixes, and \$/task is recorded-call cost as in Table~\ref{tab:code-cost}~\citep{jin2026codeprimitives}. Green marks the best and amber the second-best value in each All column.}
\label{tab:code-budget}
\fitresearchtable{\begin{tabular}{lccccc}
\toprule
\rowcolor{TabHead}
 & \multicolumn{2}{c}{\hdr{Feedback (\stg{-{}-{}-V})}} & \multicolumn{2}{c}{\hdr{LEGO (\stg{RADV})}} & \\
\hdrgapfill\cmidrule(lr){2-3}\cmidrule(lr){4-5}
\rowcolor{TabHead}
\multirow{-2}{*}{\hdr{Budget $b$}} & \hdr{All} & \hdr{\$/task} & \hdr{All} & \hdr{\$/task} & \multirow{-2}{*}{\hdr{$\Delta$}} \\
\midrule
1 & 0.2172 & 0.86 & 0.3954 & 1.13 & \sn{+0.1782} \\
2 & 0.2420 & 1.05 & 0.4412 & 1.38 & \sn{+0.1992} \\
3 & 0.2785 & 1.21 & 0.4663 & 1.61 & \sn{+0.1878} \\
4 & 0.3066 & 1.36 & 0.5071 & 1.82 & \sn{+0.2005} \\
\tabours 5 (default) & \second{0.3180} & 1.55 & \second{0.5134} & 2.08 & \sn{+0.1954} \\
8 & \best{0.3437} & 2.07 & \best{0.5493} & 2.79 & \sn{+0.2056} \\
\bottomrule
\end{tabular}}
\end{researchtable}
\end{minipage}
\end{table}

Table~\ref{tab:code-uncertainty} evaluates four configurations over three decoding seeds and gives paired contrasts on seed 1, with bootstrap intervals over the 522 tasks. LEGO averages $0.5116\pm0.0024$ over the seeds, and its paired gain over feedback with diagnosis is $+0.1665$, with 95\% interval $[+0.1503,+0.1847]$~\citep{jin2026codeprimitives}.
\begin{table}[!htbp]
\centering
\researchtablesetup
\caption{\textbf{Repeated-seed and paired uncertainty for repository construction.} All configurations use the same 522 tasks; panel (a) gives means and standard deviations over three decoding seeds, and panel (b) gives paired contrasts on seed 1 with 95\% bootstrap confidence intervals over tasks, Wilcoxon signed-rank $p$-values and Holm correction across the four contrasts~\citep{jin2026codeprimitives}. Green marks the best and amber the second-best value in each column of panel (a).}
\label{tab:code-uncertainty}
\fitresearchtable{\begin{tabular}{lcccc}
\toprule
\rowcolor{TabHead}
\multicolumn{5}{c}{\hdr{(a) Configuration means over decoding seeds}} \\
\hdrgapfill\cmidrule(lr){1-5}
\rowcolor{TabHead}
\hdr{Configuration} & \hdr{Seed 1} & \hdr{Seed 2} & \hdr{Seed 3} & \hdr{Mean\,\textpm\,s.d.} \\
\midrule
Feedback (\stg{-{}-{}-V})             & 0.3180 & 0.3202 & 0.3150 & 0.3177\,\textpm\,0.0026 \\
Feedback + Diagnosis (\stg{-{}-DV})   & 0.3469 & 0.3434 & 0.3493 & 0.3465\,\textpm\,0.0030 \\
RAG + Edit (\stg{RE\,DV})             & \second{0.4093} & \second{0.4036} & \second{0.4121} & \second{0.4083\,\textpm\,0.0043} \\
\tabours LEGO (\stg{RADV})            & \best{0.5134} & \best{0.5125} & \best{0.5089} & \best{0.5116\,\textpm\,0.0024} \\
\midrule
\rowcolor{TabHead}
\multicolumn{5}{c}{\hdr{(b) Paired contrasts on seed 1}} \\
\hdrgapfill\cmidrule(lr){1-5}
\rowcolor{TabHead}
\hdr{Contrast} & \hdr{$\Delta$} & \hdr{95\% CI} & \hdr{$p$} & \hdr{$p_{\mathrm{Holm}}$} \\
\midrule
LEGO \textminus{} Feedback                       & \sn{+0.1954} & [\sn{+0.1762}, \sn{+0.2134}] & $<$\,10\textexp{-40} & $<$\,10\textexp{-39} \\
LEGO \textminus{} (Feedback + Diagnosis)         & \sn{+0.1665} & [\sn{+0.1503}, \sn{+0.1847}] & $<$\,10\textexp{-30} & $<$\,10\textexp{-29} \\
LEGO \textminus{} (RAG + Edit)                   & \sn{+0.1041} & [\sn{+0.0868}, \sn{+0.1190}] & $<$\,10\textexp{-16} & $<$\,10\textexp{-15} \\
LEGO \textminus{} LEGO, re-mined disjoint corpus & \sn{+0.0606} & [\sn{+0.0472}, \sn{+0.0768}] & 2.1\,$\times$\,10\textexp{-9} & 2.1\,$\times$\,10\textexp{-9} \\
\bottomrule
\end{tabular}}
\end{table}

Table~\ref{tab:code-external} gives the four independently developed repository agents of Section~\ref{sec:code-experiments} by difficulty band, with their recorded cost per task. OpenHands and Claude Code cost \$2.41 and \$2.16 per task, against \$2.08 for LEGO, and SWE-agent and Agentless cost \$1.86 and \$0.94. Every agent loses most of its delivery from D1 to D5, as execution feedback does, and LEGO leads all of them in every band~\citep{jin2026codeprimitives}.
\begin{table}[!htbp]
\centering
\researchtablesetup
\caption{\textbf{Independently developed repository agents on all 522 tasks.} Delivery score by difficulty band and over all tasks, the shares of tasks that reach the identity ceiling (Ceiling) and that score zero (Zero), and recorded-call cost per task (USD). Each agent runs under its own harness with the backbone of column two, with only the task set, environment, and grading held fixed, and SWE-agent and Agentless run through an issue-to-patch adapter~\citep{jin2026codeprimitives}. Green marks the best and amber the second-best value in each column. Lower is better for Zero and \$/task.\fillcap}
\label{tab:code-external}
\fitresearchtable{\begin{tabular}{llccccccccc}
\toprule
\rowcolor{TabHead}
\hdr{System} & \hdr{Backbone} & \hdr{D1} & \hdr{D2} & \hdr{D3} & \hdr{D4} & \hdr{D5} & \hdr{All} & \hdr{Ceiling (\%)} & \hdr{Zero (\%)} & \hdr{\$/task} \\
\midrule
OpenHands   & GPT-5.6 Terra   & \second{0.556} & \second{0.401} & 0.269 & 0.198 & \second{0.129} & \second{0.3263} & 9.4 & 41.6 & 2.41 \\
Claude Code & Claude Sonnet 5 & 0.543 & 0.394 & 0.271 & 0.190 & 0.104 & 0.3170 & \second{9.6} & \second{39.8} & 2.16 \\
SWE-agent   & GPT-5.6 Terra   & 0.482 & 0.345 & 0.238 & 0.152 & 0.081 & 0.2750 & 6.7 & 46.2 & 1.86 \\
Agentless   & GPT-5.6 Terra   & 0.421 & 0.276 & 0.181 & 0.098 & 0.043 & 0.2181 & 5.4 & 55.4 & \best{0.94} \\
\midrule
Feedback    & GPT-5.6 Terra   & 0.525 & 0.382 & \second{0.274} & \second{0.206} & \second{0.129} & 0.3180 & 8.8 & 41.0 & \second{1.55} \\
\tabours LEGO & GPT-5.6 Terra & \best{0.686} & \best{0.583} & \best{0.483} & \best{0.409} & \best{0.338} & \best{0.5134} & \best{22.6} & \best{18.8} & 2.08 \\
\bottomrule
\end{tabular}}
\end{table}

\FloatBarrier

\subsection{Math Primitives: Per-Problem Method Ranks}

\begin{figure}[!htbp]
\centering
\includegraphics[width=0.92\linewidth]{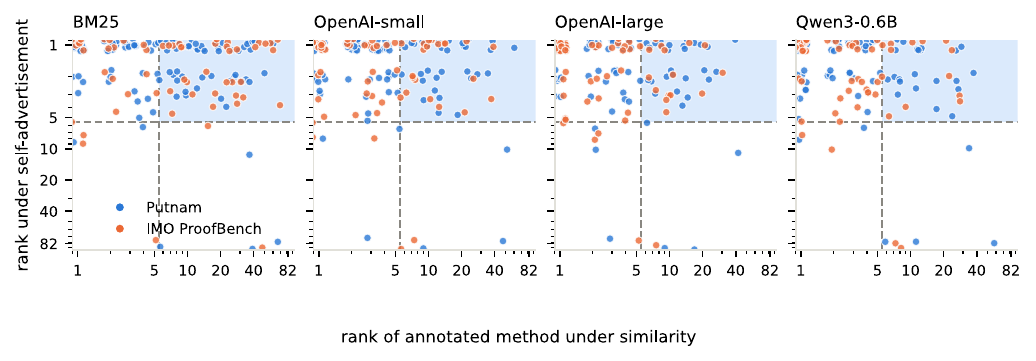}
\caption{\textbf{Per-problem rank of the highest-ranked annotated method.} Each dot is one Putnam or IMO ProofBench problem, placed vertically by the rank of its highest-ranked annotated method under self-advertisement with GPT-5.6 Luna, shared across panels, and horizontally by that method's rank under the panel's similarity retriever (BM25, OpenAI-small, OpenAI-large, or Qwen3-Emb-0.6B); dashed lines mark the top five, and the shaded region holds annotated methods that similarity ranks outside its top five but self-advertisement places inside~\citep{anonymous2026library}.\fillcap[.4][1]}
\label{fig:method-ranks}
\end{figure}

Section~\ref{sec:method-experiments} compares self-advertisement with similarity retrieval and reranking (Table~\ref{tab:method-selection}) and reports downstream proof completion (Table~\ref{tab:method-proof}). Figure~\ref{fig:method-ranks} shows where the coverage gains of Table~\ref{tab:method-selection} come from~\citep{anonymous2026library}. Each dot is one problem, placed by the rank of its highest-ranked annotated method under self-advertisement and by that method's rank under one similarity retriever. Dots in the shaded region are methods that similarity ranks outside its top five but self-advertisement places inside. Many of them lie outside the top ten of their panel's retriever. In the three embedding panels, this places them outside the pool of the reranker built on that retriever; BM25 has no reranker. This is the candidate-pool ceiling of Proposition~\ref{prop:coverage}(ii) at work: no reranking backbone can recover a method that its pool omits, whereas self-advertisement judges every contract from its proposal.
\FloatBarrier

\subsection{Knowledge Primitives: Grounding, Preservation, and Transfer}

% Pair float: Table~\ref{tab:knowledge-grounding} (left) and Table~\ref{tab:knowledge-preservation} (right), 9pt panels.
% Replaces \input{tables/restored/knowledge-grounding} and \input{tables/restored/knowledge-preservation} (appendix).
% Each body keeps its own \caption and \label, so numbering and \ref's are unchanged.
\begin{table}[!htbp]
\centering\researchpaneltrue\researchpanelsize{9}
\begin{minipage}[t]{0.545\linewidth}% Pair body (tables/pairs/knowledge-grounding+knowledge-preservation.tex); a float of its own if input directly.
\begin{researchtable}
\centering
\researchtablesetup
\caption{\textbf{Grounding audit of the stored source-model units.} Sampled units (\%) of the initial $M_0$, self-directed $M_S$ and task-guided $M_T$ source models of one document, one code and one API source, judged against their cited source elements: supported if all claims are entailed, partial if only the main claim is, unsupported if the evidence does not establish the claim, and contradicted if it states otherwise~\citep{jin2026sourcelearn}.}
\label{tab:knowledge-grounding}
\fitresearchtable{\begin{tabular}{lcccc}
\toprule
\rowcolor{TabHead}
\hdr{Model} & \hdr{Supported} & \hdr{Partial} & \hdr{Unsupported} & \hdr{Contradicted} \\
\midrule
\tabgroup{5}{MultiDoc2Dial (StudentAid)} \\
$M_0$ & 79.2 & 19.2 & 1.7 & 0.0 \\
$M_S$ & 75.0 & 20.8 & 4.2 & 0.0 \\
$M_T$ & 63.3 & 22.5 & 13.3 & 0.8 \\
\midrule
\tabgroup{5}{SWE-QA (pytest)} \\
$M_0$ & 89.2 & 10.8 & 0.0 & 0.0 \\
$M_S$ & 86.7 & 13.3 & 0.0 & 0.0 \\
$M_T$ & 65.0 & 31.7 & 3.3 & 0.0 \\
\midrule
\tabgroup{5}{APIBench (HuggingFace)} \\
$M_0$ & 65.0 & 32.5 & 1.7 & 0.8 \\
$M_S$ & 76.7 & 18.3 & 5.0 & 0.0 \\
$M_T$ & 84.2 & 15.8 & 0.0 & 0.0 \\
\bottomrule
\end{tabular}}
\end{researchtable}
\end{minipage}\hfill
\begin{minipage}[t]{0.430\linewidth}% Pair body (tables/pairs/knowledge-grounding+knowledge-preservation.tex); a float of its own if input directly.
\begin{researchtable}
\centering
\researchtablesetup
\caption{\textbf{Meaning preservation under source-model reconstruction.} Units removed by a learning stage are compared with the new model on the same entity, a claim counting as preserved if it remains represented after rewording, merging or condensation (\%); $M_0\to M_S$ is self-directed and $M_S\to M_T$ task-guided learning~\citep{jin2026sourcelearn}.}
\label{tab:knowledge-preservation}
\fitresearchtable{\begin{tabular}{lccc}
\toprule
\rowcolor{TabHead}
\hdr{Dropped by} & \hdr{Preserved} & \hdr{Lost} & \hdr{Contradicted} \\
\midrule
\tabgroup{4}{MultiDoc2Dial (StudentAid)} \\
$M_0\!\to\!M_S$ & 98.3 & 1.7 & 0.0 \\
$M_S\!\to\!M_T$ & 98.3 & 0.0 & 1.7 \\
\midrule
\tabgroup{4}{SWE-QA (pytest)} \\
$M_0\!\to\!M_S$ & 93.3 & 5.0 & 1.7 \\
$M_S\!\to\!M_T$ & 100.0 & 0.0 & 0.0 \\
\midrule
\tabgroup{4}{APIBench (HuggingFace)} \\
$M_0\!\to\!M_S$ & 96.7 & 3.3 & 0.0 \\
$M_S\!\to\!M_T$ & 100.0 & 0.0 & 0.0 \\
\bottomrule
\end{tabular}}
\end{researchtable}
\end{minipage}
\end{table}

\begin{table}[!htbp]
\centering
\researchtablesetup
\caption{\textbf{Source-model gains inside and outside the regions touched by guidance tasks.} Accuracy (\%) with GPT-5.6 Luna, where seen regions overlap at least one guidance task and unseen regions none, with the change from Hybrid RAG in the same stratum in pp in parentheses~\citep{jin2026sourcelearn}. Green marks the best and amber the second-best value in each row within each stratum.}
\label{tab:knowledge-transfer}
\fitresearchtable{\begin{tabular}{lcccccccc}
\toprule
\rowcolor{TabHead}
 & \multicolumn{2}{c}{\hdr{Hybrid RAG}} & \multicolumn{2}{c}{\hdr{$M_0$}} & \multicolumn{2}{c}{\hdr{$M_S$}} & \multicolumn{2}{c}{\hdr{$M_T$ (SourceLearn)}} \\
\hdrgapfill\cmidrule(lr){2-3}\cmidrule(lr){4-5}\cmidrule(lr){6-7}\cmidrule(lr){8-9}
\rowcolor{TabHead}
\multirow{-2}{*}{\hdr{Benchmark}} & \hdr{Seen} & \hdr{Unseen} & \hdr{Seen} & \hdr{Unseen} & \hdr{Seen} & \hdr{Unseen} & \hdr{Seen} & \hdr{Unseen} \\
\midrule
MultiDoc2Dial & 64.2 & 65.3 & 69.0\dpp{+4.8} & \second{72.3}\dpp{+7.0} & \second{77.4}\dpp{+13.2} & 72.0\dpp{+6.7} & \best{78.9}\dpp{+14.7} & \best{76.7}\dpp{+11.4} \\
SWE-QA & 57.3 & 56.4 & \second{70.3}\dpp{+13.0} & 61.1\dpp{+4.7} & 66.2\dpp{+8.9} & \second{64.6}\dpp{+8.2} & \best{72.9}\dpp{+15.6} & \best{66.3}\dpp{+9.9} \\
NarrativeQA & 65.8 & 66.2 & 74.2\dpp{+8.4} & 74.0\dpp{+7.8} & \second{77.2}\dpp{+11.4} & \second{78.4}\dpp{+12.2} & \best{80.1}\dpp{+14.3} & \best{79.9}\dpp{+13.7} \\
\midrule
All QA & 62.4 & 62.9 & 71.6\dpp{+9.2} & 69.6\dpp{+6.7} & \second{73.3}\dpp{+10.9} & \second{72.2}\dpp{+9.3} & \best{77.3}\dpp{+14.9} & \best{75.0}\dpp{+12.1} \\
\bottomrule
\end{tabular}}
\end{table}

\begin{figure}[!htbp]
\centering
\includegraphics[width=\linewidth]{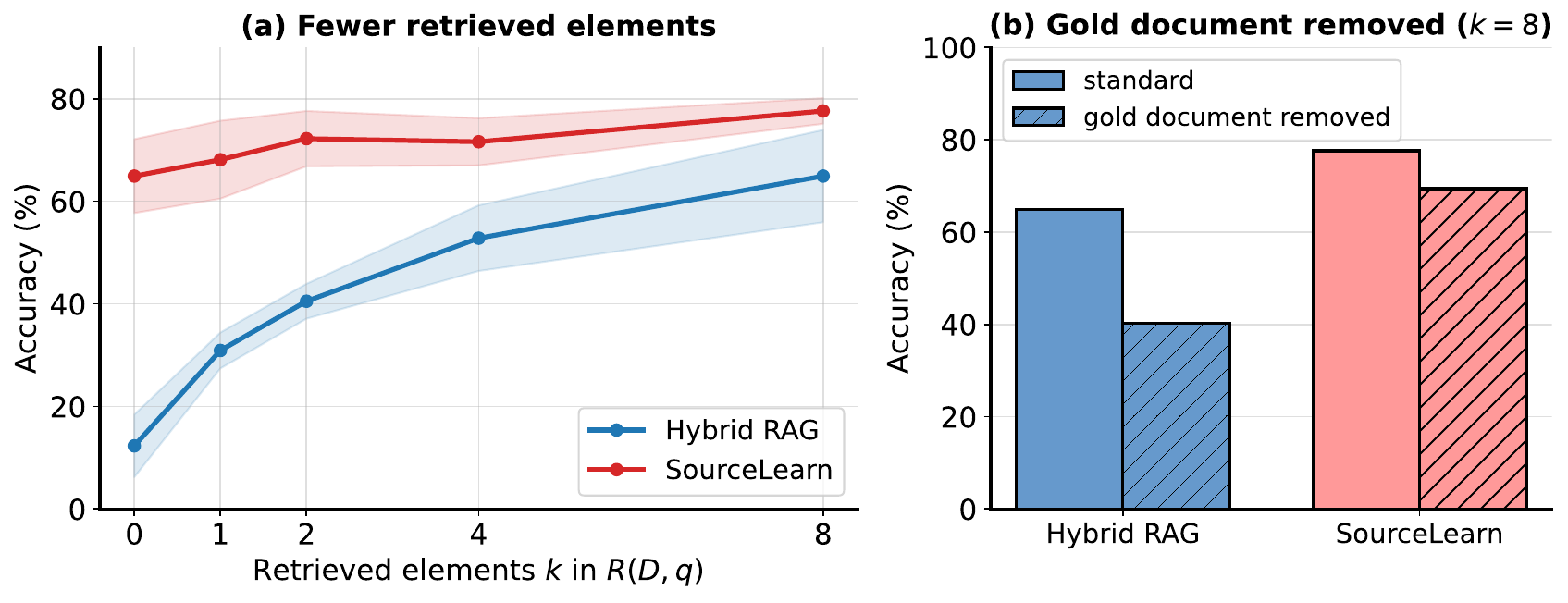}
\caption{\textbf{Missing-evidence robustness.} Accuracy (\%) of Hybrid RAG and SourceLearn on MultiDoc2Dial with GPT-5.6 Luna as fewer source elements of the Hybrid RAG evidence $R(D,q)$ are retrieved (a) and with the gold document removed (b)~\citep{jin2026sourcelearn}.}
\label{fig:knowledge-degradation}
\end{figure}
\begin{figure}[!htbp]
\centering
\includegraphics[width=\linewidth,trim=0bp 0bp 379.558bp 0bp,clip]{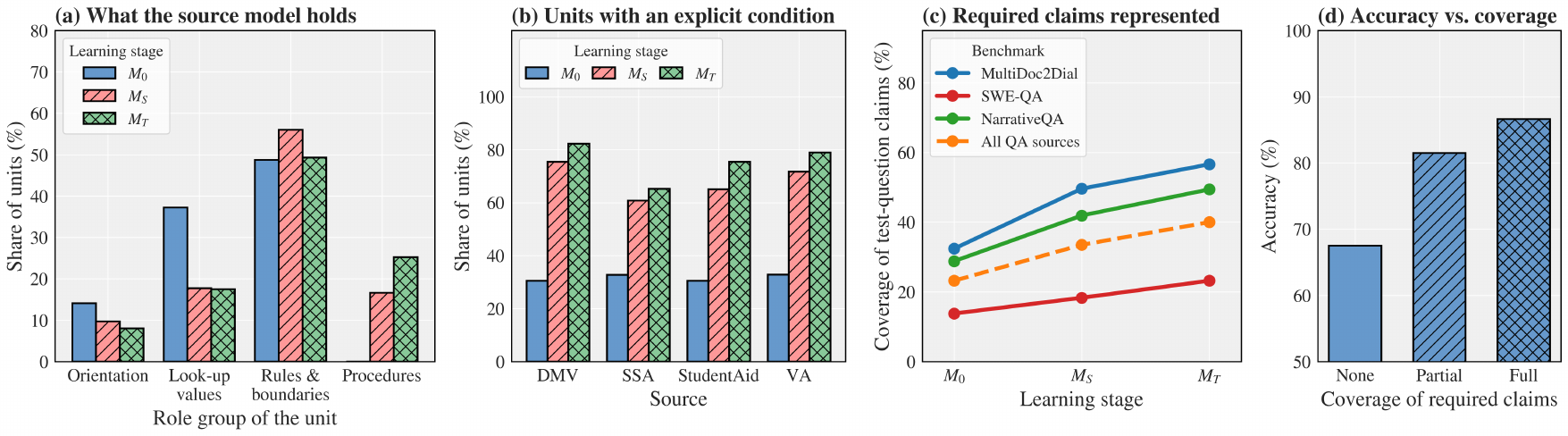}
\par\medskip
\includegraphics[width=0.821\linewidth,trim=462.535bp 0bp 0bp 0bp,clip]{figures/knowledge-reconstruction.pdf}
\caption{\textbf{Representation changes through source learning.} Share of units by role group (a) and with an explicit condition (b) in $M_0$, $M_S$, and $M_T$, coverage of the claims that test questions require at each stage (c), and accuracy by the coverage of a question's required claims (d)~\citep{jin2026sourcelearn}.\fillcap}
\label{fig:knowledge-reconstruction}
\end{figure}
Section~\ref{sec:knowledge-experiments} compares SourceLearn with retrieval and memory baselines on five benchmarks and at a matched context budget (Tables~\ref{tab:knowledge-main} and~\ref{tab:knowledge-budget}). The material below audits the learned source models, splits the gains by region, and reduces the evidence. Tables~\ref{tab:knowledge-grounding} and~\ref{tab:knowledge-preservation} are SourceLearn's post-hoc audits of grounding and source authority, the principle that learning signals determine what is reconsidered while persistent updates are reconstructed from the authoritative source (Eq.~\eqref{eq:knowledge-grounded})~\citep{jin2026sourcelearn}. The first judges sampled units of $M_0$, $M_S$, and $M_T$ against their cited source elements, and the second checks whether units removed by learning remain represented in the revised model. Table~\ref{tab:knowledge-transfer} splits test questions by whether their source regions overlapped the guidance tasks, for GPT-5.6 Luna. The gain over Hybrid RAG on unseen regions shows that the benefit of the learned source model extends beyond the regions that the guidance tasks touched.

Figure~\ref{fig:knowledge-degradation} reduces the retrieved evidence or removes the gold document on MultiDoc2Dial. SourceLearn degrades more slowly than Hybrid RAG, so the source model supplies source understanding when task-local retrieval is incomplete. Figure~\ref{fig:knowledge-reconstruction} reports what the source model holds at each learning stage and the share of units with an explicit condition. It also reports how much of the test questions' required content is represented, and the association between represented content and accuracy (Appendix~\ref{app:ledger}).
\FloatBarrier

\subsection{Agent Primitives: Structure Selection and Cost}

Section~\ref{sec:agent-experiments} compares the primitive-based MAS with a single agent, text-based and latent MAS, and each single primitive applied to every query, and it removes positional re-encoding (Tables~\ref{tab:agent-main} and~\ref{tab:agent-rope}). The eight benchmarks of Table~\ref{tab:agent-main} combine mathematical problem solving in AIME, MATH, and GSM8K~\citep{mathai_aime2025,maxwelljia_aime2024,hendrycks2021measuring,cobbe2021training}; executable code evaluation through HumanEval+ and MBPP+, the stricter test suites built on HumanEval and MBPP~\citep{chen2021evaluatinglargelanguagemodels,austin2021program,liu2023your}; and medical and graduate-level question answering~\citep{jin2021disease,rein2024gpqa}. The benchmark mean therefore aggregates heterogeneous evaluation procedures, and the individual columns report each of them.
% Pair float: Table~\ref{tab:agent-organizer} (left) and Table~\ref{tab:agent-pool} (right), 8.5pt panels.
% Replaces \input{tables/restored/agent-organizer} and \input{tables/restored/agent-pool} (appendix).
% Each body keeps its own \caption and \label, so numbering and \ref's are unchanged.
\begin{table}[!htbp]
\centering\researchpaneltrue\researchpanelsize{8.5}
\begin{minipage}[t]{0.495\linewidth}% Pair body (tables/pairs/agent-organizer+agent-pool.tex); a float of its own if input directly.
\begin{researchtable}
\centering
\researchtablesetup
\caption{\textbf{The LLM Organizer against random structure selection.} Accuracy (\%) of the composed MAS when the GPT-5.2 Organizer is replaced by Claude-4 or by random selection of MAS structures from the Knowledge Pool (R1/Llama-70B is DeepSeek-R1-Distill-Llama-70B), with the change from GPT-5.2 in pp in parentheses~\citep{jin2026agentprimitives}.}
\label{tab:agent-organizer}
\fitresearchtable{\begin{tabular}{llccc}
\toprule
\rowcolor{TabHead}
\hdr{Model} & \hdr{Organizer} & \hdr{AIME25} & \hdr{HumanEval+} & \hdr{MedQA} \\
\midrule
\tabours  & GPT-5.2 & 73.3 & 82.3 & 76.7 \\
 & Claude-4 & 73.3\dpp{0.0} & 81.1\dpp{-1.2} & 76.2\dpp{-0.5} \\
 \multirow{-3}{*}{Qwen3-8B} & Random   & 66.7\dpp{-6.6} & 77.4\dpp{-4.9} & 70.6\dpp{-6.1} \\
\midrule
\tabours  & GPT-5.2 & 56.7 & 85.3 & 81.9 \\
 & Claude-4 & 56.7\dpp{0.0} & 83.5\dpp{-1.8} & 81.9\dpp{0.0} \\
 \multirow{-3}{*}{R1/Llama-70B} & Random   & 50.0\dpp{-6.7} & 78.6\dpp{-6.7} & 69.5\dpp{-12.4} \\
\bottomrule
\end{tabular}}
\end{researchtable}
\end{minipage}\hfill
\begin{minipage}[t]{0.480\linewidth}% Pair body (tables/pairs/agent-organizer+agent-pool.tex); a float of its own if input directly.
\begin{researchtable}
\centering
\researchtablesetup
\caption{\textbf{Removing the Organizer's Knowledge Pool.} Accuracy (\%) of the composed MAS with (w/) and without (w/o) the Knowledge Pool, whose 45 entries pair previously observed queries with effective MAS structures and guide the Organizer's composition, models as in Table~\ref{tab:agent-organizer}, with the change from w/ in pp in parentheses~\citep{jin2026agentprimitives}.}
\label{tab:agent-pool}
\fitresearchtable{\begin{tabular}{llccc}
\toprule
\rowcolor{TabHead}
\hdr{Model} & \hdr{Pool} & \hdr{AIME25} & \hdr{HumanEval+} & \hdr{MedQA} \\
\midrule
\tabours  & w/ & 73.3 & 82.3 & 76.7 \\
 \multirow{-2}{*}{Qwen3-8B} & w/o & 63.3\dpp{-10.0} & 77.4\dpp{-4.9} & 71.6\dpp{-5.1} \\
\midrule
\tabours  & w/ & 56.7 & 85.3 & 81.9 \\
 \multirow{-2}{*}{R1/Llama-70B} & w/o & 50.0\dpp{-6.7} & 73.4\dpp{-11.9} & 75.9\dpp{-6.0} \\
\bottomrule
\end{tabular}}
\end{researchtable}
\end{minipage}
\end{table}

\begin{table}[!htbp]
\centering
\researchtablesetup
\caption{\textbf{Accuracy and estimated inference cost of the methods of Table~\ref{tab:agent-mas}.} Cost is the Agent Primitives study's estimated USD per query from market output-token prices, accuracy (Acc.) is in \%, efficiency (Eff.) is accuracy per \$0.01, and the two composed-MAS rows differ in the Organizer model~\citep{jin2026agentprimitives}.}
\label{tab:agent-cost}
\fitresearchtable{\begin{tabular}{lcccccccccccc}
\toprule
\rowcolor{TabHead}
 & \multicolumn{3}{c}{\hdr{MATH}} & \multicolumn{3}{c}{\hdr{GSM8K}} & \multicolumn{3}{c}{\hdr{HumanEval+}} & \multicolumn{3}{c}{\hdr{GPQA}} \\
\hdrgapfill\cmidrule(lr){2-4}\cmidrule(lr){5-7}\cmidrule(lr){8-10}\cmidrule(lr){11-13}
\rowcolor{TabHead}
\multirow{-2}{*}{\hdr{Method}} & \hdr{Cost} & \hdr{Acc.} & \hdr{Eff.} & \hdr{Cost} & \hdr{Acc.} & \hdr{Eff.} & \hdr{Cost} & \hdr{Acc.} & \hdr{Eff.} & \hdr{Cost} & \hdr{Acc.} & \hdr{Eff.} \\
\midrule
Single            & 0.0024 & 50.6 & 210.8 & 0.0016 & 92.4 & 573.9 & 0.0037 & 75.8 & 204.9 & 0.0117 & 36.7 & 31.4 \\
CoT               & 0.0033 & 53.2 & 161.2 & 0.0025 & 92.8 & 369.4 & 0.0041 & 77.0 & 186.6 & 0.0130 & 35.3 & 27.2 \\
Self-Consistency  & 0.0204 & 61.6 &  30.2 & 0.0153 & 95.0 &  62.1 & 0.0352 & 75.8 &  21.5 & 0.0729 & 37.2 &  5.1 \\
LLM-Debate        & 0.0081 & 61.4 &  75.8 & 0.0101 & 91.6 &  90.9 & 0.0151 & 74.5 &  49.3 & 0.0250 & 34.4 & 13.8 \\
Self-Refine       & 0.0055 & 58.5 & 106.1 & 0.0038 & 90.8 & 240.5 & 0.0086 & 62.7 &  73.1 & 0.0268 & 38.3 & 14.3 \\
Quality-Diversity & 0.0132 & 60.5 &  45.9 & 0.0090 & 93.0 & 103.5 & 0.0205 & 70.2 &  34.2 & 0.0352 & 33.6 &  9.5 \\
SPP               & 0.0052 & 51.7 &  99.8 & 0.0034 & 92.8 & 273.7 & 0.0067 & 73.3 & 109.3 & 0.0198 & 35.1 & 17.7 \\
AgentVerse        & 0.0080 & 55.6 &  69.6 & 0.0061 & 93.4 & 152.3 & 0.0161 & 73.9 &  45.9 & 0.0248 & 40.2 & 16.2 \\
GPTSwarm          & 0.0074 & 55.4 &  74.9 & 0.0061 & 93.2 & 153.6 & 0.0152 & 73.9 &  48.6 & 0.0232 & 36.5 & 15.7 \\
DyLAN             & 0.0085 & 59.6 &  70.1 & 0.0062 & 91.2 & 147.1 & 0.0165 & 75.8 &  45.9 & 0.0283 & 36.0 & 12.7 \\
MAS-GPT           & 0.0082 & 68.7 &  83.5 & 0.0065 & 93.4 & 143.7 & 0.0174 & 78.9 &  45.4 & 0.0257 & 37.6 & 14.6 \\
\midrule
\tabours Primitives-based MAS, GPT-5.2 Organizer     & 0.0056 & 72.4 & 129.3 & 0.0048 & 93.8 & 195.4 & 0.0118 & 82.3 &  69.7 & 0.0171 & 53.2 & 31.1 \\
\tabours Primitives-based MAS, Llama-3-70B Organizer & 0.0028 & 72.4 & 258.6 & 0.0020 & 93.8 & 469.0 & 0.0063 & 82.3 & 130.6 & 0.0122 & 53.2 & 43.6 \\
\bottomrule
\end{tabular}}
\end{table}

\begin{table}[!htbp]
\centering
\researchtablesetup
\caption{\textbf{Output tokens per query of the methods of Table~\ref{tab:agent-mas}.} Counts with Llama-3-70B-Instruct workers cover the whole pipeline, including the Organizer's 200--400 output tokens per query; Mean averages the four benchmarks, and the two composed-MAS rows differ in the Organizer model~\citep{jin2026agentprimitives}. Green marks the best and amber the second-best value in each column. Lower is better for every column.\fillcap}
\label{tab:agent-tokens}
\fitresearchtable{\begin{tabular}{lccccc}
\toprule
\rowcolor{TabHead}
\hdr{Method} & \hdr{MATH} & \hdr{GSM8K} & \hdr{HumanEval+} & \hdr{GPQA} & \hdr{Mean} \\
\midrule
Single            & \best{1,375} & \best{934} & \best{2,129} & \best{6,674} & \best{2,778} \\
CoT               & 1,874  & 1,447 & \second{2,371} & 7,408  & \second{3,275} \\
Self-Consistency  & 11,673 & 8,742 & 20,114 & 41,675 & 20,551 \\
LLM-Debate        & 4,612  & 5,747 & 8,635  & 14,278 & 8,318 \\
Self-Refine       & 3,162  & 2,149 & 4,891  & 15,350 & 6,388 \\
Quality-Diversity & 7,526  & 5,137 & 11,710 & 20,127 & 11,125 \\
SPP               & 2,957  & 1,934 & 3,832  & 11,345 & 5,017 \\
AgentVerse        & 4,572  & 3,504 & 9,216  & 14,165 & 7,864 \\
GPTSwarm          & 4,216  & 3,473 & 8,698  & 13,267 & 7,414 \\
DyLAN             & 4,873  & 3,541 & 9,427  & 16,188 & 8,507 \\
MAS-GPT           & 4,714  & 3,706 & 9,916  & 14,672 & 8,252 \\
\midrule
\tabours Primitives-based MAS, GPT-5.2 Organizer     & \second{1,524} & \second{1,017} & 3,779 & \second{6,882} & 3,301 \\
\tabours Primitives-based MAS, Llama-3-70B Organizer & 1,609 & 1,134 & 3,618 & 6,974 & 3,334 \\
\bottomrule
\end{tabular}}
\end{table}

Tables~\ref{tab:agent-organizer} and~\ref{tab:agent-pool} hold the Organizer ablations~\citep{jin2026agentprimitives}. They replace the GPT-5.2 Organizer by Claude-4~\citep{anthropic2025claude} or by random selection of structures from the Knowledge Pool, and they remove the Knowledge Pool. They thus vary the model and the information behind the Organizer's choice of primitives, down to a random choice, and the study concludes that the main benefit comes from problem-aware structure selection rather than from a specific Organizer model. Table~\ref{tab:agent-mas} compares the primitive-based MAS with eight independently designed MAS methods and two single-model prompting baselines on Llama-3-70B-Instruct~\citep{grattafiori2024llama}, separately from the backbones of Table~\ref{tab:agent-main}. The prompting baselines are Chain-of-Thought~\citep{wei2022chain} and Self-Consistency~\citep{wang2022self}; the MAS methods include Self-Refine~\citep{madaan2023self}, a Quality-Diversity comparator~\citep{lu2024intelligent}, and SPP~\citep{wang2024unleashing}. Table~\ref{tab:agent-cost} adds the cost accounting for the same methods. Its row for a Llama-3-70B Organizer repeats the GPT-5.2 accuracies, an inconsistency of the Agent Primitives study that Appendix~\ref{app:ledger} records. Table~\ref{tab:agent-tokens} gives the output tokens behind these costs. With either Organizer, the primitive-based MAS uses fewer tokens than every MAS method on each benchmark: 3,301 per query on average with the GPT-5.2 Organizer, against 8,252 for MAS-GPT, $2.5\times$ fewer.
\FloatBarrier

\subsection{Dev-Primitives: Model Allocation and On-Demand Activation}

\begin{table}[!htbp]
\centering
\researchtablesetup
\caption{\textbf{Assigning models separately to activation, the Dev-Primitives and diagnosis.} The lower blocks keep the Dev-Primitives on Qwen3-8B and scale activation, diagnosis or both, with the change from the homogeneous Qwen3-8B configuration in pp in parentheses~\citep{jin2026devprimitives}.}
\label{tab:dev-roles}
\fitresearchtable{\begin{tabular}{lllcccc}
\toprule
\rowcolor{TabHead}
 & & & \hdr{SWE-bench} & \hdr{SWE Refactor} & \hdr{Terminal-Bench} & \hdr{DevOps-Gym} \\
\hdrgapfill\cmidrule(lr){4-4}\cmidrule(lr){5-5}\cmidrule(lr){6-6}\cmidrule(lr){7-7}
\rowcolor{TabHead}
\multirow{-2}{*}{\hdr{Activation}} & \multirow{-2}{*}{\hdr{Dev-Primitives}} & \multirow{-2}{*}{\hdr{Diagnosis}} & \hdr{Resolved (\%)} & \hdr{Composite (\%)} & \hdr{Resolved (\%)} & \hdr{Avg. (\%)} \\
\midrule
\tabgroup{7}{Homogeneous backbones} \\
Qwen3-8B        & Qwen3-8B     & Qwen3-8B        & 80.6 & 10.5 & 27.6 & 37.48 \\
GPT-5.6 Luna    & GPT-5.6 Luna & GPT-5.6 Luna    & 95.6 & 16.0 & 29.1 & 46.64 \\
DeepSeek-V4     & DeepSeek-V4  & DeepSeek-V4     & 82.8 & 15.0 & 39.4 & 53.08 \\
GPT-5.6 Sol     & GPT-5.6 Sol  & GPT-5.6 Sol     & 97.0 & 31.0 & 51.8 & 55.41 \\
\midrule
\tabgroup{7}{Diagnosis scaling} \\
Qwen3-8B        & Qwen3-8B     & DeepSeek-V4     & 84.8\dpp{+4.2} & 14.5\dpp{+4.0} & 44.5\dpp{+16.9} & 41.62\dpp{+4.14} \\
Qwen3-8B        & Qwen3-8B     & GPT-5.5         & 86.4\dpp{+5.8} & 16.5\dpp{+6.0} & 46.1\dpp{+18.5} & 43.25\dpp{+5.77} \\
Qwen3-8B        & Qwen3-8B     & Claude Sonnet 5 & 86.0\dpp{+5.4} & 16.0\dpp{+5.5} & 45.5\dpp{+17.9} & 43.01\dpp{+5.53} \\
Qwen3-8B        & Qwen3-8B     & GPT-5.6 Sol     & 87.2\dpp{+6.6} & 17.5\dpp{+7.0} & 47.9\dpp{+20.3} & 44.36\dpp{+6.88} \\
\midrule
\tabgroup{7}{Activation scaling} \\
GPT-5.6 Luna    & Qwen3-8B     & Qwen3-8B        & 82.4\dpp{+1.8} & 12.0\dpp{+1.5} & 32.7\dpp{+5.1}  & 39.82\dpp{+2.34} \\
DeepSeek-V4     & Qwen3-8B     & Qwen3-8B        & 83.6\dpp{+3.0} & 13.0\dpp{+2.5} & 34.8\dpp{+7.2}  & 40.77\dpp{+3.29} \\
GPT-5.5         & Qwen3-8B     & Qwen3-8B        & 84.4\dpp{+3.8} & 13.5\dpp{+3.0} & 36.7\dpp{+9.1}  & 41.58\dpp{+4.10} \\
Claude Sonnet 5 & Qwen3-8B     & Qwen3-8B        & 84.0\dpp{+3.4} & 13.0\dpp{+2.5} & 36.1\dpp{+8.5}  & 41.20\dpp{+3.72} \\
GPT-5.6 Sol     & Qwen3-8B     & Qwen3-8B        & 85.2\dpp{+4.6} & 14.0\dpp{+3.5} & 38.2\dpp{+10.6} & 42.31\dpp{+4.83} \\
\midrule
\tabgroup{7}{Activation + diagnosis scaling} \\
GPT-5.6 Luna    & Qwen3-8B     & GPT-5.5         & 89.4\dpp{+8.8}  & 20.0\dpp{+9.5}  & 47.6\dpp{+20.0} & 46.02\dpp{+8.54} \\
GPT-5.5         & Qwen3-8B     & GPT-5.5         & 91.2\dpp{+10.6} & 22.5\dpp{+12.0} & 49.1\dpp{+21.5} & 48.73\dpp{+11.25} \\
Claude Sonnet 5 & Qwen3-8B     & Claude Sonnet 5 & 92.4\dpp{+11.8} & 23.5\dpp{+13.0} & 48.8\dpp{+21.2} & 49.66\dpp{+12.18} \\
GPT-5.6 Sol     & Qwen3-8B     & GPT-5.5         & 93.6\dpp{+13.0} & 25.0\dpp{+14.5} & 49.7\dpp{+22.1} & 51.44\dpp{+13.96} \\
GPT-5.6 Sol     & Qwen3-8B     & GPT-5.6 Sol     & 94.2\dpp{+13.6} & 26.5\dpp{+16.0} & 50.6\dpp{+23.0} & 52.30\dpp{+14.82} \\
\bottomrule
\end{tabular}}
\end{table}

% Pair float: Table~\ref{tab:dev-cost} (left) and Table~\ref{tab:dev-activation} (right), 8.5pt panels.
% Replaces \input{tables/restored/dev-cost} and \input{tables/restored/dev-activation} (appendix).
% Each body keeps its own \caption and \label, so numbering and \ref's are unchanged.
\begin{table}[!htbp]
\centering\researchpaneltrue\researchpanelsize{8.5}
\begin{minipage}[t]{0.484\linewidth}% Pair body (tables/pairs/dev-cost+dev-activation.tex); a float of its own if input directly.
\begin{researchtable}
\centering
\researchtablesetup
\caption{\textbf{Cost and resolution of mixed model assignments on Terminal-Bench 4.0.} Configurations name the activation / Dev-Primitive / diagnosis models, cost is the total inference expenditure of the evaluation, and the last row is HERMES with GPT-5.6 Sol in every role~\citep{jin2026devprimitives}.}
\label{tab:dev-cost}
\fitresearchtable{\begin{tabular}{lcc}
\toprule
\rowcolor{TabHead}
\hdr{Configuration} & \hdr{Resolved (\%)} & \hdr{Cost (\$k)} \\
\midrule
Qwen3-8B / Qwen3-8B / Qwen3-8B                   & 27.6 & 0.70 \\
Qwen3-8B / Qwen3-8B / GPT-5.5                    & 46.1 & 1.56 \\
GPT-5.6 Sol / Qwen3-8B / GPT-5.5                 & 49.7 & 2.13 \\
GPT-5.6 Sol / Qwen3-8B / GPT-5.6 Sol             & 50.6 & 2.51 \\
\tabours GPT-5.6 Sol / GPT-5.6 Sol / GPT-5.6 Sol & 51.8 & 3.40 \\
\bottomrule
\end{tabular}}
\end{researchtable}
\end{minipage}\hfill
\begin{minipage}[t]{0.491\linewidth}% Pair body (tables/pairs/dev-cost+dev-activation.tex); a float of its own if input directly.
\begin{researchtable}
\centering
\researchtablesetup
\caption{\textbf{Primitive calls with and without on-demand activation.} Average calls per task, including repeated calls for communication and revision; without on-demand activation all Dev-Primitives identified in the initial analysis stay active throughout, and cost is relative to HERMES~\citep{jin2026devprimitives}.}
\label{tab:dev-activation}
\fitresearchtable{\begin{tabular}{lccc}
\toprule
\rowcolor{TabHead}
 & \multicolumn{2}{c}{\hdr{Avg. Primitive Calls}} & \hdr{Relative Cost} \\
\hdrgapfill\cmidrule(lr){2-3}\cmidrule(lr){4-4}
\rowcolor{TabHead}
\multirow{-2}{*}{\hdr{Benchmark}} & \hdr{HERMES} & \hdr{w/o Activation} & \hdr{w/o Activation} \\
\midrule
SWE-bench Verified & 9.6  & 28.7 & 1.61$\times$ \\
SWE Refactor Bench & 25.1 & 64.3 & 1.88$\times$ \\
Terminal-Bench 4.0 & 14.2 & 39.6 & 1.73$\times$ \\
DevOps-Gym         & 11.7 & 33.2 & 1.67$\times$ \\
\bottomrule
\end{tabular}}
\end{researchtable}
\end{minipage}
\end{table}

Section~\ref{sec:dev-experiments} compares HERMES with same-backbone harnesses, with a single editor over the same components, and with variants that remove one mechanism at a time (Tables~\ref{tab:dev-swe}, \ref{tab:dev-terminal}, \ref{tab:dev-controls}, \ref{tab:dev-ablation}, and~\ref{tab:dev-ops}). Table~\ref{tab:dev-roles} assigns models independently to activation, the Dev-Primitives, and diagnosis; its values in parentheses are improvements, in pp, over the homogeneous Qwen3-8B configuration~\citep{jin2026devprimitives}. With Qwen3-8B Dev-Primitives, a stronger diagnosis model alone raises Terminal-Bench resolution from 27.6\% to between 44.5\% and 47.9\%. A stronger activation model alone reaches at most 38.2\%. Table~\ref{tab:dev-cost} adds the Terminal-Bench~4.0 inference cost of five of these configurations. Table~\ref{tab:dev-activation} counts primitive calls with and without on-demand activation on all four benchmarks, which shows how much instantiating only the activated Dev-Primitives $\mathcal{P}_{\mathcal{A}}$ saves within one task (Section~\ref{sec:composition-state}). Its ablation keeps every component found by the initial analysis active for the whole trajectory. Repeated calls of one primitive after messages or revisions make the call counts exceed the number of distinct active components.
\FloatBarrier

\subsection{Navigation Primitives: Scaling and Activity Diagnostics}

\begin{table}[!htbp]
\centering
\researchtablesetup
\caption{\textbf{Per-length results of the controlled information-space scaling.} Active coordination is the mean number of participating workers or members per query, and calls and cost are per query~\citep{jin2026antman}. Green marks the best and amber the second-best value in each column within each length. Lower is better for active coordination, calls, and cost.}
\label{tab:navigation-full-scaling}
\fitresearchtable{\begin{tabular}{llccccc}
\toprule
\rowcolor{TabHead}
\hdr{Method} & \hdr{Length} & \makecell[b]{\hdr{Active}\\\hdr{Coord.}} & \makecell[b]{\hdr{Calls}\\\hdr{/ Query}} & \makecell[b]{\hdr{Cost}\\\hdr{/ Query (\$)}} & \hdr{EM} & \hdr{F1} \\
\midrule
\tabgroup{7}{Multi-agent baselines} \\
 & 32K  & 19.00  & \second{22.83} & 0.0983 & 0.533 & 0.731 \\
          & 64K  & 37.00  & 39.50  & 0.1797 & 0.533 & \second{0.737} \\
          & 128K & 73.10  & 75.60  & 0.3549 & \second{0.500} & \second{0.670} \\
          \multirow{-4}{*}{LongAgent} & 512K & 290.50 & 292.50 & 1.3974 & \second{0.567} & \second{0.754} \\
\midrule
 & 32K  & \second{5.10}  & \best{7.10}    & \second{0.0809} & \second{0.567} & \second{0.750} \\
          & 64K  & \second{10.00} & \best{12.00}   & \second{0.1613} & \second{0.567} & 0.734 \\
          & 128K & \second{20.00} & \best{22.00}   & \second{0.3216} & 0.400 & 0.564 \\
          \multirow{-4}{*}{CoA} & 512K & \second{78.20} & \second{80.23} & \second{1.3027} & 0.500 & 0.746 \\
\midrule
\tabours  & 32K  & \best{2.57} & 27.47          & \best{0.0773} & \best{0.600} & \best{0.775} \\
\tabours        & 64K  & \best{2.87} & \second{29.53} & \best{0.1069} & \best{0.667} & \best{0.784} \\
\tabours        & 128K & \best{3.07} & \second{28.30} & \best{0.1276} & \best{0.600} & \best{0.768} \\
\tabours        \multirow{-4}{*}{ANTMAN} & 512K & \best{3.17} & \best{33.10} & \best{0.5332} & \best{0.667} & \best{0.840} \\
\bottomrule
\end{tabular}}
\end{table}

\begin{table}[!htbp]
\centering
\researchtablesetup
\caption{\textbf{Relevant and irrelevant worker activity as required evidence grows.} ANTMAN's trajectory-level worker activity at a fixed 512K space, split by whether it contributes to acquiring required evidence; the split attributes the dip in active coordination from 1 to 4 units (Table~\ref{tab:navigation-demand}) to exploration overhead, and at 16 units ANTMAN's study reports 8.2 worker activities against 8.0 actively coordinating workers~\citep{jin2026antman}.\fillcap}
\label{tab:navigation-demand-diagnostics}
\fitresearchtable{\begin{tabular}{lccc}
\toprule
\rowcolor{TabHead}
\makecell[lb]{\hdr{Required}\\\hdr{Evidence (units)}} & \makecell[b]{\hdr{Worker}\\\hdr{Activity}} & \makecell[b]{\hdr{Relevant}\\\hdr{Activity}} & \makecell[b]{\hdr{Irrelevant}\\\hdr{Activity}} \\
\midrule
\tabgroup{4}{ANTMAN} \\
1  & 5.0 & 0.9 & 4.1 \\
4  & 4.3 & 3.2 & 1.1 \\
16 & 8.2 & 8.0 & 0.2 \\
\bottomrule
\end{tabular}}
\end{table}

Section~\ref{sec:navigation-experiments} reports ANTMAN in structured information spaces, under a growing space and a growing demand, and with its need tracking, rerouting, or recovery removed (Tables~\ref{tab:navigation-main}, \ref{tab:navigation-scaling}, \ref{tab:navigation-demand}, and~\ref{tab:navigation-ablation}). The structured-navigation comparison in Table~\ref{tab:navigation-main} includes RepoDistill~\citep{yin-etal-2026-repodistill}. Table~\ref{tab:navigation-full-scaling} gives every space size and baseline row of the controlled intervention from 32K to 512K, which examines the behaviour that Proposition~\ref{prop:activation} predicts at approximately fixed demand~\citep{jin2026antman}. It reports active coordination, calls and cost per query, and answer quality (EM and F1) for each space size. Table~\ref{tab:navigation-demand-diagnostics} decomposes trajectory-level worker activity under growing demand, measured by required evidence in place of $H(q)$, into activity that contributes required evidence and activity that does not. This decomposition explains why active coordination dips from 5.0 to 4.3 workers between one and four required units: at one unit, most activity is exploration that finds no required evidence. Worker activity is a trajectory-level count and differs from the number of distinct workers in Table~\ref{tab:navigation-demand}.
\FloatBarrier

\subsection{Harness Primitives: Catalogs, Transfer, and Selection}

Section~\ref{sec:harness-experiments} compares STITCH's task-conditioned composition with the seed harness and with fixed and random compositions, and it repeats the comparison with actors not used in development (Tables~\ref{tab:harness-swe}, \ref{tab:harness-terminal}, and~\ref{tab:harness-transfer}). MemoHarness~\citep{huang2026memoharnessagentharnesseslearn} is a harness-optimization comparator in Tables~\ref{tab:harness-swe} and~\ref{tab:harness-terminal}, evaluated with the same mini-SWE-agent seed as STITCH~\citep{jin2026stitch}. The test-time libraries hold eight and six entries (Appendix~\ref{app:ledger}).

\begin{figure}[!htbp]
\centering
\begin{minipage}[t]{0.49\linewidth}\vspace{0pt}\includegraphics[width=\linewidth]{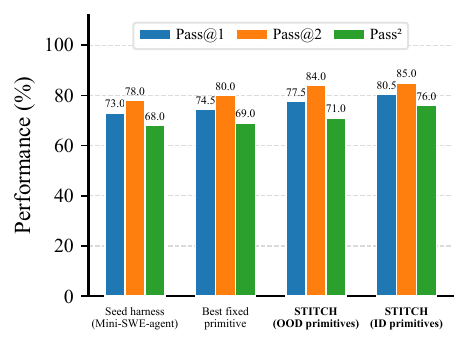}\end{minipage}\hfill
\begin{minipage}[t]{0.49\linewidth}\vspace{0pt}\includegraphics[width=\linewidth]{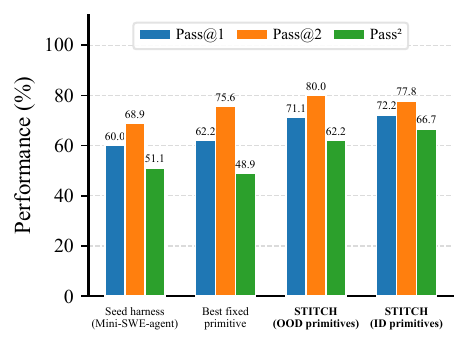}\end{minipage}
\caption{\textbf{Harness Primitive transfer across task domains.} Left, SWE-bench Verified, whose out-of-domain primitives were developed on Terminal-Bench~2.0; right, Terminal-Bench~2.0, whose out-of-domain primitives were developed on SWE-bench Verified. Implementations and composition contracts are retained while scopes are adjusted using development tasks; in-domain and out-of-domain libraries are compared with the seed and with the best fixed primitive that STITCH reports for this out-of-domain evaluation~\citep{jin2026stitch}.}
\label{fig:harness-domain-transfer}
\end{figure}

\begin{figure}[!htbp]
\centering
\begin{minipage}[t]{0.37\linewidth}\vspace{0pt}\includegraphics[width=\linewidth]{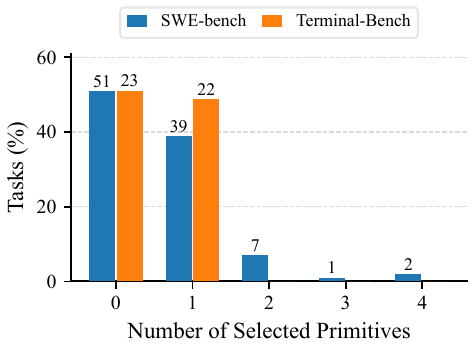}\end{minipage}\hfill
\begin{minipage}[t]{0.61\linewidth}\vspace{0pt}\includegraphics[width=\linewidth]{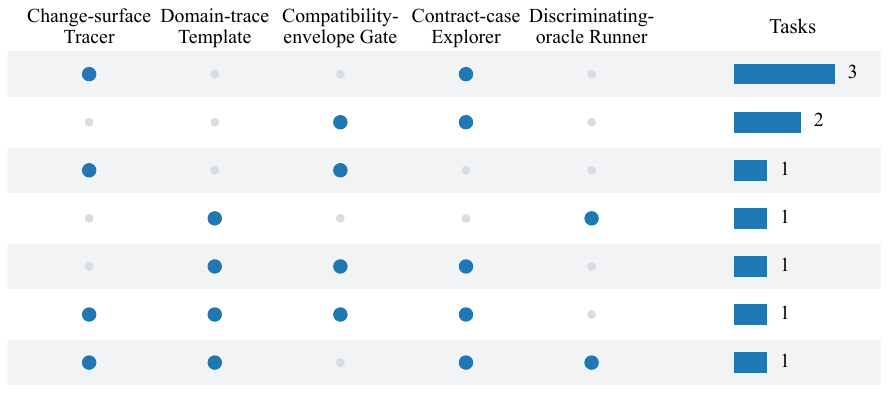}\end{minipage}
\caption{\textbf{Task-specific Harness Primitive selection patterns.} Share of tasks by the number of primitives that STITCH's composer selects (left; bar labels give task counts) and the multi-primitive combinations it selects, with their task counts (right), complementing the activation frequencies in Figure~\ref{fig:harness-overhead}~\citep{jin2026stitch}.}
\label{fig:harness-selection-patterns}
\end{figure}

Figure~\ref{fig:harness-domain-transfer} evaluates each benchmark with only the primitives developed on the other one. Implementations and contracts are kept, and scopes are adjusted on development tasks. With these out-of-domain primitives, STITCH's Pass@1 stays below its in-domain result but above the seed on both benchmarks. Its Pass@2 and Pass\textsuperscript{2} also stay above the seed, and on Terminal-Bench~2.0 its Pass@2 exceeds the in-domain run. Figure~\ref{fig:harness-selection-patterns} shows how many primitives the composer selects per task and which combinations it selects. The composer keeps the seed graph for about half of the tasks and otherwise mostly selects one primitive. This sparse selection complements the selection and activation rates of Figure~\ref{fig:harness-overhead}.
\FloatBarrier

\section{Study Settings and Evidence Ledger}
\label{app:ledger}
\runningsection{Study settings}

The quantitative record of this paper is the eight companion manuscripts. For each study, this appendix records its primitive in the study's own terms, its evaluated settings, the evidence that Sections~\ref{sec:instantiations} and~\ref{sec:analyses} use, and the scope of that evidence. Each subsection carries the identifier (\sid{S01}--\sid{S08}) of the companion manuscript's source snapshot from which its values are copied, the same identifier that the bibliography uses; because the identifiers name manuscripts, they do not follow the family order. The subsections follow the family order of Section~\ref{sec:instantiations}.

\subsection{\sid{S02}: Tool Primitives}

\textbf{Primitive.} A Tool Primitive is an LLM-based wrapper that encapsulates one ToolFace schema--function pair $(s_i,f_i)$, $\mathcal{P}_i(x;c)=\mathcal{M}([s_i;c;x])$, and all Tool Primitives share one Qwen3-8B backbone~\citep{jin2026toolprimitives}. HEART's harness has three mechanisms: request decomposition plans the task, value resolution grounds each step's values, and result validation judges every step's result. On a failed validation, the structured feedback enriches the context, and targeted revision re-plans for at most $B=3$ rounds. ToolFace contains 25,519 tools, built by collecting tools and manually authoring their schemas, and the set of Tool Primitives is induced by ToolFace and grows with it.

\textbf{Controls.} The matched-backbone table fixes Qwen3-8B and ToolFace (Table~\ref{tab:tool-matched}). ToolBench pass rates for ReAct, retrieval-only, planner--executor, DFSDT, and HEART are 33.5, 41.0, 52.6, 55.8, and 75.1\%, and their NESTFUL full accuracies are 0.08, 0.10, 0.19, 0.13, and 0.44. The budget-matched self-refinement control without Tool Primitives reaches 34.2\% and 0.15 (Table~\ref{tab:tool-budget}). Removing result validation gives 47.6\% and 0.15 in the component ablation (Table~\ref{tab:tool-ablation}).

\textbf{Scope.} The matched-backbone comparison holds Qwen3-8B and ToolFace fixed to isolate the harness (Section~\ref{sec:tool-experiments}), so it separates HEART from its backbone and catalogue. In the $\tau^2$-Bench airline domain, where the reference systems run on their own backbones, HEART succeeds in all four trials on 0.63 of the tasks against 0.50 for Claude-4.6-Sonnet (Table~\ref{tab:tool-tau}). The primitive-specific controls are the budget-matched control and the removal of ToolFace and the Tool Primitives, which leaves 16.1\%. The audit of result validation samples 200 decisions from ToolBench and NESTFUL trajectories (Table~\ref{tab:tool-verifier}). HEART's robustness study injects a malicious tool document into the retrieved candidates to manipulate tool selection, following ToolHijacker~\citep{shi2025prompt}, and reports a 0.0\% attack-success rate on ToolBench, because tool schemas stay in each primitive's own prompt and never enter the caller's~\citep{jin2026toolprimitives}.

\subsection{\sid{S04}: Code Primitives}

\textbf{Primitive.} A Code Primitive is a reusable executable software component equipped with a resident LLM, $P_i=(C_i,I_i,D_i,V_i,X_i,M_i)$: the implementation, the exposed interface and behavioural contract, the dependency closure, the carried validation tests, supporting context and provenance, and the resident LLM~\citep{jin2026codeprimitives}. One adaptation model instantiates every $M_i$. The resident LLM assesses the component's relevance to a requirement and, once the primitive is selected, adapts the component and reruns $V_i$ on it. CodeFace contains 1,424 entries, 871 mined, 343 harvested from verified construction runs, and 210 web-sourced, each admitted with executable closure, a passing component suite, and a recorded source repository, commit, and license.

\textbf{Benchmark.} LEGO-REPO contains 522 frozen reconstruction tasks from installable Python repositories, spanning seven domains, 22 capability tracks, and five difficulty bands. The benchmark removes an implementation while retaining its task interfaces, tests, and environment. Native passing-test counts are normalized between a floor and the original implementation's ceiling and clipped to $[0,1]$, so a delivery score is not binary task success. A score counts only for an admissible run, in which execution completes, provenance confirms that the tests exercised the generated code, and the span between floor and ceiling is at least five tests; any other run scores zero (Eq.~\eqref{eq:delivery-score}).

\textbf{Controls and Costs.} Feedback, feedback plus diagnosis, context-only reuse, unchanged import-and-call, structured-schema adaptation, and LEGO score 0.3180, 0.3469, 0.3544, 0.3430, 0.4974, and 0.5134 (Tables~\ref{tab:code-paradigms} and~\ref{tab:code-consumption}). With GPT-5.6 Terra in every model position, LEGO costs \$2.08 per task; with GPT-OSS-20B in the adaptation and diagnosis positions, it costs \$1.58 and scores 0.4880 (Table~\ref{tab:code-cost}). The DeepSeek-V3.2 row of Table~\ref{tab:code-cost} is one configuration of the study's sweep over thirteen backbones, in which adaptation and diagnosis stay on GPT-5.6 Terra.\fillpar[0.6][1]

\textbf{Scope.} Frozen native tests supply the acceptance specification. The field controls of Table~\ref{tab:code-fields} add fields cumulatively in one order, so their gains are increments in that order. In a model allocation, only the adaptation position belongs to the primitive. The study reports source-repository exclusion and corpus-disjoint controls.

\subsection{\sid{S05}: Math Primitives}

\textbf{Primitive.} The library holds 82 Method Contracts, one per reusable proof method extracted from Putnam 2000--2014 solutions, and each pairs a selection side, which the selector reads, with an execution side, which goes only to the prover~\citep{anonymous2026library}. The execution side holds a Lean scaffold stated for arbitrary parameters, a compiled worked instance, and Mathlib anchors. It also states the contract's required conditions $\Pre_c$ as Lean hypotheses and the residual obligations $\Obl_c$ left after the method's step. Lean provides the proof language and kernel~\citep{demoura2021lean}, and Mathlib provides the reusable mathematical library~\citep{mathlib2020}. GPT-5.6 Sol extracts the methods and proposes merges, all of which the authors review, and then drafts both sides of each contract. Every execution-side Lean artifact is compiled in Lean~4.27.0 with Mathlib commit \texttt{a3a10db}, the third of the three construction steps, and all 82 contracts pass. One batched call of a shared backbone writes, for each contract, a target, an action, required facts, cited verified material, an offer flag, a role, and a score. Residual obligations follow from the contract and its binding and are not proposal fields.\fillpar

\textbf{Evaluation.} Method selection uses 120 Putnam 2015--2025 problems and 60 IMO ProofBench problems. Reference methods are annotated from official solutions with model assistance and human checking, and fixed before selector evaluation. Each reranker reorders the top 10 contracts of an embedding ranking with the same backbone. For GPT-5.6 Luna, self-advertisement reaches Putnam hit@5 of 0.950, against 0.842 for Qwen3 with reranking, whose hit@1 is 0.683 (Table~\ref{tab:method-selection}). OpenAI-large with reranking has the best hit@1, 0.692, at hit@5 0.825, and self-advertisement's hit@1 is 0.658. IMO ProofBench hit@5 is 0.883 for these rerankers and 0.917 for self-advertisement. Adding the top-ranked contract to the prompt of AxProverBase under ten Lean compilations changes its proof rates from 5.8 to 12.5\% on Putnam and from 10.0 to 15.0\% on IMO ProofBench (Table~\ref{tab:method-proof}); pass@20 receives twice that budget.

\textbf{Scope.} Hits are scored against the annotated reference methods. A checked scaffold leaves its residual obligations to the prover, which completes 12.5\% of Putnam and 15.0\% of IMO ProofBench proofs with the contract within ten compilations (Table~\ref{tab:method-proof}). The downstream comparison adds contract content as well as selection. The coverage gain belongs to self-advertisement as a whole, commitment gate included. The study traces much of it to annotated methods that similarity ranks outside its top five (Figure~\ref{fig:method-ranks}). Ranking improvements depend on the criterion, and an additional runoff raises some metrics and lowers others.

\subsection{\sid{S06}: Knowledge Primitives}

\textbf{Primitive.} A Knowledge Primitive is SourceLearn's persistent, revisable source model $M$ of one authoritative source $D$, organized around source entities, whose representations capture properties, mechanisms, governing conditions, procedures, exceptions, distinctions, and relations to other entities, and whose units cite the source elements they rest on~\citep{jin2026sourcelearn}. At task time, source-model activation $A_B(M,q)$ supplies the whole model, or its task-relevant regions within $B=24$k tokens, to the task solver beside the Hybrid RAG evidence $R(D,q)$. The model changes only in the offline learning stages before test tasks, and every update after construction follows grounded source reconstruction: every proposed unit passes the grounding gate, which requires its cited source excerpts to entail it, and a region rewrite is committed only if the preservation check finds no previously supported meaning lost, unless a stronger statement subsumes it.

\textbf{Evaluation.} Five benchmarks are MultiDoc2Dial, NarrativeQA, SWE-QA, APIBench, and AppWorld, and three backends yield fifteen settings. All methods share the backend, the embedding model, and the answer or agent prompt, including the AppWorld scaffold. SourceLearn and AWM receive the same Hybrid RAG evidence, whereas RAPTOR and HippoRAG~2 use their own retrieval. Guidance and evaluation use fixed disjoint splits, with AppWorld using its official train/test-normal partition. SourceLearn improves over Hybrid RAG in fourteen settings and leads all evaluated methods in thirteen (Table~\ref{tab:knowledge-main}). The one setting without improvement is GPT-OSS-120B on AppWorld, from 30.4 to 27.4. On GPT-OSS-120B SWE-QA, SourceLearn improves on Hybrid RAG, but HippoRAG~2 leads with 56.7 against 53.4.

\textbf{Scope.} All QA results use one LLM judge, GPT-5.6 Luna, which is also an evaluated backend. Grounding measures support by the cited source. Learned units enter only through grounded source reconstruction, whose grounding gate is itself a model judgment. In the grounding audit, contradicted units stay at or below 1\%, but some Task-Guided units on the document and code sources are only partially supported or unsupported: full support of StudentAid units falls from 79.2\% in $M_0$ to 75.0\% in $M_S$ and 63.3\% in $M_T$ (Table~\ref{tab:knowledge-grounding}). The learned model is therefore not a replacement for the source, which stays authoritative and directly accessible through $R(D,q)$. SourceLearn also receives more source context than Hybrid RAG; at matched budgets it leads on three of the four sources (Table~\ref{tab:knowledge-budget}). Figure~\ref{fig:knowledge-reconstruction}(d) reports accuracy by the coverage of a question's required claims, an association across test questions.

\subsection{\sid{S01}: Agent Primitives}

\textbf{Primitive.} Review, Voting and Selection, and Planning and Execution each encapsulate a recurring multi-agent computation pattern as a reusable latent operator behind the external interface of a standard LLM agent; their internal agents follow task-agnostic role system prompts and coordinate exclusively through KV cache exchange~\citep{jin2026agentprimitives}. The study's Input-Output Alignment Assumption states that conditioning on a sender's KV cache approximates conditioning on its token sequence when the agents share model parameters and positional encoding, and the study re-indexes transferred keys under RoPE. An Organizer, GPT-5.2 by default, composes the three primitives per query, guided by the entries it retrieves from a Knowledge Pool of 45 MAS structures, and the system executes the selected primitives according to the composition plan.

\textbf{Evaluation and Extracted Evidence.} The main table reports average accuracy over AIME25, AIME24, MATH, GSM8K, HumanEval+, MBPP+, MedQA, and GPQA-Diamond (Table~\ref{tab:agent-main}). From the single agent to the composed MAS, average accuracy rises from 58.8\% to 75.3\% with Qwen3-8B, from 71.0\% to 77.6\% with DeepSeek-R1-Distill-Qwen-32B, and from 70.1\% to 76.4\% with DeepSeek-R1-Distill-Llama-70B.

\textbf{Scope.} The Organizer's backbone and its variable structure and agent count are part of the evaluated pipeline. Internal communication runs between agents that share one backbone. The study also compares text with cache communication in two stress tests, an auxiliary instruction injected into a long context and unrelated reasoning sentences injected as noise, which Section~\ref{sec:agent-experiments} summarizes. On DeepSeek-R1-Distill-Llama-70B, the LatentMAS baseline reaches 60.0\% average accuracy, below the single agent's 70.1\%, which the study attributes to its aggressive chunking of latent reasoning steps, in contrast to the conservative latent communication of Agent Primitives. Table~\ref{tab:agent-cost} reproduces an inconsistency of the study. Its Llama-3-70B Organizer row repeats the GPT-5.2 accuracies, 72.4, 93.8, 82.3, and 53.2\%, and derives its efficiencies from them and its own costs, whereas the study's Organizer-sensitivity table reports 72.0, 91.8, 81.0, and 52.2\% for that Organizer. The cost table prices the methods of the MAS comparison (Table~\ref{tab:agent-mas}).

\subsection{\sid{S03}: Dev-Primitives}

\textbf{Primitive.} Each component $a_i$ of the repository $\mathcal{R}=\{a_i\}_{i=1}^{N}$, a file by default, is associated with a Dev-Primitive $P_i$ that wraps it with a resident LLM; given its local objective $x_i$ and the communication context $\mathcal{C}_i$ of directed messages, it returns the optionally modified artifact and its messages~\citep{jin2026devprimitives}. The set of primitives is induced by the repository, a persistent file that an active Dev-Primitive creates is registered as a new Dev-Primitive, and each persists across the revision rounds of a task. Bug diagnosis is one LLM call over the executed repository that returns pass or fail and, on fail, the structured feedback $\phi=(e,c,u)$. The revision $\Pi'=\operatorname{ACTIVATE}(q,\mathcal{R}',\Pi,\phi)$ reactivates only the implicated primitives, within $B=3$ revision rounds after the initial round by default.

\textbf{Evaluation and Extracted Evidence.} SWE-bench Verified uses 500 instances. With GPT-5.6 Sol, the mini-SWE-agent comparator reaches $96.20\pm0.86$\% and HERMES $97.00\pm0.76$\%, reported as mean and standard error (Table~\ref{tab:dev-swe}). Recorded cost is \$1.15 versus \$1.80, and latency 182.37 versus 205.34 seconds. The DevOps-Gym comparison to Codex uses the same backbone and reports a four-stage average of 49.89 versus 55.41 (Table~\ref{tab:dev-ops}). On Terminal-Bench~4.0, over five runs of the 66 tasks with GPT-5.6 Sol, HERMES resolves 51.8\% against 33.0\% under Codex at medium effort and 55.5\% against 37.3\% at maximum effort (Table~\ref{tab:dev-terminal}). On the 20 SWE Refactor Bench migrations, the composite rises with GPT-5.6 Sol at medium effort from 6.5\% under Codex to 31.0\%.

\textbf{Scope.} The four-stage average is the unweighted mean of the stage success rates (Table~\ref{tab:dev-ops}), and task settings and repeated-run procedures vary across the benchmarks. In the backbone-scaling study (Table~\ref{tab:dev-roles}), only the configuration that keeps GPT-5.6 Sol for activation and diagnosis changes the Dev-Primitive model alone; it reaches 50.6\% against 51.8\% on Terminal-Bench. The configuration at 49.7\% also moves diagnosis to GPT-5.5. HERMES's Terminal-Bench version and task set differ from those of \sid{S08}.

\subsection{\sid{S07}: Navigation Primitives}

\textbf{Primitive.} A Navigation Primitive is a territory-backed worker of ANTMAN's static substrate map, and ANTMAN coordinates these scope-specialized workers by the information needs that remain unresolved~\citep{jin2026antman}. A deterministic, substrate-specific partitioning procedure splits the query's information space into $M$ territories, each with a WorkerCard that summarizes its scope and content, and within a substrate the workers share one search interface and, in each configuration, one backbone. The revisable Need Graph records each need's status, evidence, attempt history, and progress, and it controls worker selection, routing, and recovery. The substrate map stays fixed during execution, and what carries across queries is the partitioning procedure, the worker interface, and the coordination policy.\fillpar[0.6][1]

\textbf{Evaluation.} ANTMAN and all baselines use GPT-4.1; ANTMAN-H keeps the GPT-4.1 orchestrator with Qwen3-8B elsewhere. The structured-navigation comparison covers 108 RepoProbe questions across eight repositories, a frozen 80-question SWE-QA-Pro set, and GAIA-Text-103 (Table~\ref{tab:navigation-main}). ANTMAN's scores are 38.21, 79.78, and 58.3; GAIA iterative sparse retrieval scores 45.6 under the shared tool interface. GAIA uses a single territory-backed worker, so the coordination mechanisms it tests are need tracking and recovery. The controlled distractor-space intervention keeps the underlying needs approximately fixed while increasing space from 32K to 512K. Under it, active coordination grows $1.23\times$, against $15.29\times$ for LongAgent and $15.33\times$ for CoA (Table~\ref{tab:navigation-scaling}). A separate demand intervention fixes 512K and increases required evidence from one to four to sixteen units. Multi-document QA scores 30 questions from each of HotpotQA, 2WikiMultiHopQA, and MuSiQue for nine methods with one frozen answer-extraction and evaluation pipeline.

\textbf{Scope.} The demand intervention's calls are 35.5, 43.6, and 86.2 and costs \$0.107, \$0.111, and \$0.293 per query. The space intervention uses ten questions with evidence at early, middle, and late positions, and the demand intervention ten matched 512K contexts at each of its three demand levels~\citep{jin2026antman}. Full ANTMAN scores 60.00\% on the 40-question GAIA subset and 81.43 on the 40-question SWE-QA-Pro subset, sampled with seed 42, of Table~\ref{tab:navigation-ablation}, against 58.3\% on GAIA-Text-103 and 79.78 on the 80-question SWE-QA-Pro set (Table~\ref{tab:navigation-main}). Methods run in ANTMAN's evaluation harness share its question sets and model setting, and all GAIA methods share the tool interface.\fillpar[0.6][1]

\subsection{\sid{S08}: Harness Primitives}

\textbf{Primitive.} A Harness Primitive is an implemented, reusable control operation with an application scope and a composition contract~\citep{jin2026stitch}. Before the actor's first request, a harness composer on the actor's backbone selects primitives in a composition intent. A deterministic compiler connects them to the seed graph and supplies the dependencies that their contracts require, and the candidate is checked for supported primitives, compatibility with the execution specification $c_x$, valid connections, reachable termination, and bounded loops. An invalid candidate returns diagnostics to the composer for bounded correction, and if correction fails, the seed harness is retained. A primitive is accepted into the library after mechanism tests and a positive matched-run effect on development tasks.

\textbf{Evaluation.} SWE-bench Verified uses a 100-task evaluation subset and Terminal-Bench~2.0 a 45-task subset, each with two attempts per task and development tasks disjoint from evaluation. The actor and selector is GPT-5.6 Luna, and the development model is GPT-5.6 Sol. On SWE-bench Verified, seed and STITCH Pass@1 are 73.0 and 80.5\%; Pass@2 is 78.0 and 85.0\%; and success on both attempts is 68.0 and 76.0\% (Table~\ref{tab:harness-swe}). On Terminal-Bench~2.0, seed and STITCH Pass@1 are 60.0 and 72.2\%; Pass@2 68.9 and 77.8\%; and success on both attempts 51.1 and 66.7\% (Table~\ref{tab:harness-terminal}). Codex reaches 79.0\% Pass@1 on SWE-bench Verified and 56.7\% on Terminal-Bench~2.0.

\textbf{Scope.} These are sampled subsets and differ from \sid{S03}'s settings. The catalogs and fixed baselines cover six SWE-bench Verified and five Terminal-Bench~2.0 primitives, and the library-size curves of Figure~\ref{fig:harness-library} extend to test-time libraries of eight and six entries. The cross-actor evaluation (Table~\ref{tab:harness-transfer}, SWE-bench Verified only) uses actors held out from primitive development and adjusts the application scopes on development tasks. The 543- and 638-fold cost comparisons on SWE-bench Verified and Terminal-Bench~2.0 use an optimistic lower bound that counts only the output tokens of the generated implementation and excludes input, reasoning, and debugging costs.

\section{Evaluation Protocol for Cross-Family Composition}
\label{app:evaluation}
\runningsection{Evaluation protocol}

This appendix specifies an evaluation of \ours{} as one system that composes primitives of several families. Its experiments measure what the three results of Section~\ref{sec:analysis} predict, namely activation against realized demand (Proposition~\ref{prop:activation}), the gap between task-conditioned and fixed composition (Proposition~\ref{prop:fixed-vs-adaptive}), and the reach of a retrieval stage (Proposition~\ref{prop:coverage}), together with the reliability of the verification that each family brings. The companion studies measure these quantities within a single family: ANTMAN grows space and demand separately, STITCH compares fixed, random, and task-conditioned harnesses, and the study of self-advertised method selection compares self-advertisement with reranking of a similarity pool. The protocol specifies their measurement for compositions that span families.

\subsection{Primary Hypothesis and Task Construction}

The primary hypothesis is that composing primitives through \ours{} improves task success or total cost at a fixed evaluation standard, relative to comparable access to the same underlying assets. A benchmark should therefore contain requests that need at least two families and a task-level evaluation that cannot be replaced by adding the families' own scores. Suitable task templates include documented API migrations with repository regression tests, source-grounded workflows with mocked external actions, and formalized transformations that need both method selection and a Lean-checked proof of the selected method's required conditions and residual obligations.

Tasks should use frozen artifacts and environments, with separate development, library-validation, and held-out request sets. Library construction must not use held-out outcomes. Donor provenance, any overlap between donors and task repositories, and possible public benchmark exposure should be documented. Reference outcomes of guidance tasks should not enter persistent state directly: as in SourceLearn, they serve only to locate source evidence and derive source requirements; updates are reconstructed from the source. A task's success criteria and its grading process must be fixed before systems are compared. During a run, verification may use only the tests, builds, and evidence available in the task's environment, as in LEGO and HERMES, while held-out tests and grading outcomes are reserved for final scoring, and revision (Eq.~\eqref{eq:revise}) changes the plan but not the criteria.

\subsection{Comparators and Controlled Changes}

\begin{enumerate}
\item \textbf{Monolithic Execution}: one task-level agent uses the same underlying assets and authorized tools.
\item \textbf{Retrieved Context}: the same components, source material, and method descriptions are available as context, without primitive-local execution or adaptation.
\item \textbf{Executable Reuse}: components are invocable, but a task-level script manages selection, dependencies, evidence, and revision.\fillpar*
\item \textbf{Fixed Composition}: the best fixed subset of primitives, chosen on development tasks and applied to all tasks, which plays the role of the best static harness $H^*_{\mathrm{fixed}}$ of Proposition~\ref{prop:fixed-vs-adaptive}.
\item \textbf{Full \ours{}}: library selection and composition and evidence-based execution and improvement as described in Section~\ref{sec:harness-engineering}, with each family's own mechanisms for selection, verification, and revision (Section~\ref{sec:instantiations}) and each primitive's permissions enforced where it runs.
\item \textbf{Mechanism Ablations}: remove one mechanism at a time, such as relevance assessment, the passing of requirements between primitives, verification, or revision, while keeping the underlying assets fixed and grading each run by the preregistered task-level evaluation.
\end{enumerate}

The evaluation should hold the backbone, tool access, underlying assets, and total inference budget fixed, and add quality--cost curves for allocation strategies that one matched budget cannot represent fairly. An oracle selection condition, which activates on each task the primitives whose single-primitive runs improve it, approximates the oracle task-conditioned activation $\mathbf{z}^*(q)$ of Proposition~\ref{prop:fixed-vs-adaptive}; it estimates how much error is attributable to selection and is reported separately from deployable performance. Besides the shared assets, all comparators receive the same tests and source evidence.

\subsection{Measurements}

The primary quality measure is end-to-end task success under the preregistered task-level evaluation. The verdicts of each family's own verification should be reported beside it, together with the runs that end because the revision budget $B$ is spent. Most of these verdicts are pass or fail, such as those of HEART's result validation, of LEGO's and HERMES's diagnosis, of Lean's compilation, and of STITCH's validity check of each candidate graph. ANTMAN's need-resolution assessment is three-valued (resolved, partial, or unresolved), and its runs end in a synthesized answer that the benchmark grades (Section~\ref{sec:navigation-experiments}). Hidden checks stronger than the task-level evaluation, run on the composed system's own task distribution, should measure how often each verification passes a wrong result. HEART's audit of result validation on 200 sampled decisions measures this reliability within one family (Table~\ref{tab:tool-verifier}), and SourceLearn's grounding and preservation audits measure how well the stored units of its source models are supported by their cited source and preserve earlier meaning (Tables~\ref{tab:knowledge-grounding} and~\ref{tab:knowledge-preservation}). A composed system can route harder requests to a check than the check's own study did, so its rates must be measured under the composed distribution. False passes matter most, because they separate a more reliable system from one that reports success more often by weakening a check.

For Proposition~\ref{prop:fixed-vs-adaptive}, single-primitive arms, which apply one focused primitive with its required dependencies and shared supporting operations to all tasks, as STITCH's fixed arms do, estimate each primitive's effect $\Delta_i(q)$ against the arm without primitives. These estimates give the prevalence $\mu_i$ of the primitive's application scope, its mean in-scope gain $\bar g_i$, and its mean out-of-scope penalty $\bar\kappa_i$, and the final activations under Full \ours{} give the rates $\alpha_i$ and $\beta_i$. Together they predict $\Gamma_{\mathrm{fixed}}$ and the utility-weighted selection loss of Full \ours{} (Eq.~\eqref{eq:fixed-gap}). Comparing the predictions with the measured gaps between the oracle selection condition, Fixed Composition, and Full \ours{} tests the additive model on compositions that span families, where primitives of different families can help only together or interfere.

Library-size experiments should vary the number of irrelevant primitives in the library independently of task demand and relevant coverage, as ANTMAN's space intervention enlarges the distractor space at approximately fixed demand. They should measure candidate recall, assessment cost, the number of distinct active primitives against realized demand, and task quality. Where COMPOSE activates a primitive only by routing one unresolved need to it, as ANTMAN does, the routing count $K(q)$ and the realized demand $D(q)$ are observable, so each run can be checked against Eq.~\eqref{eq:activation}; elsewhere, required evidence can stand in for demand, as in ANTMAN's demand intervention. This separates selective execution, which Proposition~\ref{prop:activation} bounds, from discovery cost, which can grow with the library. For Proposition~\ref{prop:coverage}, each retrieval stage should report its own hit@$m$ and recall@$m$ beside the final hit@$k$ and recall@$k$, which Eq.~\eqref{eq:coverage-ceiling} bounds for every $k\le m$. This separates misses of retrieval from misses of assessment, as Figure~\ref{fig:method-ranks} does for Math Primitives and Table~\ref{tab:tool-retrieval} does for the retrieval stage of Tool Primitives.

Costs should include library construction and validation, indexing, relevance assessment, adaptation, execution, retries, final checks, hosting, and revalidation after updates. Model calls and tokens should be reported as well as any monetary values, with the recorded pricing convention and model labels. Latency should include queueing and p50/p95 values, together with cancellation and timeout rates. Cross-request reuse should report the one-time cost of building and validating the library separately from the per-task costs, together with the number of tasks that share it, as STITCH separates the offline development of its primitives from a composition cost of a few percent of one actor attempt.\fillpar

Diagnostic measurements should distinguish observed failures from causal attribution. Injected or independently verified defects can measure whether diagnosis implicates the primitives that own them, the set $\mathcal{I}_t$ of Eq.~\eqref{eq:verify}, and whether revision then reactivates only those primitives, as LEGO and HERMES do.

\subsection{Stress Conditions and Statistical Analysis}

Concurrency experiments should run concurrent tasks over one shared library, which no task changes (Section~\ref{sec:building}), with conflicting and disjoint edits. In LEGO and HERMES, a primitive edits only the component it owns and passes cross-component requirements, within the same round, to the primitives that must satisfy them; conflicting edits test this ownership when primitives of different families touch the same artifact. Drift experiments should revise API schemas, source documentation, runtime dependencies, and model compatibility while measuring invalidation and revalidation. Fault experiments should include unavailable dependencies, stale dependency records such as a Code Primitive's dependency closure $D_i$ or a Harness Primitive's composition contract, wrong relevance assessments, truncated evidence, malicious tool output, and uncertain external completion in a mock environment. These interventions test each family's own checks without irreversible real-world effects.

Task runs should be paired across conditions, with repeated seeds or attempts where the system is stochastic. Confidence intervals should come from task-level outcomes with an appropriate paired resampling or test, not from unpaired, rounded table means. Software task families and source groups may require clustered resampling. Sample size and stopping rules should rest on a prespecified, practically meaningful effect and a pilot variance estimate.

An evaluation supports the integrated system only if its task-level outcome and complete cost accounting justify composing primitives through \ours{}. A failure of interoperability or economics is reported as a specific finding about the system, separately from the overall average.

\end{document}